\documentclass[pdflatex,sn-nature,iicol]{sn-jnl}

\usepackage{graphicx}%
\usepackage{amsmath,amssymb,amsfonts}%
\usepackage{amsthm}%
\usepackage{mathrsfs}%
\usepackage[title]{appendix}%
\usepackage{listings}%
\usepackage{subcaption}
\usepackage{booktabs, multirow}%
\usepackage{soul}%
\usepackage[table]{xcolor}%
\usepackage{pifont}
\usepackage{natbib}
\usepackage{algorithm}%
\usepackage{algorithmicx}%
\usepackage{algpseudocode}%
\usepackage{pdflscape}
\usepackage{longtable}
\hypersetup{final=true}

\newcommand{\ourframework}{\textsc{DocXEval}}
\newcommand{\cmark}{\checkmark}
\newcommand{\xmark}{$\times$}
\newcommand{\layoutlm}{LayoutLMv3~\cite{huang_layoutlmv3_2022}}
\newcommand{\lilt}{LiLT~\cite{wang_lilt_2022}}
\newcommand{\bert}{BERT~\cite{devlin_bert_2019}}
\newcommand{\roberta}{RoBERTa~\cite{liu_roberta_2019}}
\newcommand{\rvlcdip}{RVL-CDIP~\cite{harley_evaluation_2015}}
\newcommand{\tobacco}{Tobacco3482~\citep{afzal_deepdocclassifier_2015}}
\newcommand{\doclaynet}{DocLayNet~\citep{pfitzmann_doclaynet_2022}}
\newcommand{\funsd}{FUNSD~\cite{funsd}}
\newcommand{\cord}{CORD~\citep{park_cord_2019}}
\newcommand{\sroie}{SROIE~\citep{huang_icdar2019_2019}}
\newcommand{\wildreceipts}{WildReceipts~\citep{Sun2021SpatialDG}}
\newcommand{\docvqa}{DocVQA~\citep{mathew_docvqa_2021}}
\newcommand{\adam}{\textsc{Adam}}
\newcommand{\adamw}{\textsc{Adam-W}}
\newcommand{\cls}{\textsc{CLS}}
\newcommand{\ser}{\textsc{SER}}
\newcommand{\qa}{\textsc{QA}}
\newcommand{\etal}{\textit{et al.}}
\newcommand{\saliency}{Saliency (SAL)~\cite{simonyan_deep_2013}}

\newcommand{\guidedbackprop}{Guided Backpropagation (GBP)~\cite{springenberg_striving_2015}}
\newcommand{\inputxgradient}{Input$\times$Gradient (IxG)~\cite{shrikumar_learning_2017}}
\newcommand{\deeplift}{DeepLIFT (DL)~\cite{shrikumar_learning_2017}}
\newcommand{\integratedgradients}{Integrated Gradients (IG)~\cite{sundararajan_axiomatic_2017}}

\newcommand{\deepshap}{DeepSHAP (DS)~\cite{lundberg_unified_2017}}
\newcommand{\kernelshap}{KernelSHAP (KS)~\cite{lundberg_unified_2017}}
\newcommand{\lime}{LIME~\cite{ribeiro_why_2016}}

\newcommand{\gradientshap}{GradientSHAP (GS)~\cite{lundberg_unified_2017}}
\newcommand{\occlusion}{Occlusion (OCC)~\cite{zeiler_visualizing_2014}}
\newcommand{\featureablation}{Feature Ablation (FA)~\cite{zeiler_visualizing_2014}}

\newcommand{\rawattn}{Raw Attention (RA)~\cite{devlin_bert_2019}}
\newcommand{\attnrollout}{Attention Rollout (AR)~\cite{abnar_quantifying_2020}}
\newcommand{\attnflow}{Attention Flow (AF)~\cite{abnar_quantifying_2020}}
\newcommand{\saliencyf}{SAL}

\newcommand{\guidedbackpropf}{GBP}
\newcommand{\inputxgradientf}{IxG}
\newcommand{\deepliftf}{DL}
\newcommand{\integratedgradientsf}{IG}

\newcommand{\kernelshapf}{KS}
\newcommand{\limef}{LIME}

\newcommand{\occlusionf}{OCC}
\newcommand{\featureablationf}{FA}

\newcommand{\rawattnf}{RA}
\newcommand{\attnrolloutf}{AR}

\newcommand{\completenessf}{Completeness (\textsc{compl})~\citep{sundararajan_axiomatic_2017}}
\newcommand{\completeness}{\textsc{compl}}
\newcommand{\nonsensitivityf}{Non-Sensitivity (\textsc{non-sens})~\citep{nguyen2020quantitativeaspectsmodelinterpretability}}
\newcommand{\nonsensitivity}{\textsc{non-sens}}
\newcommand{\aopcf}{Area Over Perturbation Curve (\textsc{aopc})~\citep{samek_evaluating_2017}}
\newcommand{\aopc}{\textsc{aopc}}
\newcommand{\abpcf}{Area Between Perturbation Curves (\textsc{abpc})~\citep{samek_evaluating_2017}}
\newcommand{\abpc}{\textsc{abpc}}
\newcommand{\sensitivitynf}{Sensitivity-N (\textsc{sens-n})~\citep{ancona2018betterunderstandinggradientbasedattribution}}
\newcommand{\sensitivityn}{\textsc{sens-n}}
\newcommand{\faithfulnesscorrf}{Faithfulness Correlation (\textsc{faith-corr})~\citep{bhatt2020evaluatingaggregatingfeaturebasedmodel}}
\newcommand{\faithfulnesscorr}{\textsc{faith-corr}}
\newcommand{\infidelityf}{Infidelity (\textsc{infid})~\citep{yeh_fidelity_2019}}
\newcommand{\infidelity}{\textsc{infid}}
\newcommand{\faithfulnessestf}{Faithfulness Estimation (\textsc{faith-est})~\citep{alvarez_melis_towards_2018}}
\newcommand{\faithfulnessest}{\textsc{faith-est}}
\newcommand{\monotonicityf}{Monotonicity (\textsc{mon})~\citep{arya2019explanationdoesfitall}}
\newcommand{\monotonicity}{\textsc{mon}}
\newcommand{\monotonicitycorrf}{Monotonicity Correlation (\textsc{mon-corr})~\citep{nguyen2020quantitativeaspectsmodelinterpretability}}
\newcommand{\monotonicitycorr}{\textsc{mon-corr}}
\newcommand{\sufficiencyf}{Sufficiency (\textsc{suf})~\citep{deyoung-etal-2020-eraser}}
\newcommand{\sufficiency}{\textsc{suf}}
\newcommand{\effectivecomplexityf}{Effective Complexity (\textsc{eff-comp})~\citep{nguyen2020quantitativeaspectsmodelinterpretability}}
\newcommand{\effectivecomplexity}{\textsc{eff-comp}}
\newcommand{\entropycomplexityf}{Entropy-based Complexity (\textsc{ent-comp})~\citep{bhatt2020evaluatingaggregatingfeaturebasedmodel}}
\newcommand{\entropycomplexity}{\textsc{ent-comp}}
\newcommand{\sundararajancomplexityf}{Sundararajan Complexity (\textsc{s-comp})~\citep{sundararajan_axiomatic_2017}}
\newcommand{\sundararajancomplexity}{\textsc{s-comp}}
\newcommand{\sparsenessf}{Sparseness (\textsc{spar})~\citep{chalasani2020conciseexplanationsneuralnetworks}}
\newcommand{\sparseness}{\textsc{spar}}
\newcommand{\maxsensitivityf}{Max Sensitivity (\textsc{max-sens})~\citep{yeh_fidelity_2019}}
\newcommand{\maxsensitivity}{\textsc{max-sens}}
\newcommand{\avgsensitivityf}{Avg Sensitivity (\textsc{avg-sens})~\citep{yeh_fidelity_2019}}
\newcommand{\avgsensitivity}{\textsc{avg-sens}}

\theoremstyle{thmstyleone}%
\theoremstyle{thmstyletwo}%
\theoremstyle{thmstylethree}%
\newcommand{\subheading}[1]{\medskip \noindent \textbf{#1}}

\usepackage{placeins}
\begin{document}
\title[\ourframework{}: A Systematic Evaluation of Explainability in Multimodal Document Understanding
]{\ourframework{}: A Systematic Evaluation of Explainability in Multimodal Document Understanding
}
\author*[1,2]{\fnm{Saifullah} \sur{Saifullah}}\email{firstname.lastname@dfki.de}
\author[1,2]{\fnm{David} \sur{Dembinsky}}
\author[1,3]{\fnm{Stefan} \sur{Agne}}
\author[1,2]{\fnm{Andreas} \sur{Dengel}}
\author[1,3]{\fnm{Sheraz} \sur{Ahmed}}
\affil*[1]{\orgdiv{Smart Data and Knowledge Services (SDS)}, \orgname{German Research Center for Artificial Intelligence (DFKI)}, \orgaddress{\street{Trippstadter Str. 122}, \city{Kaiserslautern}, \postcode{67663}, \country{Germany}}}
\affil[2]{\orgdiv{Department of Computer Science}, \orgname{University of Kaiserslautern-Landau (RPTU)}, \orgaddress{\street{Erwin-Schr\"odinger-Str. 52},  \city{Kaiserslautern}, \postcode{67663}, \country{Germany}}}
\affil[3]{\orgname{DeepReader GmbH}, \orgaddress{ \city{Kaiserslautern}, \postcode{67663}, \country{Germany}}}
\abstract{
  Multimodal transformer models have driven unprecedented performance breakthroughs in modern document understanding. However, the inherent black-box nature of these models remains a major obstacle to their safe deployment in high-stakes sectors. Despite significant progress in explainability for vision and language tasks, explainable AI (XAI) research tailored to multimodal document understanding has received little attention. To address this gap, we present \ourframework{}, the first comprehensive explainability framework that adapts existing XAI methods and evaluation metrics to the multimodal (text, layout, and image) document understanding context, while introducing efficient multi-target adaptations of both methods and metrics for domain-specific tasks such as semantic entity recognition. Using \ourframework{}, we systematically evaluate 13 XAI methods across 3 document understanding tasks, 8 datasets, 4 models, and 17 evaluation metrics, and find that only a few methods perform consistently across all evaluation criteria, with Occlusion emerging as the most faithful and interpretable method overall.
  Through a dedicated meta-evaluation, we observe moderate to strong agreement among several faithfulness and complexity metrics while identifying certain ones that produce unreliable results.
  Finally, through multiple dataset-wide diagnostic analyses, we uncover key insights into model behavior, including the contribution of individual modalities across tasks and labels, spatial attribution patterns around target entities, and potential dataset biases.
  Overall, this work serves as both a benchmark and a practical guide for future XAI research in multimodal document understanding.
  To support reproducibility, we publicly release our code at \url{https://github.com/saifullah3396/docxeval/}, along with the \texttt{torchxai} library implementing all multimodal and multi-target method and metric adaptations at \url{https://atriaml.github.io/torchxai/}.
}
\maketitle
\section{Introduction} \label{sec:introduction} Deep learning has driven state-of-the-art performance in document understanding across a wide range of tasks, including document classification~\citep{ferrando_improving_2020,huang_layoutlmv3_2022}, information extraction~\citep{xu_layoutlm_2020,xu_layoutlmv2_2021}, document visual question answering~\citep{borchmann_due_2021}, and more~\citep{wang_document_2025,shen_layoutparser_2021}. As multimodal document understanding models become increasingly integrated into automated decision-making pipelines~\citep{powalski_going_2021,huang_layoutlmv3_2022,garncarek_lambert_2021,wang_lilt_2022,lamott_slimdoc_2025}, ensuring the transparency of their decisions has become critical, especially for deployment in high-stakes sectors where inadvertent errors can lead to devastating legal, financial, or societal consequences~\citep{bibal_legal_2021,perc_social_2019}. However, despite a vast array of explainable AI (XAI) methods~\citep{lundberg_consistent_2019,ribeiro_why_2016} and an equally diverse set of XAI evaluation metrics~\citep{samek_explainable_2017,adebayo_sanity_2018,yeh_fidelity_2019} proposed for vision and language models~\citep{vilone_explainable_2020,ras_explainable_2022,Danilevsky2020,dembinsky2025unifying}, there is little guidance on extending these methods and metrics to the multimodal document understanding context, where models typically combine multiple heterogeneous input modalities, such as text, layout, and visual/image features, in an end-to-end fashion~\citep{xu_layoutlmv2_2021,huang_layoutlmv3_2022,wang_lilt_2022,appalaraju_docformer_2021}.

Existing XAI methods and evaluation metrics are typically sensitive to numerous configurable parameters, including the choice of baselines, feature grouping strategies, and attribution aggregation schemes~\citep{samek_explainable_2017,lundberg_consistent_2019,ribeiro_why_2016}. In multimodal document understanding, where features originate from multiple modalities, these design choices become even more complex, as decisions must be made separately for each modality. For example, a meaningful baseline (i.e., one representing the absence of information) for text tokens differs fundamentally from one for continuous layout coordinates or visual patches.
Analogous decisions arise for feature grouping and aggregation. For instance, should text attributions be aggregated at the token, word, or line level, and should visual attributions be aggregated at the pixel or patch level?
Each such decision critically affects both the resulting explanations and the evaluation metrics used to assess them~\citep{adebayo_sanity_2018,yeh_fidelity_2019}, yet no principled guidelines exist for these choices in the existing XAI literature on document understanding.

A separate challenge arises from the nature of the tasks themselves. Many core document understanding tasks, such as semantic entity recognition and question answering, involve multiple predictions per input sequence. For example, in semantic entity recognition, each token is independently classified, resulting in numerous explanation targets per sample.
In contrast, existing XAI methods~\citep{ribeiro_why_2016,lundberg_unified_2017,samek_explainable_2017} and evaluation metrics~\citep{yeh_fidelity_2019,klein_navigating_nodate} are typically designed around single-target scenarios such as image classification. This makes them either computationally infeasible or conceptually ill-suited for direct application to multi-target document tasks.

In the absence of a principled framework for reasoning about these design choices in the multimodal document understanding context, several fundamental research questions (RQs) remain unanswered. \textbf{RQ1:} How should existing XAI methods and evaluation metrics be adapted to handle multiple heterogeneous input modalities and multi-target prediction tasks in document understanding? \textbf{RQ2:} Which XAI methods yield the most faithful and interpretable explanations in multimodal document understanding? \textbf{RQ3:} Given the widespread metric disagreement documented in prior work~\citep{klein_navigating_nodate,tomsett_sanity_2020}, what agreement and disagreement patterns emerge among evaluation metrics in this domain? \textbf{RQ4:} How can the resulting explanations be systematically analyzed to gain insights into model behavior across modalities?
To address these RQs, we make the following contributions:
\begin{enumerate}
  \item We introduce \ourframework{}, the first comprehensive explainability framework for multimodal document understanding that adapts $13$ XAI methods and $17$ evaluation metrics to generate and evaluate multimodal explanations, and additionally introduces efficient adaptations of both methods and metrics for multi-target tasks such as semantic entity recognition.

  \item Using \ourframework{}, we systematically benchmark these methods across $3$ tasks, $8$ datasets, and $4$ models, and find that only a few methods perform consistently across all evaluation criteria, with \occlusion{} emerging as the most faithful and interpretable overall.

  \item We conduct a dedicated meta-evaluation of metrics and find moderate to strong agreement among several faithfulness and complexity metrics, while identifying \monotonicityf{} and \infidelityf{}\footnote{Note that Infidelity in this work refers to the noisy baseline formulation proposed in the original work~\cite{yeh_fidelity_2019}.} as unreliable.

  \item Through multiple dataset-wide diagnostic analyses, we uncover key insights into model behavior, such as variable modality reliance across tasks and labels, potential reliance on task-irrelevant features, and spatial attribution patterns around target entities.
\end{enumerate}

\section{Related Work}
\subheading{Multimodal Document Understanding.}
While the broader document understanding landscape is increasingly shifting towards generative approaches~\citep{wang_document_2025,wang_docllm_2024,fujitake-2024-layoutllm}, discriminative multimodal document foundation models (predominantly transformer-based~\citep{vaswani_attention_2017}) remain a prominent choice for structured document understanding tasks~\citep{huang_layoutlmv3_2022,powalski_going_2021,shen_layoutparser_2021,lamott_slimdoc_2025}. This is primarily due to their ease of adaptation (such as fine-tuning on downstream tasks), reduced data privacy risks (since most models are small enough to be deployed in-house), and lower computational costs.
Multimodal document understanding models typically employ early or late fusion techniques to integrate multiple document modalities. For example, the LayoutLM family~\citep{xu_layoutlm_2020,xu_layoutlmv2_2021,huang_layoutlmv3_2022} fuses text, position, and layout modalities directly through the early addition of token-level embeddings, and fuses the image modality through self-attention between image and text tokens. LiLT~\citep{wang_lilt_2022}, on the other hand, omits the image modality entirely and takes text and layout as separate input tokens, which are fused through bi-directional cross-attention. In contrast, DocFormer~\citep{appalaraju_docformer_2021} proposes a unique multimodal attention layer to fuse the image, text, and layout features, and additionally uses residual connections to inject visual and layout information into each attention block.
Several similar approaches have been proposed in the literature, differing in their modality fusion and training strategies~\citep{powalski_going_2021,garncarek_lambert_2021,luo2023geolayoutlmgeometricpretrainingvisual}; however, few, if any, of these works explicitly consider model explainability.

\subheading{Explainable AI.}
The field of XAI has produced a rich body of literature over the years~\cite{ras_explainable_2022}, with numerous methods~\cite{simonyan_deep_2013,springenberg_striving_2015,ribeiro_why_2016,lundberg_unified_2017,sundararajan_axiomatic_2017,zeiler_visualizing_2014} proposed to explain the decisions of black-box AI models. These methods can be largely categorized into two families: (1) \textit{post-hoc} methods, which generate explanations post-training via an external explainer, and (2) \textit{intrinsic} methods, which build interpretability directly into the model architecture.
Among post-hoc methods, feature-attribution (FA) methods are the most prominent. These methods assign importance to input features or feature groups by estimating their relevance to the model's decision, typically resulting in importance maps that highlight both positively and negatively influencing features~\cite{simonyan_deep_2013,lundberg_unified_2017,sundararajan_axiomatic_2017,shrikumar_learning_2017}.
FA methods can be further divided into two main categories: backpropagation-based methods~\cite{simonyan_deep_2013,sundararajan_axiomatic_2017} and perturbation-based methods~\cite{ribeiro_why_2016,lundberg_unified_2017,zeiler_visualizing_2014}.

Backpropagation-based methods typically depend on propagating feature relevance backward through the model from the output to the input, utilizing either gradients~\cite{simonyan_deep_2013,sundararajan_axiomatic_2017}, modified gradient rules~\cite{zeiler_visualizing_2014,springenberg_striving_2015}, or activation differences~\cite{shrikumar_learning_2017}. Popular examples include \saliency{}, \inputxgradient{}, \integratedgradients{}, \deeplift{}, and \deepshap{}, among others.

Perturbation-based methods generate importance maps by perturbing the input features and evaluating the impact of the perturbation on the model's output~\cite{ras_explainable_2022}. These methods are especially valued for their model-agnostic nature, making them applicable to any model without requiring access to its internal architecture. However, this advantage is generally accompanied by increased computational overhead. Prominent perturbation-based methods include \featureablation{}, \occlusion{}, \lime{}, and \kernelshap{}.

Recent work has also adapted feature attribution methods to document understanding tasks~\cite{saifullah_docvce_2025,saifullah_docxclassifier_2024,pouliquen2024weaklysupervisedtraininghologram,brini_end--end_2022,souibgui_docvxqa_2025}. Saifullah~\etal~\cite{saifullah_reality_2024} adapted existing feature-attribution methods to explain document image classification models and later proposed DocXplain~\cite{saifullah_docxplain_2024}, a perturbation-based explainability framework for document image classification. Brini~\etal~\cite{brini_end--end_2022} adapted gradient-based methods to explain document layout analysis models.  Pouliquen~\etal~\cite{pouliquen2024weaklysupervisedtraininghologram} used gradient-based feature-attribution for hologram verification in documents. In a different direction, Saifullah~\etal~\cite{saifullah_docxclassifier_2024} explored intrinsic explainability for document image classification by augmenting existing models with an attention-based explainability head. Despite these advances, existing work remains focused on unimodal explanations and single-target scenarios.
\begin{figure*}[t!]

	\centering

	\includegraphics[width=\textwidth,draft=false]{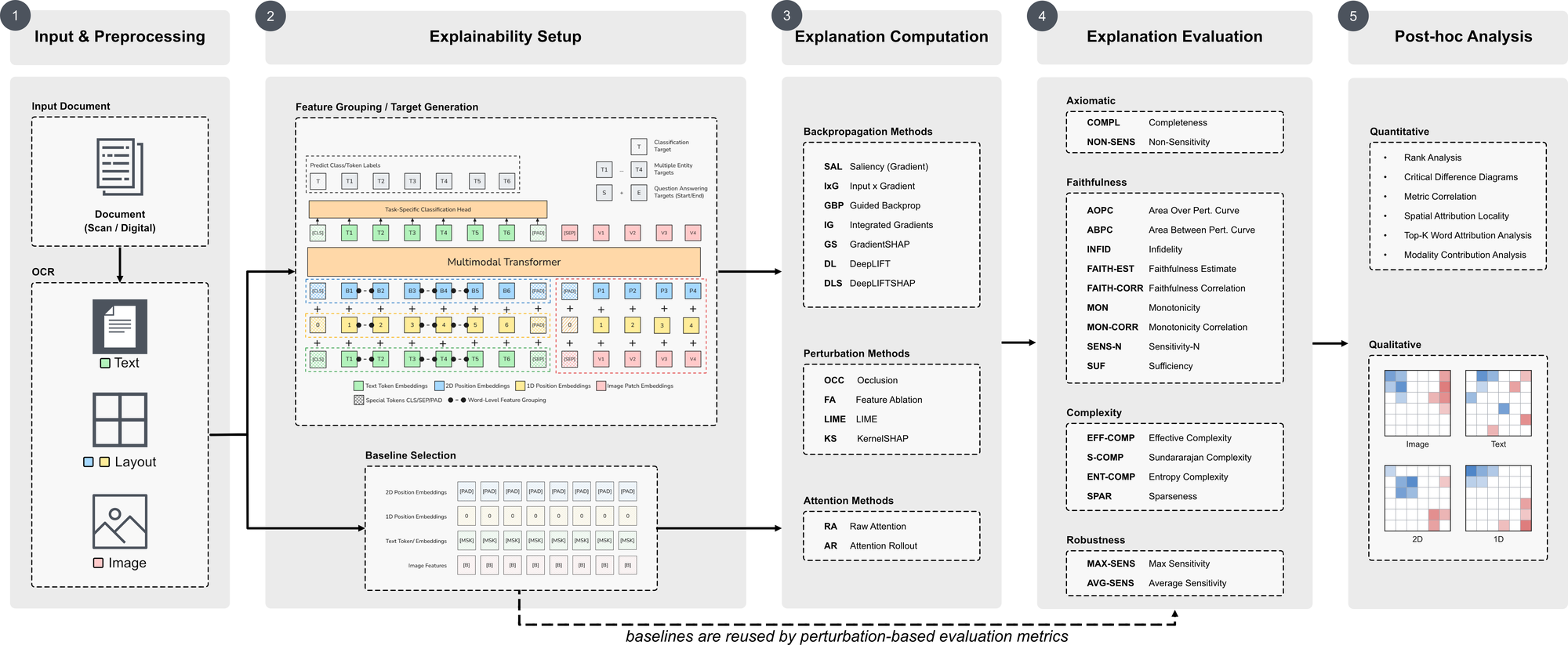}

	\caption{Overview of the \ourframework{} framework. Given a dataset and a trained model, the framework (1) extracts modality-specific inputs from each document, (2) defines feature segmentation masks, reference baselines per modality, and targets, (3) computes modality-wise attribution scores via XAI methods, (4) evaluates explanation quality across four XAI evaluation criteria, and (5) conducts a systematic qualitative and quantitative analysis of results across datasets, models, and tasks.\label{fig:framework}}
\end{figure*}

\subheading{Multimodal Explainable AI.}
Explainable AI for multimodal models has also gained significant attention in recent years. However, the field has primarily focused on vision-language tasks such as visual question answering~\citep{goyal_towards_2016,lyu_dime_2022,aflalo_vl-interpret_2022} and image captioning~\citep{parcalabescu_mm-shap_2023}. For instance, VL-InterpreT~\citep{aflalo_vl-interpret_2022} visualizes attention patterns via heatmaps for vision-language transformers. M\textsuperscript{2}Lens~\citep{wang_m2lens_2022} provides an interactive analytics system integrating feature-attribution methods~\citep{ribeiro_why_2016,lundberg_unified_2017,sundararajan_axiomatic_2017} for multimodal sentiment analysis. Other works~\citep{hessel_does_2020,lyu_dime_2022} decompose multimodal predictions into unimodal and cross-modal components for vision-language tasks to quantify interaction effects. In the document domain, Souibgui~\etal~\citep{souibgui_docvxqa_2025} proposed DocVXQA for explaining generative DocVQA models; however, it operates on visual inputs only and is not applicable to discriminative multimodal models.
While these works have advanced multimodal explainability considerably, they remain scoped to vision-language settings. Furthermore, most rely primarily on qualitative visual analysis~\cite{aflalo_vl-interpret_2022,lyu_dime_2022,wang_m2lens_2022}, and those that do employ quantitative evaluation typically use only a small subset of metrics~\cite{souibgui_docvxqa_2025}.
To the best of our knowledge, no existing work systematically adapts and evaluates XAI methods and metrics for multimodal document understanding, the central research gap that we aim to address in this work.

\section{\ourframework{}: Our Framework}
\label{sec:framework}
\subheading{Problem Formulation.} We consider a general document understanding setup, where a model $f_\theta$ is trained on a dataset $\mathcal{D} = \{(x_i, y_i)\}_{i=1}^N$ of document-label pairs for a downstream task $T$. Each document $x \in \mathcal{D}$ is represented by up to four modalities\footnote{While we consider four modalities in this work, the framework is not limited to them and can directly accommodate additional modalities depending on the task.}: textual content $\mathcal{T}$, 1D sequence positions $\mathcal{S}$, 2D layout information $\mathcal{L}$, and visual features $\mathcal{V}$, which the model jointly processes to produce task-level predictions. We assume a preprocessing function $\phi: x \mapsto (\mathcal{T}, \mathcal{S}, \mathcal{L}, \mathcal{V})$ that extracts these modalities from each document, e.g., via standard optical character recognition (OCR) pipelines commonly used in multimodal document understanding~\citep{xu_layoutlmv2_2021,powalski_going_2021,wang_lilt_2022}. Given this setting, we seek to (i) explain model predictions through modality-wise feature attributions $\mathbf{A}_{\mathcal{M}}$, and (ii) systematically evaluate them in terms of axiomatic soundness, faithfulness, complexity, and robustness. Here, $\mathbf{A}_{\mathcal{M}}$ denotes the attribution scores assigned to modality $\mathcal{M} \in \{\mathcal{T}, \mathcal{S}, \mathcal{L}, \mathcal{V}\}$ for a given prediction target.

\subheading{Framework Overview.} As illustrated in Fig.~\ref{fig:framework}, \ourframework{} operates in five steps: (1) modality-specific inputs are extracted from each document and projected to input embeddings; (2) feature segmentation masks and reference baselines are defined per modality, and task-specific attribution targets are generated; (3) adapted XAI methods are applied to produce modality-wise explanations with respect to each target; (4) the resulting explanations are evaluated through standardized metrics to generate evaluation scores for each sample-target pair; and (5) evaluation scores are aggregated per dataset and systematically analyzed across datasets, models, and tasks. We describe the main components of the framework in the following sections.

\subsection{Input and Preprocessing}
\label{sec:input_preprocessing}
The first stage of the framework prepares a uniform attribution space across models and modalities.
Multimodal document understanding models~\cite{xu_layoutlm_2020,appalaraju_docformer_2021,powalski_going_2021} generally differ in how they represent inputs. For example, text may be tokenized using different vocabularies, spatial positioning may be encoded as a bounding-box or polygon coordinates, and visual features may be derived from CNN-based embeddings~\citep{appalaraju_docformer_2021,xu_layoutlm_2020} or patch-based projections~\citep{huang_layoutlmv3_2022}. Irrespective of these differences, however, the models typically map their discrete token-level inputs to continuous embedding representations via modality-specific embedding functions. We denote these as $\pi_{\mathcal{M}}: \mathcal{M} \rightarrow \mathbb{R}^{d}, \quad \mathcal{M} \in \{\mathcal{T}, \mathcal{S}, \mathcal{L}\}$. For token-level inputs, our framework operates at this embedding level, meaning the attribution scores are computed over the full embedding space of each token. This provides three main advantages: (1) a consistent attributable unit across models regardless of tokenization scheme or input representation; (2) a common dimensionality across all token-level modalities, ensuring each modality has the same number of attributable features per token; and (3) a uniform input representation across both gradient-based and perturbation-based methods\footnote{Gradient-based methods require continuous representations for computing attribution scores, and while perturbation-based methods do not strictly require them, a uniform input representation across both ensures a consistent comparison and evaluation.}. For visual inputs $\mathcal{V}$, our framework operates directly on raw image pixels rather than model-specific visual embeddings (e.g., patch projections), as raw images are inherently continuous and can be directly segmented into a set of attributable features.

\subsection{Feature Segmentation and Aggregation}
\label{sec:feature_segmentation}
In the second stage, we define a feature segmentation mask for each modality to group features into a smaller set of semantically meaningful units. This is necessary for three reasons:
\begin{enumerate}
	\item 	Since our framework operates on token-level embeddings (each of dimension $\mathbb{R}^d$) and raw image space (typically of dimensions $\mathbb{R}^{C \times H \times W}$), the resulting attributable feature space becomes very large. For instance, an input sequence of length $L=512$ with each token of dimension $\mathbb{R}^d$, combined with visual features of dimension $\mathbb{R}^{224 \times 224 \times 3}$, yields hundreds of thousands of features across modalities, making explanations expensive to compute and evaluate.
	\item Perturbation-based methods inherently operate over segmented features, producing coarser explanations with lower complexity scores, whereas gradient-based methods produce fine-grained per-feature attributions. Therefore, aggregating explanations over the same set of features is necessary for a fair cross-method evaluation.
	\item Aggregating explanations into human-understandable units improves interpretability; for instance, word-level attributions are more meaningful than subword-level scores, and image patch attributions are more interpretable than pixel-level maps.
\end{enumerate}

\subheading{Segmentation Strategy.}
Formally, we define feature segmentation through a mask that assigns each input feature to a semantically coherent group. For discrete modalities $\mathcal{M} \in \{\mathcal{T}, \mathcal{S}, \mathcal{L}\}$ with $L$ tokens and embedding dimension $d$, we define a feature-level mask $m_{\mathcal{M}} \in \mathbb{N}^{L \times d}$ where all $d$ features of token $j$ share the same group assignment $k$. The feature set for group $k$ in modality $\mathcal{M}$ is then:
\begin{equation}
\label{eq:1}
\mathcal{F}_{\mathcal{M}, k} = \{\pi_{\mathcal{M}}(t_j, l) \mid m_{\mathcal{M}}(j, l) = k\}
\end{equation}
where $\pi_{\mathcal{M}}(t_j, l)$ denotes the $l$-th element of the embedding of token $j$.
Similarly, for visual modality $\mathcal{V}$, we define a feature-level mask $m_{\mathcal{V}} \in \mathbb{N}^{H \times W \times C}$ where all $C$ channels of pixel $(x, y)$ share the same group assignment $p$. The feature set for visual group $p$ is:
\begin{equation}
\label{eq:2}
\mathcal{F}_{\mathcal{V}, p} = \{\mathcal{V}(x, y, c) \mid m_{\mathcal{V}}(x, y, c) = p\}
\end{equation}
In this work, we employ word-level grouping for discrete modalities, where each group $k$ in Eq.~\eqref{eq:1} corresponds to a complete word, and patch-level grouping for visual features, where each group $p$ in Eq.~\eqref{eq:2} corresponds to a spatial image region.

\subheading{Feature Aggregation.}
Given fine-grained attribution scores computed at the embedding element or pixel level, we aggregate them to group-level scores for downstream evaluation and analysis. For a modality $\mathcal{M}$ with scalar attribution scores $a_{\mathcal{M},i} \in \mathbb{R}$ assigned to each individual feature $i$, we first sum all attribution scores within a group and then normalize by a scaling factor $w_{\mathcal{M}, k}$:
\begin{equation}
\begin{split}
a_{\mathcal{M}, k} &= w_{\mathcal{M}, k} \sum_{i \in \mathcal{G}_{\mathcal{M}, k}} a_{\mathcal{M}, i}, \\
w_{\mathcal{M}, k} &= \frac{\min_{\mathcal{M}', k'} |\mathcal{G}_{\mathcal{M}', k'}|}{|\mathcal{G}_{\mathcal{M}, k}|}
\end{split}
\end{equation}
where $\mathcal{G}_{\mathcal{M}, k} = \{i \mid m_{\mathcal{M}, i} = k\}$ denotes the set of scalar feature indices in group $k$. The scaling factor $w_{\mathcal{M}, k}$ is defined as the ratio of the smallest group size across all modalities to the size of the current group.
This ensures that attribution scores for the smallest feature group remain unscaled, while those for larger groups are scaled down proportionally to prevent their larger feature counts from inflating their importance. For instance, in our setting, a single-token word with embedding dimension $d=768$ and an image patch of dimension $16\times16\times3=768$ both receive a weight of $1.0$, retaining their original scores, whereas a word spanning $n$ tokens has its attributions downscaled by a factor of $n$.

\subsection{Baseline Selection}
\label{sec:baseline_selection}
\begin{figure*}[t]
	\centering
	\includegraphics[width=0.49\textwidth,trim=0 543.6bp 0 9.6bp,clip,draft=false]{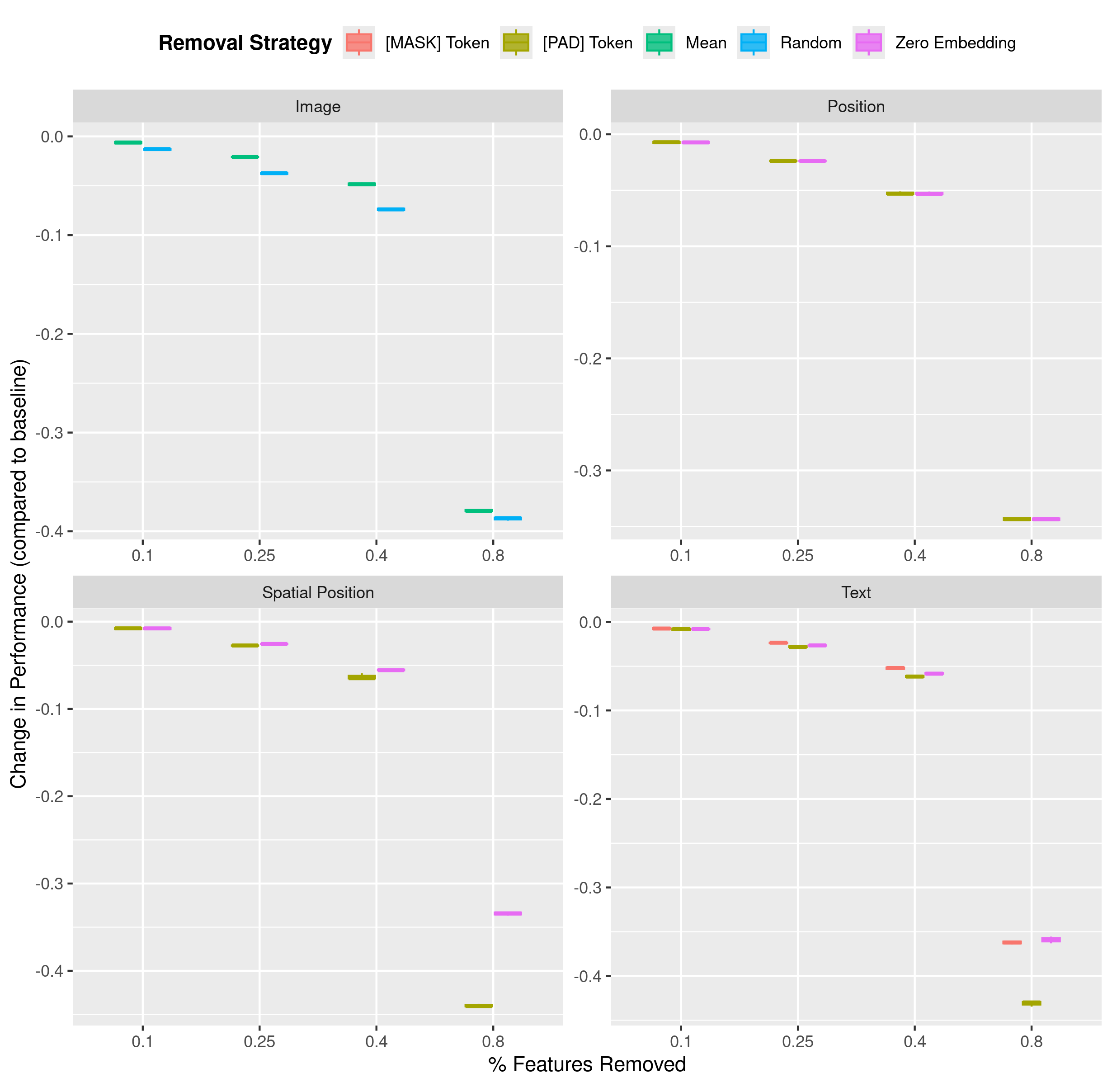}\\[6pt]
	\begin{subfigure}[t]{0.49\textwidth}
		\centering
		\includegraphics[width=\textwidth,trim=0 0 0 43.2bp,clip,draft=false]{images/methodology/pert_rvlcdip.png}
		\caption{\rvlcdip{} (classification)}
		\label{fig:baseline_ablation_rvlcdip}
	\end{subfigure}
	\hfill
	\begin{subfigure}[t]{0.49\textwidth}
		\centering
		\includegraphics[width=\textwidth,trim=0 0 0 43.2bp,clip,draft=false]{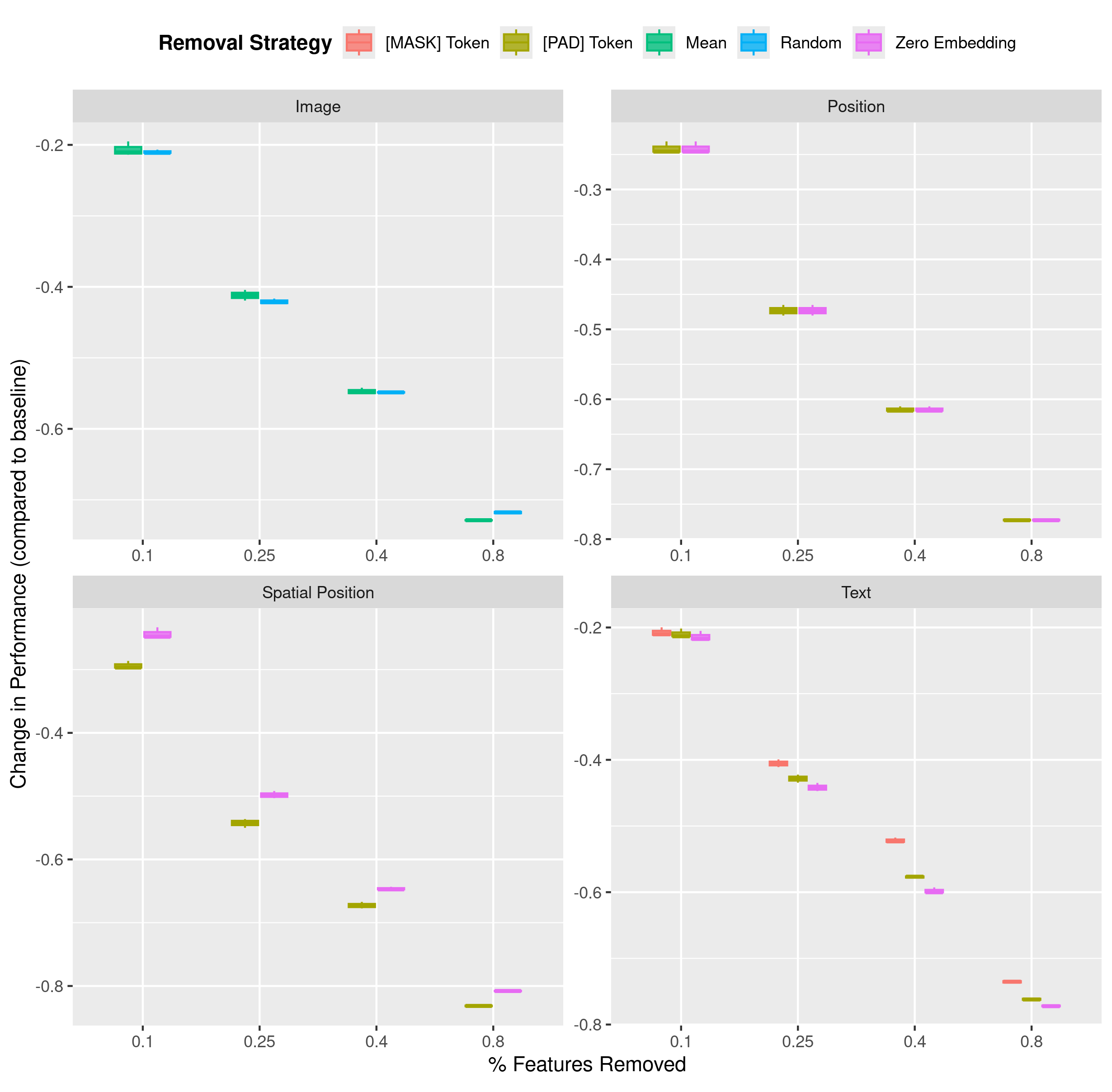}
		\caption{\funsd{} (SER)}
		\label{fig:baseline_ablation_funsd}
	\end{subfigure}
	\caption{
		Modality-wise model sensitivity to different baselines, shown for \layoutlm{} on \rvlcdip{} and \funsd{}. The model is generally less sensitive to the \textsc{Mean} baseline for the image modality on both datasets, whereas for spatial position, it is less sensitive to the \textsc{Zero Embedding} baseline compared to the \textsc{[PAD]} baseline. For the text modality, the model is less sensitive to the \textsc{[MASK]} token compared to other baselines.
	}
	\label{fig:baseline_ablation}
\end{figure*}

Many feature attribution methods~\citep{sundararajan_axiomatic_2017,ribeiro_why_2016,lundberg_unified_2017} and evaluation metrics~\citep{klein_navigating_nodate} require a reference baseline in their computation. For example, perturbation-based XAI methods and evaluation metrics assess the impact of input features on model output by replacing their values with the baseline. However, selecting a suitable baseline is not straightforward as it is possible that replacing a feature with a baseline produces an out-of-distribution input (unseen during training), and therefore the resulting impact on the model output may not reflect actual information removal but rather a distribution shift~\citep{hase_out--distribution_2021}.
In a multimodal setup, this becomes further complicated, as each modality requires its own baseline, and the optimal combination of modality-specific baselines for minimizing the out-of-distribution effect is not obvious.

To address this, we propose a perturbation-based baseline selection strategy which extends the \texttt{Replace} strategy proposed by Hase~\cite{hase_out--distribution_2021} to the multimodal context. In particular, for each modality $\mathcal{M}$, we define a set of baseline candidates $\mathcal{B}_{\mathcal{M}} = \{b_1^{\mathcal{M}}, \ldots, b_{n}^{\mathcal{M}}\}$ representing different mechanisms for removing information from that modality, and construct the Cartesian product $\mathcal{C} = \prod_{\mathcal{M}} \mathcal{B}_{\mathcal{M}}$ over all modality baselines. Then, for each combination $\mathbf{c} \in \mathcal{C}$ and varying perturbation levels $\alpha \in \{0.1, 0.25, 0.4, 0.8\}$, we compute the performance of the pretrained model on the perturbed test set, where for each sample we randomly replace a fraction $\alpha$ of feature groups (as defined in Section~\ref{sec:feature_segmentation}) with their corresponding baseline values.
We repeat this process over $R=3$ random seeds, and compute the mean performance:
\begin{equation}
	\bar{f}(\mathbf{c}, \alpha) = \frac{1}{R} \sum_{r=1}^{R} f(\mathbf{c}, \alpha, r)
\end{equation}
where $f(\mathbf{c}, \alpha, r)$ refers to a task-specific metric such as accuracy or F1-score that determines the model performance. To isolate the effect of a specific baseline $b_i^{\mathcal{M}}$ for modality $\mathcal{M}$ on model performance, we take the average over all combinations containing that baseline, effectively marginalizing out the influence of other modalities' baselines:
\begin{equation}
	\bar{f}(b_i^{\mathcal{M}}, \alpha) = \frac{1}{|\mathcal{C}_i^{\mathcal{M}}|} \sum_{\mathbf{c} \in \mathcal{C}_i^{\mathcal{M}}} \bar{f}(\mathbf{c}, \alpha)
\end{equation}
where $\mathcal{C}_i^{\mathcal{M}} = \{\mathbf{c} \in \mathcal{C} : \mathbf{c}_{\mathcal{M}} = b_i^{\mathcal{M}}\}$. Finally, we compute the model sensitivity to baseline $b_i^{\mathcal{M}}$ as  $\bar{f}(b_i^{\mathcal{M}}, \alpha) - \bar{f}_0$, where $\bar{f}_0$ denotes the unperturbed model performance.

Fig.~\ref{fig:baseline_ablation} shows the resulting modality-wise sensitivity of \layoutlm{} to different baseline candidates: \textsc{[PAD]}, \textsc{[MASK]}, \textsc{Zero Embedding}, \textsc{Mean} image, and \textsc{Random} image. As shown, sensitivity varies across modalities and strategies, with differences becoming more pronounced at higher perturbation levels.
To select the baselines for downstream explainability evaluation, for each modality, we compute the average sensitivity of each strategy across perturbation levels and choose the strategy that results in the lowest overall model sensitivity. Intuitively, since all baselines inherently remove information, greater model sensitivity is more likely attributable to out-of-distribution inputs than to genuine information removal.

\subsection{Attribution Targets}
\label{sec:attribution_targets}
Our framework is designed to work across three types of document understanding tasks: (1) classification (\cls{}), where the model outputs a distribution $p(y \mid x)$ over $C$ document labels; (2) semantic entity recognition (\ser{}), where the model outputs distributions $p(y_i \mid x)$ over $C$ entity types for each token; and (3) question answering (\qa{}), where given a question $q$, the model independently outputs distributions $p(s \mid x,q)$ and $p(e \mid x,q)$ over start and end answer span positions. For each task, we define an attribution target with respect to which the explanations are computed. For document classification, the target is defined as the probability of the predicted class $\arg\max p(y \mid x)$. For SER, a target is defined for each word as the probability of the predicted entity label $\arg\max p(y_i \mid x)$ of its first subword token. This results in $K$ attribution targets per sample, where $K$ is the total number of words in the sample. For \qa{}, the targets are defined as the probabilities of the predicted start and end span positions $\arg\max p(s \mid x,q)$ and $\arg\max p(e \mid x,q)$, respectively.

\subsection{XAI Method Adaptations}
\label{sec:xai_method_adaptations}

\begin{algorithm}[t]
	\caption{Batched Multi-Target Backprop}
	\label{alg:batched_gradients}
	\begin{algorithmic}[1]
		\Require Forward function $f$, input $x$, targets $\{c_1, \ldots, c_K\}$, batch size $b$
		\Ensure Gradients $\{\nabla_{x} f_{c_k}(x)\}_{k=1}^K$
		\State $\mathbf{y} \gets f(x)$ \Comment{Single forward pass}
		\State $\mathbf{E} \gets \texttt{eye}(K)$ \Comment{Unit vectors for each target}
		\State $\mathbf{G} \gets \texttt{zeros}(K, |x|)$ \Comment{Initialize gradient storage}
		\For{$j = 0$ \textbf{to} $K-1$ \textbf{step} $b$}
		\State $\mathbf{V} \gets \mathbf{E}[j : j+b]$ \Comment{Batch of unit vectors}
		\State $\mathbf{G}[j : j+b] \gets \texttt{autograd.grad}(\mathbf{y}, x, \mathbf{V})$ \Comment{Accumulate gradients with graph retention}
		\EndFor
		\State \Return $\mathbf{G}$
	\end{algorithmic}
\end{algorithm}

This section describes how we adapt the existing XAI methods in our framework to work with multimodal document inputs and multi-target tasks.

\subheading{Backpropagation-based methods.}
In single-target setups such as classification, applying backpropagation-based methods is straightforward. Simple backpropagation-based methods such as \saliency{}, \guidedbackprop{}, and \inputxgradient{} compute the backward pass with respect to each input modality, while methods that require reference baselines, such as \integratedgradients{} and \deeplift{}, use the predetermined modality-wise baselines to compute attribution scores.

In multi-target tasks such as SER, each document contains $K$ attribution targets (e.g., corresponding to $K$ predicted word entities), requiring $K$ attributions $\{\mathbf{A}_{\mathcal{M}}^{(c_1)}, \ldots, \mathbf{A}_{\mathcal{M}}^{(c_K)}\}$ to be computed independently for each target. A naive approach would require $K$ separate forward-backward passes per sample. To avoid this, after a single forward pass, we compute the Jacobian $\mathbf{J}_{\mathcal{M}}$ over all $K$ targets simultaneously. For instance, \saliency{} scores over $K$ targets of a single sample can be computed as follows:
\begin{equation}
\mathbf{J}_{\mathcal{M}} = \begin{bmatrix} \nabla_{\mathcal{M}} f_{c_1}(x)^T \\ \vdots \\ \nabla_{\mathcal{M}} f_{c_K}(x)^T \end{bmatrix} \in \mathbb{R}^{K \times |\mathcal{M}|}
\end{equation}
where each row $k$ contains the attribution for target $c_k$ with respect to modality $\mathcal{M}$.
In practice, we compute the above Jacobian with $K/b$ batched backward passes using PyTorch \texttt{autograd} (see Algorithm~\ref{alg:batched_gradients}), where $b$ is the batch size for the gradient computation. We integrate all backpropagation-based methods with this modified backward pass to accommodate multiple targets in each computation run.

\subheading{Perturbation-based methods.}
For single-target scenarios, applying perturbation-based methods is also straightforward with predetermined modality-wise feature segmentation masks and reference baselines. For instance, \featureablation{} directly perturbs grouped features by replacing them with their corresponding reference baselines. \occlusion{} works similarly, but instead of using a perturbation mask, uses a sliding window and strides over each modality to perturb the data. \lime{} and \kernelshap{} generate perturbed neighborhood samples around the input using the modality-wise feature masks and reference baselines, and then fit local surrogate models to approximate feature contributions.

For multi-target tasks, since each perturbed forward pass already produces predictions for all $K$ targets, we extend these methods to simultaneously accumulate the attribution scores over all $K$ targets at each perturbation step, resulting in $K$ attribution maps in a single computation run. For surrogate-based methods, such as \lime{} and \kernelshap{}, we create the neighborhood dataset once for all targets and then train $K$ surrogate models independently.

\subheading{Attention-based methods.}
Attention-based methods cannot directly produce attribution maps over the embedding space of all modalities, as the accessible modalities depend on the model architecture; for instance, \layoutlm{} produces attention scores over text and visual tokens, whereas \lilt{} produces attention scores over both text and layout tokens. We therefore restrict attention-based attributions to the modalities that are explicitly represented as separate tokens in each model. Moreover, since attention methods only provide token-level attribution scores, we distribute each score uniformly across the embedding dimension $d$ so that the resulting output lies in the same space as the outputs of other XAI methods. This ensures a consistent downstream evaluation across all methods.

\medskip \noindent For all single-target scenarios, we use the method implementations available in \texttt{captum}\footnote{\url{https://captum.ai/}} and extend them with multi-target support in the \texttt{torchxai}\footnote{\url{https://atriaml.github.io/torchxai/}} library.

\subsection{XAI Metric Adaptations}
\begin{table*}[t]
    \centering

    \caption{Overview of the 17 evaluation metrics adapted in this work. \textit{Perturbation Type} indicates whether the metric requires ordered (features removed in attribution-ranked order), unordered (random subset removal), or no perturbation of the input. FS refers to whether a metric supports computation over modality-wise segmented features. \textsc{MT} indicates whether the metric supports efficient multi-target computation. Chunking indicates whether a metric supports perturbation over feature chunks to improve computational efficiency.  \label{tab:metrics}}
   
    \rowcolors{2}{gray!25}{white}
    \resizebox{\textwidth}{!}{ %
        \begin{tabular}{lrrrrrrrr}
            Type                                          & Metric                                    & Perturbation Type & Requires Model & Requires Baseline & FS support & MT Support & Chunking \\\midrule
            \cellcolor{white}                             & \completenessf{} ($\downarrow$)           & N/A               & \cmark         & \cmark            & N/A        & \cmark     & \xmark   \\
            \cellcolor{white}\multirow{-2}{*}{Axiomatic}  & \nonsensitivityf{} ($\downarrow$)         & Unordered         & \cmark         & \cmark            & \cmark     & \cmark     & \cmark   \\\midrule
            \cellcolor{white}                             & \aopcf{} ($\uparrow$)                     & Ordered           & \cmark         & \cmark            & \cmark     & \xmark     & \cmark   \\
            \cellcolor{white}                             & \abpcf{} ($\uparrow$)                     & Ordered           & \cmark         & \cmark            & \cmark     & \xmark     & \cmark   \\
            \cellcolor{white}                             & \sufficiencyf{} ($\downarrow$)            & Ordered           & \cmark         & \cmark            & \cmark     & \xmark     & \cmark   \\

            \cellcolor{white}                             & \sensitivitynf{} ($\downarrow$)           & Unordered         & \cmark         & \cmark            & \cmark     & \cmark     & \cmark   \\

            \cellcolor{white}                             & \infidelityf{} ($\downarrow$)             & Unordered         & \cmark         & \xmark            & N/A        & \cmark     & N/A      \\

            \cellcolor{white}                             & \faithfulnessestf{} ($\uparrow$)          & Ordered           & \cmark         & \cmark            & \cmark     & \xmark     & \cmark   \\

            \cellcolor{white}                             & \faithfulnesscorrf{} ($\uparrow$)         & Unordered         & \cmark         & \cmark            & \cmark     & \cmark     & \cmark   \\
            \cellcolor{white}                             & \monotonicityf{} ($\uparrow$)             & Ordered           & \cmark         & \cmark            & \cmark     & \xmark     & \cmark   \\
            \multirow{-9}{*}{Faithfulness}                & \monotonicitycorrf{} ($\uparrow$)         & Unordered         & \cmark         & \cmark            & \cmark     & \cmark     & \cmark   \\\midrule
            \cellcolor{white}                             &
            \effectivecomplexityf{} ($\downarrow$)        & Ordered                                   & \cmark            & \cmark         & \cmark            & \xmark     & \cmark                \\
            \cellcolor{white}                             & \entropycomplexityf{} ($\downarrow$)      & N/A               & \xmark         & \xmark            & \cmark     & N/A        & N/A      \\
            \cellcolor{white}                             & \sundararajancomplexityf{} ($\downarrow$) & N/A               & \xmark         & \xmark            & \cmark     & N/A        & N/A      \\
            \multirow{-4}{*}{Complexity}                  & \sparsenessf{} ($\uparrow$)               & N/A               & \xmark         & \xmark            & \cmark     & N/A        & N/A      \\\midrule
            \cellcolor{white}                             & \maxsensitivityf{} ($\downarrow$)         & Unordered         & \cmark         & \xmark            & N/A        & \cmark     & N/A      \\
            \cellcolor{white}\multirow{-2}{*}{Robustness} & \avgsensitivityf{} ($\downarrow$)         & Unordered         & \cmark         & \xmark            & N/A        & \cmark     & N/A      \\
                    \bottomrule
        \end{tabular}
    }
\end{table*}

\label{sec:explainability-evaluation}
This section describes how we adapt the existing XAI evaluation metrics to multimodal and multi-target settings.

\subheading{Multimodal adaptation.}
Since each XAI method produces separate attribution maps per modality $\mathbf{A}_{\mathcal{M}}, \mathcal{M} \in \{\mathcal{T}, \mathcal{S}, \mathcal{L}, \mathcal{V}\}$, we first perform modality-wise feature aggregation as described in Section~\ref{sec:feature_segmentation}, so that each feature group carries a single score.
All metrics are then computed over these aggregated scores, treating each feature group independently of its originating modality.
For perturbation-based metrics, the perturbation order is determined by ranking all feature groups jointly across modalities, and the corresponding features are then perturbed using their modality-specific segmentation masks and reference baselines defined in previous sections.
There are two exceptions to this general pipeline: \completenessf{} is computed on the raw attribution scores rather than group-aggregated scores, as the axiom requires the exact sum over all individual attributions, and \infidelityf{} adds noise directly to each modality input independently, rather than operating on the aggregated features, to keep the implementation close to the original formulation.

\begin{table*}[!t]\centering\small
    \centering%
	\caption{Overview of the eight datasets used in this work. For each dataset, we provide the number of classes or entity types, dataset splits, the size of the subset used for XAI evaluation, and the task-specific performance metric.\label{tab:datasets}}
    \rowcolors{2}{gray!15}{white}

    \begin{tabular}{llrrrrrl}
        \textbf{Task}                           & \textbf{Dataset} & \textbf{Classes} & \textbf{Train} & \textbf{Validation} & \textbf{Test} & \textbf{XAI Eval. Set}& \textbf{Metric} \\\midrule

        \cellcolor{white}                       & \tobacco{}       & 10                 & 2504           & 278                 & 700      & 700     & Accuracy        \\
        \cellcolor{white}                       & \rvlcdip{}       & 16                 & 320000         & 40000               & 40000    & 1000     & Accuracy        \\
        \multirow{-3}{*}{\cls}                  & \doclaynet{}     & 11                 & 69103          & 6480                & 4999    & 1000      & Accuracy        \\\midrule

        \cellcolor{white}                       & \funsd{}         & 4                  & 149            & —                   & 50       & 50     & F1              \\
        \cellcolor{white}                       & \cord{}          & 30                 & 800            & 100                 & 100  & 100         & F1              \\
        \cellcolor{white}                       & \sroie{}         & 5                  & 626            & —                   & 347       & 100    & F1              \\
        \cellcolor{white}\multirow{-4}{*}{\ser} & \wildreceipts{}  & 13                 & 1267           & —                   & 472         & 100  & F1              \\\midrule
        \cellcolor{white}\qa                    & \docvqa{}        & —                  & 10194          & 1286                & 1287       & 1000   & ANLS            \\
        \bottomrule
    \end{tabular}
\end{table*}

\subheading{Multi-target adaptation.}
As with the explanation methods, adapting the metrics for multi-target settings is necessary to reduce computational costs. Metrics that use unordered or random perturbations (such as \nonsensitivity{}, \monotonicitycorr{}, \sensitivityn{}, \infidelity{}, and \faithfulnesscorr{}) can be efficiently adapted, as the effect of each perturbation can be accumulated over all $K$ targets in a single forward pass. We implement multi-target support for all of these metrics.
However, metrics that require ordered perturbations (such as \aopc{}, \sufficiency{}, \monotonicity{}, \faithfulnessest{}, and \effectivecomplexity{}) cannot be adapted for multi-target computation, since the perturbation order depends on each target's unique attribution scores. Metrics that require recomputing explanations (such as \maxsensitivity{} and \avgsensitivity{}) are adapted by using the underlying multi-target explanation mode. Metrics that do not require model forward passes (\entropycomplexity{}, \sundararajancomplexity{}, and \sparseness{}) are computed independently for each target's attribution map.

\medskip \noindent
All metrics are implemented in \texttt{PyTorch} with GPU support, batched computation, and compatibility with \texttt{captum}. We reuse the implementations of \maxsensitivityf{} and \infidelityf{} from \texttt{captum} and extend them for the multi-target case. The remaining metrics, along with their multi-target adaptations, are implemented in the \texttt{torchxai} library. Table~\ref{tab:metrics} summarizes the supported metrics; details on individual metrics and their adaptations are provided in Appendix~\ref{sec:appendix_metrics}.

\section{Experimental Setup}
\label{sec:experimental_setup}
\subheading{Datasets.}
We evaluate our framework on 8 benchmark datasets across three tasks: \cls{}, \ser{}, and \qa{}. For classification, we use the \rvlcdip{}, \tobacco{}, and \doclaynet{} datasets. For \ser{}, we use the \funsd{}, \cord{}, \sroie{}, and \wildreceipts{} datasets. For \qa{}, we use the \docvqa{} dataset. Table~\ref{tab:datasets} presents an overview of the datasets.

\subheading{Models.}
We select four models representing different modality fusion strategies: \bert{} and \roberta{} as text-only baselines ($\mathcal{T}$, $\mathcal{S}$), \lilt{} as a multimodal model processing text, positional, and layout modalities ($\mathcal{T}$, $\mathcal{S}$, $\mathcal{L}$), and \layoutlm{} as a fully multimodal model processing all four modalities ($\mathcal{T}$, $\mathcal{S}$, $\mathcal{L}$, $\mathcal{V}$). This selection enables analysis of how explainability methods behave across models with varying numbers of modalities and fusion strategies.

\begin{table*}[!t]\centering\small
\caption{
	Model performance on individual datasets. Best results are highlighted in \textbf{bold}. Training hyperparameters are reported in Appendix~\ref{tab:hyperparams}.
	\label{tab:model_performance}
}
    \rowcolors{2}{gray!15}{white}
    \resizebox{\textwidth}{!}{
        \begin{tabular}{lrrrrrrr}
            \rowcolor{gray!15}
            Task                                                 & Dataset         & Metric            & \bert{} & \roberta{} & \lilt{} & \layoutlm{}    \\\midrule
            \cellcolor{white}                                    & \tobacco{}      & Acc. $(\uparrow)$ & 86.38   & 87.81      & 88.53   & \textbf{94.27} \\
            \cellcolor{white}                                    & \rvlcdip{}      & Acc. $(\uparrow)$ & 89.70   & 90.05      & 91.20   & \textbf{94.12} \\
            \multirow{-3}{*}{Classification}   \cellcolor{white} & \doclaynet{}    & Acc. $(\uparrow)$ & 97.50   & 97.90      & 97.78   & \textbf{98.97} \\\midrule
            \cellcolor{white}                                    & \funsd{}        & F1 $(\uparrow)$   & 64.69   & 67.47      & 76.19   & \textbf{88.09} \\
            \cellcolor{white}                                    & \cord{}         & F1 $(\uparrow)$   & 95.10   & 96.28      & 96.85   & \textbf{97.01} \\
            \cellcolor{white}                                    & \sroie{}        & F1 $(\uparrow)$   & 89.63   & 93.33      & 94.12   & \textbf{95.28} \\
            \multirow{-4}{*}{SER}\cellcolor{white}               & \wildreceipts{} & F1 $(\uparrow)$   & 89.70   & 91.32      & 92.80   & \textbf{93.38} \\\midrule
            \cellcolor{white} QA                                 & \docvqa{}       & ANLS $(\uparrow)$ & 59.95   & 63.26      & 70.11   & \textbf{73.37} \\
            \bottomrule
        \end{tabular}
    }
\end{table*}

\subheading{Training details.}
We initialize all models from publicly available pretrained checkpoints on Hugging Face\footnote{\url{https://huggingface.co}} and fine-tune them using the \adam{} or \adamw{} optimizers.
We use 40 training epochs for the classification task, 50 for the DocVQA task, and 100 for the SER task, with early stopping based on validation performance (patience of 10).
Table~\ref{tab:model_performance} reports the performance achieved by each model on the individual datasets. Additional training hyperparameter details are provided in Appendix~\ref{sec:appendix_hyperparams}.

\subheading{XAI Methods.}
We adapt and investigate 13 XAI methods in this work, spanning three families: seven backpropagation-based methods, namely \saliency{}, \inputxgradient{}, \guidedbackprop{}, \integratedgradients{}, \gradientshap{}, \deeplift{}, and \deepshap{}; four perturbation-based methods, namely \featureablation{}, \occlusion{}, \lime{}, and \kernelshap{}; and two attention-based methods, namely \rawattn{} and \attnrollout{}. Within attention-based methods, we also considered \attnflow{} but excluded it due to its high computational cost when applied to large sequences.

We configure each explainability method to achieve a suitable trade-off between quality and computational efficiency. For methods that require sampling or interpolation steps, such as \integratedgradients{}, \gradientshap{}, \lime{}, and \kernelshap{}, we use 200 samples or steps. For \deepshap{}, a reference baseline distribution of $N=100$ samples is precomputed from the training set of each dataset.
For \occlusion{}, we define modality-specific sliding windows and strides: a window of 8 tokens with a stride of 4 for text, position, and layout modalities, and a window of $3 \times 16 \times 16$ with a stride of $3 \times 8 \times 8$ for the image modality.

\subheading{XAI Evaluation Details.}
\label{sec:explainability-evaluation-metrics}
We evaluate the explanations using the 17 metrics listed in Table~\ref{tab:metrics}, spanning four criteria: axiomatic soundness, faithfulness, complexity, and robustness. For \sensitivitynf{} and \faithfulnesscorrf{}, we evaluate at perturbation levels of 20\%, 40\%, and 60\%. For perturbation-based metrics, we use a chunk size of 1\% where applicable, meaning 1\% of feature groups are removed at each perturbation step. Moreover, all metrics use the same modality-wise feature segmentation masks and baselines as those used for computing the XAI attributions.
All experiments are conducted on subsets of the test sets as listed in Table~\ref{tab:datasets} (XAI Eval.\ Set), using NVIDIA H200 and RTX A6000 GPUs. Full hyperparameter configurations for explainability methods and evaluation metrics are provided in Appendix~\ref{sec:appendix_hyperparams}.

\section{Experimental Results}
\label{sec:results}
\subsection{Rank Analysis of XAI Methods}
\label{sec:results_ranking}
\begin{figure}[t]
    \centering
    \includegraphics[width=\columnwidth,draft=false]{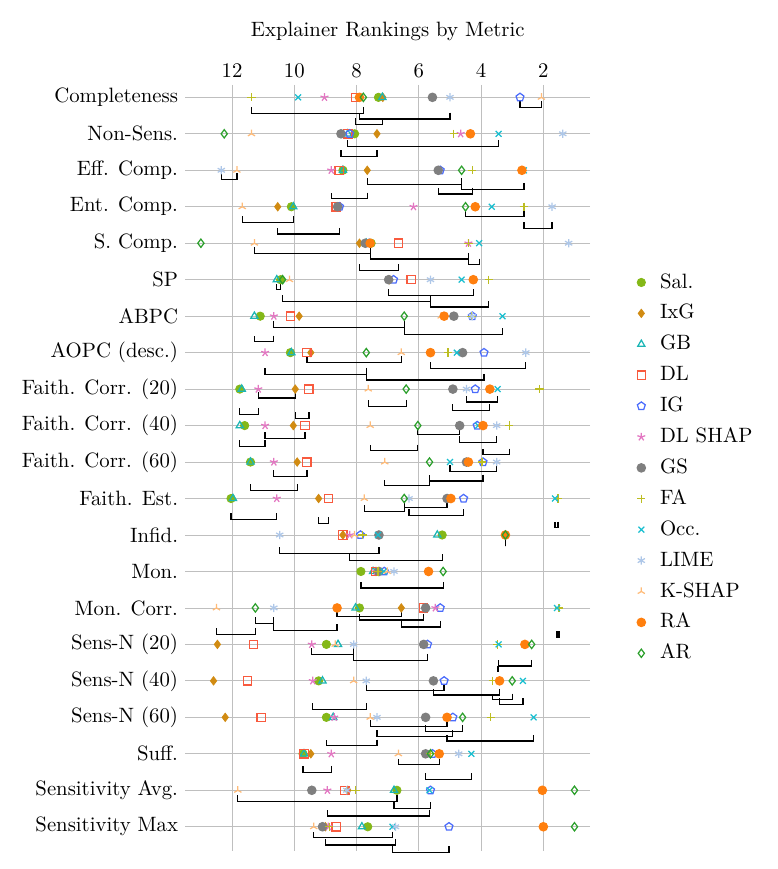}
    \caption{Critical difference diagram showing per-metric XAI method rankings across all tasks. Methods connected by a horizontal bar are statistically indistinguishable. Lower ranks indicate better performance. Task-level critical difference diagrams are additionally provided in Appendix~\ref{sec:appendix_ranking}.}
    \label{fig:cd_overall}
\end{figure}

\begin{figure}[t]
    \centering
    \includegraphics[width=\columnwidth,draft=false]{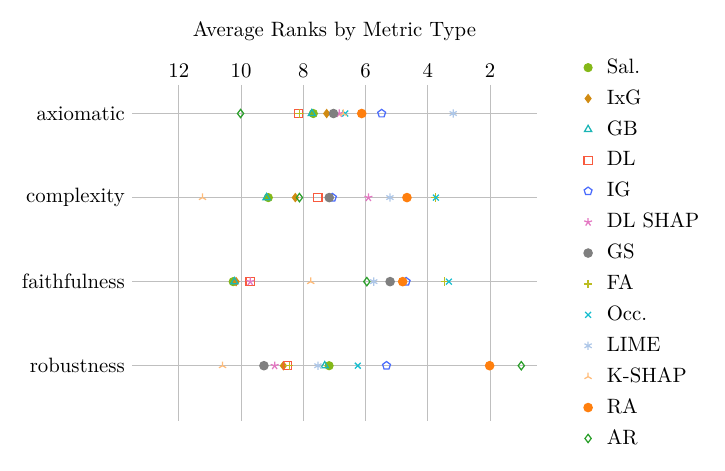}
    \caption{Aggregated mean ranks of XAI methods per evaluation criterion across all tasks, datasets, and models. Lower ranks indicate better performance. Per-task breakdowns are provided in Appendix~\ref{sec:appendix_ranking}.}
    \label{fig:avg_rank_overall}
\end{figure}

To determine which XAI methods perform best under each evaluation criterion, we conduct a rank-based analysis. For each metric, we first summarize the results per dataset via mean aggregation over the sample-level scores; for multi-target scenarios, each sample-target pair is treated as an individual data point in this computation.
We then rank all XAI methods per dataset-model setting and average the ranks across all settings. We adopt a rank-based approach because metric values can vary in scale across different setups, and therefore, direct aggregation may disproportionately favor certain setups.
However, while ranking solves the scaling issue, the rank difference between two methods may not be statistically significant. To address this, we first apply the Friedman test over aggregated metric values to check the null hypothesis that all methods perform equally. If the null hypothesis is rejected, the Holm-adjusted pairwise Wilcoxon signed-rank test is then applied (as suggested in \citep{JMLR:v17:benavoli16a}) to determine pairwise indistinguishable groups. For all statistical analyses, we assume an $\alpha$ of 0.05.
Both the mean ranks and statistical significance results are jointly visualized using critical difference (CD) diagrams~\citep{JMLR:v7:demsar06a}, where methods with statistically insignificant differences are connected by a horizontal bar.
Besides metric-level ranking, we also perform mean aggregation across metrics within each category to obtain a global method ranking per criterion.
Fig.~\ref{fig:cd_overall} shows the critical difference diagram for the XAI method rankings across all tasks, and Fig.~\ref{fig:avg_rank_overall} shows the mean ranks per evaluation criterion. Per-task results are additionally provided in Appendix~\ref{sec:appendix_ranking}.
For the \aopc{} ranking, we use the difference between \aopc{}\textsubscript{MoRF} and \aopc{}\textsubscript{RAND} at $100\%$ features removed as a single summary score. For completeness, \aopc{} curves over multiple feature removal steps are additionally provided in Appendix~\ref{sec:appendix_aopc}.

\subheading{Method Rankings and Statistical Significance.} From Fig.~\ref{fig:cd_overall}, we observe that although methods vary considerably in their raw rankings across the metrics, many of these differences are not statistically significant (as highlighted by the horizontal bar). For instance, on \completeness{}, \kernelshapf{} is ranked higher than \integratedgradientsf{}, but both are statistically indistinguishable. Likewise, on \nonsensitivityf{} and \sparsenessf{}, the vast majority of methods show no statistically significant differences despite a wide separation in their raw ranks. This effect is most extreme on \monotonicityf{} and \infidelityf{}, where all or nearly all methods form a single indistinguishable group. Upon investigation, we found that \monotonicityf{}, being a binary metric (0 or 1), yielded a value of $0$ the majority of the time, as even minor noise in an otherwise monotonic perturbation curve is sufficient to trigger non-monotonicity. Similarly, \infidelityf{} produced near-identical raw scores across all methods, with the exception of \rawattnf{} and \attnrolloutf{}, for which the scores diverged from the rest but remained comparable to each other. Based on these observations, we consider both metrics unsuitable for evaluation in this setting. Overall, we observe similar patterns in task-level method rankings, with slightly more pronounced statistical indistinguishability on some metrics, as detailed in Appendix~\ref{sec:appendix_ranking}.

\subheading{Ranking Patterns Across Metrics.}
When examining rankings across individual metrics, some important observations arise. On both \faithfulnesscorr{} and \sensitivityn{}, methods achieve very similar rankings across the 20\%, 40\%, and 60\% perturbation levels, indicating that a single configuration may suffice for these metrics.
We also observe comparable orderings across some groups of faithfulness metrics; for instance, on \aopc{}, \abpc{}, \faithfulnessest{}, \faithfulnesscorr{}, and \sufficiency{}, methods produce similar ranking patterns. This observation also holds for the robustness metrics \maxsensitivity{} and \avgsensitivity{}.

\subheading{Best-Performing Methods per Criterion.}
Despite the widespread statistical indistinguishability, certain method groups consistently rank higher or lower across metrics. Basic backpropagation-based methods, including \saliencyf{}, \guidedbackpropf{}, \inputxgradientf{}, and \deepliftf{}, occupy the lowest ranks across the majority of metrics. On \completeness{}, \integratedgradientsf{} and \kernelshapf{} perform the best, whereas \limef{} and \occlusionf{} lead on \nonsensitivity{}. These observations are also reflected in the mean ranking across axiomatic properties (Fig.~\ref{fig:avg_rank_overall}), where \limef{}, \integratedgradientsf{}, \rawattnf{}, and \occlusionf{} lead overall.
On individual complexity metrics, \kernelshapf{} and \attnrolloutf{} perform the worst, whereas \limef{}, \rawattnf{}, \occlusionf{}, and \featureablationf{} consistently rank higher than other methods. The average ranking on the complexity criterion (see Fig.~\ref{fig:avg_rank_overall}) confirms these findings, with \limef{}, \rawattnf{}, \occlusionf{}, and \featureablationf{} ranking the highest, while \kernelshapf{} ranks considerably lower than all other methods.
On faithfulness metrics, \occlusionf{}, \featureablationf{}, and \rawattnf{} show the best rankings overall. For instance, on \faithfulnessest{} and \monotonicitycorr{}, \occlusionf{} and \featureablationf{} considerably outperform other methods. Attention-based methods, \rawattnf{} and \attnrolloutf{}, show variable results on faithfulness; however, \rawattnf{} generally ranks as one of the best alongside \occlusionf{}. The average ranking on faithfulness supports these findings (see Fig.~\ref{fig:avg_rank_overall}), with \occlusionf{}, \rawattnf{}, \featureablationf{}, and \integratedgradientsf{} being the top-4 methods.
On robustness, attention-based methods considerably outperform all other method families; the average ranking on robustness confirms this, with \attnrolloutf{} ranking first, followed by \rawattnf{}, \integratedgradientsf{}, and \occlusionf{}.
Taken together, the methods that most consistently stand out as best-performing across all evaluation criteria are \occlusionf{}, \rawattnf{}, and \integratedgradientsf{}.

\begin{figure}[b]
    \centering
    \includegraphics[width=\columnwidth,draft=false]{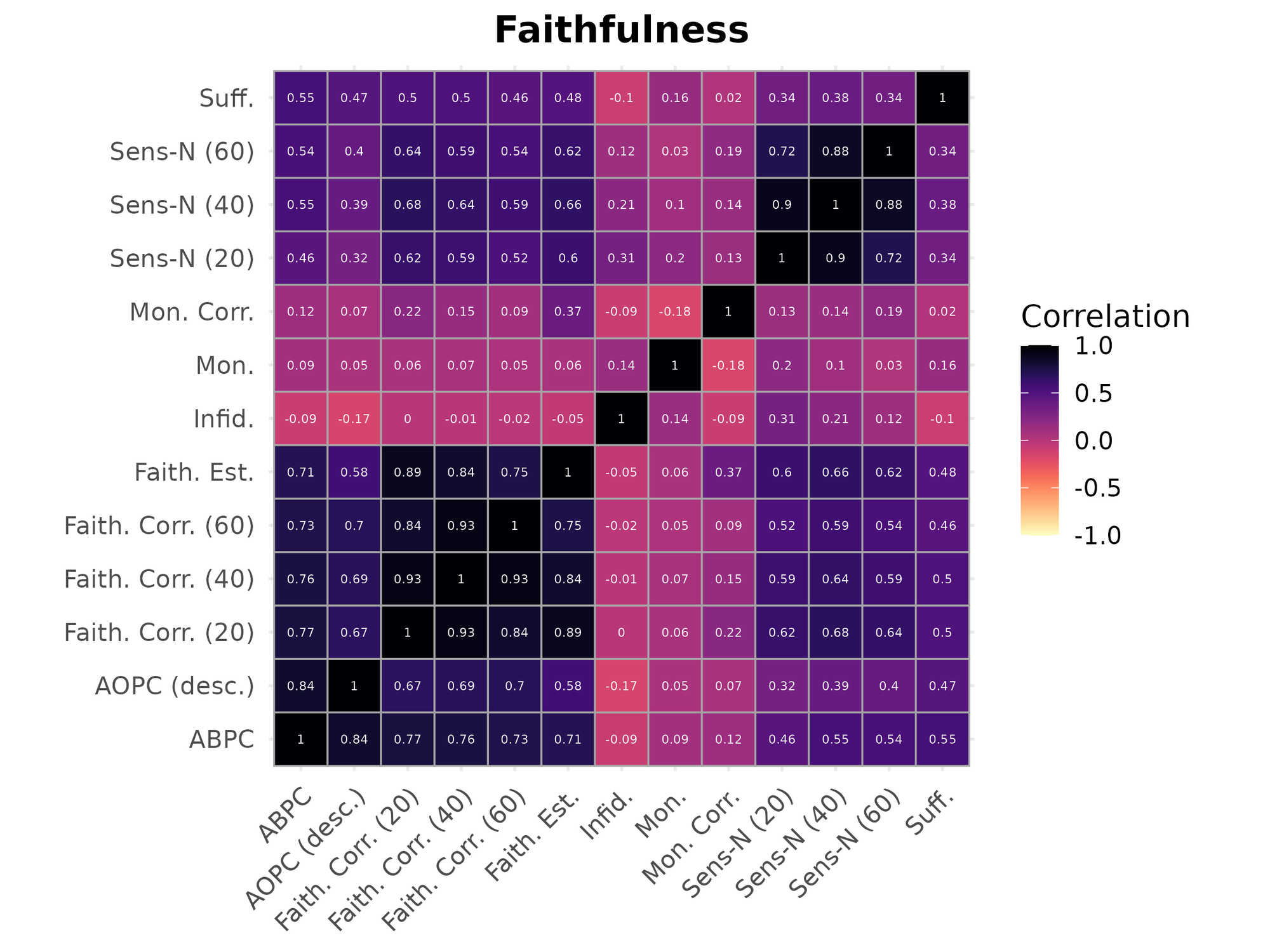}
\caption{Pairwise Spearman correlation between faithfulness metric rankings across all datasets, models, and explainer settings. Per-task correlation matrices are provided in Appendix~\ref{sec:appendix_metric_agreement}.}
    
    \label{fig:faithfulness_corr_overall}
\end{figure}

\begin{figure}[t]
    \centering
    \includegraphics[width=\columnwidth,draft=false]{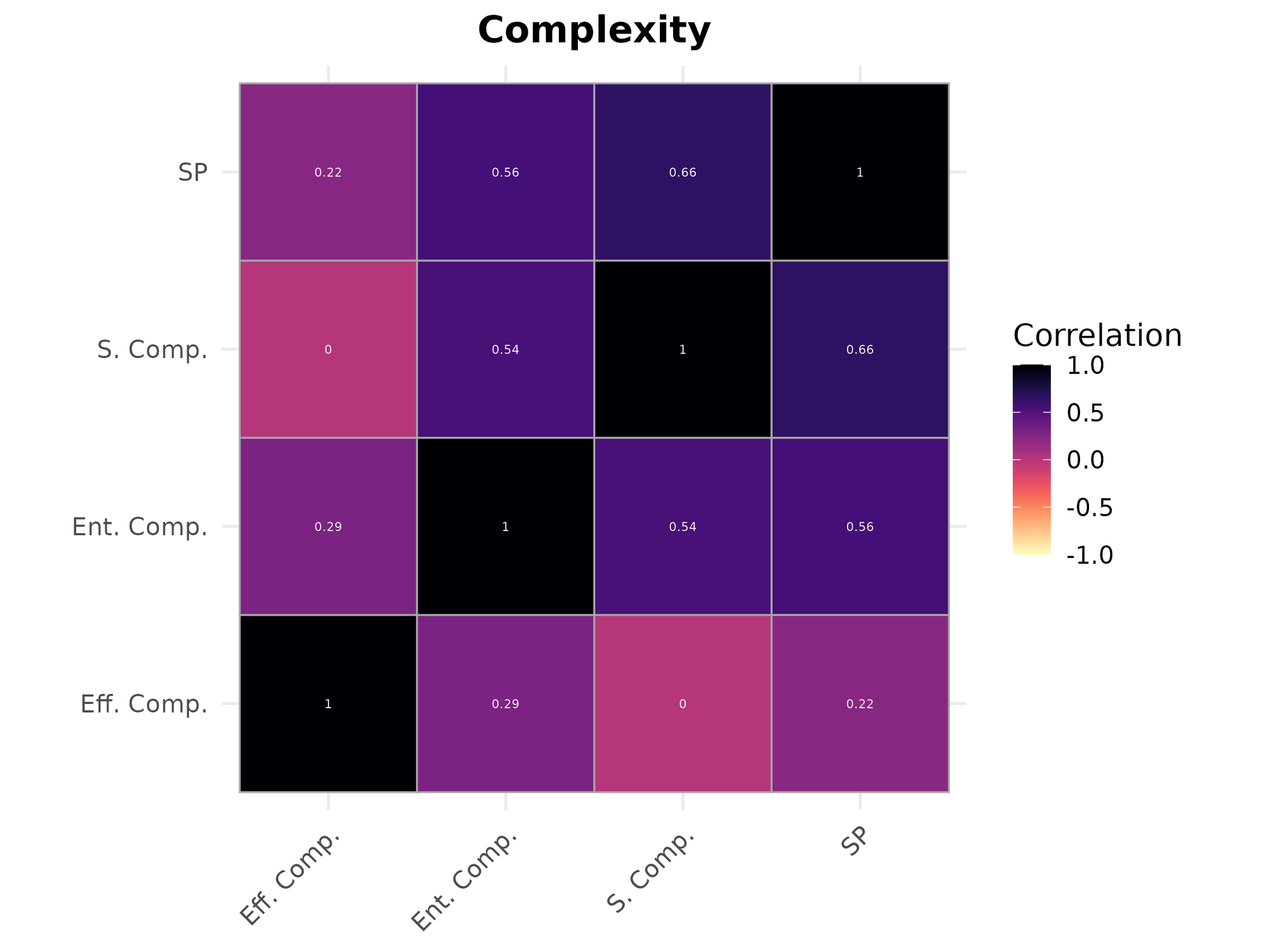}
\caption{Pairwise Spearman correlation between complexity metric rankings across all datasets, models, and explainer settings. Per-task correlation matrices are provided in Appendix~\ref{sec:appendix_metric_agreement}.}
    \label{fig:comp_corr_overall}
\end{figure}

\subsection{Metric Agreement Analysis}
\label{sec:results_agreement}
The metric disagreement problem has been well-noted in prior work~\citep{klein_navigating_nodate,tomsett_sanity_2020}, and therefore, it is important to investigate whether multiple evaluation metrics agree or disagree on the relative ranking of explanation methods. We assess metric agreement for the faithfulness and complexity categories by computing the Spearman correlation between metric rankings across all datasets, models, and explainer settings.
Such analysis can not only improve our confidence in the evaluation, but also help determine whether some metrics are redundant, e.g., when two metrics are highly correlated.
Fig.~\ref{fig:faithfulness_corr_overall} and Fig.~\ref{fig:comp_corr_overall} present the correlation heatmaps for faithfulness and complexity metrics, respectively; per-task results are additionally provided in Appendix~\ref{sec:appendix_metric_agreement}.

\subheading{Faithfulness Metrics.}
On the faithfulness criterion, both \faithfulnesscorr{} and \sensitivityn{} show strong correlation across their respective perturbation levels (20\%, 40\%, 60\%), aligning with the observations made via the critical difference diagrams.
Strong correlation is observed between \aopc{}, \faithfulnessest{}, \abpc{}, \faithfulnesscorr{}, and \sufficiency{}. \sensitivityn{} also shows a moderate correlation with other faithfulness metrics (around 0.5--0.65), whereas \monotonicitycorr{} is only weakly correlated with \faithfulnessest{}. Notably, neither \infidelity{} nor \monotonicity{} show any correlation with other faithfulness metrics. This lack of correlation coincides with the statistical indistinguishability and unreliability observed for these metrics in the critical difference diagrams.

\subheading{Complexity Metrics.}
On the complexity criterion, \sundararajancomplexity{}, \sparseness{}, and \entropycomplexity{} show a moderate correlation with each other (around 0.54--0.66).
\effectivecomplexity{}, however, shows only a very weak correlation with \entropycomplexity{} and \sparseness{}, and no correlation with \sundararajancomplexity{}.

\noindent \medskip For both criteria, the observations remained largely consistent across individual task-level results (see Appendix~\ref{sec:appendix_metric_agreement}).

\subsection{Computational Efficiency of Multi-Target Adaptations}
\label{sec:results_execution_time}
In this section, we examine the computational efficiency of our multi-target adaptations for both XAI methods and evaluation metrics.
\begin{figure*}[t]
    \centering
    \includegraphics[width=\textwidth,draft=false]{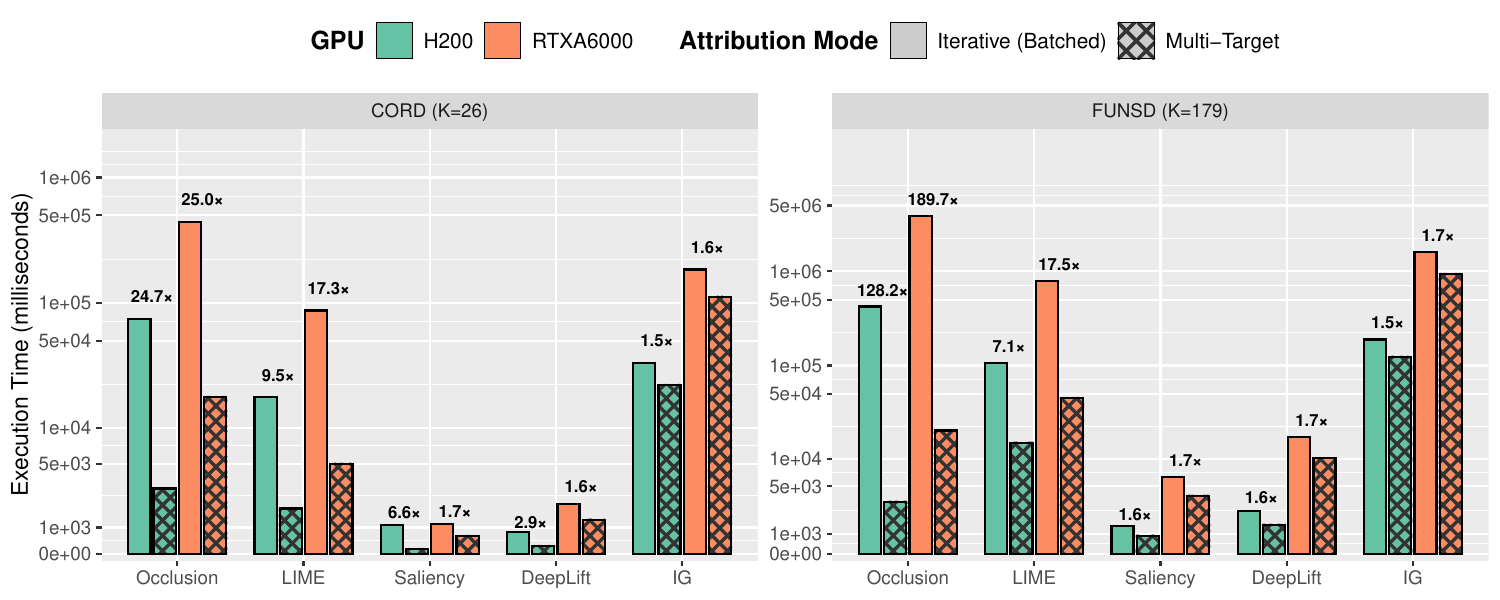}
    \caption{Execution time comparison between iterative (batched) and multi-target attribution computation on CORD ($K=26$) and FUNSD ($K=179$) datasets across two GPUs. Numbers above bars indicate the speedup factor of multi-target over iterative computation. Note the log scale on the y-axis.}
    \label{fig:mt_exec}
\end{figure*}

\subheading{XAI Methods.}
Fig.~\ref{fig:mt_exec} compares the execution time of iterative versus multi-target attribution computation for five XAI methods (\occlusionf{}, \limef{}, \saliencyf{}, \deepliftf{}, and \integratedgradientsf{}). We report the average execution time over a fixed number of samples on the CORD and FUNSD datasets across two GPUs (H200 and RTX A6000), along with the speedup factor for each method. Note that for the iterative scenario, we use batched computation to keep a fair comparison.
Perturbation-based methods such as \occlusionf{} achieve speedups roughly proportional to the number of targets per sample, reaching approximately 25$\times$ on CORD (26 targets/sample) and up to 189$\times$ on FUNSD (179 targets/sample), with minor variations between GPUs. For surrogate-based methods such as \limef{}, speedups are lower and more GPU-dependent, ranging from 7$\times$ to 17$\times$ depending on the dataset and hardware.
Backpropagation-based methods show comparatively modest but still notable gains, consistently achieving 1.5$\times$--1.7$\times$ speedups regardless of the number of targets.

\begin{figure*}[t]
    \centering
    \begin{subfigure}[t]{0.48\textwidth}
        \centering
        \includegraphics[width=\textwidth,draft=false]{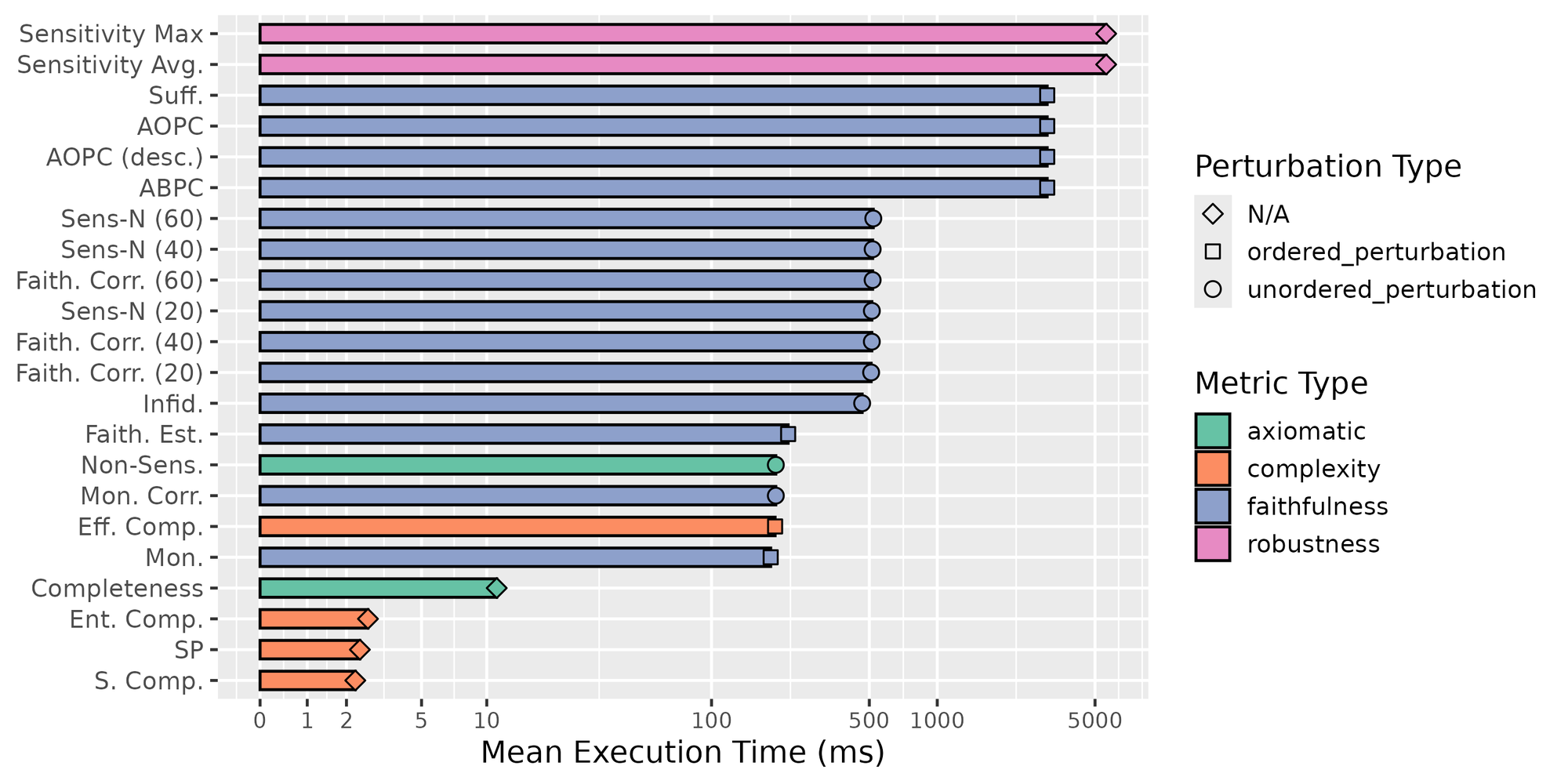}
        \caption{Classification (single-target)}
        \label{fig:metric_runtime_cls}
    \end{subfigure}
    \hfill
    \begin{subfigure}[t]{0.48\textwidth}
        \centering
        \includegraphics[width=\textwidth,draft=false]{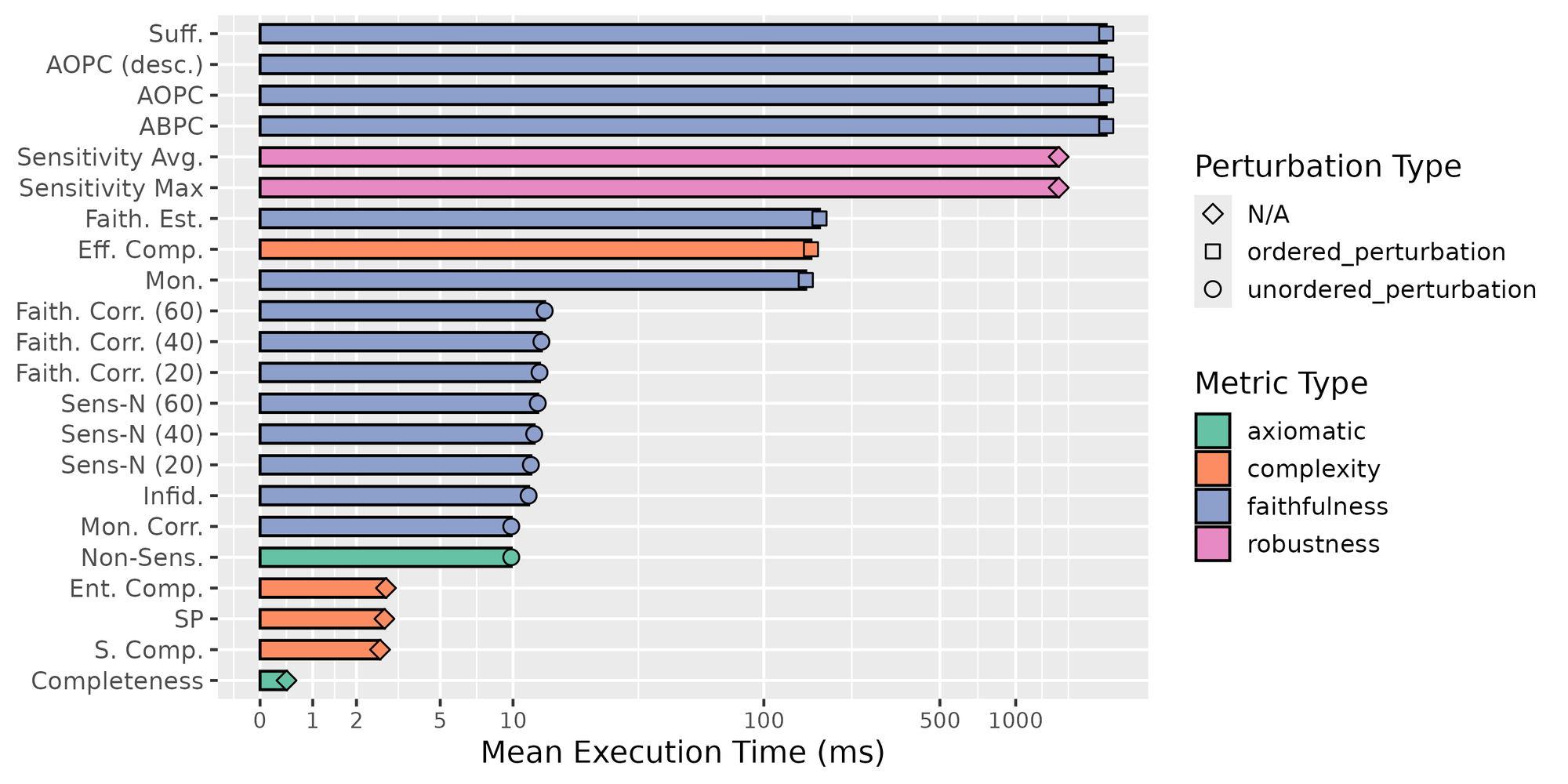}
        \caption{Semantic Entity Recognition (multi-target)}
        \label{fig:metric_runtime_ser}
    \end{subfigure}
	\caption{Mean per-sample-target computation time of each evaluation metric on \cls{} (single-target) and \ser{} (multi-target) tasks. All results are computed on H100/H200 GPUs. Unordered perturbation metrics benefit significantly from multi-target adaptations, while ordered metrics remain expensive as they cannot be adapted for multi-target computation.}
    \label{fig:metric_runtime}
\end{figure*}

\subheading{Evaluation Metrics.}
Fig.~\ref{fig:metric_runtime} compares the per-sample-target execution time of all evaluation metrics across \cls{} (single-target) and \ser{} (multi-target) settings. All results are computed on H100/H200 GPUs, with execution times averaged over all datasets and XAI methods.
Within the axiomatic metrics, \completeness{} benefits significantly from the multi-target support, with execution times reaching below $1$ms.
Complexity metrics \sundararajancomplexity{}, \sparseness{}, and \entropycomplexity{} show similar computational times on both tasks even without multi-target adaptations.
For robustness metrics, although a direct comparison is not possible as the two tasks are computed on different datasets, computation on the \ser{} task is roughly $3.8\times$ faster than on the \cls{} task, which can be attributed to the underlying use of multi-target adaptations of the explainers when computing these metrics.
The advantage of multi-target adaptations becomes most apparent when comparing metrics that use ordered perturbations (e.g., \aopc{}, \abpc{}, \sufficiency{}) against those that use unordered perturbations.
On the single-target \cls{} task, unordered metrics take approximately $193$--$521$ms while ordered metrics take approximately $3{,}069$ms, making the former roughly $6$--$16\times$ faster.
On the \ser{} task, multi-target adaptations widen this gap significantly: execution times for unordered metrics drop to approximately $10$--$13$ms per sample-target, while those for ordered metrics remain at approximately $2{,}287$ms as they cannot be adapted, yielding speedups of $176$--$229\times$.

\subsection{Qualitative Analysis}
\label{sec:results_qual}
\begin{figure*}[htbp]
\centering
\noindent
\parbox[c]{1.5em}{\centering\rotatebox{90}{\small IG (RVL-CDIP)}}
\begin{subfigure}[c]{0.8\linewidth}
\includegraphics[width=\textwidth]{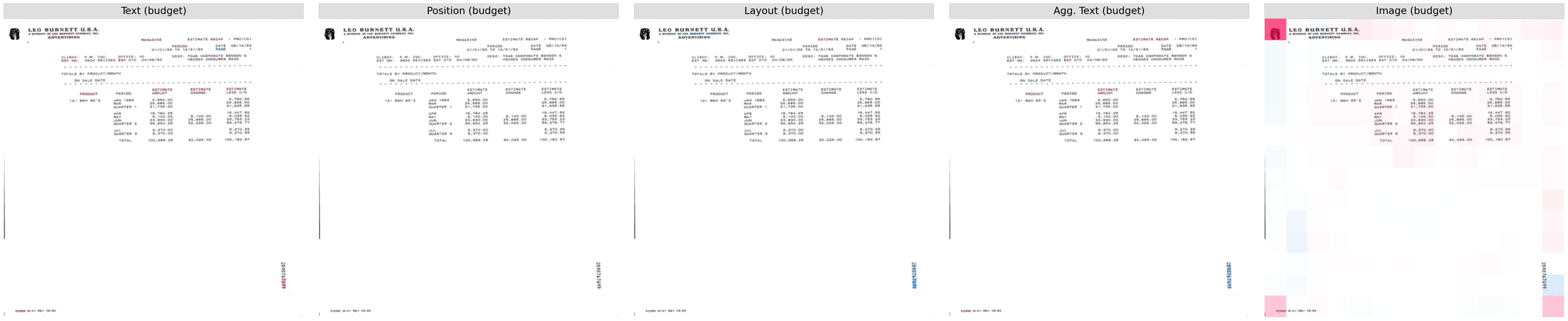}
\end{subfigure}
\par\smallskip
\noindent
\parbox[c]{1.5em}{\centering\rotatebox{90}{\small Occ. (RVL-CDIP)}}
\begin{subfigure}[c]{0.8\linewidth}
\includegraphics[width=\textwidth]{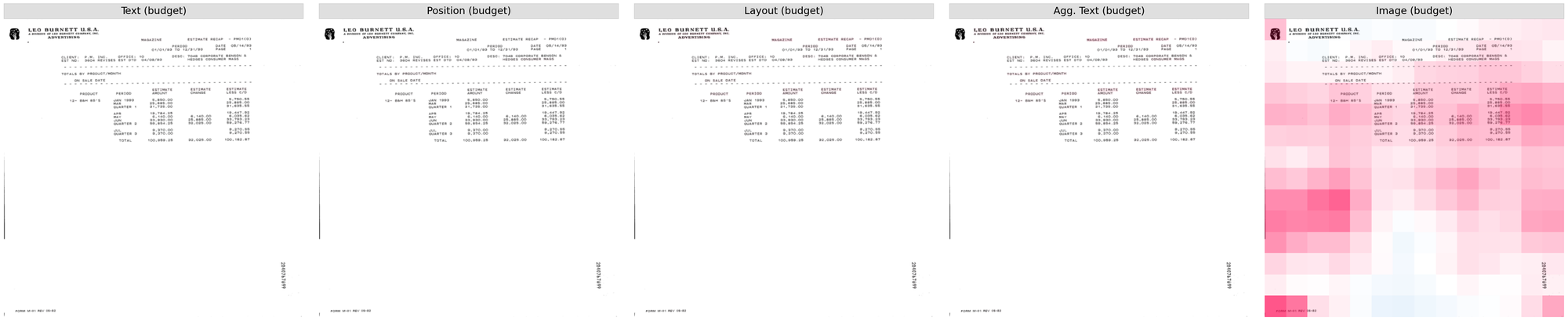}
\end{subfigure}
\par\smallskip
\noindent
\parbox[c]{1.5em}{\centering\rotatebox{90}{\small RA (RVL-CDIP)}}
\begin{subfigure}[c]{0.8\linewidth}
\includegraphics[width=0.5\textwidth]{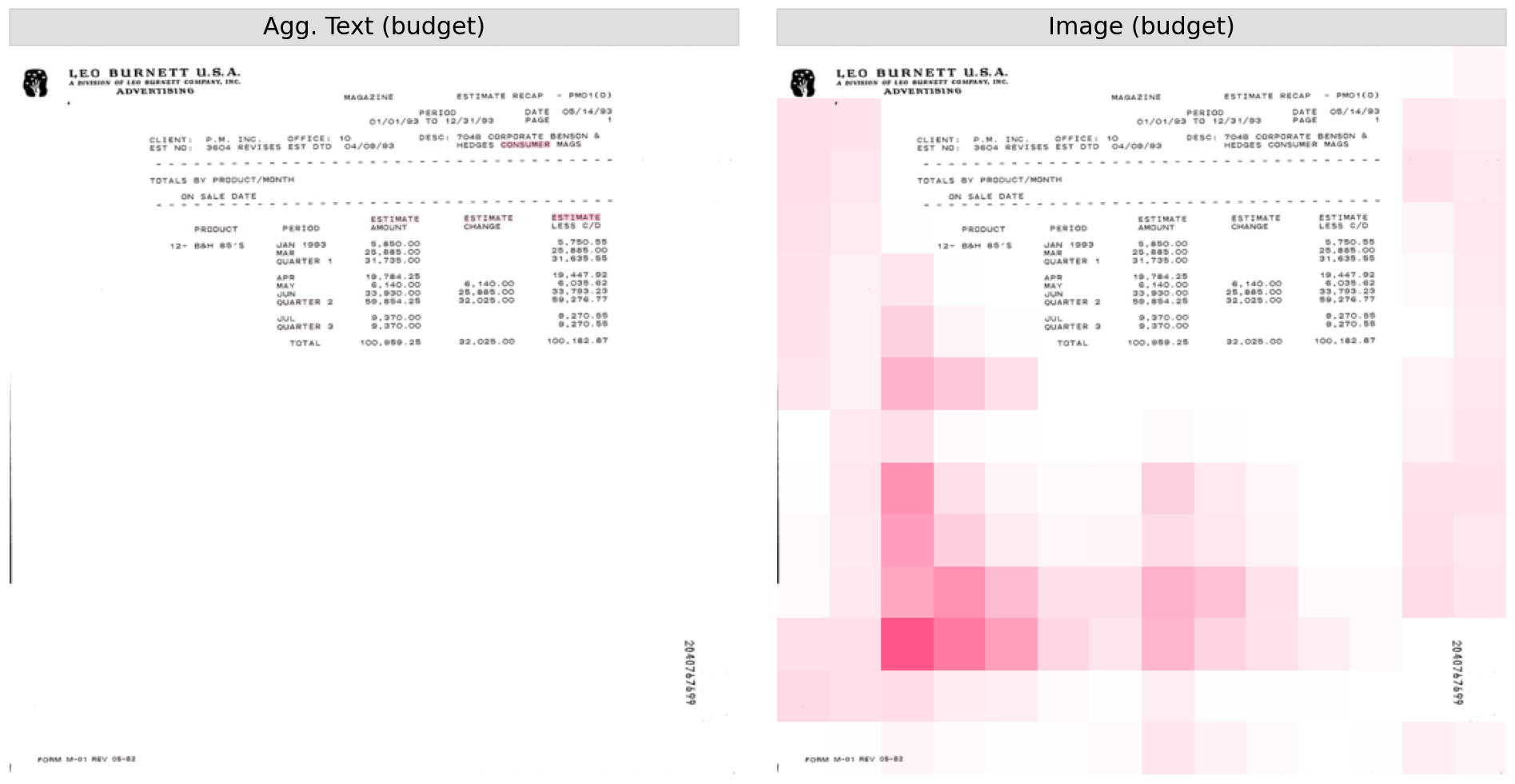}
\end{subfigure}
\par\smallskip
\noindent
\parbox[c]{1.5em}{\centering\rotatebox{90}{\small IG (DocLayNet)}}
\begin{subfigure}[c]{0.8\linewidth}
\includegraphics[width=\textwidth]{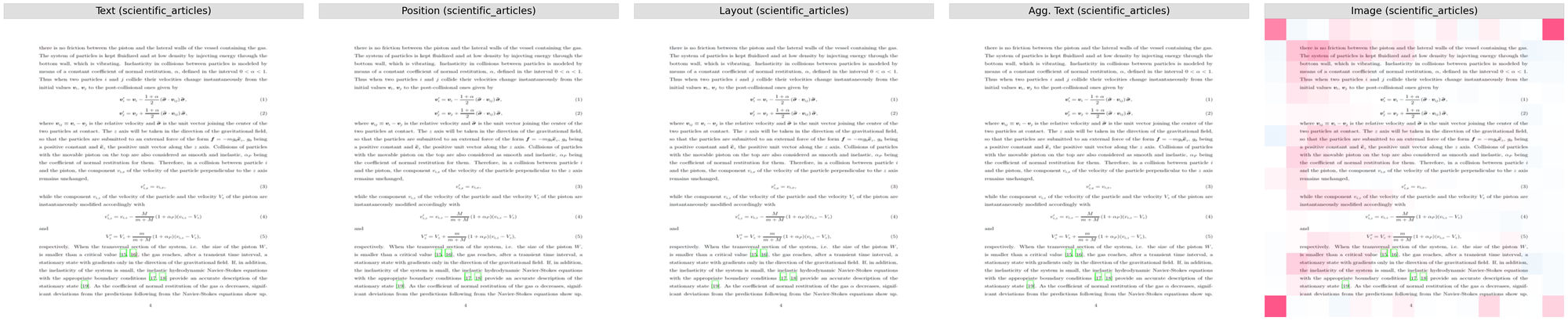}
\end{subfigure}
\par\smallskip
\noindent
\parbox[c]{1.5em}{\centering\rotatebox{90}{\small Occ. (DocLayNet)}}
\begin{subfigure}[c]{0.8\linewidth}
\includegraphics[width=\textwidth]{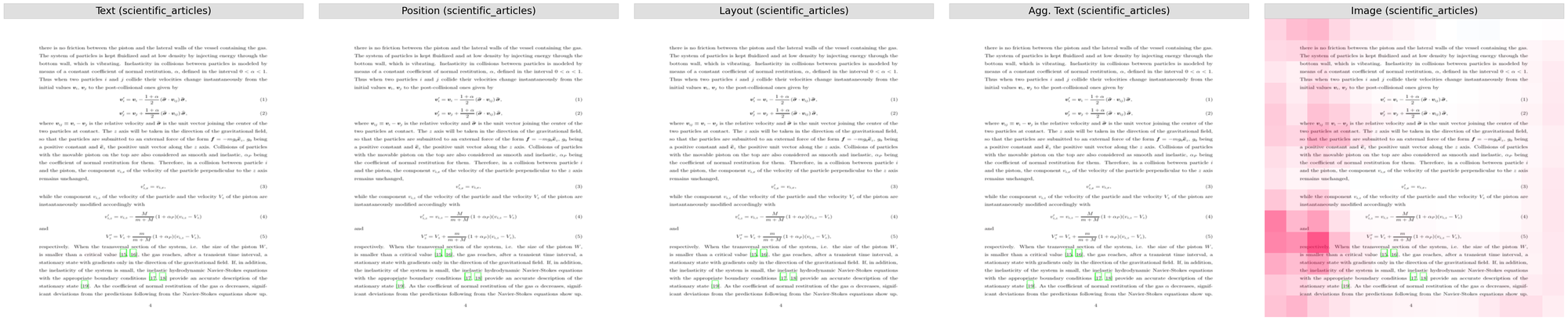}
\end{subfigure}
\par\smallskip
\noindent
\parbox[c]{1.5em}{\centering\rotatebox{90}{\small RA (DocLayNet)}}
\begin{subfigure}[c]{0.8\linewidth}
\includegraphics[width=0.5\textwidth]{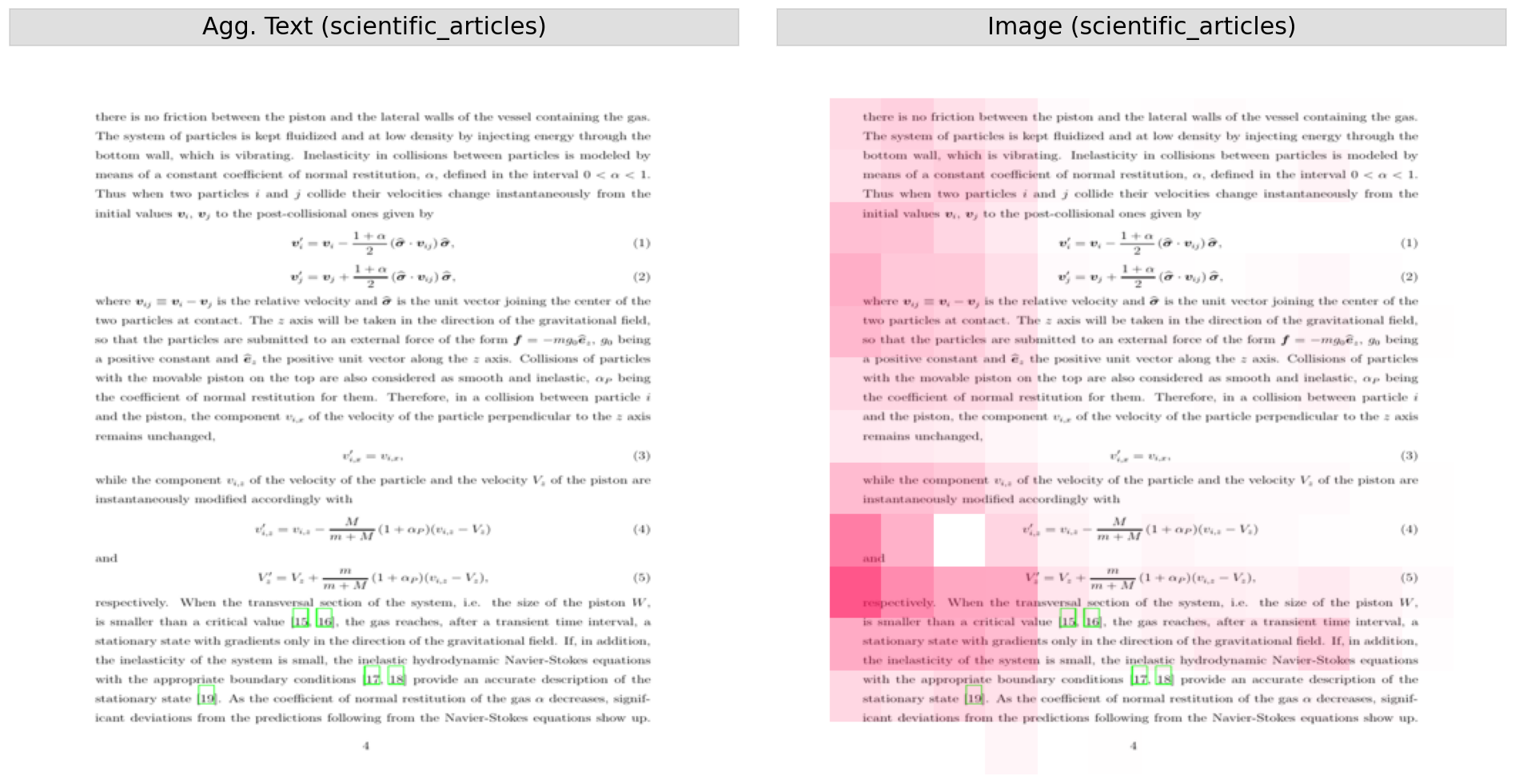}
\end{subfigure}
\caption{Modality-wise explanations of the \layoutlm{} model predictions for two randomly selected samples from \rvlcdip{} and \doclaynet{} (\cls{} task). Results are shown for three methods: \integratedgradientsf{}, \occlusionf{}, and \rawattnf{}.}
\label{fig:qual_cls}
\end{figure*}

\begin{figure*}[htbp]
\centering
\noindent
\parbox[c]{1.5em}{\centering\rotatebox{90}{\small IG (CORD)}}
\begin{subfigure}[c]{0.8\linewidth}
\includegraphics[width=\textwidth]{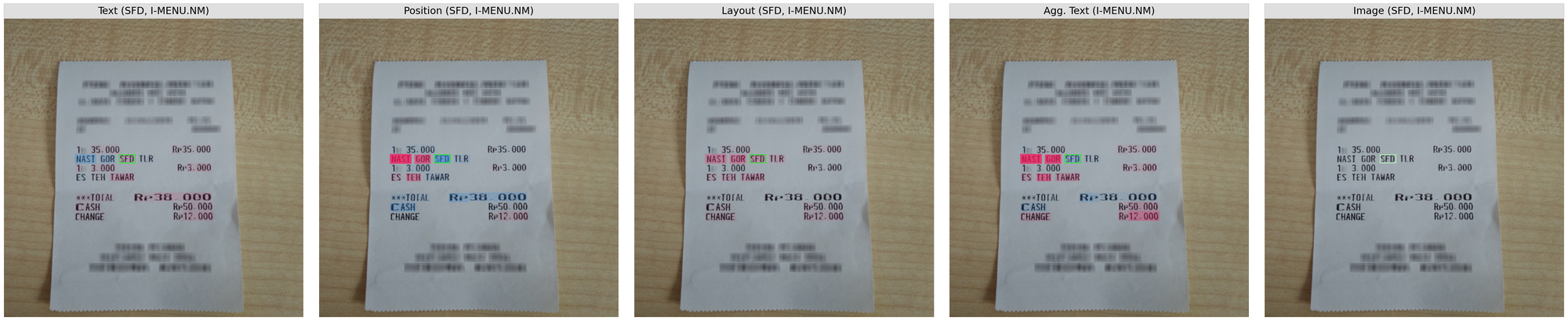}
\end{subfigure}
\par\smallskip
\noindent
\parbox[c]{1.5em}{\centering\rotatebox{90}{\small Occ. (CORD)}}
\begin{subfigure}[c]{0.8\linewidth}
\includegraphics[width=\textwidth]{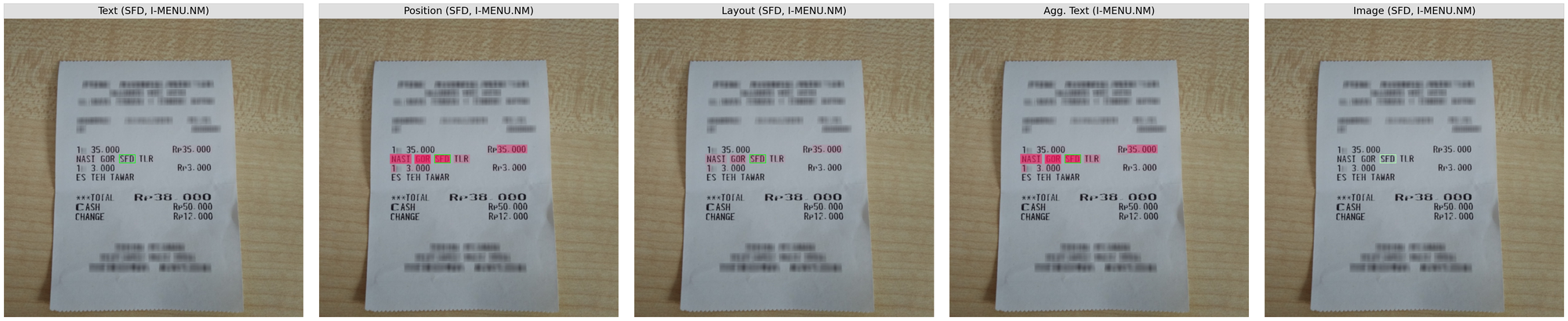}
\end{subfigure}
\par\smallskip
\noindent
\parbox[c]{1.5em}{\centering\rotatebox{90}{\small RA (CORD)}}
\begin{subfigure}[c]{0.8\linewidth}
\includegraphics[width=0.5\textwidth]{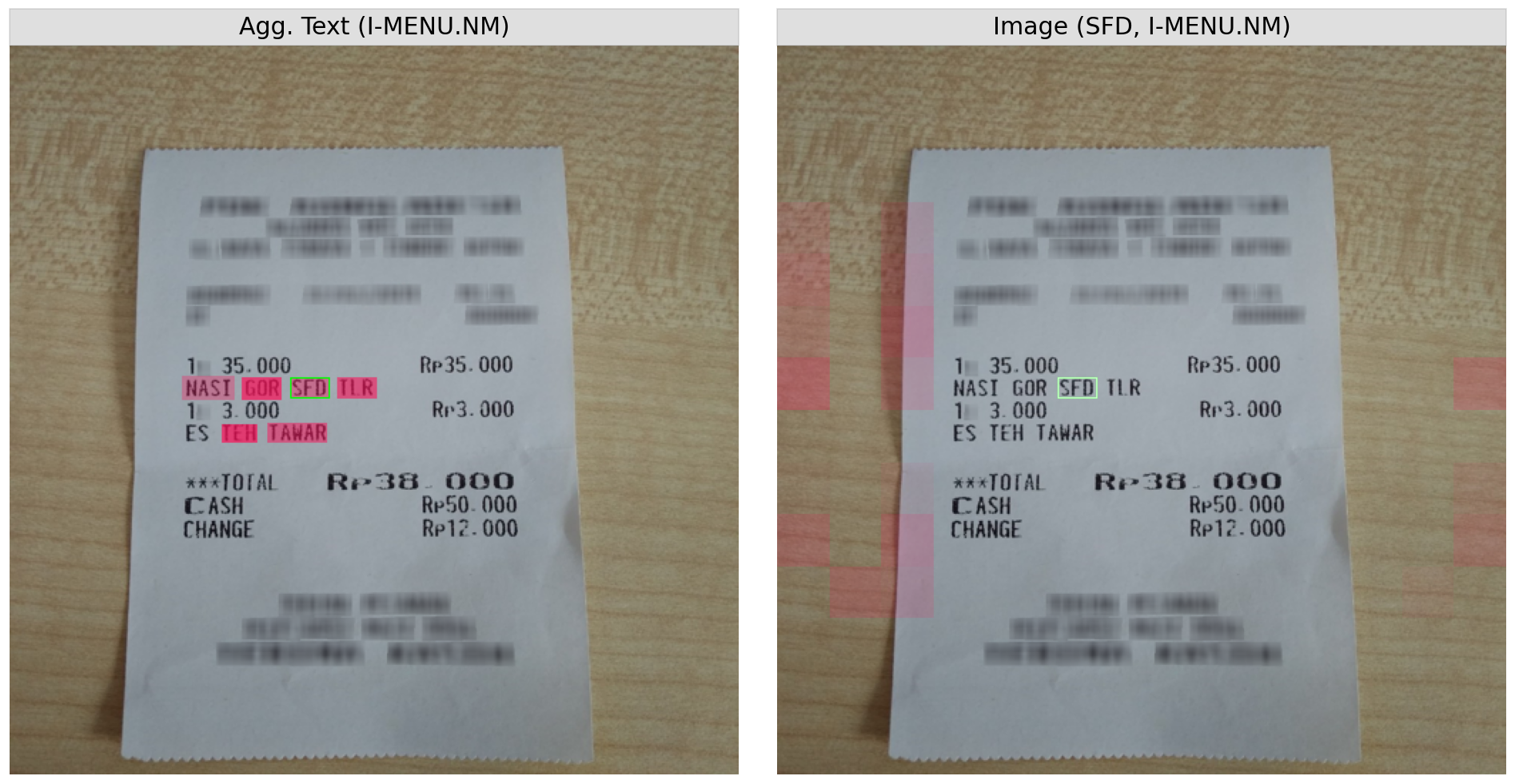}
\end{subfigure}
\par\smallskip
\noindent
\parbox[c]{1.5em}{\centering\rotatebox{90}{\small IG (FUNSD)}}
\begin{subfigure}[c]{0.8\linewidth}
\includegraphics[width=\textwidth]{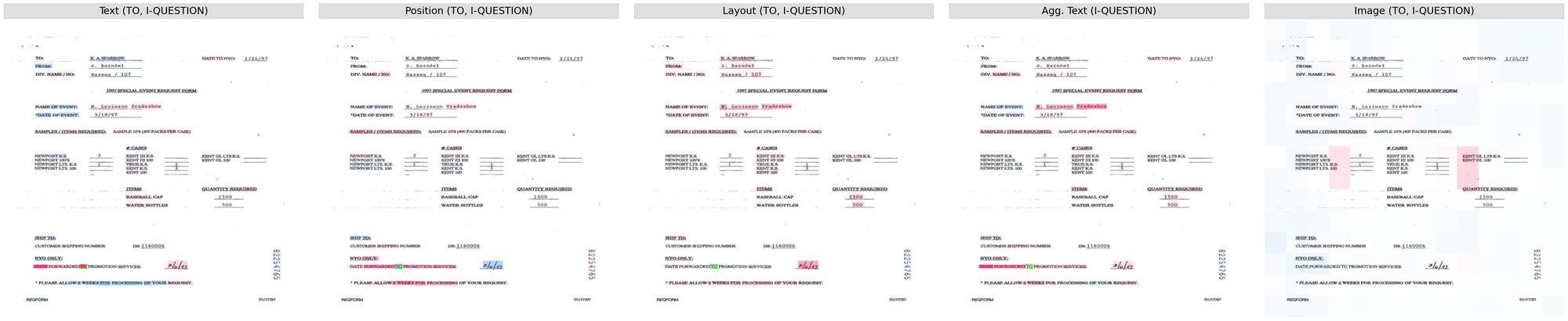}
\end{subfigure}
\par\smallskip
\noindent
\parbox[c]{1.5em}{\centering\rotatebox{90}{\small Occ. (FUNSD)}}
\begin{subfigure}[c]{0.8\linewidth}
\includegraphics[width=\textwidth]{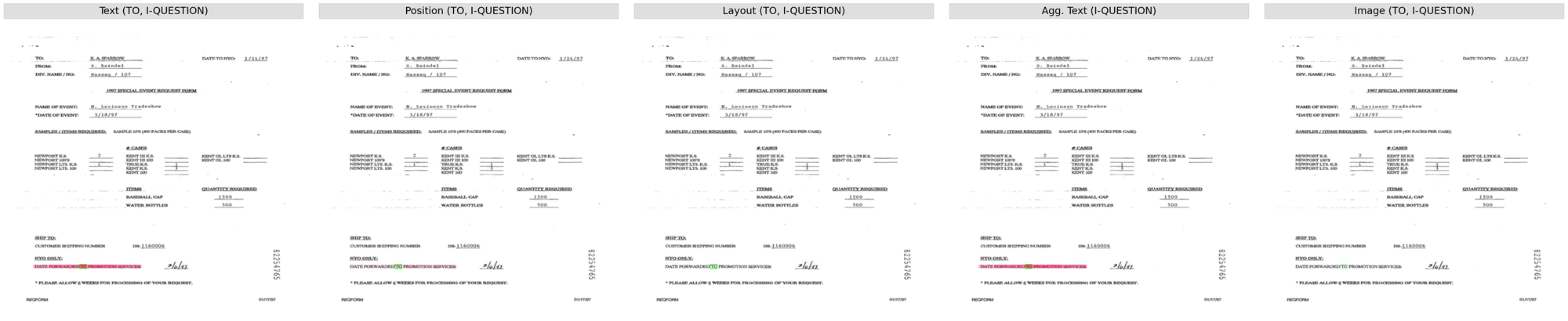}
\end{subfigure}
\par\smallskip
\noindent
\parbox[c]{1.5em}{\centering\rotatebox{90}{\small RA (FUNSD)}}
\begin{subfigure}[c]{0.8\linewidth}
\includegraphics[width=0.5\textwidth]{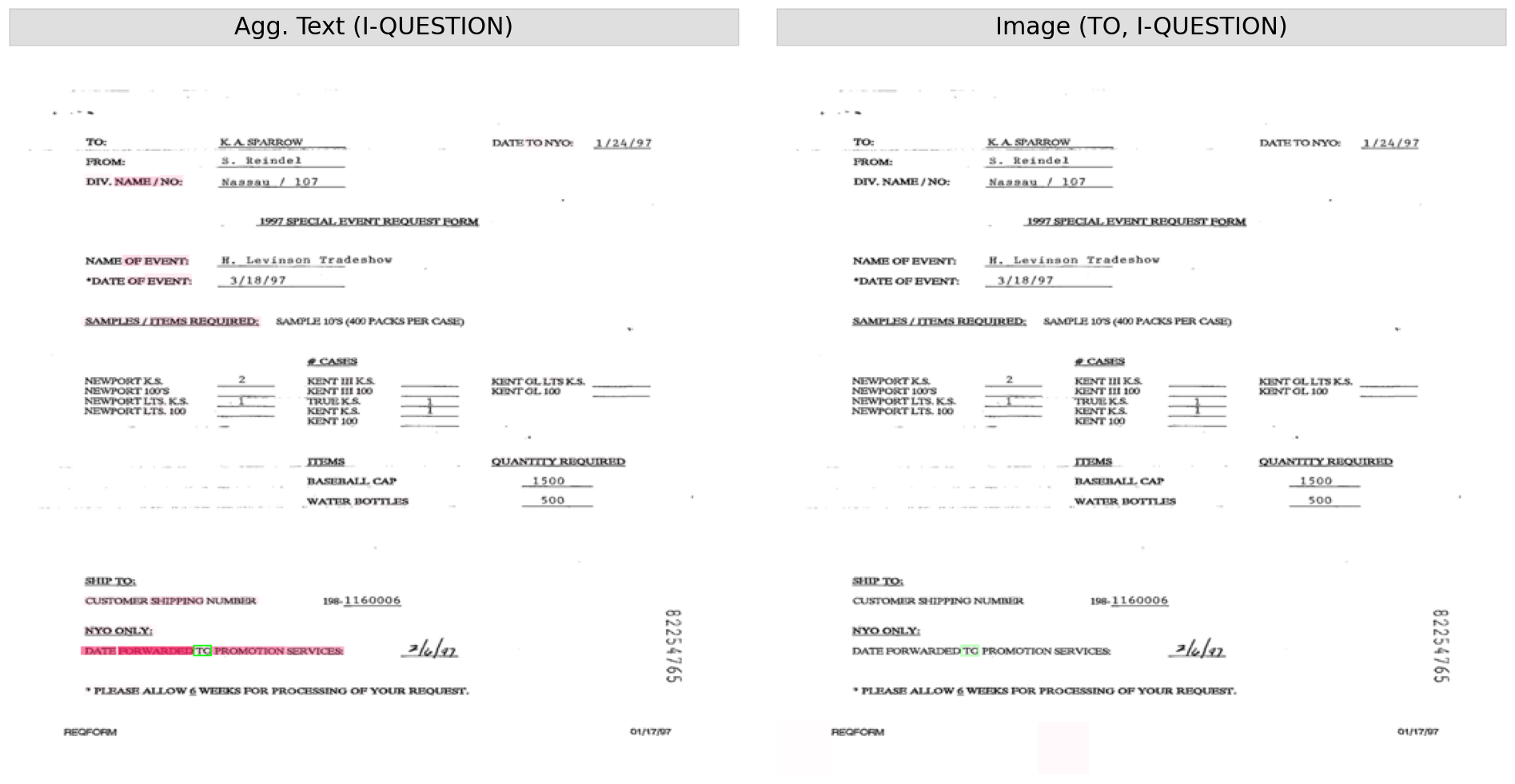}
\end{subfigure}
\caption{Modality-wise explanations of the \layoutlm{} model predictions for two randomly selected samples from \cord{} and \funsd{} (\ser{} task). Results are shown for three methods: \integratedgradientsf{}, \occlusionf{}, and \rawattnf{}, and for each sample, the target entity to be labeled is highlighted in a green bounding box.}
\label{fig:qual_ser}
\end{figure*}

\begin{figure*}[htbp]
\centering
\noindent
\parbox[c]{1.5em}{\centering\rotatebox{90}{\small IG (DocVQA)}}
\begin{subfigure}[c]{0.8\linewidth}
\includegraphics[width=\textwidth]{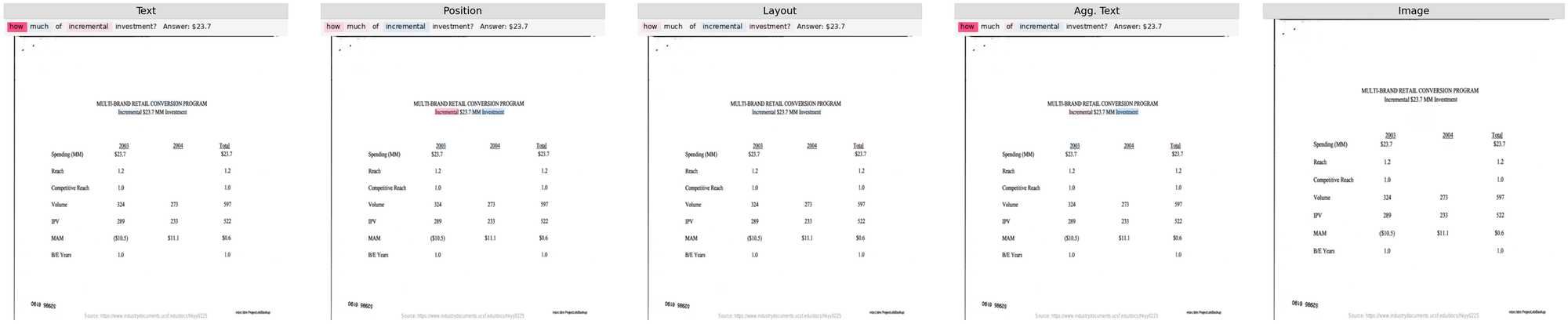}
\end{subfigure}
\par\smallskip
\noindent
\parbox[c]{1.5em}{\centering\rotatebox{90}{\small Occ. (DocVQA)}}
\begin{subfigure}[c]{0.8\linewidth}
\includegraphics[width=\textwidth]{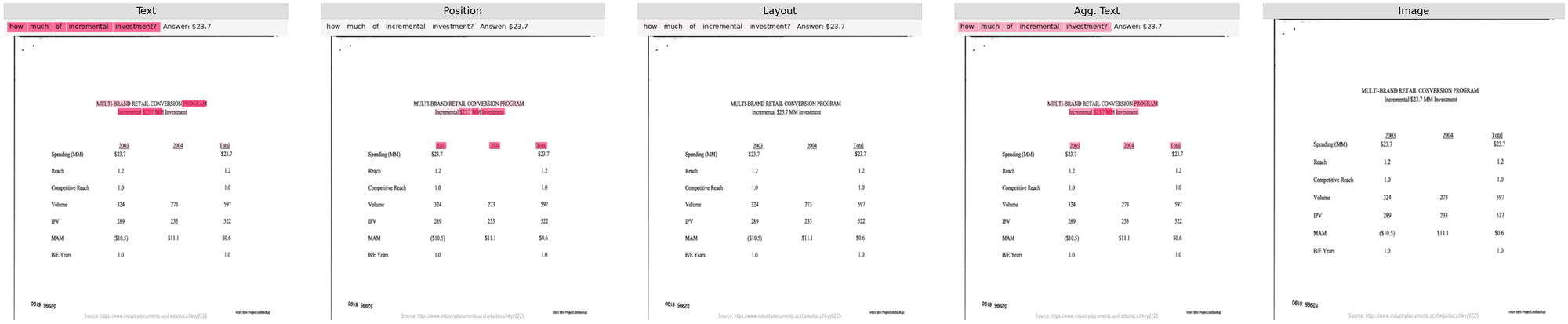}
\end{subfigure}
\par\smallskip
\noindent
\parbox[c]{1.5em}{\centering\rotatebox{90}{\small RA (DocVQA)}}
\begin{subfigure}[c]{0.8\linewidth}
\includegraphics[width=0.5\textwidth]{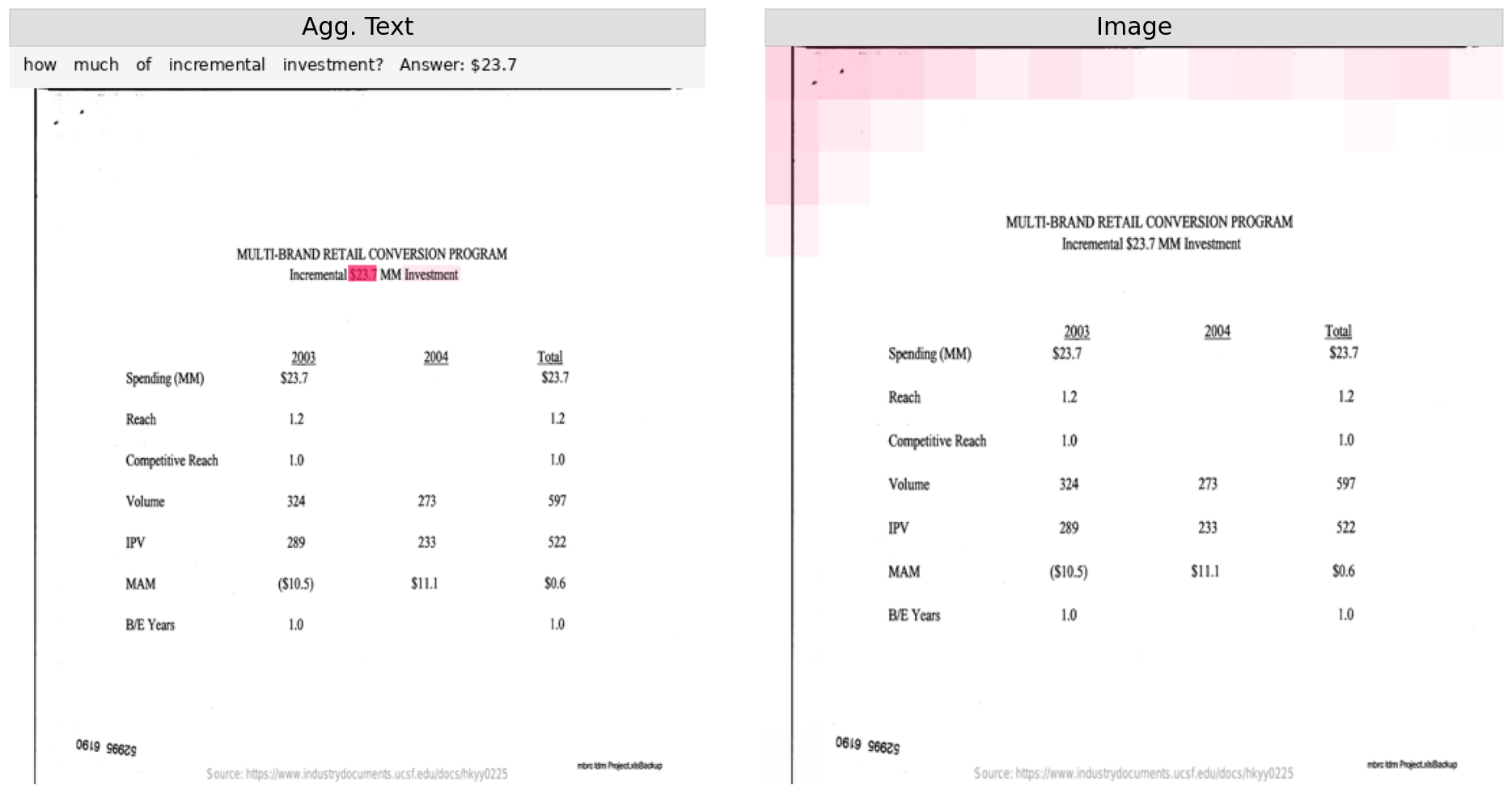}
\end{subfigure}
\par\smallskip
\noindent
\parbox[c]{1.5em}{\centering\rotatebox{90}{\small IG (DocVQA)}}
\begin{subfigure}[c]{0.8\linewidth}
\includegraphics[width=\textwidth]{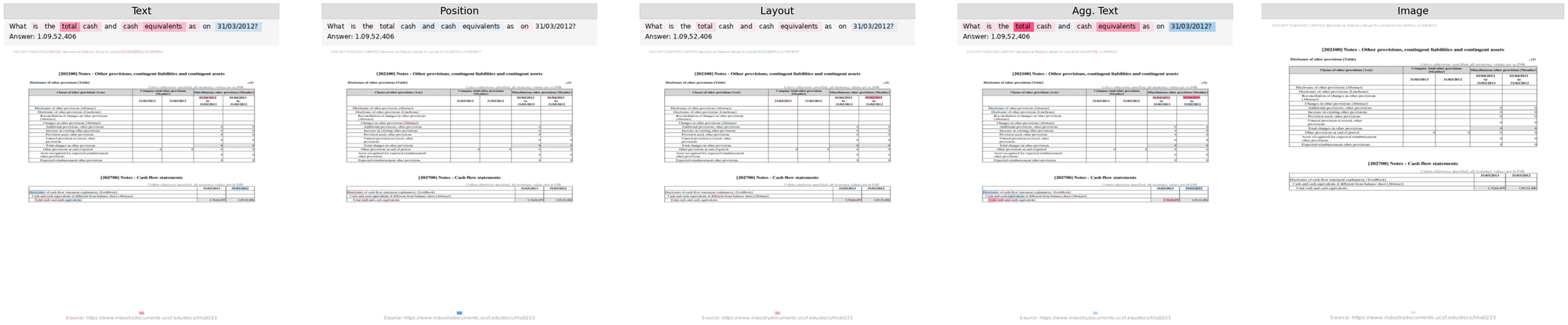}
\end{subfigure}
\par\smallskip
\noindent
\parbox[c]{1.5em}{\centering\rotatebox{90}{\small Occ. (DocVQA)}}
\begin{subfigure}[c]{0.8\linewidth}
\includegraphics[width=\textwidth]{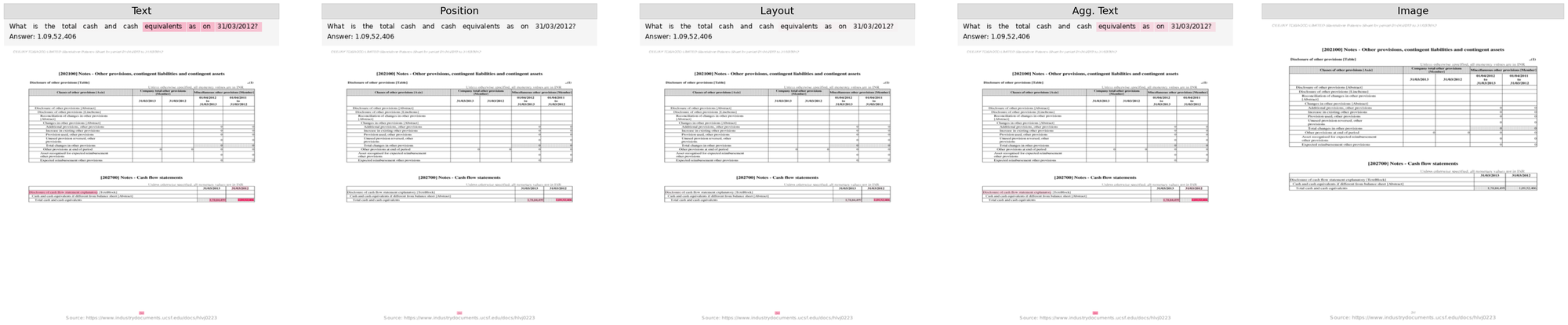}
\end{subfigure}
\par\smallskip
\noindent
\parbox[c]{1.5em}{\centering\rotatebox{90}{\small RA (DocVQA)}}
\begin{subfigure}[c]{0.8\linewidth}
\includegraphics[width=0.5\textwidth]{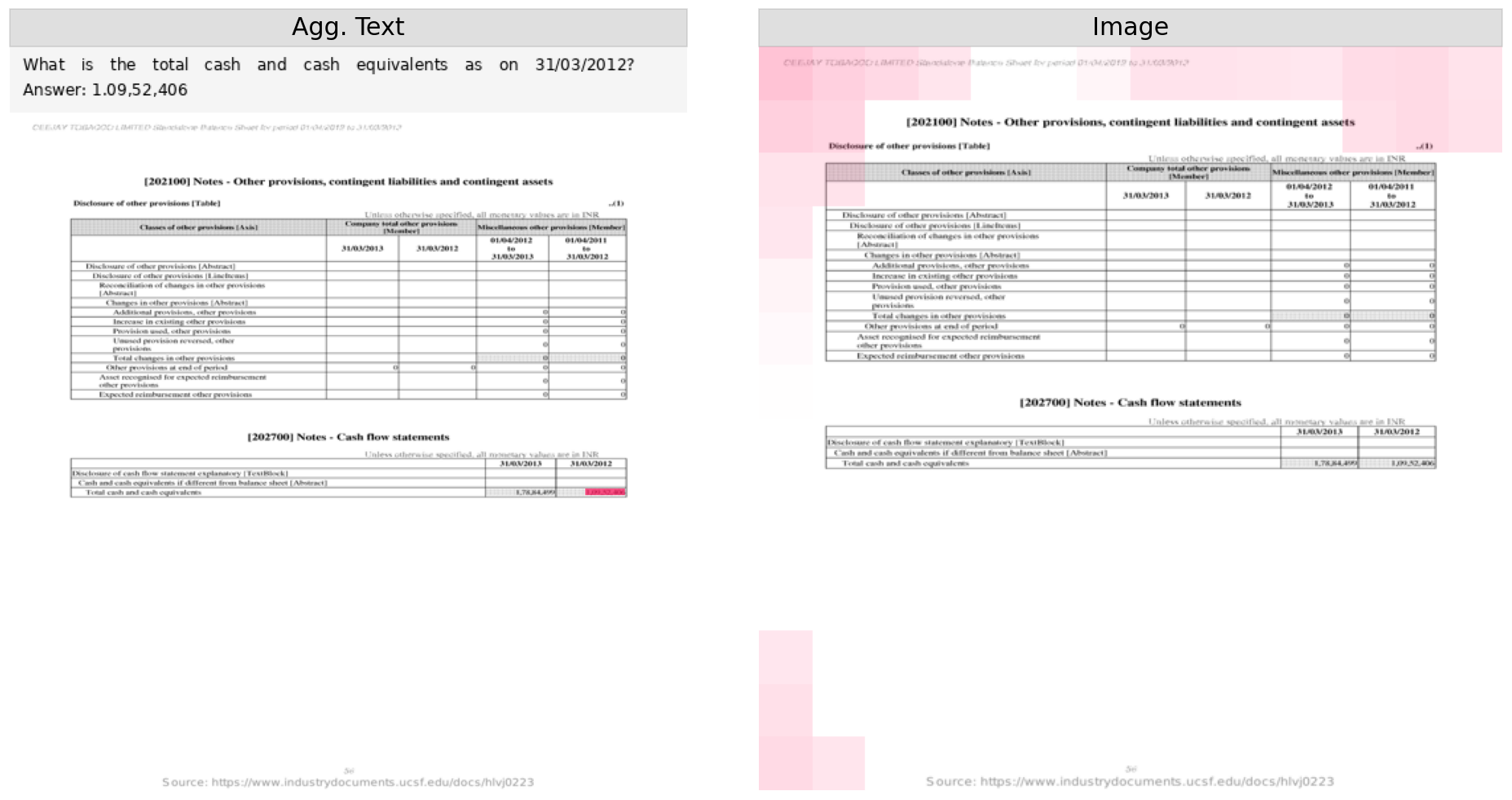}
\end{subfigure}
\caption{Modality-wise explanations of the \layoutlm{} model predictions for two randomly selected samples from \docvqa{} (\qa{} task). Results are shown for three methods: \integratedgradientsf{}, \occlusionf{}, and \rawattnf{}.}
\label{fig:qual_vqa}
\end{figure*}

To complement the quantitative evaluation, we qualitatively analyze modality-wise explanations produced by the three best-performing methods: \integratedgradientsf{}, \occlusionf{}, and \rawattnf{}. Fig.~\ref{fig:qual_cls}, Fig.~\ref{fig:qual_ser}, and Fig.~\ref{fig:qual_vqa} show randomly selected samples across all three tasks: \cls{} (\rvlcdip{} and \doclaynet{}), \ser{} (\cord{} and \funsd{}), and \qa{}, respectively. For each sample, we visualize the text, position, layout, aggregated text, and image attribution maps. Note that for \rawattnf{}, only aggregated text and image attributions are visualized, as it does not generate position-level and layout-level explanations. Additional samples are provided in Appendix~\ref{sec:appendix_qualitative}.

\subheading{Classification.}
Fig.~\ref{fig:qual_cls} presents modality-wise explanations for \layoutlm{} on two classification samples from \rvlcdip{} and \doclaynet{}. On the \rvlcdip{} sample (budget class), \integratedgradientsf{} suggests the model relies on both text and image modalities for its classification decision. One notable observation is that it assigns strong positive importance to the text modality of certain tokens (e.g., the document identity number or document header), but this is accompanied by negative importance from the layout modality of the same tokens. Interestingly, from the aggregated text-level attributions (Agg.\ Text.), it can be observed that for some tokens, these counteracting attributions cancel each other out, resulting in negligible aggregated attribution, but many words still emerge as more prominent in the aggregated view. For instance, the document identity number emerges with a strongly negative influence in the aggregated view. \occlusionf{}, in contrast to \integratedgradientsf{}, suggests the model relies heavily on the visual layout of the image as well as the spatial positioning of text, including table headers. \rawattnf{} similarly indicates strong reliance on the image modality, while for the text modality, it suggests a strong focus on table header values. Notably, all three methods assign noticeable importance to the word \texttt{ESTIMATE}, which intuitively makes sense for the budget class.
On the \doclaynet{} sample (scientific article), all three methods indicate that the model relies predominantly on the image modality for its classification decision, with no significant focus on textual modalities. Here, the focus appears to be mainly on the visual formatting of the document, particularly paragraph indentation and white space around the main text.

\subheading{Semantic Entity Recognition.}
Fig.~\ref{fig:qual_ser} presents the modality-wise explanations for the predictions of the \layoutlm{} model on two randomly selected SER samples from \cord{} and \funsd{}. For each sample, the target entity is highlighted in a green bounding box.
On the \cord{} sample, where the target entity is \texttt{I-MENU.NM}, \integratedgradientsf{} and \occlusionf{} show no focus on the image modality, while \rawattnf{} only very weakly focuses on some image regions far from the actual target. All three methods show strong focus on text in close vicinity to the target entity. \occlusionf{} in particular shows that the model strongly focuses on the 1D position modality of tokens preceding the target entity, with slight focus on the 2D positions of the same tokens. \rawattnf{} shows a similar pattern but also highlights some text that is not in the immediate vicinity of the target.
In contrast to \rawattnf{} and \occlusionf{}, \integratedgradientsf{} assigns particularly high importance to tokens far from the target token, a pattern that is difficult to interpret.
Additionally, it again shows counteracting positive and negative attributions for some tokens that zero out in the aggregated text attributions.
Similar patterns are observed on the FUNSD sample (target entity \texttt{I-QUESTION}), where \occlusionf{} and \rawattnf{} provide concise explanations focused on text tokens in the local neighborhood of the target token. \integratedgradientsf{}, on the other hand, spreads importance across the entire document, significantly increasing the explanation complexity.

\subheading{Document Visual Question Answering.}
Fig.~\ref{fig:qual_vqa} presents the modality-wise explanations for the predictions of the \layoutlm{} model on two randomly selected \docvqa{} samples. In each case, the explanation is computed with respect to the predicted start or end answer token.
On the first sample, \integratedgradientsf{} indicates a strong focus on the question tokens along with some focus on the tokens relevant to the answer token. \occlusionf{} indicates strong focus on the text and position modalities of the answer token \texttt{\$23.7} and the tokens in its close vicinity. \rawattnf{} only assigns high importance to the answer token itself and the word \texttt{Investment}, which is relevant to the prediction. Interestingly, it also indicates some focus on the top region of the image. Similar patterns are observed on the second sample, where again \rawattnf{} assigns high importance to the answer token and top image patches. \occlusionf{} assigns high importance to the tokens in the close vicinity of the answer token, and especially to its layout and position modalities. \integratedgradientsf{} assigns high importance to the question text and to tokens vertically and horizontally aligned with the answer token.

\begin{figure*}[htbp]
	\centering
	\begin{subfigure}[t]{\linewidth}
		\includegraphics[width=\textwidth]{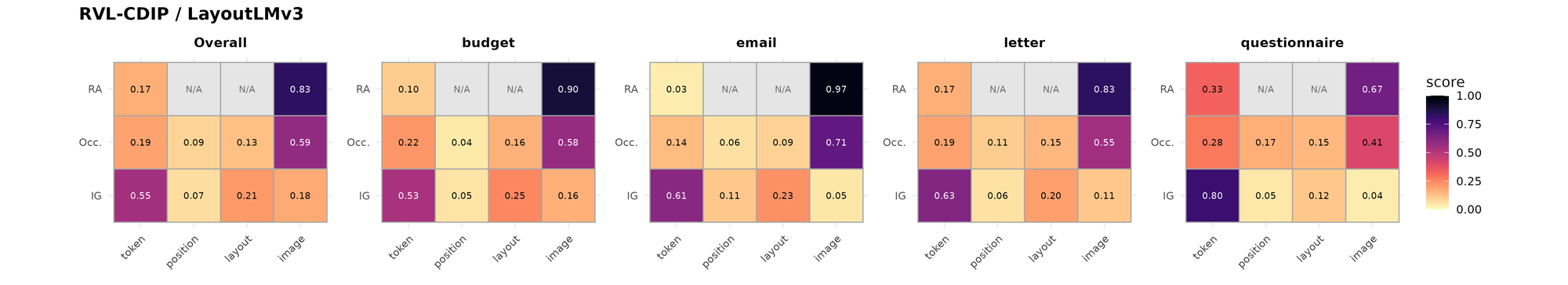}
	\end{subfigure}
	\begin{subfigure}[t]{\linewidth}
		\includegraphics[width=\textwidth]{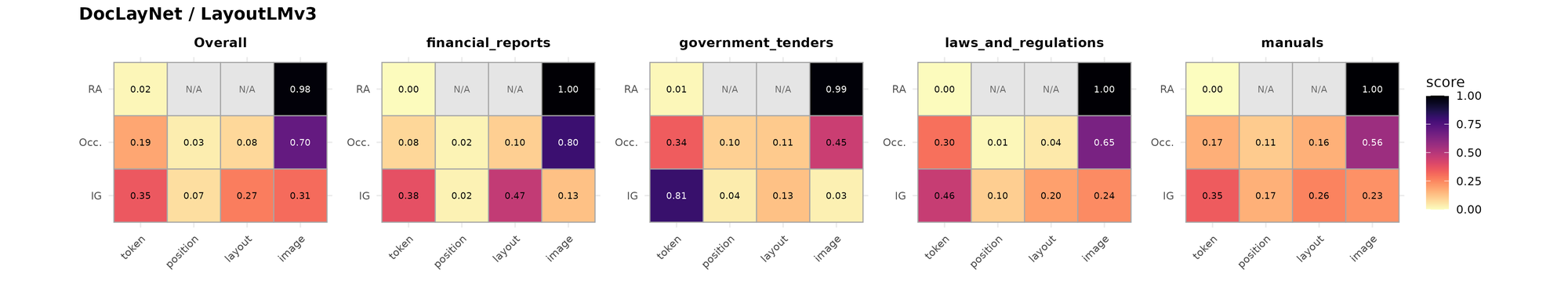}
	\end{subfigure}
	\begin{subfigure}[t]{\linewidth}
		\includegraphics[width=\textwidth]{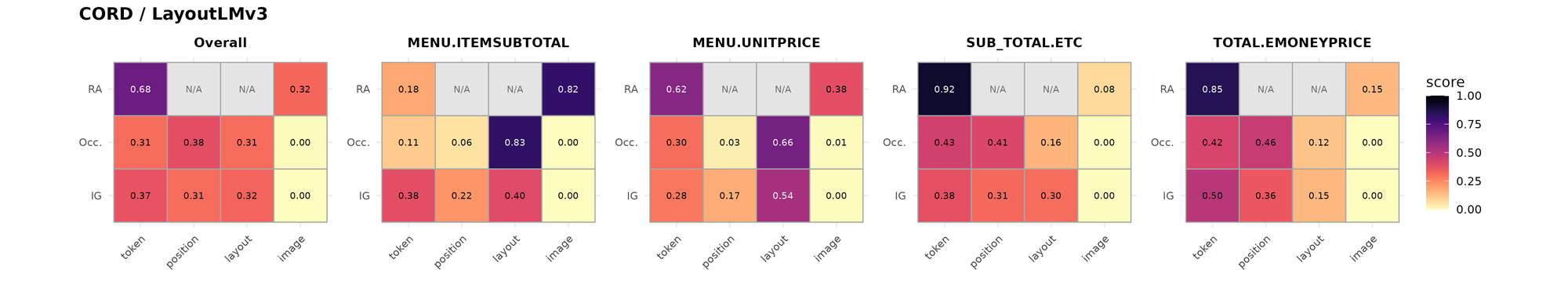}
	\end{subfigure}
	\begin{subfigure}[t]{\linewidth}
		\includegraphics[width=\textwidth]{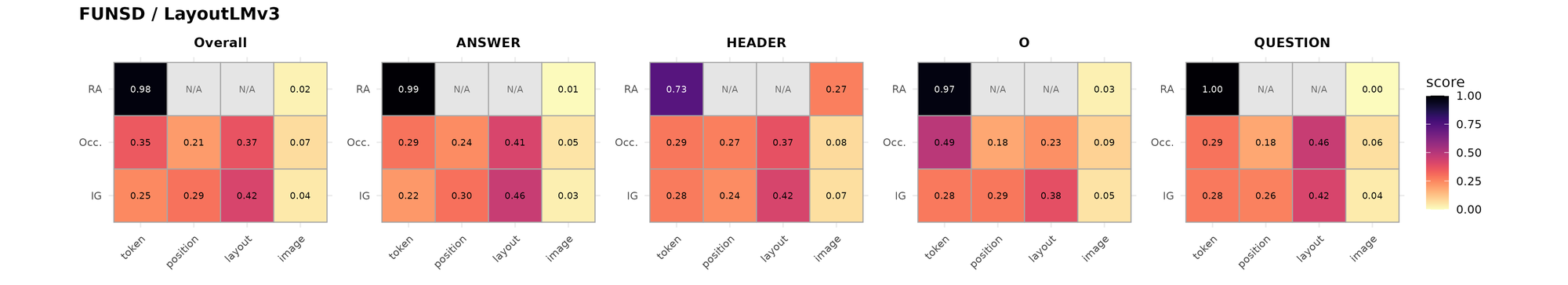}
	\end{subfigure}
	\begin{subfigure}[t]{0.6\linewidth}
		\includegraphics[width=\textwidth]{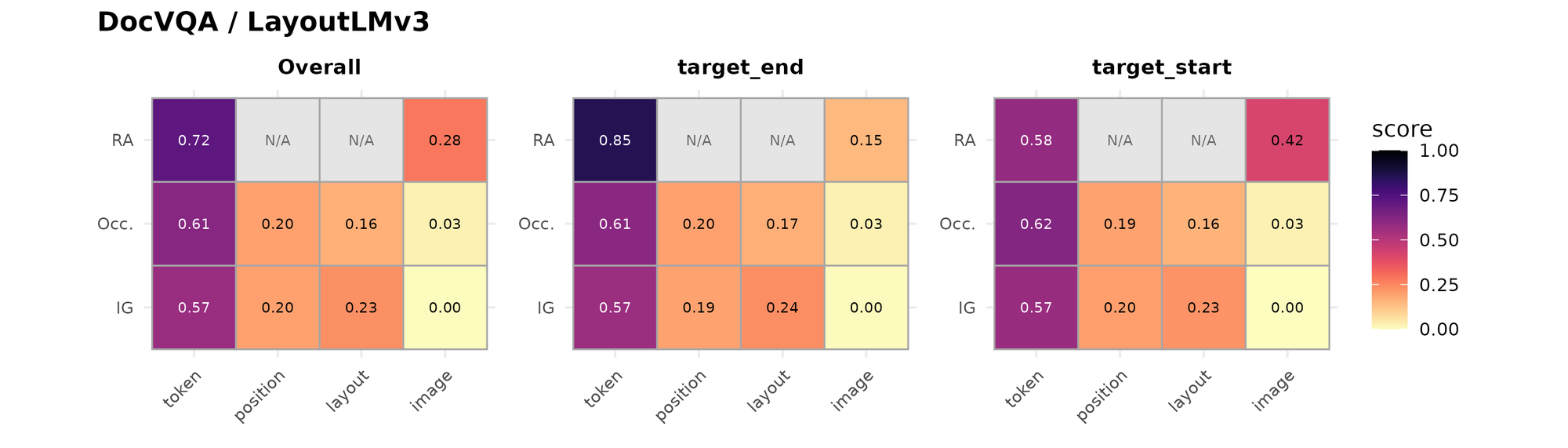}
	\end{subfigure}%
	\caption{
	Modality contributions computed over explanations generated by \integratedgradientsf{}, \occlusionf{}, and \rawattnf{} for the \layoutlm{} model. Results are reported as a dataset-wide mean (left) and as a per-label mean for four randomly selected labels (right). For \docvqa{}, contributions are reported separately for the start and end tokens of the predicted answer.
	}
	\label{fig:modality_frac}
\end{figure*}

\subsection{Global Attribution Analysis}
\label{sec:global_attribution}
While qualitative analysis of individual samples provides useful insights into local model behavior, inspecting explanations one sample at a time is neither scalable nor sufficient for drawing broader conclusions. To obtain global insights, we conduct three dataset-wide analyses: a modality contribution analysis that examines which modalities the model relies on (Section~\ref{sec:modality-contribution}), a spatial attribution locality analysis that measures how concentrated attributions are around target entities (Section~\ref{sec:results_locality}), and a top-$k$ word frequency analysis that identifies which words are most frequently attributed across classes (Section~\ref{sec:results_topk}).

\subsubsection{Modality Contribution Analysis}
\label{sec:modality-contribution}
To quantify the contribution of each modality across the dataset, we select the top-$k$ most important features in each sample's explanation and compute the fraction of features belonging to each modality. The results are then aggregated over the dataset or per class. In practice, we compute modality contribution on the top 10\% of features.
As a representative example, Fig.~\ref{fig:modality_frac} presents these fractions for the \layoutlm{} model on five datasets (\rvlcdip{}, \doclaynet{}, \cord{}, \funsd{}, and \docvqa{}), reported both as a dataset-wide mean and per label for four randomly selected labels per dataset. For \docvqa{}, fractions are reported separately for the start and end tokens of the predicted answer. Additional results are provided in Appendix~\ref{sec:appendix_modality}.

Focusing on \rvlcdip{}, \rawattnf{} indicates that the model relies heavily on the image modality (up to 80\%), with considerably lower importance on text. \occlusion{} suggests a similar pattern but with moderate reliance on text and layout modalities as well. \integratedgradientsf{}, in contrast, suggests the opposite pattern, with high importance on the text modality and moderate importance on layout.
This is consistent with our observations in the qualitative analysis, where \integratedgradientsf{} often distributed positive and negative contributions across text modalities.
Overall, we observe that the class-wise importance scores align well with the dataset-wide mean scores.

On \doclaynet{}, similar patterns are observed. Compared to the \rvlcdip{} results, \integratedgradientsf{} indicates relatively larger importance on layout and image modalities. \rawattnf{} suggests the model relies almost exclusively on the image modality. \occlusionf{} indicates reliance on both image and text, with variation among classes; for instance, for \texttt{government\_tenders}, it suggests a relatively higher focus on the text modality compared to \texttt{financial\_reports} where the image modality dominates.

On the \cord{} and \funsd{} datasets, \occlusionf{} and \integratedgradientsf{} demonstrate a similar pattern, both indicating negligible importance on the image modality and comparable importance on the text, position, and layout modalities. On \cord{}, \rawattnf{} also indicates a moderate reliance on the image modality, with notably high importance assigned to the image modality for the \texttt{MENU.ITEMSUBTOTAL} and \texttt{MENU.UNITPRICE} entity types.
On \docvqa{}, \occlusionf{} and \integratedgradientsf{} show a similar importance distribution for both start and end tokens of the predicted answer. In contrast, \rawattnf{} indicates a particularly high reliance on the image modality for the start token.

\begin{figure*}[t]
	\centering
	\begin{subfigure}[b]{\textwidth}
		\centering
		\includegraphics[width=\textwidth,draft=false]{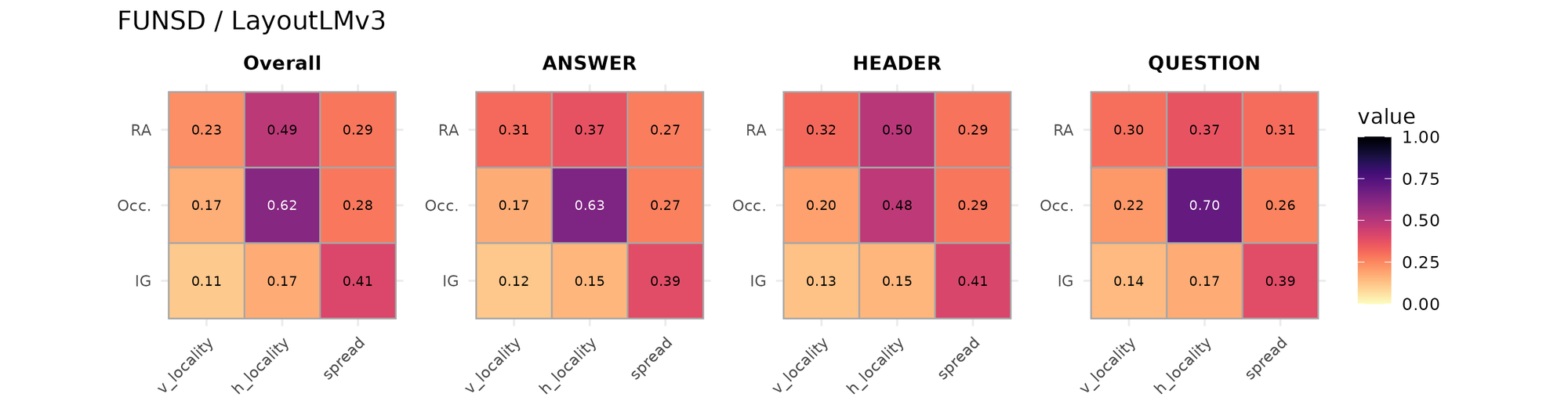}
		\label{fig:locality_funsd}
	\end{subfigure}
	\begin{subfigure}[b]{\textwidth}
		\centering
		\includegraphics[width=\textwidth,draft=false]{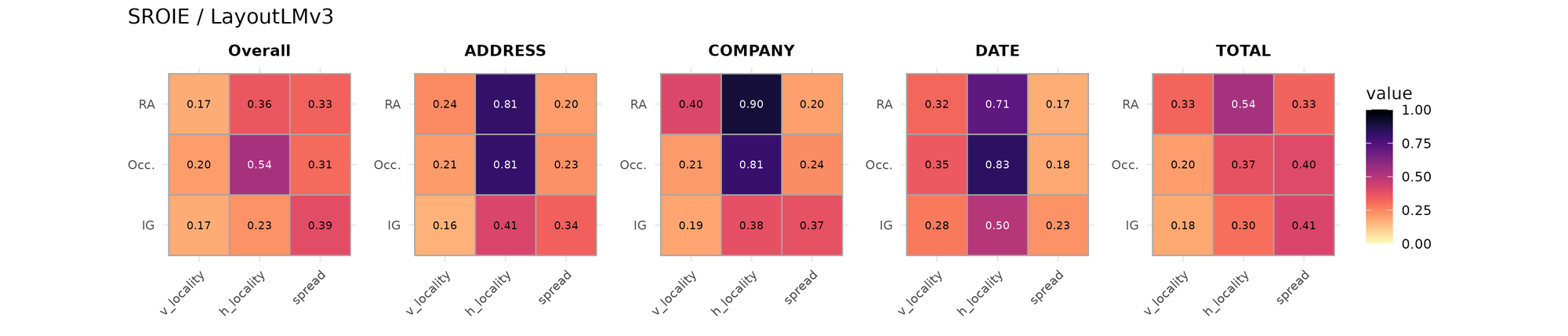}
		\label{fig:locality_sroie}
	\end{subfigure}
	\begin{subfigure}[b]{\textwidth}
		\centering
		\includegraphics[width=\textwidth,draft=false]{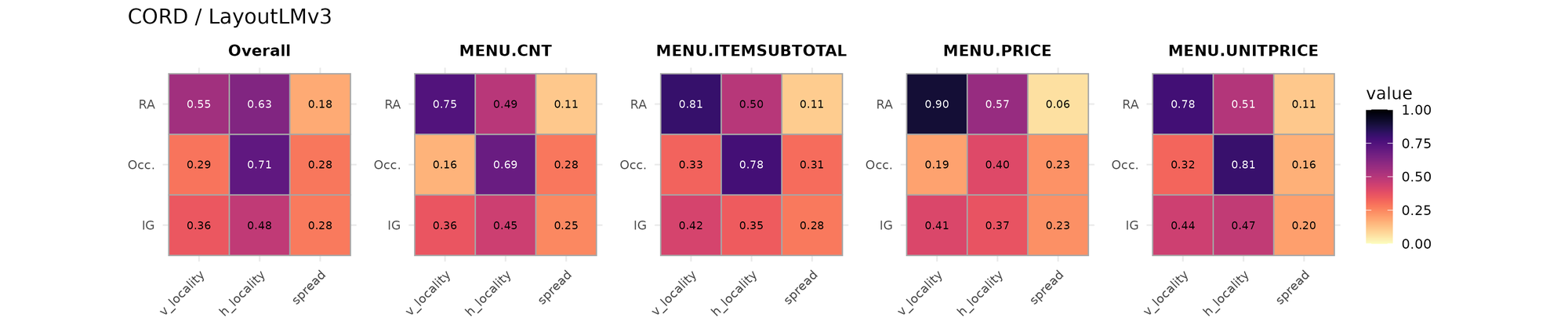}
		\label{fig:locality_cord}
	\end{subfigure}
	\begin{subfigure}[b]{\textwidth}
		\centering
		\includegraphics[width=\textwidth,draft=false]{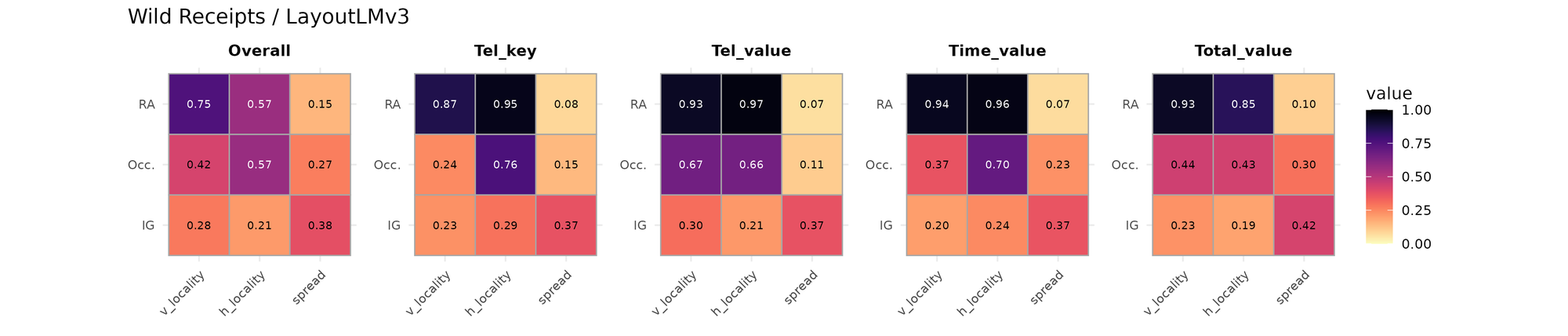}
		\label{fig:locality_wildreceipts}
	\end{subfigure}
	\caption{
		Spatial attribution locality scores computed over explanations generated by \integratedgradientsf{}, \occlusionf{}, and \rawattnf{} for the \layoutlm{} model. For each dataset, vertical locality, horizontal locality, and total spread are reported as a dataset-wide mean (left) and as a per-entity-type mean for four randomly selected entity types (right).
	}
	\label{fig:locality_ser}
\end{figure*}

\subsubsection{Spatial Attribution Locality Analysis}
\label{sec:results_locality}
In the \ser{} task, the large number of target entities per sample makes it impractical to analyze each individual sample-target pair qualitatively.
Here, spatial attribution locality analysis provides a broader view of how the model distributes importance around each target entity type; for instance, whether the model focuses entirely on the entities themselves, their vertical or horizontal surrounding tokens, or whether its focus is spread out to irrelevant entities across the document. In particular, we analyze the spatial locality of attributions by computing three quantities. \emph{Vertical locality}, the total attribution assigned to tokens in the same column as the target (i.e., vertically aligned):
\begin{equation}
L_v = \sum_{i:\,|x_i - x_t| \leq \tau_x} a_i
\end{equation}
\emph{Horizontal locality}, the total attribution assigned to tokens on the same row as the target (i.e., horizontally aligned):
\begin{equation}
L_h = \sum_{i:\,|y_i - y_t| \leq \tau_y} a_i
\end{equation}
\emph{Total spread}, the attribution-weighted root mean squared distance of all tokens from the target:
\begin{equation}
S = \sqrt{\sum_i a_i \left((x_i - x_t)^2 + (y_i - y_t)^2\right)}
\end{equation}
For all three quantities, attributions are first aggregated to the text level, clamped to non-negative values, and normalized to sum to one.
Higher vertical and horizontal locality values indicate that the method focuses on tokens along the respective axes, while lower spread values indicate that attributions are spatially compact overall. Both thresholds $\tau_x$ and $\tau_y$ are set to a fixed ratio of $0.025$ relative to the total image width and height, respectively.

Fig.~\ref{fig:locality_ser} presents the spatial attribution locality scores for the \layoutlm{} model, computed over explanations generated by \occlusionf{}, \integratedgradientsf{}, and \rawattnf{}. For each of the four \ser{} datasets, we report vertical locality, horizontal locality, and total spread as a dataset-wide mean across all samples, as well as per-entity-type means for four randomly selected entity types.

A number of interesting observations can be made from these results. On both \sroie{} and \cord{}, \occlusionf{} and \rawattnf{} show a similar pattern, with notably larger horizontal locality than vertical locality. This suggests that, when predicting entity types on these datasets, the model pays relatively greater attention to the target entity itself and to other words in its horizontal neighborhood. In contrast, \integratedgradientsf{} shows a considerably higher spread on all datasets, with no particular preference for horizontal or vertical locality.

The label-wise results further reveal how the attribution locality distribution varies across entity types. For instance, on \sroie{}, \occlusionf{} shows high horizontal locality scores for the \texttt{ADDRESS}, \texttt{COMPANY}, and \texttt{DATE} entity types, meaning that importance is primarily assigned to tokens in the same row as the target token. On the other hand, on \wildreceipts{}, \rawattnf{} shows both high vertical and horizontal locality with negligible overall spread for all listed entity types.
This indicates that the \rawattnf{} attributions for these entity types focus almost exclusively on the target words themselves, with only minimal attention to their surrounding neighborhood.

\begin{figure*}[t]
	\centering
	\begin{subfigure}[b]{0.45\textwidth}
		\centering
		\includegraphics[width=\textwidth,draft=false]{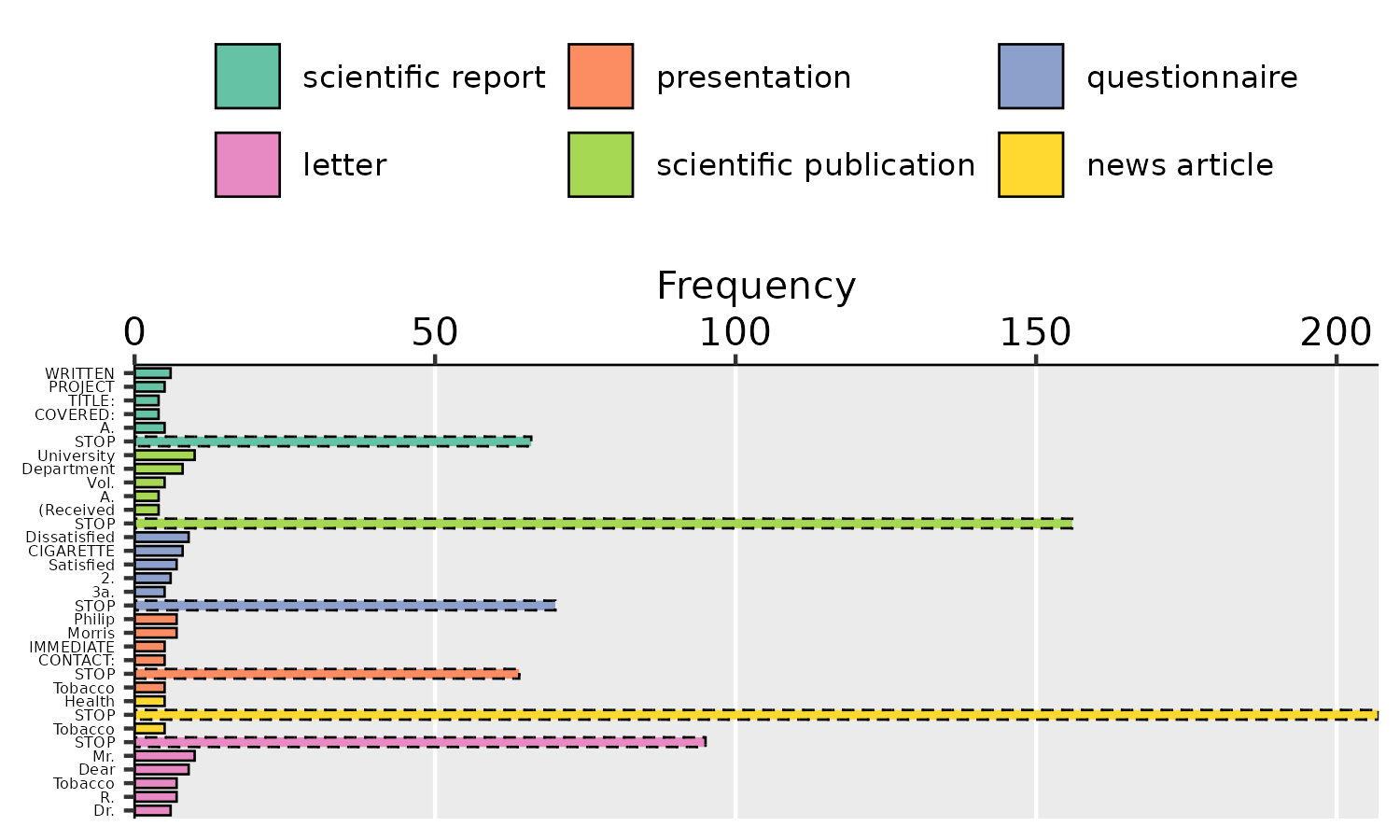}
		\caption{\rvlcdip{}}
		\label{fig:topk_rvlcdip}
	\end{subfigure}
	\begin{subfigure}[b]{0.45\textwidth}
		\centering
		\includegraphics[width=\textwidth,draft=false]{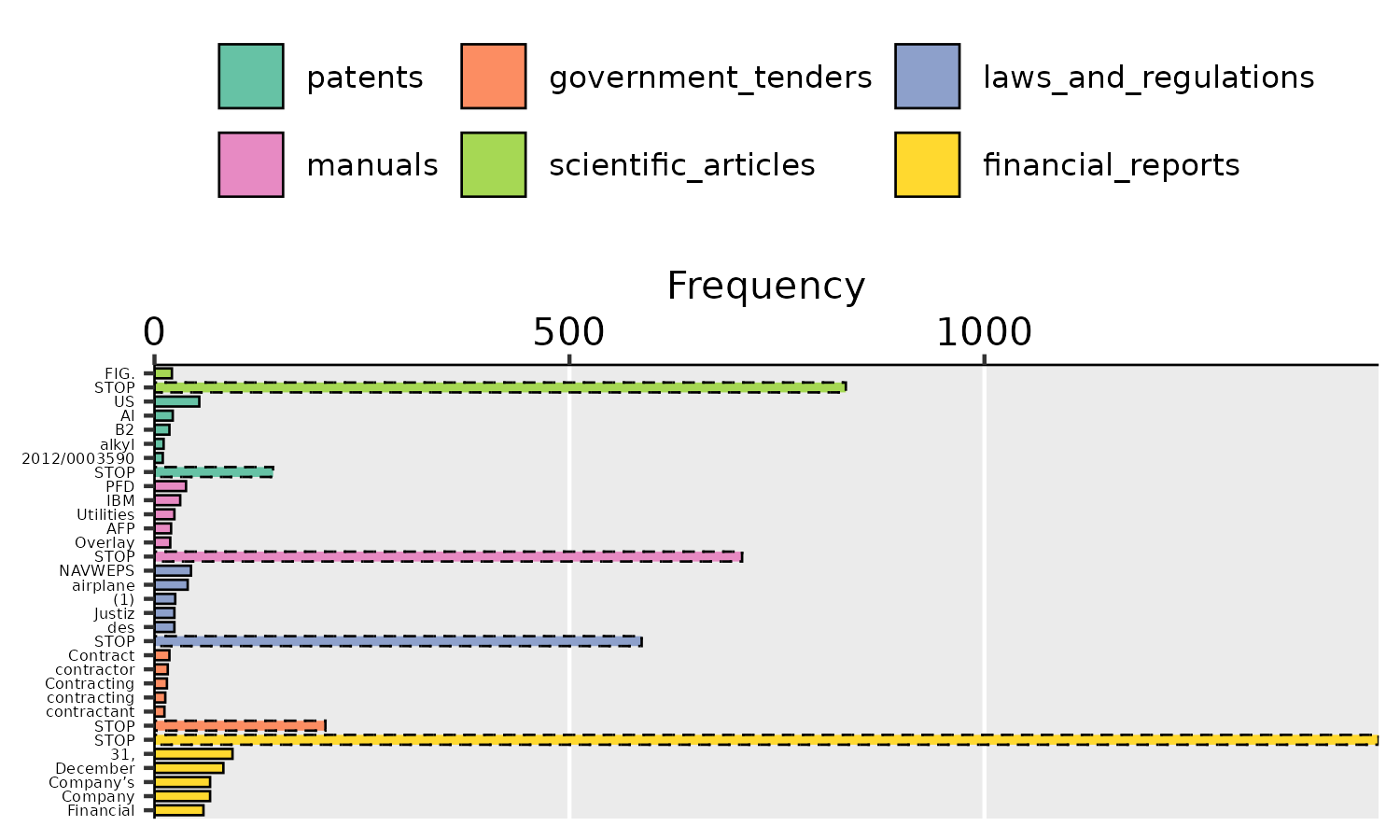}
		\caption{\doclaynet{}}
		\label{fig:topk_doclaynet}
	\end{subfigure}
	\begin{subfigure}[b]{0.45\textwidth}
	\centering
	\includegraphics[width=\textwidth,draft=false]{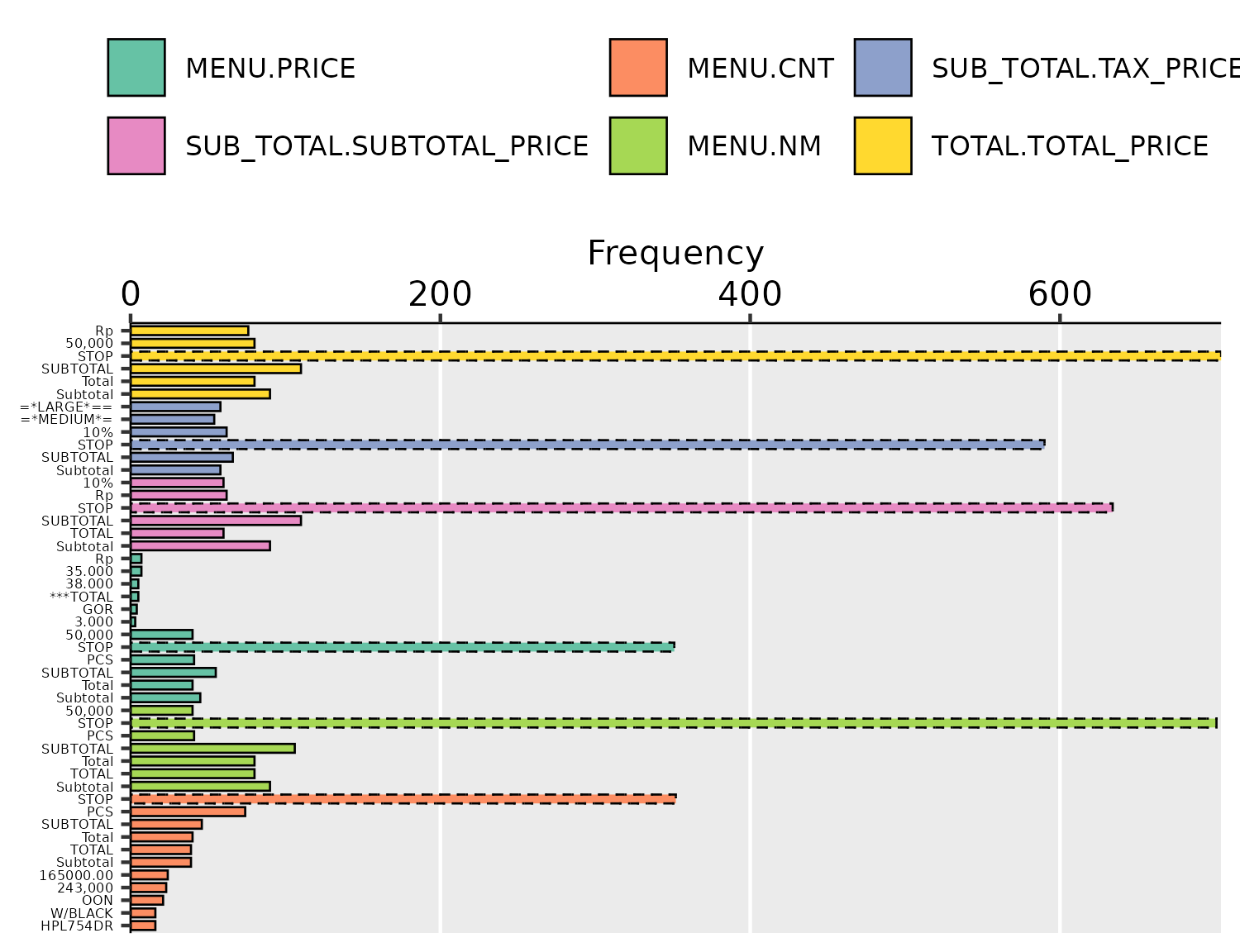}
	\caption{\cord{}}
	\label{fig:topk_cord}
	\end{subfigure}
	\begin{subfigure}[b]{0.45\textwidth}
		\centering
		\includegraphics[width=\textwidth,draft=false]{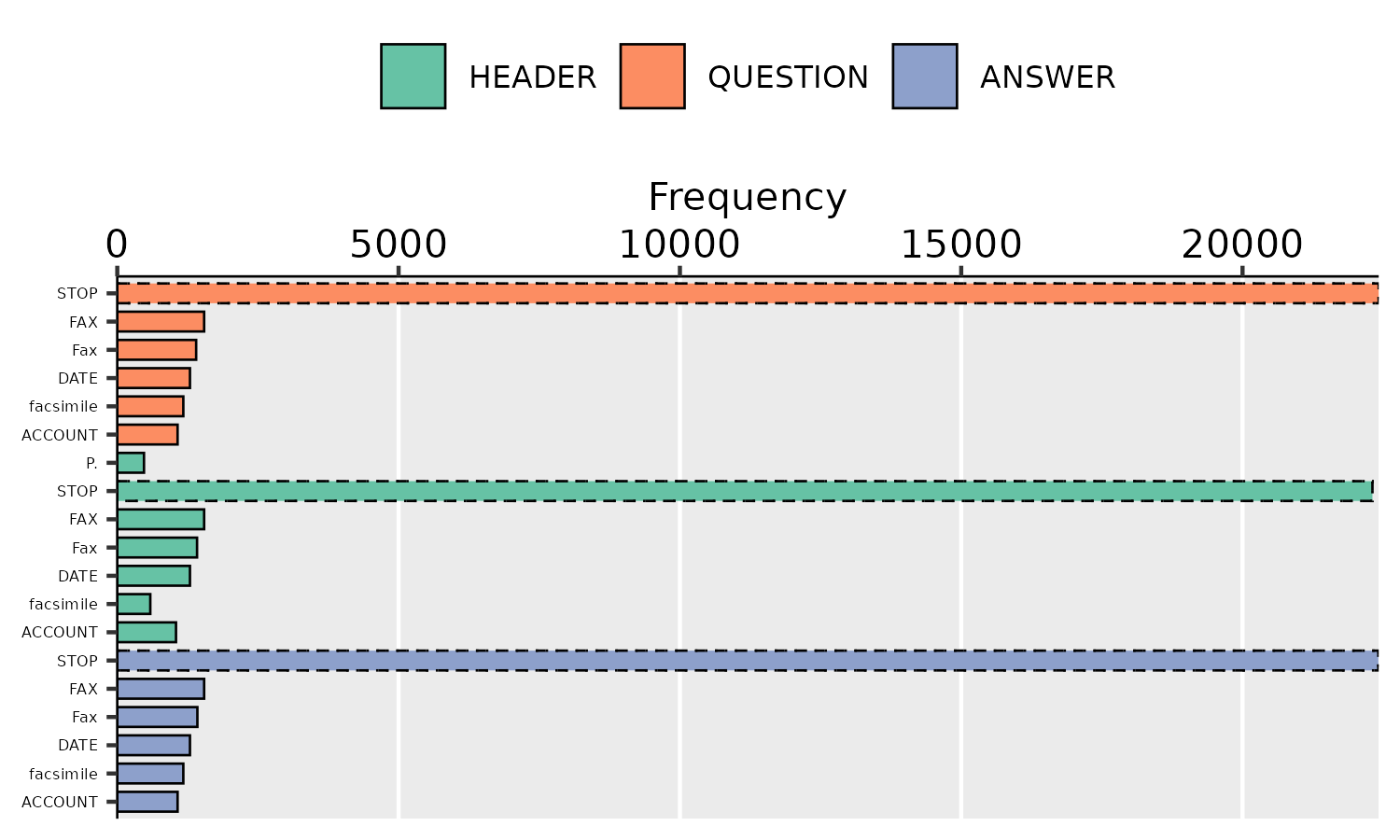}
		\caption{\funsd{}}
		\label{fig:topk_funsd}
	\end{subfigure}
	\begin{subfigure}[b]{0.3\textwidth}
		\centering
		\includegraphics[width=\textwidth,draft=false]{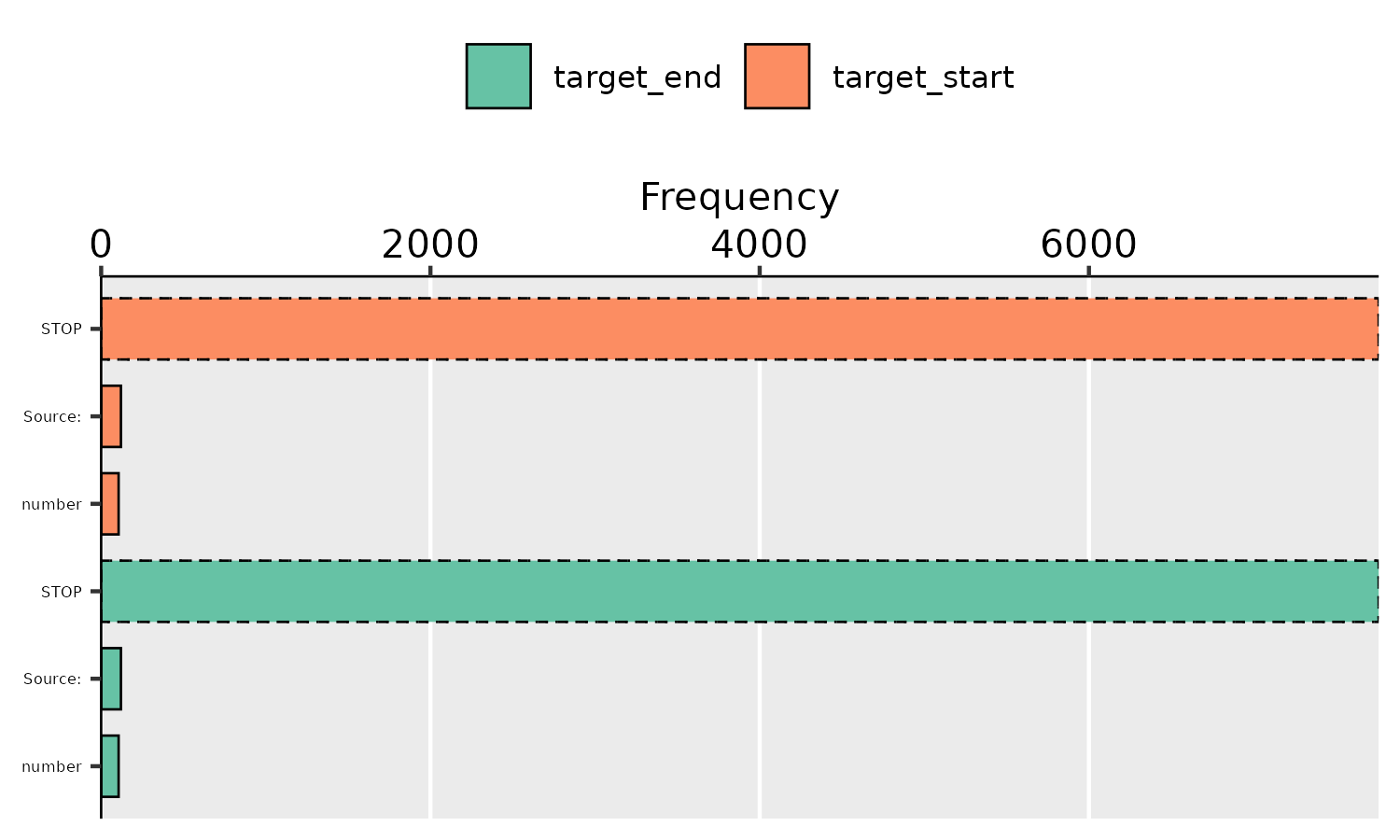}
		\caption{\docvqa{}}
		\label{fig:topk_docvqa}
	\end{subfigure}
\caption{Top-$k$ word frequency analysis for the \layoutlm{} model across five datasets. For each dataset, the most frequently attributed words (top 1\% by attribution magnitude) are shown for up to six randomly selected labels. (a--b)~\cls{} task, (c--d)~\ser{} task, (e)~\qa{} task.}
	\label{fig:topk_wordfreq}
\end{figure*}

\subsubsection{Top-k Word Frequency Analysis}
\label{sec:results_topk}
To analyze whether the model relies on specific textual patterns for its predictions, we analyze the most frequently attributed words across each dataset. For each sample, we aggregate the attribution scores across all modalities at the word level and select the top 1\% of word tokens by attribution magnitude. We then compute the frequency of each word appearing in this top 1\% across all samples in the dataset and report the most frequent words per class. For \cls{}, frequencies are reported per document class; for \ser{}, per entity type; and for \qa{}, separately for the start and end tokens of the predicted answer.
Fig.~\ref{fig:topk_wordfreq} presents the results for the \layoutlm{} model on five datasets, showing the most frequently attributed words for up to six randomly selected labels per dataset.

It is noticeable that across all datasets and tasks, stop words (labeled as \texttt{STOP}) dominate the attributions. While in some cases stop words may carry genuine structural or contextual cues relevant to the prediction (e.g., to serve as contextual separators between entities), their widespread and uniform dominance across all classes and entity types suggests that this may largely be driven by their high frequency in the data.
Interestingly, some task-relevant words do emerge across the datasets.
For instance, on \rvlcdip{}, words such as \texttt{Mr.}, \texttt{Dear}, and \texttt{Dr.} appear frequently for the letter class, while \texttt{University}, \texttt{Department}, and \texttt{Vol.} are among the most attributed words for scientific publications. Notably, words such as \texttt{Tobacco}, \texttt{Cigarette}, and \texttt{Healthy} appear consistently across multiple classes, despite having no apparent relevance to the class labels themselves.
While their prevalence in the dataset is unsurprising given its tobacco industry origins, their high attribution across unrelated classes suggests the model may be relying on dataset-specific artifacts rather than task-relevant features.

On \doclaynet{}, various forms of the word \texttt{Contract} appear consistently for the \texttt{government\_tenders} class, which aligns well with the class semantics.
For \texttt{financial\_reports}, the words \texttt{December} and \texttt{31.} appear with high frequency despite having no direct relevance to the class, alongside more plausible words such as \texttt{Company} and \texttt{Financial}.
For the manuals class, specific abbreviations such as \texttt{PDF}, \texttt{IVM}, and \texttt{AFP} emerge as high-frequency top-$k$ words, though whether these reflect genuine class-relevant patterns or spurious artifacts requires closer inspection of the dataset.

On \funsd{}, words such as \texttt{FAX}, \texttt{DATE}, \texttt{Account}, and \texttt{Facsimile} appear consistently across all entity types. On \cord{}, words such as \texttt{PCS}, \texttt{TOTAL}, and \texttt{SUBTOTAL} dominate the top-$k$ attributions, all of which appear relevant to the receipt domain. However, specific numbers and percentages such as \texttt{50,000} or \texttt{10\%} also appear with high frequency for some entity types, which may reflect a bias in the model's decision-making driven by their prevalence in the training data.
On \docvqa{}, the attributions for both start and end answer tokens are largely dominated by stop words. However, the words \texttt{source} and \texttt{number} also appear frequently among the top-$k$ attributed words.

\section{Discussion}
\label{sec:discussion}
Explainable AI has received considerable attention in deep learning over the past decade; however, its extension to the multimodal document understanding domain has remained largely unexplored.
By providing a systematic evaluation framework and a rigorous benchmark across multiple models, methods, datasets, and evaluation metrics, this study takes a significant step forward in improving the explainability of multimodal document understanding models and highlighting the challenges associated with existing methods. In the following, we discuss the key findings of this work, its limitations, broader impact, and future directions.

\subsection{Key Findings}

\subheading{Top-performing XAI methods.}
Our results showed that the majority of XAI methods, particularly basic backpropagation methods, consistently underperformed across all tasks and evaluation criteria.
Among the 13 methods evaluated, only three from different families emerged as the strongest: \occlusionf{} (perturbation-based), \rawattnf{} (attention-based), and \integratedgradientsf{} (gradient-based).
However, when considering both quantitative and qualitative evaluation jointly, only \occlusionf{} stood out as the most reliable overall.
Unlike \integratedgradientsf{}, which tended to spread importance across the entire document and produce high complexity attribution maps, \occlusionf{} produced concise and interpretable modality-wise contributions. Furthermore, \rawattnf{} can only attribute importance to the modalities explicitly represented as separate tokens in the model, a limitation that does not apply to \occlusionf{} due to its model-agnostic nature.

\subheading{Computational trade-offs of methods.}
Despite its strengths, \occlusionf{} remains one of the most expensive methods to compute, since it requires perturbations over sliding windows, and therefore its computational cost scales proportionally to the size of the input. In multi-target scenarios, however, this overhead can be effectively reduced due to shared perturbations over targets. \integratedgradientsf{}, on the other hand, is expensive in both single- and multi-target scenarios, as even with our multi-target adaptation, it only gained modest speedups of up to $1.5$--$1.7\times$. In this aspect, \rawattnf{} can be considered a particularly strong choice, as, despite its limitations, it requires no additional computation other than a forward pass for both single- and multi-target scenarios.

\subheading{Completeness of methods.}
Another important consideration is the Completeness Axiom, as it allows us to evaluate the exact contribution of each feature towards the output. However, neither \occlusionf{} nor \rawattnf{} satisfy completeness. \integratedgradientsf{} does satisfy the Completeness Axiom, but as we observed in our experiments, its high complexity made the results difficult to interpret.
Another method that satisfies Completeness is \kernelshapf{}; however, it showed considerably poorer performance on all other evaluation criteria in our experiments. This is expected, since \kernelshapf{} approximates Shapley values by fitting a surrogate model over a sampled subset of feature coalitions from a space of $2^n$ possible coalitions. In our setting, $n$ can easily grow large (up to hundreds or thousands of features) when accounting for separate text, position, layout, and image features, making reliable estimation of the Shapley values infeasible with a limited number of perturbation samples. Increasing the number of perturbations may help, but it incurs impractical computational costs. Note that these limitations apply equally to \limef{}.

\subheading{Metric reliability and agreement.}
Our analysis of the evaluation metrics revealed two important findings. First, for many XAI evaluation metrics, different methods showed statistically indistinguishable results when observed across datasets, models, and tasks. Therefore, practitioners should exercise caution when ranking one method over another, and we recommend evaluating methods across multiple datasets and models before drawing conclusions about their suitability.
Second, we observed moderate to strong agreement between several faithfulness metrics, including \aopc{}, \faithfulnessest{}, \abpc{}, \faithfulnesscorr{}, \sufficiency{}, and \sensitivityn{}.
While each of these metrics measures a different aspect, their mutual agreement can help reinforce confidence in their results.
Conversely, neither \monotonicityf{} nor \infidelityf{} showed a positive correlation with any other metric, and upon deeper inspection, we found the results of both of these metrics to be unreliable for evaluating the methods.
The complexity metrics also showed moderate mutual correlation among \entropycomplexity{}, \sparseness{}, and \sundararajancomplexity{}, while \effectivecomplexity{} showed only negligible correlation with the other three. This is expected, since \effectivecomplexity{} is computed over perturbed inputs, unlike the other complexity metrics which are computed directly on the generated explanations.

\subheading{Modality reliance across tasks and labels.}
Our modality contribution analysis of the \layoutlm{} model revealed variable reliance on individual modalities across tasks, datasets, and individual labels. For instance, on classification tasks, \occlusionf{} indicated a stronger reliance on the image modality, whereas on SER tasks, it indicated that text and layout were the primary contributors. Within datasets, label-level variation was also observed (for example, higher contribution of text modality for \texttt{government\_tenders} vs. image for \texttt{financial\_reports}). Such insights can help optimize, debug, and validate model behavior. For example, our results indicate that the image modality contributes negligibly to the SER task. Therefore, the image modality could potentially be removed as an additional input for this task to optimize inference costs. On the other hand, if strong reliance on the image modality were expected for a particular task, modality contribution analysis could serve as a means of verifying whether the model is indeed utilizing it.

\subheading{Spatial attribution locality patterns.}
Our spatial locality analysis of the \layoutlm{} model also revealed several insights into which spatial regions the model attends to for different entity types. For example, on the \sroie{} dataset, \occlusionf{} indicated greater focus on the horizontal neighborhood of target entities, especially for entity types \texttt{ADDRESS}, \texttt{DATE}, and \texttt{COMPANY}. This aligns with our expectation, as these entities are typically horizontally spread out in a document, and therefore, the model's focus on horizontally aligned tokens seems reasonable. On the other hand, for many entity types, \rawattnf{} indicated focus only on the target token itself. Such entity-level insights can be particularly useful for detecting unexpected model behavior. For instance, if a model were found to focus on spatially distant and irrelevant tokens for a given entity type, this could signal a potential spurious correlation that warrants further investigation.

\subheading{Word frequency and dataset biases.}
Our top-$k$ word frequency analysis revealed that stop words dominated the attributions across all tasks and datasets. While in the \ser{} and \qa{} tasks, this is plausible as stop words frequently appear in the neighborhood of target entities and may carry contextual relevance, the dominance of stop words in the \cls{} task may reflect a frequency bias instead, as stop words should hold no class-specific relevance. Besides stop words, we also noted a mix of class-relevant and potentially spurious words emerging across the datasets (e.g., tobacco-related terms across multiple classes in \rvlcdip{}, or specific dates for \texttt{financial\_reports} in \doclaynet{}), suggesting potential dataset biases that warrant deeper inspection. Note that the word frequency analysis can be further strengthened when combined with the modality contribution analysis. For instance, if the model relies heavily on text modality for a given class, the word frequency analysis can pinpoint which specific words drive the predictions. Conversely, if the text modality receives negligible importance, the presence of spurious words among the top attributions is less concerning, as the model's predictions are not primarily driven by text.

\subheading{Limited consensus across methods.}
Despite the utility of the above analyses, it is important to note that the three top-ranking methods showed limited consensus in their attributions, leading to different interpretations depending on the method used. Therefore, to draw reliable insights from these analyses, one should ideally rely on a method with established faithfulness. In our evaluation, \occlusionf{} consistently ranked among the top methods on this criterion and can therefore serve as a reasonable choice, though this also highlights the need for developing more faithful domain-specific XAI methods for multimodal document understanding.

\subsection{Limitations}
\label{sec:limitations}
While our study presents a comprehensive evaluation of existing XAI methods in the context of multimodal document understanding, there are some limitations that warrant discussion. First, we evaluate all methods and metrics only on four models: \bert{}, \roberta{}, \lilt{}, and \layoutlm{}, of which only two are document-specific multimodal models and only one processes all four modalities. This limitation arises primarily from the lack of open-source model architectures and pretrained weights. Several existing works that employ significantly different fusion strategies, such as DocFormer and TiLT, do not release their model weights, and reproducing them is non-trivial, as each requires complex pretraining routines that are also closed source. Secondly, all evaluated models are Transformer-based, and our findings may not generalize to other model families. That said, our framework is directly extensible to such architectures, as it treats the model as a black box, independent of the specific modalities over which attributions are computed and how they are aggregated for evaluation.

\subsection{Broader Impact}
\label{sec:broader_impact}
Despite the limitations, our work makes several contributions that extend beyond the immediate scope of this study.
As part of this work, we release code for all XAI method and evaluation metric adaptations presented in this study, along with a unified evaluation pipeline that ensures consistent metric computation across criteria.
We believe this will lower the barrier for researchers and practitioners to apply and evaluate explainability methods at scale in both single- and multi-target settings.
Furthermore, while this work focuses on multimodal document understanding, the multimodal and multi-target adaptations and evaluation suite are task-agnostic and can be directly applied to other domains. Overall, as multimodal models are increasingly deployed in high-stakes sectors such as finance, legal, and healthcare, where transparency is often legally required, our framework can serve as both a benchmark for future research and a practical guide for evaluating the explainability of such models.

\subsection{Future Work}
\label{sec:future_work}
While our results provide a first comprehensive picture, several directions remain open for future work.
First, while Occlusion emerged as the most faithful and interpretable method in our analysis, it does not satisfy completeness. KernelSHAP, which does satisfy completeness, could serve as an alternative; however, it performed considerably worse in our setting due to the large multimodal feature space.
To address this, one direction worth exploring is hierarchical decomposition of Shapley values over nested coalition structures~\cite{Owen1977}. Rather than computing Shapley values over the full set of individual features, such an approach would decompose the attribution process into multiple stages (e.g., attributing importance to blocks, then lines, and finally individual words and word modalities), reducing $n$ to a tractable number at each stage. This could preserve completeness guarantees while simultaneously producing interpretable explanations at each level of the hierarchy.
Second, while our framework considers modality-wise attributions, capturing cross-modal interactions, as investigated in DIME~\cite{lyu_dime_2022}, remains an open problem in the document domain.
Third, while we explored several attribution methods in this work, a fundamental limitation remains in that they reveal \textit{which} features the model relies on but not \textit{why}. In this regard, exploring (generative) counterfactual explanations~\cite{verma_counterfactual_2020} for multimodal document understanding may be a promising direction for uncovering the underlying causal factors.
Finally, while we proposed multiple dataset-wide analyses to obtain global insights, developing additional domain-specific diagnostic approaches and dedicated visualization interfaces, especially for multi-target tasks, would further extend the framework's utility.

\section{Conclusion}
\label{sec:conclusion}
In this work, we presented \ourframework{}, the first comprehensive framework for generating and evaluating explanations in multimodal document understanding. Using our framework, we systematically evaluated 13 XAI methods across 3 tasks, 8 datasets, 4 models, and 17 evaluation metrics, and found that only a few methods perform consistently across all criteria, with \occlusion{} emerging as the most faithful and interpretable overall. Our meta-evaluation of metrics showed moderate to strong agreement among several faithfulness and complexity metrics, while identifying Monotonicity and Infidelity as unreliable. Furthermore, through multiple dataset-wide analyses, we uncovered several insights into model behavior, including variable modality reliance across tasks and labels, spatial attribution patterns around target entities, and potential dataset biases.
Overall, we believe our findings, together with the publicly released framework, will serve as a valuable reference for future research on XAI evaluation,  especially in the multimodal document understanding context.

\backmatter
\bmhead{Acknowledgements}
We thank Adriano Lucieri for helpful discussions and feedback on this work.

\bibliography{sn-bibliography.bib}%

\begin{thebibliography}{10}
\expandafter\ifx\csname url\endcsname\relax
  \def\url#1{\burl{#1}}\fi
\expandafter\ifx\csname urlprefix\endcsname\relax\def\urlprefix{URL }\fi
\providecommand{\bibinfo}[2]{#2}
\providecommand{\eprint}[2][]{\url{#2}}
\providecommand{\doi}[1]{\url{https://doi.org/#1}}
\bibcommenthead

\bibitem{ferrando_improving_2020}
\bibinfo{author}{Ferrando, J.} \emph{et~al.}
\newblock \bibinfo{editor}{Krzhizhanovskaya, V.~V.} \emph{et~al.} (eds)
  \emph{\bibinfo{title}{Improving {Accuracy} and {Speeding} {Up} {Document}
  {Image} {Classification} {Through} {Parallel} {Systems}}}.
\newblock (eds \bibinfo{editor}{Krzhizhanovskaya, V.~V.} \emph{et~al.})
  \emph{\bibinfo{booktitle}{Computational {Science} – {ICCS} 2020}},
  \bibinfo{pages}{387--400} (\bibinfo{publisher}{Springer International
  Publishing}, \bibinfo{address}{Cham}, \bibinfo{year}{2020}).

\bibitem{huang_layoutlmv3_2022}
\bibinfo{author}{Huang, Y.}, \bibinfo{author}{Lv, T.}, \bibinfo{author}{Cui,
  L.}, \bibinfo{author}{Lu, Y.} \& \bibinfo{author}{Wei, F.}
\newblock \emph{\bibinfo{title}{{LayoutLMv3}: {Pre}-{Training} for {Document}
  {AI} with {Unified} {Text} and {Image} {Masking}}}, {MM} '22,
  \bibinfo{pages}{4083--4091} (\bibinfo{publisher}{Association for Computing
  Machinery}, \bibinfo{address}{New York, NY, USA}, \bibinfo{year}{2022}).
\newblock \urlprefix\url{https://doi.org/10.1145/3503161.3548112}.
\newblock \bibinfo{note}{Event-place: Lisboa, Portugal}.

\bibitem{xu_layoutlm_2020}
\bibinfo{author}{Xu, Y.} \emph{et~al.}
\newblock \bibinfo{editor}{Gupta, R.}, \bibinfo{editor}{Liu, Y.},
  \bibinfo{editor}{Tang, J.} \& \bibinfo{editor}{Prakash, B.~A.} (eds)
  \emph{\bibinfo{title}{{LayoutLM}: pre-training of text and layout for
  document image understanding}}.
\newblock (eds \bibinfo{editor}{Gupta, R.}, \bibinfo{editor}{Liu, Y.},
  \bibinfo{editor}{Tang, J.} \& \bibinfo{editor}{Prakash, B.~A.})
  \emph{\bibinfo{booktitle}{Proc. 26th {ACM} {SIGKDD} {Conference} on
  {Knowledge} {Discovery} and {Data} {Mining}}}, \bibinfo{pages}{1192--1200}
  (\bibinfo{address}{virtual}, \bibinfo{year}{2020}).

\bibitem{xu_layoutlmv2_2021}
\bibinfo{author}{Xu, Y.} \emph{et~al.}
\newblock \emph{\bibinfo{title}{{LayoutLMv2}: {Multi}-modal {Pre}-training for
  {Visually}-rich {Document} {Understanding}}}, \bibinfo{pages}{2579--2591}
  (\bibinfo{publisher}{Association for Computational Linguistics (ACL)},
  \bibinfo{year}{2021}).
\newblock \bibinfo{note}{\_eprint: 2012.14740}.

\bibitem{borchmann_due_2021}
\bibinfo{author}{Borchmann, L.} \emph{et~al.}
\newblock \bibinfo{editor}{Vanschoren, J.} \& \bibinfo{editor}{Yeung, S.-K.}
  (eds) \emph{\bibinfo{title}{{DUE}: end-to-end document understanding
  benchmark}}.
\newblock (eds \bibinfo{editor}{Vanschoren, J.} \& \bibinfo{editor}{Yeung,
  S.-K.}) \emph{\bibinfo{booktitle}{{NeurIPS} {Datasets} and {Benchmarks}}},
  Vol.~\bibinfo{volume}{1} (\bibinfo{address}{virtual}, \bibinfo{year}{2021}).

\bibitem{wang_document_2025}
\bibinfo{author}{Wang, W.} \emph{et~al.}
\newblock \bibinfo{title}{Document intelligence in the era of large language
  models: A survey} (\bibinfo{year}{2025}).
\newblock \urlprefix\url{https://arxiv.org/abs/2510.13366}.
\newblock
  \bibinfo{eprint}{{\href{https://arxiv.org/abs/2510.13366}{{arXiv:2510.13366}}}}.

\bibitem{shen_layoutparser_2021}
\bibinfo{author}{Shen, Z.} \emph{et~al.}
\newblock \bibinfo{editor}{Lladós, J.}, \bibinfo{editor}{Lopresti, D.} \&
  \bibinfo{editor}{Uchida, S.} (eds) \emph{\bibinfo{title}{{LayoutParser}: {A}
  {Unified} {Toolkit} for {Deep} {Learning} {Based} {Document} {Image}
  {Analysis}}}.
\newblock (eds \bibinfo{editor}{Lladós, J.}, \bibinfo{editor}{Lopresti, D.} \&
  \bibinfo{editor}{Uchida, S.}) \emph{\bibinfo{booktitle}{Document {Analysis}
  and {Recognition} – {ICDAR} 2021}}, \bibinfo{pages}{131--146}
  (\bibinfo{publisher}{Springer International Publishing},
  \bibinfo{address}{Cham}, \bibinfo{year}{2021}).

\bibitem{powalski_going_2021}
\bibinfo{author}{Powalski, R.} \emph{et~al.}
\newblock \bibinfo{editor}{Lladós, J.}, \bibinfo{editor}{Lopresti, D.} \&
  \bibinfo{editor}{Uchida, S.} (eds) \emph{\bibinfo{title}{Going {Full}-{TILT}
  {Boogie} on {Document} {Understanding} with {Text}-{Image}-{Layout}
  {Transformer}}}.
\newblock (eds \bibinfo{editor}{Lladós, J.}, \bibinfo{editor}{Lopresti, D.} \&
  \bibinfo{editor}{Uchida, S.}) \emph{\bibinfo{booktitle}{Document {Analysis}
  and {Recognition} – {ICDAR} 2021}}, \bibinfo{pages}{732--747}
  (\bibinfo{publisher}{Springer International Publishing},
  \bibinfo{address}{Cham}, \bibinfo{year}{2021}).

\bibitem{garncarek_lambert_2021}
\bibinfo{author}{Garncarek, L.} \emph{et~al.}
\newblock \bibinfo{title}{ in \textit{{LAMBERT}: layout-aware language modeling
  for information extraction}} (eds \bibinfo{editor}{Lladós, J.},
  \bibinfo{editor}{Lopresti, D.} \& \bibinfo{editor}{Uchida, S.})
  \emph{\bibinfo{booktitle}{Document analysis and recognition - {ICDAR} 2021}},
  Vol. \bibinfo{volume}{12821} of \emph{\bibinfo{series}{{LNCS}}}
  \bibinfo{pages}{532--547} (\bibinfo{publisher}{Springer},
  \bibinfo{address}{Cham}, \bibinfo{year}{2021}).

\bibitem{wang_lilt_2022}
\bibinfo{author}{Wang, J.}, \bibinfo{author}{Jin, L.} \& \bibinfo{author}{Ding,
  K.}
\newblock \bibinfo{editor}{Muresan, S.}, \bibinfo{editor}{Nakov, P.} \&
  \bibinfo{editor}{Villavicencio, A.} (eds) \emph{\bibinfo{title}{{LiLT}: a
  simple yet effective language-independent layout transformer for structured
  document understanding}}.
\newblock (eds \bibinfo{editor}{Muresan, S.}, \bibinfo{editor}{Nakov, P.} \&
  \bibinfo{editor}{Villavicencio, A.}) \emph{\bibinfo{booktitle}{{ACL}}},
  \bibinfo{pages}{7747--7757} (\bibinfo{address}{Dublin, Ireland},
  \bibinfo{year}{2022}).

\bibitem{lamott_slimdoc_2025}
\bibinfo{author}{Lamott, M.}, \bibinfo{author}{Shakir, M.~A.},
  \bibinfo{author}{Ulges, A.}, \bibinfo{author}{Weweler, Y.-N.} \&
  \bibinfo{author}{Shafait, F.}
\newblock \bibinfo{title}{{SlimDoc}: lightweight distillation of document
  transformer models}.
\newblock \emph{\bibinfo{journal}{Ijdar}} \textbf{\bibinfo{volume}{28}},
  \bibinfo{pages}{457--473} (\bibinfo{year}{2025}).

\bibitem{bibal_legal_2021}
\bibinfo{author}{Bibal, A.}, \bibinfo{author}{Lognoul, M.},
  \bibinfo{author}{De~Streel, A.} \& \bibinfo{author}{Frénay, B.}
\newblock \bibinfo{title}{Legal requirements on explainability in machine
  learning}.
\newblock \emph{\bibinfo{journal}{Artificial Intelligence and Law}}
  \textbf{\bibinfo{volume}{29}}, \bibinfo{pages}{149--169}
  (\bibinfo{year}{2021}).
\newblock
  \urlprefix\url{https://www.scopus.com/pages/publications/85088816026}.

\bibitem{perc_social_2019}
\bibinfo{author}{Perc, M.}, \bibinfo{author}{Ozer, M.} \&
  \bibinfo{author}{Hojnik, J.}
\newblock \bibinfo{title}{Social and juristic challenges of artificial
  intelligence}.
\newblock \emph{\bibinfo{journal}{Palgrave Communications}}
  \textbf{\bibinfo{volume}{5}}, \bibinfo{pages}{61} (\bibinfo{year}{2019}).
\newblock \urlprefix\url{https://www.nature.com/articles/s41599-019-0278-x}.
\newblock \bibinfo{note}{Publisher: Palgrave}.

\bibitem{lundberg_consistent_2019}
\bibinfo{author}{Lundberg, S.~M.}, \bibinfo{author}{Erion, G.~G.} \&
  \bibinfo{author}{Lee, S.-I.}
\newblock \bibinfo{title}{Consistent {Individualized} {Feature} {Attribution}
  for {Tree} {Ensembles}} (\bibinfo{year}{2019}).
\newblock \bibinfo{note}{\_eprint: 1802.03888}.

\bibitem{ribeiro_why_2016}
\bibinfo{author}{Ribeiro, M.}, \bibinfo{author}{Singh, S.} \&
  \bibinfo{author}{Guestrin, C.}
\newblock \bibinfo{editor}{DeNero, J.}, \bibinfo{editor}{Finlayson, M.} \&
  \bibinfo{editor}{Reddy, S.} (eds) \emph{\bibinfo{title}{“{Why} {Should} {I}
  {Trust} {You}?”: {Explaining} the {Predictions} of {Any} {Classifier}}}.
\newblock (eds \bibinfo{editor}{DeNero, J.}, \bibinfo{editor}{Finlayson, M.} \&
  \bibinfo{editor}{Reddy, S.}) \emph{\bibinfo{booktitle}{Proceedings of the
  2016 {Conference} of the {North} {American} {Chapter} of the {Association}
  for {Computational} {Linguistics}: {Demonstrations}}},
  \bibinfo{pages}{97--101} (\bibinfo{publisher}{Association for Computational
  Linguistics}, \bibinfo{address}{San Diego, California},
  \bibinfo{year}{2016}).
\newblock \urlprefix\url{https://aclanthology.org/N16-3020}.
\newblock \bibinfo{note}{ArXiv:1602.04938 [cs]}.

\bibitem{samek_explainable_2017}
\bibinfo{author}{Samek, W.}, \bibinfo{author}{Wiegand, T.} \&
  \bibinfo{author}{Müller, K.-R.}
\newblock \bibinfo{title}{Explainable {Artificial} {Intelligence}:
  {Understanding}, {Visualizing} and {Interpreting} {Deep} {Learning} {Models}}
  (\bibinfo{year}{2017}).
\newblock \urlprefix\url{http://arxiv.org/abs/1708.08296}.
\newblock \bibinfo{note}{ArXiv:1708.08296 [cs]}.

\bibitem{adebayo_sanity_2018}
\bibinfo{author}{Adebayo, J.} \emph{et~al.}
\newblock \bibinfo{editor}{Bengio, S.} \emph{et~al.} (eds)
  \emph{\bibinfo{title}{Sanity {Checks} for {Saliency} {Maps}}}.
\newblock (eds \bibinfo{editor}{Bengio, S.} \emph{et~al.})
  \emph{\bibinfo{booktitle}{Advances in {Neural} {Information} {Processing}
  {Systems}}}, Vol.~\bibinfo{volume}{31} (\bibinfo{publisher}{Curran
  Associates, Inc.}, \bibinfo{year}{2018}).
\newblock
  \urlprefix\url{https://proceedings.neurips.cc/paper_files/paper/2018/file/294a8ed24b1ad22ec2e7efea049b8737-Paper.pdf}.

\bibitem{yeh_fidelity_2019}
\bibinfo{author}{Yeh, C.-K.}, \bibinfo{author}{Hsieh, C.-Y.},
  \bibinfo{author}{Suggala, A.}, \bibinfo{author}{Inouye, D.~I.} \&
  \bibinfo{author}{Ravikumar, P.~K.}
\newblock \bibinfo{editor}{Wallach, H.} \emph{et~al.} (eds)
  \emph{\bibinfo{title}{On the ({In})fidelity and {Sensitivity} of
  {Explanations}}}.
\newblock (eds \bibinfo{editor}{Wallach, H.} \emph{et~al.})
  \emph{\bibinfo{booktitle}{Advances in {Neural} {Information} {Processing}
  {Systems}}}, Vol.~\bibinfo{volume}{32} (\bibinfo{publisher}{Curran
  Associates, Inc.}, \bibinfo{year}{2019}).
\newblock
  \urlprefix\url{https://proceedings.neurips.cc/paper_files/paper/2019/file/a7471fdc77b3435276507cc8f2dc2569-Paper.pdf}.

\bibitem{vilone_explainable_2020}
\bibinfo{author}{Vilone, G.} \& \bibinfo{author}{Longo, L.}
\newblock \bibinfo{title}{Explainable artificial intelligence: a systematic
  review}.
\newblock \emph{\bibinfo{journal}{arXiv preprint arXiv:2006.00093}}
  (\bibinfo{year}{2020}).
\newblock \urlprefix\url{http://arxiv.org/abs/2006.00093}.
\newblock \bibinfo{note}{ArXiv:2006.00093 [cs]}.

\bibitem{ras_explainable_2022}
\bibinfo{author}{Ras, G.}, \bibinfo{author}{Xie, N.}, \bibinfo{author}{van
  Gerven, M.} \& \bibinfo{author}{Doran, D.}
\newblock \bibinfo{title}{Explainable {Deep} {Learning}: {A} {Field} {Guide}
  for the {Uninitiated}}.
\newblock \emph{\bibinfo{journal}{J. Artif. Int. Res.}}
  \textbf{\bibinfo{volume}{73}}, \bibinfo{pages}{329–397}
  (\bibinfo{year}{2022}).
\newblock \urlprefix\url{https://doi.org/10.1613/jair.1.13200}.
\newblock \bibinfo{note}{Place: El Segundo, CA, USA Publisher: AI Access
  Foundation}.

\bibitem{Danilevsky2020}
\bibinfo{author}{Danilevsky, M.} \emph{et~al.}
\newblock \emph{\bibinfo{title}{A survey of the state of explainable ai for
  natural language processing}}, \bibinfo{pages}{447–459}
  (\bibinfo{publisher}{Association for Computational Linguistics},
  \bibinfo{year}{2020}).
\newblock \urlprefix\url{http://dx.doi.org/10.18653/v1/2020.aacl-main.46}.
\newblock \bibinfo{note}{ArXiv:2010.00711 [cs]}.

\bibitem{dembinsky2025unifying}
\bibinfo{author}{Dembinsky, D.} \emph{et~al.}
\newblock \bibinfo{title}{Unifying vxai: a systematic review and framework for
  the evaluation of explainable ai}.
\newblock \emph{\bibinfo{journal}{arXiv preprint arXiv:2506.15408}}
  (\bibinfo{year}{2025}).

\bibitem{appalaraju_docformer_2021}
\bibinfo{author}{Appalaraju, S.}, \bibinfo{author}{Jasani, B.},
  \bibinfo{author}{Kota, B.~U.}, \bibinfo{author}{Xie, Y.} \&
  \bibinfo{author}{Manmatha, R.}
\newblock \emph{\bibinfo{title}{{DocFormer}: end-to-end transformer for
  document understanding}}, \bibinfo{pages}{993--1003}
  (\bibinfo{address}{virtual}, \bibinfo{year}{2021}).

\bibitem{lundberg_unified_2017}
\bibinfo{author}{Lundberg, S.~M.} \& \bibinfo{author}{Lee, S.-I.}
\newblock \emph{\bibinfo{title}{A {Unified} {Approach} to {Interpreting}
  {Model} {Predictions}}}, {NIPS}'17, \bibinfo{pages}{4768--4777}
  (\bibinfo{publisher}{Curran Associates Inc.}, \bibinfo{address}{Red Hook, NY,
  USA}, \bibinfo{year}{2017}).
\newblock \bibinfo{note}{Event-place: Long Beach, California, USA}.

\bibitem{klein_navigating_nodate}
\bibinfo{author}{Klein, L.} \emph{et~al.}
\newblock \emph{\bibinfo{title}{Navigating the {Maze} of {Explainable} {AI}:
  {A} {Systematic} {Approach} to {Evaluating} {Methods} and {Metrics}}},
  Vol.~\bibinfo{volume}{37} (\bibinfo{publisher}{Curran Associates, Inc.},
  \bibinfo{year}{2024}).

\bibitem{tomsett_sanity_2020}
\bibinfo{author}{Tomsett, R.}, \bibinfo{author}{Harborne, D.},
  \bibinfo{author}{Chakraborty, S.}, \bibinfo{author}{Gurram, P.} \&
  \bibinfo{author}{Preece, A.}
\newblock \bibinfo{title}{Sanity {Checks} for {Saliency} {Metrics}}.
\newblock \emph{\bibinfo{journal}{Proceedings of the AAAI Conference on
  Artificial Intelligence}} \textbf{\bibinfo{volume}{34}},
  \bibinfo{pages}{6021--6029} (\bibinfo{year}{2020}).

\bibitem{zeiler_visualizing_2014}
\bibinfo{author}{Zeiler, M.~D.} \& \bibinfo{author}{Fergus, R.}
\newblock \bibinfo{editor}{Fleet, D.}, \bibinfo{editor}{Pajdla, T.},
  \bibinfo{editor}{Schiele, B.} \& \bibinfo{editor}{Tuytelaars, T.} (eds)
  \emph{\bibinfo{title}{Visualizing and {Understanding} {Convolutional}
  {Networks}}}.
\newblock (eds \bibinfo{editor}{Fleet, D.}, \bibinfo{editor}{Pajdla, T.},
  \bibinfo{editor}{Schiele, B.} \& \bibinfo{editor}{Tuytelaars, T.})
  \emph{\bibinfo{booktitle}{Computer {Vision} – {ECCV} 2014}},
  \bibinfo{pages}{818--833} (\bibinfo{publisher}{Springer International
  Publishing}, \bibinfo{address}{Cham}, \bibinfo{year}{2014}).

\bibitem{arya2019explanationdoesfitall}
\bibinfo{author}{Arya, V.} \emph{et~al.}
\newblock \bibinfo{title}{One explanation does not fit all: A toolkit and
  taxonomy of ai explainability techniques} (\bibinfo{year}{2019}).
\newblock \urlprefix\url{https://arxiv.org/abs/1909.03012}.
\newblock
  \bibinfo{eprint}{{\href{https://arxiv.org/abs/1909.03012}{{arXiv:1909.03012}}}}.

\bibitem{wang_docllm_2024}
\bibinfo{author}{Wang, D.} \emph{et~al.}
\newblock \bibinfo{editor}{Ku, L.-W.}, \bibinfo{editor}{Martins, A.} \&
  \bibinfo{editor}{Srikumar, V.} (eds) \emph{\bibinfo{title}{{D}oc{LLM}: A
  layout-aware generative language model for multimodal document
  understanding}}.
\newblock (eds \bibinfo{editor}{Ku, L.-W.}, \bibinfo{editor}{Martins, A.} \&
  \bibinfo{editor}{Srikumar, V.}) \emph{\bibinfo{booktitle}{Proceedings of the
  62nd Annual Meeting of the Association for Computational Linguistics (Volume
  1: Long Papers)}}, \bibinfo{pages}{8529--8548}
  (\bibinfo{publisher}{Association for Computational Linguistics},
  \bibinfo{address}{Bangkok, Thailand}, \bibinfo{year}{2024}).
\newblock \urlprefix\url{https://aclanthology.org/2024.acl-long.463/}.

\bibitem{fujitake-2024-layoutllm}
\bibinfo{author}{Fujitake, M.}
\newblock \bibinfo{editor}{Calzolari, N.} \emph{et~al.} (eds)
  \emph{\bibinfo{title}{{L}ayout{LLM}: Large language model instruction tuning
  for visually rich document understanding}}.
\newblock (eds \bibinfo{editor}{Calzolari, N.} \emph{et~al.})
  \emph{\bibinfo{booktitle}{Proceedings of the 2024 Joint International
  Conference on Computational Linguistics, Language Resources and Evaluation
  (LREC-COLING 2024)}}, \bibinfo{pages}{10219--10224} (\bibinfo{publisher}{ELRA
  and ICCL}, \bibinfo{address}{Torino, Italia}, \bibinfo{year}{2024}).
\newblock \urlprefix\url{https://aclanthology.org/2024.lrec-main.892/}.

\bibitem{vaswani_attention_2017}
\bibinfo{author}{Vaswani, A.} \emph{et~al.}
\newblock \bibinfo{editor}{Guyon, I.} \emph{et~al.} (eds)
  \emph{\bibinfo{title}{Attention is {All} you {Need}}}.
\newblock (eds \bibinfo{editor}{Guyon, I.} \emph{et~al.})
  \emph{\bibinfo{booktitle}{Advances in {Neural} {Information} {Processing}
  {Systems}}}, Vol.~\bibinfo{volume}{30} (\bibinfo{publisher}{Curran
  Associates, Inc.}, \bibinfo{year}{2017}).
\newblock
  \urlprefix\url{https://proceedings.neurips.cc/paper_files/paper/2017/file/3f5ee243547dee91fbd053c1c4a845aa-Paper.pdf}.

\bibitem{luo2023geolayoutlmgeometricpretrainingvisual}
\bibinfo{author}{Luo, C.}, \bibinfo{author}{Cheng, C.}, \bibinfo{author}{Zheng,
  Q.} \& \bibinfo{author}{Yao, C.}
\newblock \bibinfo{title}{Geolayoutlm: Geometric pre-training for visual
  information extraction} (\bibinfo{year}{2023}).
\newblock \urlprefix\url{https://arxiv.org/abs/2304.10759}.
\newblock
  \bibinfo{eprint}{{\href{https://arxiv.org/abs/2304.10759}{{arXiv:2304.10759}}}}.

\bibitem{simonyan_deep_2013}
\bibinfo{author}{Simonyan, K.}, \bibinfo{author}{Vedaldi, A.} \&
  \bibinfo{author}{Zisserman, A.}
\newblock \bibinfo{title}{Deep {Inside} {Convolutional} {Networks}:
  {Visualising} {Image} {Classification} {Models} and {Saliency} {Maps}}.
\newblock \emph{\bibinfo{journal}{CoRR}}
  \textbf{\bibinfo{volume}{abs/1312.6034}} (\bibinfo{year}{2013}).
\newblock \urlprefix\url{https://api.semanticscholar.org/CorpusID:1450294}.

\bibitem{springenberg_striving_2015}
\bibinfo{author}{Springenberg, J.~T.}, \bibinfo{author}{Dosovitskiy, A.},
  \bibinfo{author}{Brox, T.} \& \bibinfo{author}{Riedmiller, M.}
\newblock \emph{\bibinfo{title}{Striving for {Simplicity}: {The} {All}
  {Convolutional} {Net}}} (\bibinfo{year}{2015}).
\newblock
  \urlprefix\url{http://lmb.informatik.uni-freiburg.de/Publications/2015/DB15a}.

\bibitem{sundararajan_axiomatic_2017}
\bibinfo{author}{Sundararajan, M.}, \bibinfo{author}{Taly, A.} \&
  \bibinfo{author}{Yan, Q.}
\newblock \emph{\bibinfo{title}{Axiomatic attribution for deep networks}},
  \bibinfo{pages}{3319--3328} (\bibinfo{publisher}{PMLR},
  \bibinfo{year}{2017}).

\bibitem{shrikumar_learning_2017}
\bibinfo{author}{Shrikumar, A.}, \bibinfo{author}{Greenside, P.} \&
  \bibinfo{author}{Kundaje, A.}
\newblock \emph{\bibinfo{title}{Learning {Important} {Features} through
  {Propagating} {Activation} {Differences}}}, {ICML}'17,
  \bibinfo{pages}{3145--3153} (\bibinfo{publisher}{JMLR.org},
  \bibinfo{address}{Sydney, NSW, Australia}, \bibinfo{year}{2017}).

\bibitem{saifullah_docvce_2025}
\bibinfo{author}{Saifullah, S.}, \bibinfo{author}{Agne, S.},
  \bibinfo{author}{Dengel, A.} \& \bibinfo{author}{Ahmed, S.}
\newblock \bibinfo{title}{{DocVCE}: diffusion-based visual counterfactual
  explanations for document image classification} (\bibinfo{year}{2025}).
\newblock \urlprefix\url{http://arxiv.org/abs/2508.04233}.
\newblock \bibinfo{note}{ArXiv:2508.04233 [cs]}.

\bibitem{saifullah_docxclassifier_2024}
\bibinfo{author}{Saifullah, S.}, \bibinfo{author}{Agne, S.},
  \bibinfo{author}{Dengel, A.} \& \bibinfo{author}{Ahmed, S.}
\newblock \bibinfo{title}{{DocXclassifier}: towards a robust and interpretable
  deep neural network for document image classification}.
\newblock \emph{\bibinfo{journal}{International Journal on Document Analysis
  and Recognition (IJDAR)}} \textbf{\bibinfo{volume}{27}},
  \bibinfo{pages}{447--473} (\bibinfo{year}{2024}).
\newblock \urlprefix\url{https://link.springer.com/10.1007/s10032-024-00483-w}.

\bibitem{pouliquen2024weaklysupervisedtraininghologram}
\bibinfo{author}{Pouliquen, G.}, \bibinfo{author}{Chiron, G.},
  \bibinfo{author}{Chazalon, J.}, \bibinfo{author}{Géraud, T.} \&
  \bibinfo{author}{Awal, A.~M.}
\newblock \bibinfo{title}{Weakly supervised training for hologram verification
  in identity documents} (\bibinfo{year}{2024}).
\newblock \urlprefix\url{https://arxiv.org/abs/2404.17253}.
\newblock
  \bibinfo{eprint}{{\href{https://arxiv.org/abs/2404.17253}{{arXiv:2404.17253}}}}.

\bibitem{brini_end--end_2022}
\bibinfo{author}{Brini, I.}, \bibinfo{author}{Mehri, M.},
  \bibinfo{author}{Ingold, R.} \& \bibinfo{author}{Essoukri Ben~Amara, N.}
\newblock \bibinfo{editor}{Nguyen, N.~T.}, \bibinfo{editor}{Manolopoulos, Y.},
  \bibinfo{editor}{Chbeir, R.}, \bibinfo{editor}{Kozierkiewicz, A.} \&
  \bibinfo{editor}{Trawiński, B.} (eds) \emph{\bibinfo{title}{An
  {End}-to-{End} {Framework} for {Evaluating} {Explainable}
  {D}{\textless}b{\textgreater}{\textless}/b{\textgreater}eep {Models}:
  {Application} to {Historical} {Document} {Image} {Segmentation}}}.
\newblock (eds \bibinfo{editor}{Nguyen, N.~T.}, \bibinfo{editor}{Manolopoulos,
  Y.}, \bibinfo{editor}{Chbeir, R.}, \bibinfo{editor}{Kozierkiewicz, A.} \&
  \bibinfo{editor}{Trawiński, B.}) \emph{\bibinfo{booktitle}{Computational
  {Collective} {Intelligence}}}, \bibinfo{pages}{106--119}
  (\bibinfo{publisher}{Springer International Publishing},
  \bibinfo{address}{Cham}, \bibinfo{year}{2022}).

\bibitem{souibgui_docvxqa_2025}
\bibinfo{author}{Souibgui, M.~A.} \emph{et~al.}
\newblock \emph{\bibinfo{title}{{DocVXQA}: context-aware visual explanations
  for document question answering}} (\bibinfo{year}{2025}).
\newblock \urlprefix\url{https://openreview.net/forum?id=wex0vL4c2Y}.

\bibitem{saifullah_reality_2024}
\bibinfo{author}{{Saifullah}}, \bibinfo{author}{Agne, S.},
  \bibinfo{author}{Dengel, A.} \& \bibinfo{author}{Ahmed, S.}
\newblock \bibinfo{title}{The reality of high performing deep learning models:
  a case study on document image classification}.
\newblock \emph{\bibinfo{journal}{IEEE Access}} \textbf{\bibinfo{volume}{12}},
  \bibinfo{pages}{103537--103564} (\bibinfo{year}{2024}).
\newblock
  \urlprefix\url{https://ieeexplore.ieee.org/abstract/document/10591981}.

\bibitem{saifullah_docxplain_2024}
\bibinfo{author}{Saifullah, S.}, \bibinfo{author}{Agne, S.},
  \bibinfo{author}{Dengel, A.} \& \bibinfo{author}{Ahmed, S.}
\newblock \bibinfo{editor}{Barney~Smith, E.~H.}, \bibinfo{editor}{Liwicki, M.}
  \& \bibinfo{editor}{Peng, L.} (eds) \emph{\bibinfo{title}{{DocXplain}: a
  novel model-agnostic explainability method for document image
  classification}}.
\newblock (eds \bibinfo{editor}{Barney~Smith, E.~H.}, \bibinfo{editor}{Liwicki,
  M.} \& \bibinfo{editor}{Peng, L.}) \emph{\bibinfo{booktitle}{Document
  {Analysis} and {Recognition} - {ICDAR} 2024}}, \bibinfo{pages}{103--123}
  (\bibinfo{publisher}{Springer Nature Switzerland}, \bibinfo{address}{Cham},
  \bibinfo{year}{2024}).

\bibitem{goyal_towards_2016}
\bibinfo{author}{Goyal, Y.}, \bibinfo{author}{Mohapatra, A.},
  \bibinfo{author}{Parikh, D.} \& \bibinfo{author}{Batra, D.}
\newblock \bibinfo{title}{Towards {Transparent} {AI} {Systems}: {Interpreting}
  {Visual} {Question} {Answering} {Models}} (\bibinfo{year}{2016}).
\newblock \urlprefix\url{http://arxiv.org/abs/1608.08974}.
\newblock \bibinfo{note}{ArXiv:1608.08974 [cs]}.

\bibitem{lyu_dime_2022}
\bibinfo{author}{Lyu, Y.}, \bibinfo{author}{Liang, P.~P.},
  \bibinfo{author}{Deng, Z.}, \bibinfo{author}{Salakhutdinov, R.} \&
  \bibinfo{author}{Morency, L.-P.}
\newblock \bibinfo{title}{{DIME}: fine-grained interpretations of multimodal
  models via disentangled local explanations} (\bibinfo{year}{2022}).
\newblock \urlprefix\url{http://arxiv.org/abs/2203.02013}.
\newblock \bibinfo{note}{ArXiv:2203.02013 [cs]}.

\bibitem{aflalo_vl-interpret_2022}
\bibinfo{author}{Aflalo, E.} \emph{et~al.}
\newblock \bibinfo{title}{{VL}-{InterpreT}: {An} {Interactive} {Visualization}
  {Tool} for {Interpreting} {Vision}-{Language} {Transformers}}
  (\bibinfo{year}{2022}).
\newblock \urlprefix\url{http://arxiv.org/abs/2203.17247}.
\newblock \bibinfo{note}{ArXiv:2203.17247 [cs]}.

\bibitem{parcalabescu_mm-shap_2023}
\bibinfo{author}{Parcalabescu, L.} \& \bibinfo{author}{Frank, A.}
\newblock \emph{\bibinfo{title}{{MM}-{SHAP}: {A} {Performance}-agnostic
  {Metric} for {Measuring} {Multimodal} {Contributions} in {Vision} and
  {Language} {Models} \& {Tasks}}}, \bibinfo{pages}{4032--4059}
  (\bibinfo{year}{2023}).
\newblock \urlprefix\url{http://arxiv.org/abs/2212.08158}.
\newblock \bibinfo{note}{ArXiv:2212.08158 [cs]}.

\bibitem{wang_m2lens_2022}
\bibinfo{author}{Wang, X.} \emph{et~al.}
\newblock \bibinfo{title}{{M2Lens}: {Visualizing} and {Explaining} {Multimodal}
  {Models} for {Sentiment} {Analysis}}.
\newblock \emph{\bibinfo{journal}{IEEE Transactions on Visualization and
  Computer Graphics}} \textbf{\bibinfo{volume}{28}}, \bibinfo{pages}{802--812}
  (\bibinfo{year}{2022}).
\newblock \urlprefix\url{http://arxiv.org/abs/2107.08264}.
\newblock \bibinfo{note}{ArXiv:2107.08264 [cs]}.

\bibitem{hessel_does_2020}
\bibinfo{author}{Hessel, J.} \& \bibinfo{author}{Lee, L.}
\newblock \bibinfo{editor}{Webber, B.}, \bibinfo{editor}{Cohn, T.},
  \bibinfo{editor}{He, Y.} \& \bibinfo{editor}{Liu, Y.} (eds)
  \emph{\bibinfo{title}{Does my multimodal model learn cross-modal
  interactions? {It}'s harder to tell than you might think!}}
\newblock (eds \bibinfo{editor}{Webber, B.}, \bibinfo{editor}{Cohn, T.},
  \bibinfo{editor}{He, Y.} \& \bibinfo{editor}{Liu, Y.})
  \emph{\bibinfo{booktitle}{Proceedings of the 2020 {Conference} on {Empirical}
  {Methods} in {Natural} {Language} {Processing} ({EMNLP})}},
  \bibinfo{pages}{861--877} (\bibinfo{publisher}{Association for Computational
  Linguistics}, \bibinfo{address}{Online}, \bibinfo{year}{2020}).
\newblock \urlprefix\url{https://aclanthology.org/2020.emnlp-main.62/}.

\bibitem{harley_evaluation_2015}
\bibinfo{author}{Harley, A.~W.}, \bibinfo{author}{Ufkes, A.} \&
  \bibinfo{author}{Derpanis, K.~G.}
\newblock \emph{\bibinfo{title}{Evaluation of deep convolutional nets for
  document image classification and retrieval}}, \bibinfo{pages}{991--995}
  (\bibinfo{year}{2015}).

\bibitem{funsd}
\bibinfo{author}{Jaume, G.}, \bibinfo{author}{Kemal~Ekenel, H.} \&
  \bibinfo{author}{Thiran, J.-P.}
\newblock \emph{\bibinfo{title}{Funsd: A dataset for form understanding in
  noisy scanned documents}}, Vol.~\bibinfo{volume}{2}, \bibinfo{pages}{1--6}
  (\bibinfo{year}{2019}).

\bibitem{hase_out--distribution_2021}
\bibinfo{author}{Hase, P.}, \bibinfo{author}{Xie, H.} \&
  \bibinfo{author}{Bansal, M.}
\newblock \emph{\bibinfo{title}{The {Out}-of-{Distribution} {Problem} in
  {Explainability} and {Search} {Methods} for {Feature} {Importance}
  {Explanations}}}, Vol.~\bibinfo{volume}{34}, \bibinfo{pages}{3650--3666}
  (\bibinfo{publisher}{Curran Associates, Inc.}, \bibinfo{year}{2021}).
\newblock
  \urlprefix\url{https://papers.neurips.cc/paper_files/paper/2021/hash/1def1713ebf17722cbe300cfc1c88558-Abstract.html}.

\bibitem{nguyen2020quantitativeaspectsmodelinterpretability}
\bibinfo{author}{phi Nguyen, A.} \& \bibinfo{author}{Martínez, M.~R.}
\newblock \bibinfo{title}{On quantitative aspects of model interpretability}
  (\bibinfo{year}{2020}).
\newblock \urlprefix\url{https://arxiv.org/abs/2007.07584}.
\newblock
  \bibinfo{eprint}{{\href{https://arxiv.org/abs/2007.07584}{{arXiv:2007.07584}}}}.

\bibitem{samek_evaluating_2017}
\bibinfo{author}{Samek, W.}, \bibinfo{author}{Binder, A.},
  \bibinfo{author}{Montavon, G.}, \bibinfo{author}{Lapuschkin, S.} \&
  \bibinfo{author}{Müller, K.-R.}
\newblock \bibinfo{title}{Evaluating the {Visualization} of {What} a {Deep}
  {Neural} {Network} {Has} {Learned}}.
\newblock \emph{\bibinfo{journal}{IEEE Transactions on Neural Networks and
  Learning Systems}} \textbf{\bibinfo{volume}{28}}, \bibinfo{pages}{2660--2673}
  (\bibinfo{year}{2017}).

\bibitem{deyoung-etal-2020-eraser}
\bibinfo{author}{DeYoung, J.} \emph{et~al.}
\newblock \bibinfo{editor}{Jurafsky, D.}, \bibinfo{editor}{Chai, J.},
  \bibinfo{editor}{Schluter, N.} \& \bibinfo{editor}{Tetreault, J.} (eds)
  \emph{\bibinfo{title}{{ERASER}: {A} benchmark to evaluate rationalized {NLP}
  models}}.
\newblock (eds \bibinfo{editor}{Jurafsky, D.}, \bibinfo{editor}{Chai, J.},
  \bibinfo{editor}{Schluter, N.} \& \bibinfo{editor}{Tetreault, J.})
  \emph{\bibinfo{booktitle}{Proceedings of the 58th Annual Meeting of the
  Association for Computational Linguistics}}, \bibinfo{pages}{4443--4458}
  (\bibinfo{publisher}{Association for Computational Linguistics},
  \bibinfo{address}{Online}, \bibinfo{year}{2020}).
\newblock \urlprefix\url{https://aclanthology.org/2020.acl-main.408/}.

\bibitem{ancona2018betterunderstandinggradientbasedattribution}
\bibinfo{author}{Ancona, M.}, \bibinfo{author}{Ceolini, E.},
  \bibinfo{author}{Öztireli, C.} \& \bibinfo{author}{Gross, M.}
\newblock \bibinfo{title}{Towards better understanding of gradient-based
  attribution methods for deep neural networks} (\bibinfo{year}{2018}).
\newblock \urlprefix\url{https://arxiv.org/abs/1711.06104}.
\newblock
  \bibinfo{eprint}{{\href{https://arxiv.org/abs/1711.06104}{{arXiv:1711.06104}}}}.

\bibitem{alvarez_melis_towards_2018}
\bibinfo{author}{Alvarez~Melis, D.} \& \bibinfo{author}{Jaakkola, T.}
\newblock \emph{\bibinfo{title}{Towards {Robust} {Interpretability} with
  {Self}-{Explaining} {Neural} {Networks}}}, Vol.~\bibinfo{volume}{31}
  (\bibinfo{publisher}{Curran Associates, Inc.}, \bibinfo{year}{2018}).
\newblock
  \urlprefix\url{https://proceedings.neurips.cc/paper_files/paper/2018/hash/3e9f0fc9b2f89e043bc6233994dfcf76-Abstract.html}.

\bibitem{bhatt2020evaluatingaggregatingfeaturebasedmodel}
\bibinfo{author}{Bhatt, U.}, \bibinfo{author}{Weller, A.} \&
  \bibinfo{author}{Moura, J. M.~F.}
\newblock \bibinfo{title}{Evaluating and aggregating feature-based model
  explanations} (\bibinfo{year}{2020}).
\newblock \urlprefix\url{https://arxiv.org/abs/2005.00631}.
\newblock
  \bibinfo{eprint}{{\href{https://arxiv.org/abs/2005.00631}{{arXiv:2005.00631}}}}.

\bibitem{chalasani2020conciseexplanationsneuralnetworks}
\bibinfo{author}{Chalasani, P.}, \bibinfo{author}{Chen, J.},
  \bibinfo{author}{Chowdhury, A.~R.}, \bibinfo{author}{Jha, S.} \&
  \bibinfo{author}{Wu, X.}
\newblock \bibinfo{title}{Concise explanations of neural networks using
  adversarial training} (\bibinfo{year}{2020}).
\newblock \urlprefix\url{https://arxiv.org/abs/1810.06583}.
\newblock
  \bibinfo{eprint}{{\href{https://arxiv.org/abs/1810.06583}{{arXiv:1810.06583}}}}.

\bibitem{afzal_deepdocclassifier_2015}
\bibinfo{author}{Afzal, M.~Z.} \emph{et~al.}
\newblock \emph{\bibinfo{title}{Deepdocclassifier: {Document} classification
  with deep {Convolutional} {Neural} {Network}}}, \bibinfo{pages}{1111--1115}
  (\bibinfo{year}{2015}).

\bibitem{pfitzmann_doclaynet_2022}
\bibinfo{author}{Pfitzmann, B.}, \bibinfo{author}{Auer, C.},
  \bibinfo{author}{Dolfi, M.}, \bibinfo{author}{Nassar, A.~S.} \&
  \bibinfo{author}{Staar, P.}
\newblock \emph{\bibinfo{title}{{DocLayNet}: {A} {Large} {Human}-{Annotated}
  {Dataset} for {Document}-{Layout} {Segmentation}}}, {KDD} ’22
  (\bibinfo{publisher}{ACM}, \bibinfo{year}{2022}).
\newblock \urlprefix\url{http://dx.doi.org/10.1145/3534678.3539043}.

\bibitem{park_cord_2019}
\bibinfo{author}{Park, S.} \emph{et~al.}
\newblock \emph{\bibinfo{title}{{CORD}: a consolidated receipt dataset for
  post-{OCR} parsing}} (\bibinfo{address}{Vancouver, Canada},
  \bibinfo{year}{2019}).

\bibitem{huang_icdar2019_2019}
\bibinfo{author}{Huang, Z.} \emph{et~al.}
\newblock \emph{\bibinfo{title}{{ICDAR2019} competition on scanned receipt
  {OCR} and information extraction}}, \bibinfo{pages}{1516--1520}
  (\bibinfo{address}{Sydney, Australia}, \bibinfo{year}{2019}).

\bibitem{Sun2021SpatialDG}
\bibinfo{author}{Sun, H.}, \bibinfo{author}{Kuang, Z.}, \bibinfo{author}{Yue,
  X.}, \bibinfo{author}{Lin, C.} \& \bibinfo{author}{Zhang, W.}
\newblock \bibinfo{title}{Spatial dual-modality graph reasoning for key
  information extraction}.
\newblock \emph{\bibinfo{journal}{ArXiv}}
  \textbf{\bibinfo{volume}{abs/2103.14470}} (\bibinfo{year}{2021}).

\bibitem{mathew_docvqa_2021}
\bibinfo{author}{Mathew, M.}, \bibinfo{author}{Karatzas, D.} \&
  \bibinfo{author}{Jawahar, C.~V.}
\newblock \emph{\bibinfo{title}{{DocVQA}: a dataset for {VQA} on document
  images}} (\bibinfo{address}{Waikoloa, HI}, \bibinfo{year}{2021}).

\bibitem{devlin_bert_2019}
\bibinfo{author}{Devlin, J.}, \bibinfo{author}{Chang, M.-W.},
  \bibinfo{author}{Lee, K.} \& \bibinfo{author}{Toutanova, K.}
\newblock \bibinfo{editor}{Burstein, J.}, \bibinfo{editor}{Doran, C.} \&
  \bibinfo{editor}{Solorio, T.} (eds) \emph{\bibinfo{title}{{BERT}:
  {Pre}-training of {Deep} {Bidirectional} {Transformers} for {Language}
  {Understanding}}}.
\newblock (eds \bibinfo{editor}{Burstein, J.}, \bibinfo{editor}{Doran, C.} \&
  \bibinfo{editor}{Solorio, T.}) \emph{\bibinfo{booktitle}{Proceedings of the
  2019 {Conference} of the {North} {American} {Chapter} of the {Association}
  for {Computational} {Linguistics}: {Human} {Language} {Technologies},
  {Volume} 1 ({Long} and {Short} {Papers})}}, \bibinfo{pages}{4171--4186}
  (\bibinfo{publisher}{Association for Computational Linguistics},
  \bibinfo{address}{Minneapolis, Minnesota}, \bibinfo{year}{2019}).
\newblock \urlprefix\url{https://aclanthology.org/N19-1423}.

\bibitem{liu_roberta_2019}
\bibinfo{author}{Liu, Y.} \emph{et~al.}
\newblock \bibinfo{title}{Roberta: a robustly optimized bert pretraining
  approach}.
\newblock \emph{\bibinfo{journal}{arXiv Preprint arXiv:1907.11692}}
  (\bibinfo{year}{2019}).

\bibitem{abnar_quantifying_2020}
\bibinfo{author}{Abnar, S.} \& \bibinfo{author}{Zuidema, W.}
\newblock \bibinfo{editor}{Jurafsky, D.}, \bibinfo{editor}{Chai, J.},
  \bibinfo{editor}{Schluter, N.} \& \bibinfo{editor}{Tetreault, J.} (eds)
  \emph{\bibinfo{title}{Quantifying {Attention} {Flow} in {Transformers}}}.
\newblock (eds \bibinfo{editor}{Jurafsky, D.}, \bibinfo{editor}{Chai, J.},
  \bibinfo{editor}{Schluter, N.} \& \bibinfo{editor}{Tetreault, J.})
  \emph{\bibinfo{booktitle}{Proceedings of the 58th {Annual} {Meeting} of the
  {Association} for {Computational} {Linguistics}}},
  \bibinfo{pages}{4190--4197} (\bibinfo{publisher}{Association for
  Computational Linguistics}, \bibinfo{address}{Online}, \bibinfo{year}{2020}).
\newblock \urlprefix\url{https://aclanthology.org/2020.acl-main.385}.

\bibitem{JMLR:v17:benavoli16a}
\bibinfo{author}{Benavoli, A.}, \bibinfo{author}{Corani, G.} \&
  \bibinfo{author}{Mangili, F.}
\newblock \bibinfo{title}{Should we really use post-hoc tests based on
  mean-ranks?}
\newblock \emph{\bibinfo{journal}{Journal of Machine Learning Research}}
  \textbf{\bibinfo{volume}{17}}, \bibinfo{pages}{1--10} (\bibinfo{year}{2016}).
\newblock \urlprefix\url{http://jmlr.org/papers/v17/benavoli16a.html}.

\bibitem{JMLR:v7:demsar06a}
\bibinfo{author}{Dem{\v{s}}ar, J.}
\newblock \bibinfo{title}{Statistical comparisons of classifiers over multiple
  data sets}.
\newblock \emph{\bibinfo{journal}{Journal of Machine Learning Research}}
  \textbf{\bibinfo{volume}{7}}, \bibinfo{pages}{1--30} (\bibinfo{year}{2006}).
\newblock \urlprefix\url{http://jmlr.org/papers/v7/demsar06a.html}.

\bibitem{Owen1977}
\bibinfo{author}{Owen, G.}
\newblock \emph{\bibinfo{title}{Values of Games with a Priori Unions}},
  \bibinfo{pages}{76–88} (\bibinfo{publisher}{Springer Berlin Heidelberg},
  \bibinfo{year}{1977}).
\newblock \urlprefix\url{http://dx.doi.org/10.1007/978-3-642-45494-3_7}.

\bibitem{verma_counterfactual_2020}
\bibinfo{author}{Verma, S.}, \bibinfo{author}{Dickerson, J.~P.} \&
  \bibinfo{author}{Hines, K.~E.}
\newblock \bibinfo{title}{Counterfactual {Explanations} for {Machine}
  {Learning}: {A} {Review}}.
\newblock \emph{\bibinfo{journal}{ArXiv}}
  \textbf{\bibinfo{volume}{abs/2010.10596}} (\bibinfo{year}{2020}).
\newblock \urlprefix\url{https://api.semanticscholar.org/CorpusID:224818450}.

\end{thebibliography}
\newpage
\begin{appendices}
  \onecolumn
  \section{Metric Descriptions and Adaptations}
\label{sec:appendix_metrics}
In this section, we provide detailed descriptions of the evaluation metrics used in this work. Unless otherwise stated, all metrics are computed over group-aggregated attribution scores, treating each feature group independently of its originating modality. For perturbation-based metrics, the perturbation order is determined by ranking all feature groups jointly across modalities, and the corresponding features are then perturbed using their modality-specific segmentation masks and reference baselines. Note that for all perturbation-based metrics, feature groups associated with special tokens such as [CLS], [PAD], or [SEP] are excluded from perturbation.

\subsection{Axiomatic Metrics}
\label{sec:app:axiomatic}
\subheading{Completeness.}
\completenessf{} is an axiomatic property of an explanation method that measures whether the attributions add up to the difference between the model output at the input and at the baseline:
\begin{equation}
    \completeness(x) = \left| \sum_{\mathcal{M}} \sum_{i} \mathbf{a}_i^{\mathcal{M}} - \left( f(x) - f(x') \right) \right|
\end{equation}
where $\mathbf{a}_i^{\mathcal{M}}$ are the raw attribution scores for feature $i$ in modality $\mathcal{M}$, $f(x)$ is the model output at the input, and $f(x')$ is the model output at the baseline. A score of zero indicates perfect completeness, and some methods, like \integratedgradients{}, satisfy completeness by construction. Unlike other metrics, \completenessf{} is computed on the raw attribution scores instead of group-aggregated scores, as the axiom requires the exact sum over all individual attributions. For multi-target tasks, we adapt the metric to efficiently compute the difference between the model output at the input and at the baseline over all $T$ targets in a single step.

\subheading{Non-Sensitivity.}
\nonsensitivityf{} is another axiomatic property that determines whether features with zero attribution have a negligible influence on the model output. Formally, let $A_0 \subset \{1, \ldots, N\}$ denote the subset of feature indices with zero attribution, i.e., $a_i = 0$. Further, let $X_0 = \{i \in \{1, \ldots, N\} \mid \mathbb{E}(\ell(y^*, f_i)) = 0\}$ denote the subset of features that the model $f$ is not functionally dependent on. The non-sensitivity score is then defined as $|A_0 \triangle X_0|$, where $|\cdot|$ is the cardinality of a set and $\triangle$ denotes the symmetric difference.
In our adaptation, we compute \nonsensitivity{} over the unified feature space using group-aggregated attribution scores $\{a_k\}_{k=1}^{K}$ and instantiate the otherwise abstract loss function $\ell$ as the relative variance of the model output under perturbation of each feature group $k$:
\begin{equation}
    v_k = \frac{\frac{1}{N} \sum_{n=1}^{N} \left( f(x_k'^{(n)}) - f(x) \right)^2}{f(x)^2}
    \label{app:eq:relvar}
\end{equation}
where $f(x_k'^{(n)})$ is the model output with the $k$-th feature group perturbed and $f(x)$ is the model output on the unperturbed input. $N$ denotes the number of perturbations per feature group. When using a random baseline function, $N > 1$ is necessary to obtain a stable variance estimate; however, with a fixed reference baseline, as in our case, each perturbation is deterministic, and therefore we use $N=1$. We can then compute $A_0 = \{k : |a_k| < \epsilon_a\}$ and $X_0 = \{k : v_k < \epsilon_v\}$, where $\epsilon_a$ and $\epsilon_v$ are thresholds that define near-zero values. Furthermore, as the number of features can vary drastically across multimodal sequential inputs, we normalize the non-sensitivity score by the number of features to ensure comparability across setups:
\begin{equation}
    \nonsensitivity(x) = \frac{|A_0 \triangle X_0|}{K}
\end{equation}
where $K$ is the total number of feature groups. Lower \nonsensitivity{} values indicate better alignment between zero-attribution features and features that do not affect the model output. Note that \nonsensitivity{} is computed jointly with \monotonicitycorr{} (described in Section~\ref{sec:app:faithfulness}), as both metrics share the same per-feature perturbation results. Furthermore, to reduce computational cost, we apply chunking, where rather than perturbing each feature individually, a fixed percentage of feature groups is perturbed together in each step.  For multi-target tasks, we adapt the metric to efficiently compute the relative variances $v_k$ over all $T$ targets of a single sample simultaneously.

\subsection{Complexity Metrics}
\label{sec:app:complexity}
Complexity metrics assess how easily a user can interpret the explanations, with simpler explanations that concentrate importance on fewer features being generally preferred. We adapt all complexity metrics to be computed over the group-aggregated attribution scores. For complexity metrics that do not require a model forward pass, multi-target adaptations are not necessary; instead, these metrics are computed independently for each sample-target pair.

\subheading{Effective Complexity.}
\effectivecomplexityf{} measures the minimum number of top-attributed features sufficient to influence the model output. Formally, let $a^{(i)}$ be the attributions ordered increasingly by their absolute value, and $x^{(i)}$ the corresponding features. Let $M_k = \{x^{(N-k)}, \ldots, x^{(N)}\}$ be the set of top $k$ features. The effective complexity is defined as:
\[
k^* = \operatorname*{argmin}_{k \in \{1, \ldots, N\}} |M_k| \quad \text{s.t.} \quad \mathbb{E}(\ell(y^*, f_{-M_k}) | \mathbf{x}^*_{M_k}) < \epsilon
\]
where $f_{-M_k}$ is the restriction of the model $f$ to the non-important features, given fixed values for the important features in $M_k$.
In our adaptation, \effectivecomplexity{} follows a similar perturbation strategy to \nonsensitivity{} and instantiates loss as the relative variance of the model output under perturbation of each feature group $k$ (see Eq.~\ref{app:eq:relvar}). To find $M_k$, we remove feature groups cumulatively in ascending order of their absolute attribution scores $\{a_k\}_{k=1}^K$ and find the index $k^*$ at which the relative variance exceeds a threshold $\epsilon_v$. The normalized effective complexity is the ratio of top-k features to the total number of features:
\begin{equation}
    \effectivecomplexity(x) = \frac{K - k^*}{K}
\end{equation}
Lower \effectivecomplexity{} values indicate that the explanation concentrates importance on fewer important features. To reduce the computational costs, we also apply chunking to compute \effectivecomplexity{}, where a fixed percentage of feature groups are perturbed together in each step. Furthermore, since \effectivecomplexity{} requires ordered perturbation, multi-target adaptation is not possible; the metric is therefore computed independently for each sample-target pair.

\subheading{Entropy-based Complexity.}
\entropycomplexityf{} measures the total entropy of the attribution distribution over feature groups. In our adaptation, we first compute the normalized absolute group-aggregated attribution scores:
\[
p_k = \frac{|a_k|}{\sum_{k} |a_{k}|}
\]
We then compute normalized entropy over feature groups to keep the entropy scores comparable across samples of varying feature sets:
\[
\entropycomplexity(x) = \frac{-\sum_{k} p_k \log p_k}{\log K}
\]
where $K$ is the total number of feature groups. A value near 0 indicates that the attribution mass is concentrated on very few feature groups, while a value near 1 indicates a near-uniform spread.

\subheading{Sundararajan Complexity.}
\sundararajancomplexityf{} measures the fraction of features with non-negligible attribution. In our adaptation, we first compute the normalized absolute group-aggregated attribution scores:
\begin{equation}
    \hat{a}_k = \frac{|a_k|}{\sum_{k'} |a_{k'}|}
\end{equation}
The complexity score is then defined as the fraction of feature groups whose normalized attribution exceeds a threshold $\epsilon$:
\begin{equation}
    \sundararajancomplexity(x) = \frac{|\{k : \hat{a}_k > \epsilon\}|}{K}
\end{equation}
where $K$ is the total number of feature groups. Lower values indicate sparser, more concise explanations.

\subheading{Sparseness.}
\sparsenessf{} measures the concentration of attribution scores using the Gini index. In our adaptation, we compute the Gini index over the absolute group-aggregated attribution scores:
\begin{equation}
    \sparseness(x) = G(\{|a_1|, \ldots, |a_K|\})
\end{equation}
where $K$ is the total number of feature groups. Higher values indicate sparser attributions, meaning the explanation concentrates importance on fewer features.

\subsection{Faithfulness Metrics}
\label{sec:app:faithfulness}
Faithfulness metrics assess whether the attributions accurately reflect the model's decision-making process.

\subheading{Area Over the Perturbation Curve.}
\aopcf{} measures the average change in model output as features are progressively removed in order of their attributed importance. Given an input $x$ and an ordered sequence of feature removals, the MoRF (Most Relevant First) perturbation curve is defined recursively as $x^{(0)} = x$ and $x^{(k)} = g(x^{(k-1)}, r_k), \forall \; 1 \leq k \leq L$, where $g$ indicates removal of the feature at location $r_k$. The AOPC metric is then computed as:
\begin{equation}
    \aopc = \frac{1}{L+1} \left\langle \sum_{k=0}^{L} f(x^{(0)}) - f(x^{(k)}) \right\rangle_{p(x)}
\end{equation}
where $L$ is the total number of removal steps and $\langle \cdot \rangle_{p(x)}$ denotes the mean over the dataset. A higher $\aopc_{\text{MoRF}}$ indicates that the most important features, as ranked by the attribution, have a larger impact on the model output, suggesting more faithful explanations.
In our adaptation, feature groups are removed cumulatively based on the perturbation order determined in the aggregated feature space, and the removal is performed by replacing the corresponding features with the modality-wise reference baselines. Furthermore, to ensure comparability across samples with varying numbers of feature groups, we remove features in fixed-size chunks, where each chunk contains an equal percentage of features, resulting in the same number of removal steps $L$ regardless of the feature count. In practice, the \aopc{} curve is computed in three orderings: descending (MoRF), ascending (LeRF), and random (averaged over $n$ random permutations). Since \aopc{} requires ordered cumulative perturbation and the perturbation order differs for each target, efficient multi-target computation is not possible.

\subheading{Area Between the Perturbation Curves.}
\abpcf{} measures the area between the descending (MoRF) and ascending (LeRF) perturbation curves:
\begin{equation}
    \abpc = \frac{1}{L+1} \left\langle \sum_{k=0}^{L} f(x_{\text{LeRF}}^{(k)}) - f(x_{\text{MoRF}}^{(k)}) \right\rangle_{p(x)}
\end{equation}
A higher \abpc{} indicates that removing high-importance features degrades model performance quickly, while removing low-importance features has comparatively little effect. In our implementation, we derive the \abpc{} score directly from the \aopc{} computation and report scores at full feature removal $L$.

\subheading{Sufficiency.}
\sufficiencyf{} measures whether the top-attributed features alone are sufficient to retain the model's prediction. In the original formulation~\citep{deyoung-etal-2020-eraser}, sufficiency is computed by retaining only the top-ranked features and measuring the change in model output.
In our adaptation, we compute sufficiency similarly to \aopc{} but with an inverse perturbation direction: starting from the reference baseline, features are progressively added back in descending order of their attribution scores, and the area under the resulting perturbation curve is computed. Lower \sufficiency{} scores indicate better faithfulness.

\subheading{Infidelity.}
\infidelityf{} measures the discrepancy between the importance assigned to features and their observed impact on the model output under perturbation. Given an explanation functional $\Phi$, a model $f$, and a random perturbation variable $\mathbf{I} \in \mathbb{R}^d$, infidelity is defined as:
\begin{equation}
    \text{INFD}(\Phi, f, x) = \mathbb{E}_{\mathbf{I} \sim \mu_I} \left[ \left( \mathbf{I}^\top \Phi(f, x) - (f(x) - f(x - \mathbf{I})) \right)^2 \right]
\end{equation}
where $\mathbf{I}$ represents meaningful perturbations of interest and can be instantiated in various ways. In our implementation, we use the noisy baseline variant where $\mathbf{I} = x - z_0$ with $z_0 = x_0 + \epsilon$ and $\epsilon \sim \mathcal{N}(0, \sigma^2)$. Unlike other faithfulness metrics, \infidelity{} applies random Gaussian noise directly to each modality input, rather than operating on the aggregated feature space, to keep the implementation close to the original formulation. Lower \infidelity{} values indicate that the attribution scores more faithfully reflect the impact of features on the model output. For multi-target tasks, we adapt the \infidelity{} metric such that for each input perturbation, the differences between the model outputs on the unperturbed and perturbed inputs are computed for all targets $T$ simultaneously.

\subheading{Sensitivity-N.}
\sensitivitynf{} is a specific instantiation of infidelity where the perturbation corresponds to the subset-of-difference-to-baseline variant~\cite{yeh_fidelity_2019}. An attribution method satisfies Sensitivity-N when, for any random subset of features $S_k \subseteq [d]$ of cardinality $N$, the sum of the attributions equals the change in model output caused by removing those features:
\begin{equation}
    \sensitivityn(\Phi, f, x) = \mathbb{E}_{S_k} [ ( \sum_{i \in S_k} \Phi(f, x)_i - ( f(x) - f(x_{[x_{S_k} = x_0]}) ))^2 ]
\end{equation}
where the expectation is estimated over $P$ randomly sampled subsets $S_k$ of $N$ feature groups, and $x_{[x_{S_k} = x_0]}$ denotes the input with features in $S_k$ replaced by the reference baseline. Unlike \infidelity{}, we compute \sensitivityn{} over the segmented feature space with group-aggregated attribution scores. The parameter $N$ can be specified either as an absolute count or as a percentage of the total number of feature groups. We evaluate \sensitivityn{} at perturbation levels of $N=\{20\%, 40\%, 60\%\}$. For multi-target tasks, we apply the same adaptation as for \infidelity{}, that is, the differences between the model outputs on the unperturbed and perturbed inputs are computed for all targets $T$ simultaneously.

\subheading{Monotonicity Correlation.}
\monotonicitycorrf{} measures whether features with higher attribution scores have a proportionally larger effect on the model output when perturbed. In our adaptation of \monotonicitycorr{}, for each feature group $k$, the relative variance $v_k$ of the output under perturbation is computed as described for \nonsensitivity{} (see Eq.~\ref{app:eq:relvar}). The monotonicity correlation is then defined as the Spearman rank correlation between the absolute group-aggregated attribution scores and the corresponding output variances:
\begin{equation}
    \monotonicitycorr(x) = \rho_s\left( \{|a_k|\}_{k=1}^{K}, \{v_k\}_{k=1}^{K} \right)
\end{equation}
A value close to 1 indicates that the attribution ranking is well-aligned with the actual feature importance as measured by the model's sensitivity to perturbations. As noted in Section~\ref{sec:app:axiomatic}, \monotonicitycorr{} is computed jointly with \nonsensitivity{}, sharing the same per-feature perturbation results. For multi-target tasks, perturbations are applied once and the output changes are accumulated over all $T$ targets simultaneously.

\subheading{Monotonicity.}
\monotonicityf{} evaluates whether the model output increases monotonically as features are progressively added to the input from least important to most important. We compute \monotonicity{} by starting from a reference baseline input and introducing feature groups in ascending order of their attribution scores, recording the model output at each step.
If the output increases monotonically across all steps, the score is $1$; otherwise, it is $0$. Since this metric uses ordered perturbations, multi-target adaptation is not possible and the metric is computed independently for each sample-target pair.

\subheading{Faithfulness Estimation.}
\faithfulnessestf{} assesses faithfulness by measuring the correlation between the individual feature attributions and the observed effect of removing a feature on the model output. In our adaptation, each feature group $k$ is independently removed by replacing it with the corresponding modality-wise reference baseline, and the resulting change in model output is recorded. The faithfulness estimation score is the Pearson correlation between the attribution scores and the corresponding output changes:
\begin{equation}
    \faithfulnessest(x) = r\left( \{a_k\}_{k=1}^{K}, \{f(x) - f(x_{\setminus k})\}_{k=1}^{K} \right)
\end{equation}
where $a_k$ is the group-aggregated attribution score for feature group $k$ and $x_{\setminus k}$ denotes the input with feature group $k$ replaced by the reference baseline. A value close to 1 indicates that the attribution scores faithfully predict the impact of removing individual features. This metric uses ordered perturbations, and therefore multi-target adaptation is not possible and the metric is computed independently for each sample-target pair.

\subheading{Faithfulness Correlation.}
\faithfulnesscorrf{} works in a similar fashion to \sensitivityn{}, but instead of computing the mean squared error, it measures the Pearson correlation between the sum of attributions over a randomly selected subset of features and the corresponding change in model output. In our adaptation, for each sample, we generate $R$ randomly sampled subsets $S_r$ containing $N$ feature groups:
\begin{equation}
    \faithfulnesscorr(x) = r\left( \left\{\sum_{i \in S_r} a_i\right\}_{r=1}^{R}, \left\{f(x) - f(x_{[x_{S_r} = x_0]})\right\}_{r=1}^{R} \right)
\end{equation}
where $a_i$ is the group-aggregated attribution score for feature group $i$, $x_{[x_{S_r} = x_0]}$ denotes the input with features in $S_r$ replaced by the reference baseline, and $R$ is the number of random perturbation samples. A value close to 1 indicates that the attribution scores faithfully predict the effect of removing feature subsets. We evaluate \faithfulnesscorr{} at perturbation levels of $N=\{20\%, 40\%, 60\%\}$ of the total number of feature groups. For multi-target tasks, perturbations are applied once and the output changes are accumulated over all $T$ targets simultaneously.

\subsection{Robustness Metrics}
\label{sec:app:robustness}
Robustness metrics measure the stability of explanations under small input perturbations.

\subheading{Max Sensitivity.}
\maxsensitivityf{} measures the maximum change in the explanation within a local neighborhood of the input. Given an explanation functional $\Phi$, a model $f$, and an input neighborhood radius $r$, max sensitivity is defined as:
\begin{equation}
\maxsensitivity(\Phi, f, x, r) = \max_{\|y - x\| \leq r} \|\Phi(f, y) - \Phi(f, x)\|
\end{equation}
In practice, we approximate this by sampling $N$ random perturbations, applied independently to each input modality by adding uniform random noise in $[-r, r]$. Lower \maxsensitivityf{} values indicate more robust explanations. Since this metric requires recomputing explanations for each perturbation, the underlying multi-target explanation mode is used for multi-target tasks.

\subheading{Avg. Sensitivity.}
\avgsensitivityf{} is defined identically to \maxsensitivityf{} but replaces the maximum with the mean over the sampled perturbations. Lower values indicate more robust explanations. The same multi-target adaptation as \maxsensitivity{} applies.

  \newpage
\section{Training and Evaluation Hyperparameters}
\label{sec:appendix_hyperparams}

\subsection{Training Hyperparameters}
\label{sec:appendix_training_hyperparams}
Table~\ref{tab:hyperparams} lists the training hyperparameters for all model-task combinations used in this work. We use \adam{} or \adamw{} with linear warmup for optimization and apply early stopping based on validation performance on task-specific metrics with a patience of $10$. All models are trained with a dropout rate of $0.1$. For \layoutlm{}, input images are resized to $224 \times 224$ following the original work.

\begin{table*}[h]
    \centering
    
    \caption{Training hyperparameters for all model-task combinations used in this work. Modality columns indicate which input modalities each model processes: $\mathcal{T}$ (text), $\mathcal{S}$ (position), $\mathcal{L}$ (layout), $\mathcal{V}$ (image).    \label{tab:hyperparams}}

    \rowcolors{2}{gray!15}{white}
    \resizebox{0.8\textwidth}{!}{
\begin{tabular}{lrrrrrrrrrrrr}
\rowcolor{gray!15}
    \textbf{Task} &\textbf{Model} &\textbf{Modality} &\rotatebox{90}{\textbf{Learning Rate}} &\rotatebox{90}{\textbf{Batch Size}} &\rotatebox{90}{\textbf{Epochs}} &\rotatebox{90}{\textbf{Optimizer}} &\rotatebox{90}{\textbf{Weight Decay}} &\rotatebox{90}{\textbf{Warmup Ratio}} &\rotatebox{90}{\textbf{Dropout}} &\rotatebox{90}{\textbf{Early Stopping}} &\rotatebox{90}{\textbf{Image Size}} \\\midrule
	\cellcolor{white}&\bert{} &$\mathcal{T}$; $\mathcal{S}$ &1.00e-5 &32/64 &40 &\adam{} &0 &1000 &0.1 &\cmark &— \\
	\cellcolor{white}    &\roberta{} &$\mathcal{T}$; $\mathcal{S}$ &1.00e-5 &32/64 &40 &\adam{} &0 &1000 &0.1 &\cmark &— \\
	\cellcolor{white}    &\lilt{} &$\mathcal{T}$; $\mathcal{S}$; $\mathcal{L}$ &1.00e-5 &32/64 &40 &\adam{} &0 &1000 &0.1 &\cmark &— \\
	\cellcolor{white}    \multirow{-4}{*}{\cls}    &\layoutlm{} &$\mathcal{T}$; $\mathcal{S}$; $\mathcal{L}$; $\mathcal{V}$ &1.00e-5 &32/64 &40 &\adam{} &0 &1000 &0.1 &\cmark &224$\times$224 \\\midrule
		\cellcolor{white}&\bert{} &$\mathcal{T}$; $\mathcal{S}$ &2.00e-5 &16 &100 &\adamw{} &0.01 &100 &0.1 &\cmark &— \\
	\cellcolor{white}    &\roberta{} &$\mathcal{T}$; $\mathcal{S}$ &2.00e-5 &16 &100 &\adamw{} &0.01 &100 &0.1 &\cmark &— \\
	\cellcolor{white}    &\lilt{} &$\mathcal{T}$; $\mathcal{S}$; $\mathcal{L}$ &2.00e-5 &16 &100 &\adamw{} &0.01 &100 &0.1 &\cmark &— \\
	\cellcolor{white}    \multirow{-4}{*}{\ser}     &\layoutlm{} &$\mathcal{T}$; $\mathcal{S}$; $\mathcal{L}$; $\mathcal{V}$ &2.00e-5 &16 &100 &\adamw{} &0.01 &100 &0.1 &\cmark &224$\times$224 \\\midrule
		\cellcolor{white}&\bert{} &$\mathcal{T}$; $\mathcal{S}$ &5.00e-5 &32 &50 &\adamw{} &0.01 &1000 &0.1 &\cmark &— \\
	\cellcolor{white}    &\roberta{} &$\mathcal{T}$; $\mathcal{S}$ &5.00e-5 &32 &50 &\adamw{} &0.01 &1000 &0.1 &\cmark &— \\
	\cellcolor{white}    &\lilt{} &$\mathcal{T}$; $\mathcal{S}$; $\mathcal{L}$ &5.00e-5 &32 &50 &\adamw{} &0.01 &1000 &0.1 &\cmark &— \\
	\cellcolor{white}    \multirow{-4}{*}{\qa}     &\layoutlm{} &$\mathcal{T}$; $\mathcal{S}$; $\mathcal{L}$; $\mathcal{V}$ &5.00e-5 &32 &50 &\adamw{} &0.01 &1000 &0.1 &\cmark &224$\times$224 \\
    \bottomrule
\end{tabular}

    }

\end{table*}

\subsection{Explainer Hyperparameters}
\label{sec:appendix_explainer_hyperparams}
Table~\ref{tab:explainer_hyperparams} lists the hyperparameters used for the XAI methods. We use the same set of hyperparameters across all tasks. For methods that require sampling or interpolation steps, such as \integratedgradients{}, \gradientshap{}, \lime{}, and \kernelshap{}, we use 200 samples or steps.
For \deepshap{}, a reference baseline distribution of 100 samples is precomputed from the training set of each dataset.\begin{table*}[!h]
	\centering

\caption{Hyperparameters for the evaluation of XAI methods used in this work. Methods that are not listed have no configurable parameters.    \label{tab:explainer_hyperparams}}%

\rowcolors{2}{gray!15}{white}
\resizebox{0.6\textwidth}{!}{ %
    \begin{tabular}{lrrr}
        \rowcolor{gray!15}
        Method &Hyperparameter &Value \\\midrule
        \integratedgradients{} &\texttt{n\_steps} &200 \\
        \gradientshap{} &\texttt{n\_samples} &200 \\
        \deepshap{} &\texttt{num\_baselines} &100 \\
        \occlusion{} &\texttt{sliding\_window (token)} &8$\times$d \\
        \occlusion{} &\texttt{stride (token)} &4$\times$d \\
        \occlusion{} &\texttt{sliding\_window (image)} &16$\times$16$\times$3 \\
        \occlusion{} &\texttt{stride (image)} &8$\times$8$\times$3 \\
        \lime{} &\texttt{n\_samples} &200 \\
        \lime{} &\texttt{alpha} &0.01 \\
        \kernelshap{} &\texttt{n\_samples} &200 \\
        \bottomrule
    \end{tabular}
}

\end{table*}
 Note that increasing the number of samples in the baseline may improve the performance of \deepshap{}, but this comes at extreme computational cost, as the method requires $N$ backward passes for each input sample (or input sample-target pair when using the naive multi-target implementation), where $N$ is the size of the baseline distribution.
For \occlusion{}, we define modality-specific sliding windows and strides. The \texttt{alpha} parameter in \lime{} denotes the $L_1$ penalty coefficient of the LASSO regression used to fit the surrogate model. Note that for methods that require interpolation or sampling, an additional parameter \texttt{internal\_batch\_size} is used that determines the internal computation batch size. Similarly, for backpropagation-based methods in multi-target scenarios, an additional parameter \texttt{grad\_batch\_size} is used that determines the batch of targets for which backpropagation is performed simultaneously. Both parameters are configured based on the available GPU memory.
For all single-target scenarios, we use the method implementations available in \texttt{captum}\footnote{\url{https://captum.ai/}}, with the exception of \deepshap{} and \gradientshap{}, both of which we extend for batched computation. For multi-target scenarios, we extend all \texttt{captum} implementations to support the multi-target setup. These extended implementations are made publicly available in the \texttt{torchxai}\footnote{\url{https://atriaml.github.io/torchxai/}} library.

\subsection{Metric Hyperparameters}
\label{sec:appendix_metric_hyperparams}
Table~\ref{tab:metric_hyperparams} lists the hyperparameters for all XAI metrics used in the evaluation. Perturbation-based metrics that support chunking use a chunk size of 1\%. The parameter \texttt{max\_features\_per\_batch} determines the batch size for metrics that require a model forward pass. For feature grouping and removal, all metrics use the same modality-wise feature segmentation masks and baselines as those used in the computation of the XAI attributions.
\begin{table*}[!htp]\centering\small

\caption{Hyperparameters for all XAI metrics used in the evaluation. Metrics that are not listed (e.g., \completeness{}, \entropycomplexity{}, \sparseness{}) have no configurable parameters. \nonsensitivity{} and \monotonicitycorr{} share the same hyperparameters as they are computed jointly. Similarly, \abpc{} and \sufficiency{} share hyperparameters with \aopc{}.    \label{tab:metric_hyperparams}}

    \resizebox{0.7\textwidth}{!}{ %
    \begin{tabular}{lrrrr}
\rowcolor{gray!15}
        Type &Metric &Hyperparameter &Value \\\midrule
        \multirow{5}{*}{\rotatebox{90}{Axiomatic}} &\cellcolor{gray!15} &\cellcolor{gray!15}\texttt{n\_perturbations\_per\_feature} &1\cellcolor{gray!15} \\
        &\cellcolor{gray!15} &\texttt{chunk\_size}\cellcolor{gray!15} &1\% \cellcolor{gray!15}\\
        &\cellcolor{gray!15} &$\epsilon_a$ \cellcolor{gray!15}&1.00e-3 \cellcolor{gray!15}\\
        &\cellcolor{gray!15} &$\epsilon_v$ \cellcolor{gray!15}&1.00e-1 \cellcolor{gray!15}\\
        &\cellcolor{gray!15}\multirow{-5}{*}{Non-Sensitivity (\textsc{ns}) ($\downarrow$)} &\texttt{max\_features\_per\_batch}\cellcolor{gray!15} &100\cellcolor{gray!15} \\\midrule
        \multirow{25}{*}{\rotatebox{90}{Faithfulness}} &\cellcolor{gray!15} &\cellcolor{gray!15}\texttt{total\_feature\_bins} &200 \cellcolor{gray!15}\\
        & \cellcolor{gray!15}&\texttt{n\_random\_perms} \cellcolor{gray!15}&3 \cellcolor{gray!15}\\
        &\multirow{-3}{*}{AOPC (\textsc{aopc})\cellcolor{gray!15} ($\uparrow$)} &\texttt{max\_features\_per\_batch} \cellcolor{gray!15}&100 \cellcolor{gray!15}\\\cmidrule{2-4}
        &\cellcolor{gray!15} &\texttt{total\_feature\_bins} \cellcolor{gray!15}&200\cellcolor{gray!15} \\
        & \cellcolor{gray!15}&\cellcolor{gray!15}\texttt{n\_random\_perms} &3\cellcolor{gray!15}\\
        &\multirow{-3}{*}{ABPC (\textsc{abpc}) ($\uparrow$)}\cellcolor{gray!15} &\cellcolor{gray!15}\texttt{max\_features\_per\_batch} &100\cellcolor{gray!15} \\\cmidrule{2-4}
        &\cellcolor{gray!15} &\cellcolor{gray!15}\texttt{total\_feature\_bins} &200 \cellcolor{gray!15}\\
        &\cellcolor{gray!15} &\cellcolor{gray!15}\texttt{n\_random\_perms} \cellcolor{gray!15}&3 \cellcolor{gray!15}\\
        & \multirow{-3}{*}{Sufficiency (\textsc{suf}) ($\downarrow$)}\cellcolor{gray!15} &\texttt{max\_features\_per\_batch} \cellcolor{gray!15}&100 \cellcolor{gray!15}\\\cmidrule{2-4}
        & \cellcolor{gray!15}&\cellcolor{gray!15}\texttt{n\_perturb\_samples} &\cellcolor{gray!15}200 \cellcolor{gray!15}\\
        &\cellcolor{gray!15}\multirow{-2}{*}{Sensitivity-N (\textsc{sn})($\downarrow$)}  &\cellcolor{gray!15}\texttt{max\_examples\_per\_batch} &\cellcolor{gray!15}100 \\\cmidrule{2-4}
        &\cellcolor{gray!15} &\cellcolor{gray!15}\texttt{n\_perturb\_samples} &200 \cellcolor{gray!15}\\
        & \cellcolor{gray!15}&\texttt{noise\_scale} \cellcolor{gray!15}&0.1 \cellcolor{gray!15}\\
        & \multirow{-3}{*}{Infidelity (\textsc{inf}) ($\downarrow$)}\cellcolor{gray!15}&\cellcolor{gray!15}\texttt{max\_examples\_per\_batch} &\cellcolor{gray!15}100 \\\cmidrule{2-4}
        &\cellcolor{gray!15} &\texttt{chunk\_size} \cellcolor{gray!15}&1\cellcolor{gray!15}  \\
        & \multirow{-2}{*}{Faithfulness Estimation (\textsc{fe})\cellcolor{gray!15} ($\uparrow$)}&\cellcolor{gray!15}\texttt{max\_features\_per\_batch} &100\cellcolor{gray!15} \\\cmidrule{2-4}
        & \cellcolor{gray!15}&\cellcolor{gray!15}\texttt{n\_perturb\_samples} &\cellcolor{gray!15}200 \\
        &\multirow{-2}{*}{Faithfulness Correlation (\textsc{fc}) ($\uparrow$)}\cellcolor{gray!15} &\cellcolor{gray!15}\texttt{max\_examples\_per\_batch} &\cellcolor{gray!15}100 \\\cmidrule{2-4}
        & \cellcolor{gray!15}&\cellcolor{gray!15}\texttt{chunk\_size} &\cellcolor{gray!15}1\% \\
        & \multirow{-2}{*}{Monotonicity (\textsc{m}) ($\uparrow$)}\cellcolor{gray!15}&\texttt{max\_features\_per\_batch} \cellcolor{gray!15}&100\cellcolor{gray!15} \\\cmidrule{2-4}
        &\cellcolor{gray!15} &\cellcolor{gray!15}\texttt{n\_perturbations\_per\_feature} &1 \cellcolor{gray!15}\\
        &\cellcolor{gray!15} &\texttt{chunk\_size}\cellcolor{gray!15} &1\% \cellcolor{gray!15}\\
        &\cellcolor{gray!15} &$\epsilon_a$ \cellcolor{gray!15}&1.00e-3 \cellcolor{gray!15}\\
        &\cellcolor{gray!15} &$\epsilon_v$ \cellcolor{gray!15}&1.00e-1 \cellcolor{gray!15}\\
        &\multirow{-5}{*}{Monotonicity Correlation (\textsc{mc}) ($\uparrow$)}\cellcolor{gray!15} &\texttt{max\_features\_per\_batch} \cellcolor{gray!15}&100 \cellcolor{gray!15}\\
        \midrule
        \multirow{5}{*}{\rotatebox{90}{Complexity}} & \cellcolor{gray!15}&\texttt{n\_perturbations\_per\_feature}\cellcolor{gray!15} &1 \cellcolor{gray!15}\\
        & \cellcolor{gray!15}&\texttt{chunk\_size}\cellcolor{gray!15} &1\% \cellcolor{gray!15}\\
        & \cellcolor{gray!15}&$\epsilon_v$\cellcolor{gray!15} &1.00e-1 \cellcolor{gray!15}\\
        &\cellcolor{gray!15}\multirow{-4}{*}{Effective Complexity (\textsc{ec}) ($\downarrow$)} &\cellcolor{gray!15}\texttt{max\_features\_per\_batch} &100 \cellcolor{gray!15}\\\cmidrule{2-4}
        &Sundararajan Complexity (\textsc{sc}) \cellcolor{gray!15} ($\downarrow$) &\cellcolor{gray!15}$\epsilon$ &1.00e-3\cellcolor{gray!15} \\\midrule
        \multirow{8}{*}{\rotatebox{90}{Robustness}} &\cellcolor{gray!15} &\cellcolor{gray!15}\texttt{perturb\_radius} &0.02 \cellcolor{gray!15}\\
        & \cellcolor{gray!15}&\cellcolor{gray!15}\texttt{n\_perturb\_samples} &10 \cellcolor{gray!15}\\
        & \cellcolor{gray!15}&\texttt{norm} \cellcolor{gray!15}&Frobenius \cellcolor{gray!15}\\
        &\cellcolor{gray!15} \multirow{-4}{*}{Max Sensitivity (\textsc{ms}) ($\downarrow$)}&\cellcolor{gray!15}\texttt{max\_examples\_per\_batch} &1\cellcolor{gray!15} \\\cmidrule{2-4}
        &\cellcolor{gray!15} &\texttt{perturb\_radius} \cellcolor{gray!15}&0.02 \cellcolor{gray!15}\\
        & \cellcolor{gray!15}&\cellcolor{gray!15}\texttt{n\_perturb\_samples} &10\cellcolor{gray!15} \\
        &\cellcolor{gray!15} &\cellcolor{gray!15}\texttt{norm} &Frobenius \cellcolor{gray!15}\\
        &\multirow{-4}{*}{Avg Sensitivity (\textsc{as}) ($\downarrow$)}\cellcolor{gray!15} &\cellcolor{gray!15}\texttt{max\_examples\_per\_batch} &1 \cellcolor{gray!15}\\
        \bottomrule
    \end{tabular}
}

\end{table*}

We use the implementations of \maxsensitivityf{} and \infidelityf{} from \texttt{captum}, and extend them for multi-target scenarios. All remaining metric implementations, along with the multi-target adaptations, are made publicly available in the \texttt{torchxai} library.
  \section{Additional Results}
\subsection{Per-Task XAI Method Rankings}
\label{sec:appendix_ranking}

\begin{figure*}[!ht]
    \centering
    \begin{subfigure}[t]{0.3\textwidth}
        \centering
        \includegraphics[width=\textwidth,draft=false]{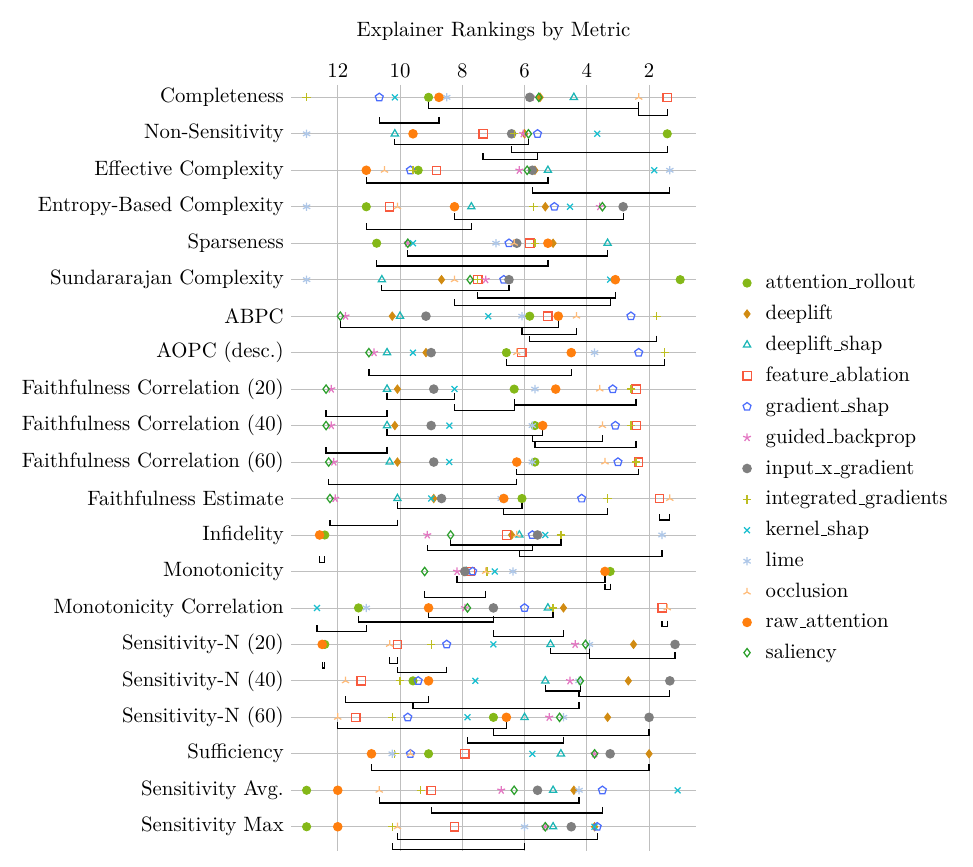}
        \caption{Classification}
        \label{fig:cd_cls}
    \end{subfigure}
    \begin{subfigure}[t]{0.3\textwidth}
        \centering
        \includegraphics[width=\textwidth,draft=false]{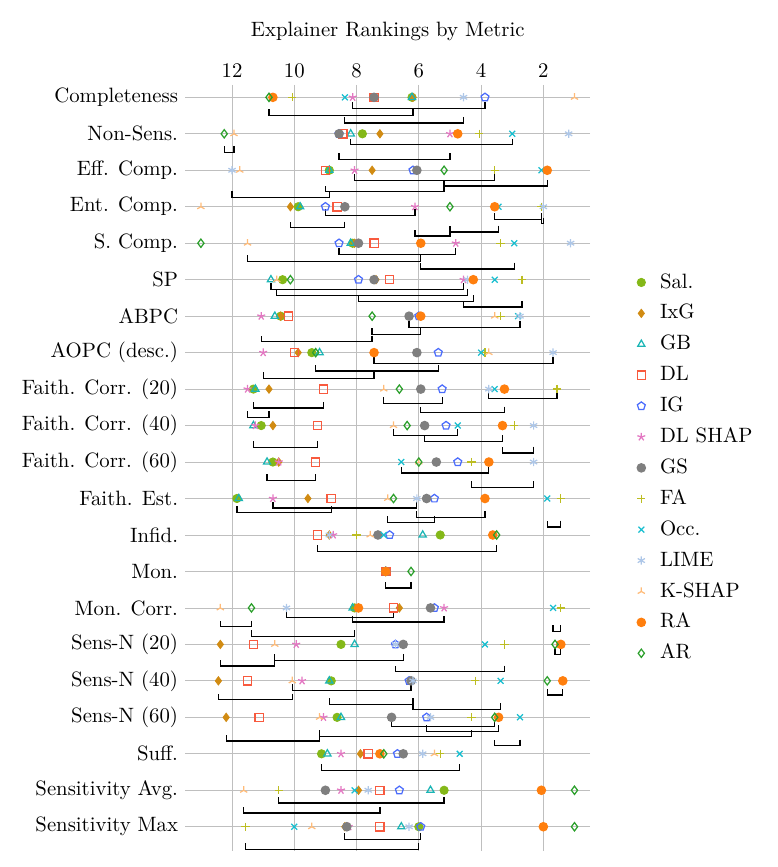}
        \caption{Semantic Entity Recognition}
        \label{fig:cd_ser}
    \end{subfigure}
    \begin{subfigure}[t]{0.3\textwidth}
        \centering
        \includegraphics[width=\textwidth,draft=false]{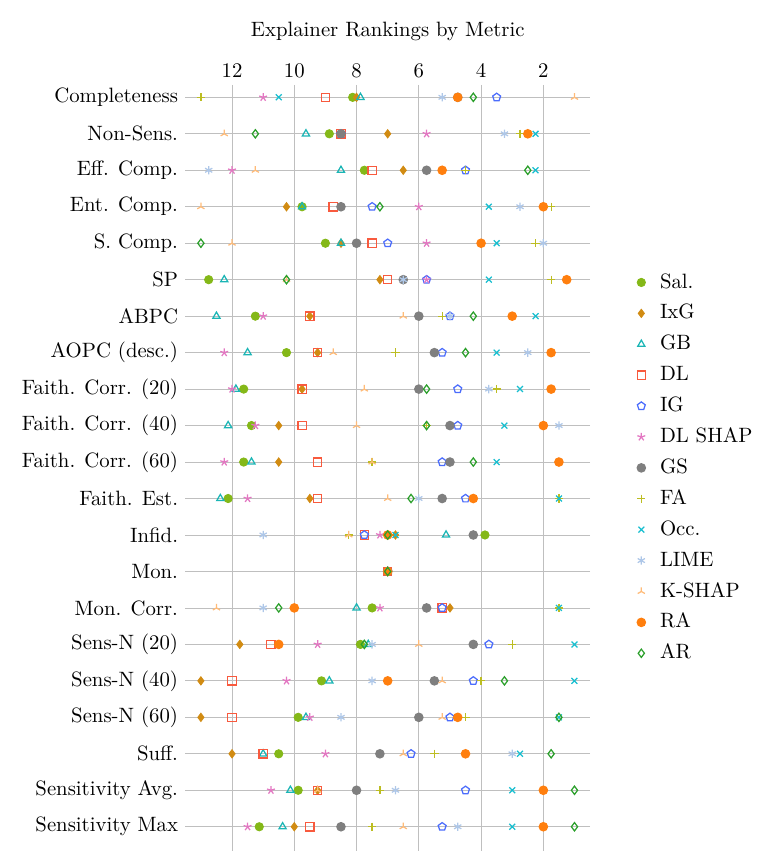}
        \caption{Document VQA}
        \label{fig:cd_vqa}
    \end{subfigure}
    \caption{Per-metric critical difference diagrams for each task. Methods connected by a horizontal bar are not statistically significantly different (Wilcoxon signed-rank test with Holm correction, $\alpha = 0.05$). Lower ranks indicate better performance.}
    \label{fig:cd_per_task}
\end{figure*}

Fig.~\ref{fig:cd_per_task} presents the critical difference diagrams for each task individually.
For the \qa{} task (Fig.~\ref{fig:cd_vqa}), only rankings are reported, as we were unable to perform pairwise significance testing due to the limited number of observations. The overall patterns largely mirror the aggregated results over all tasks reported in Fig.~\ref{fig:cd_overall}; however, the statistical indistinguishability among methods for some metrics was more pronounced on individual tasks. For instance, on the \ser{} task, both \monotonicity{} and \infidelity{} show large statistical indistinguishability across all methods, which suggests that these metrics do not provide any meaningful results for evaluation. Similar results can be seen on the \cls{} task for \monotonicity{} and \sparseness{} metrics.

\begin{figure*}[!ht]
    \centering
    \begin{subfigure}[t]{0.3\textwidth}
        \centering
        \includegraphics[width=\textwidth]{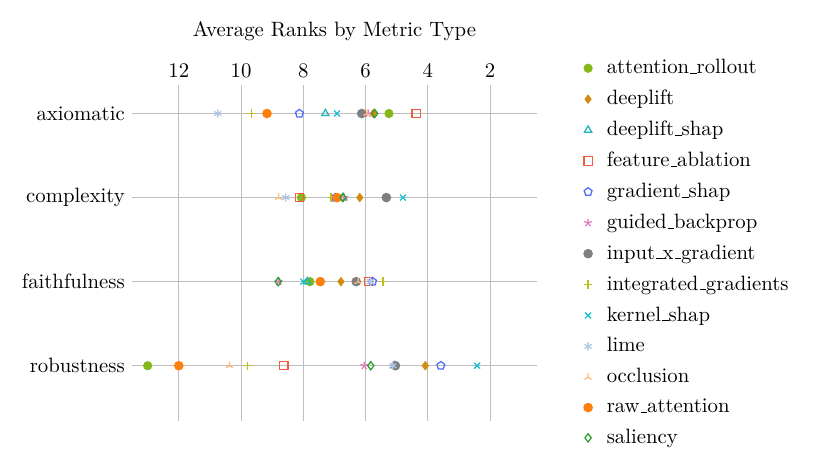}
        \caption{Classification}
        \label{fig:avg_rank_cls}
    \end{subfigure}
    \begin{subfigure}[t]{0.3\textwidth}
        \centering
        \includegraphics[width=\textwidth]{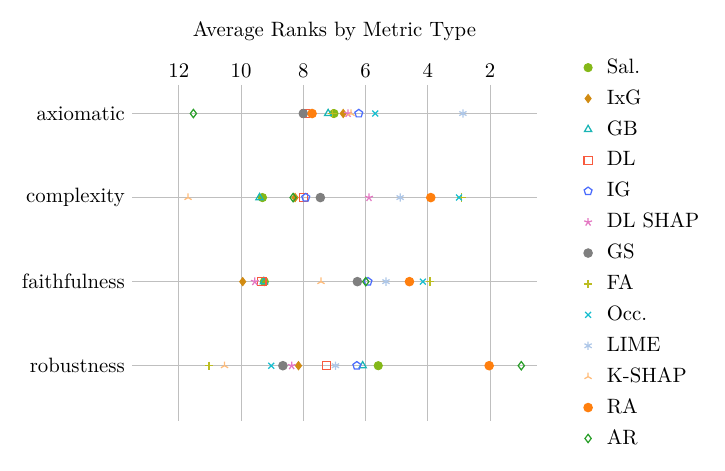}
        \caption{Semantic Entity Recognition}
        \label{fig:avg_rank_ser}
    \end{subfigure}
    \begin{subfigure}[t]{0.3\textwidth}
        \centering
        \includegraphics[width=\textwidth]{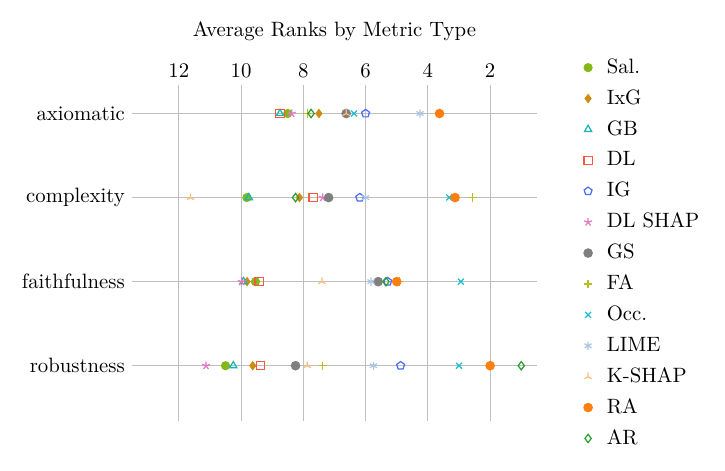}
        \caption{Document VQA}
        \label{fig:avg_rank_vqa}
    \end{subfigure}
    \caption{Aggregated mean ranks of XAI methods per evaluation criterion for each task. Lower ranks indicate better performance.}
    \label{fig:avg_rank_per_task}
\end{figure*}

Fig.~\ref{fig:avg_rank_per_task} shows the aggregated mean ranks per evaluation criterion for each task. The per-task patterns are broadly consistent with the overall aggregated ranking (Fig.~\ref{fig:avg_rank_overall}) with minor task-specific variations.

\begin{figure*}[t]
    \centering
    \begin{subfigure}[t]{0.3\textwidth}
        \centering
        \includegraphics[width=\textwidth,draft=false]{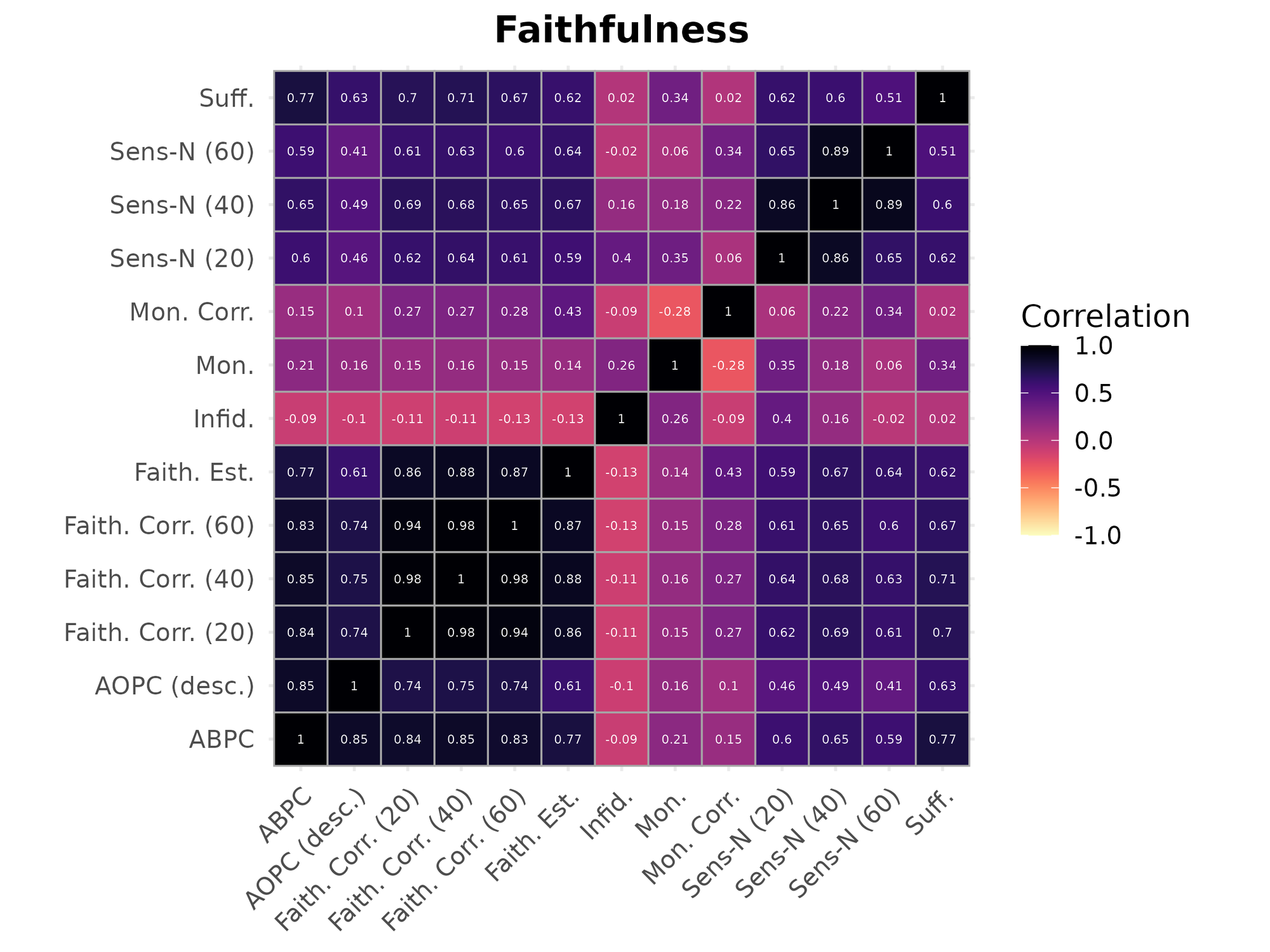}
        \caption{\cls{}}
        \label{fig:faithfulness_corr_cls}
    \end{subfigure}
    \begin{subfigure}[t]{0.3\textwidth}
        \centering
        \includegraphics[width=\textwidth,draft=false]{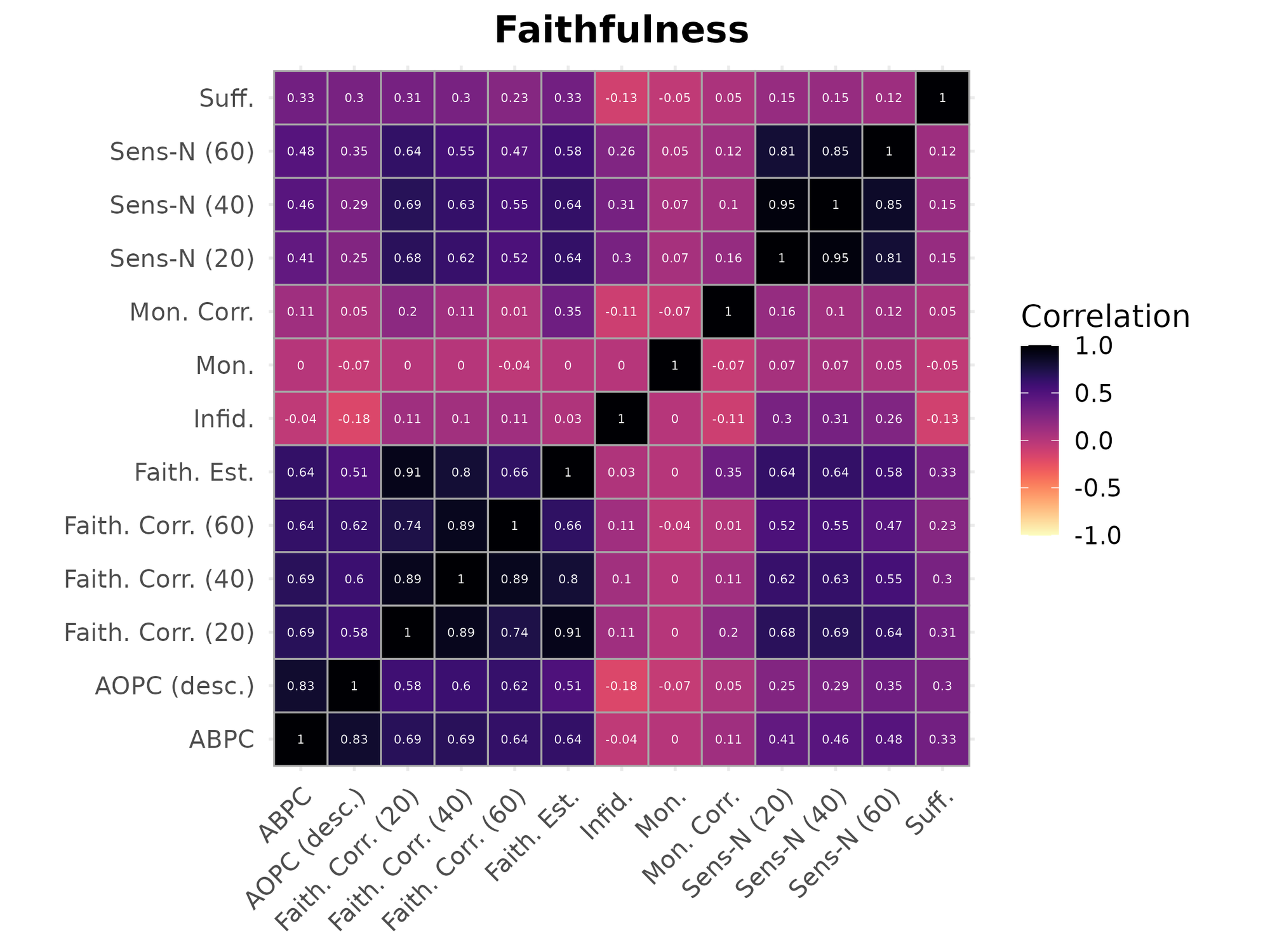}
        \caption{\ser{}}
        \label{fig:faithfulness_corr_ser}
    \end{subfigure}
    \begin{subfigure}[t]{0.3\textwidth}
        \centering
        \includegraphics[width=\textwidth,draft=false]{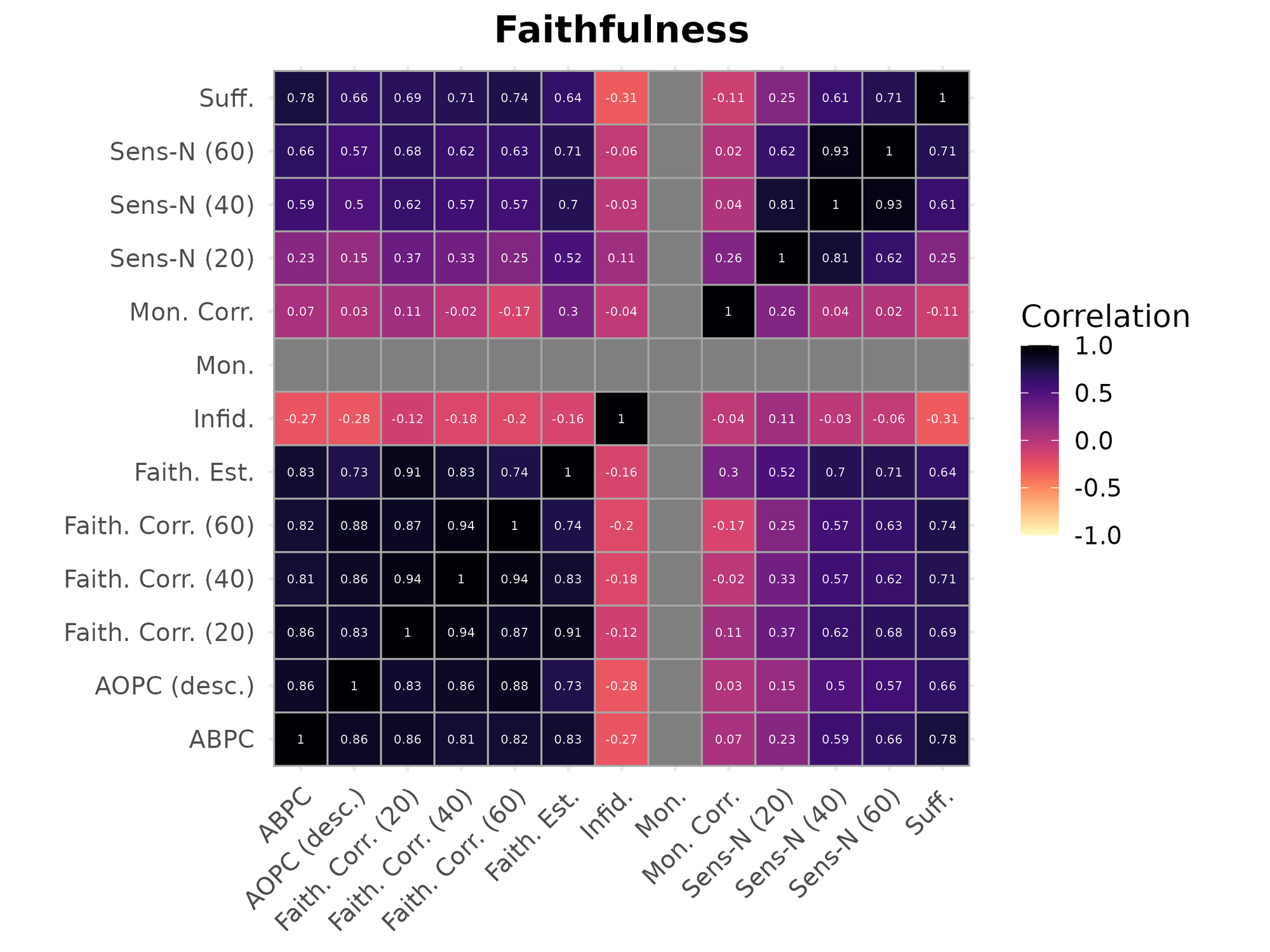}
        \caption{\qa{}}
        \label{fig:faithfulness_corr_qa}
    \end{subfigure}
\caption{Per-task pairwise Spearman correlation between faithfulness metric rankings across respective datasets, models, and explainer settings. On the \qa{} task, \monotonicityf{} produced undefined values and is therefore excluded.}
    \label{fig:corr_faithfulness_per_task}
\end{figure*}

\begin{figure*}[t]
    \centering
    \begin{subfigure}[t]{0.3\textwidth}
        \centering
        \includegraphics[width=\textwidth,draft=false]{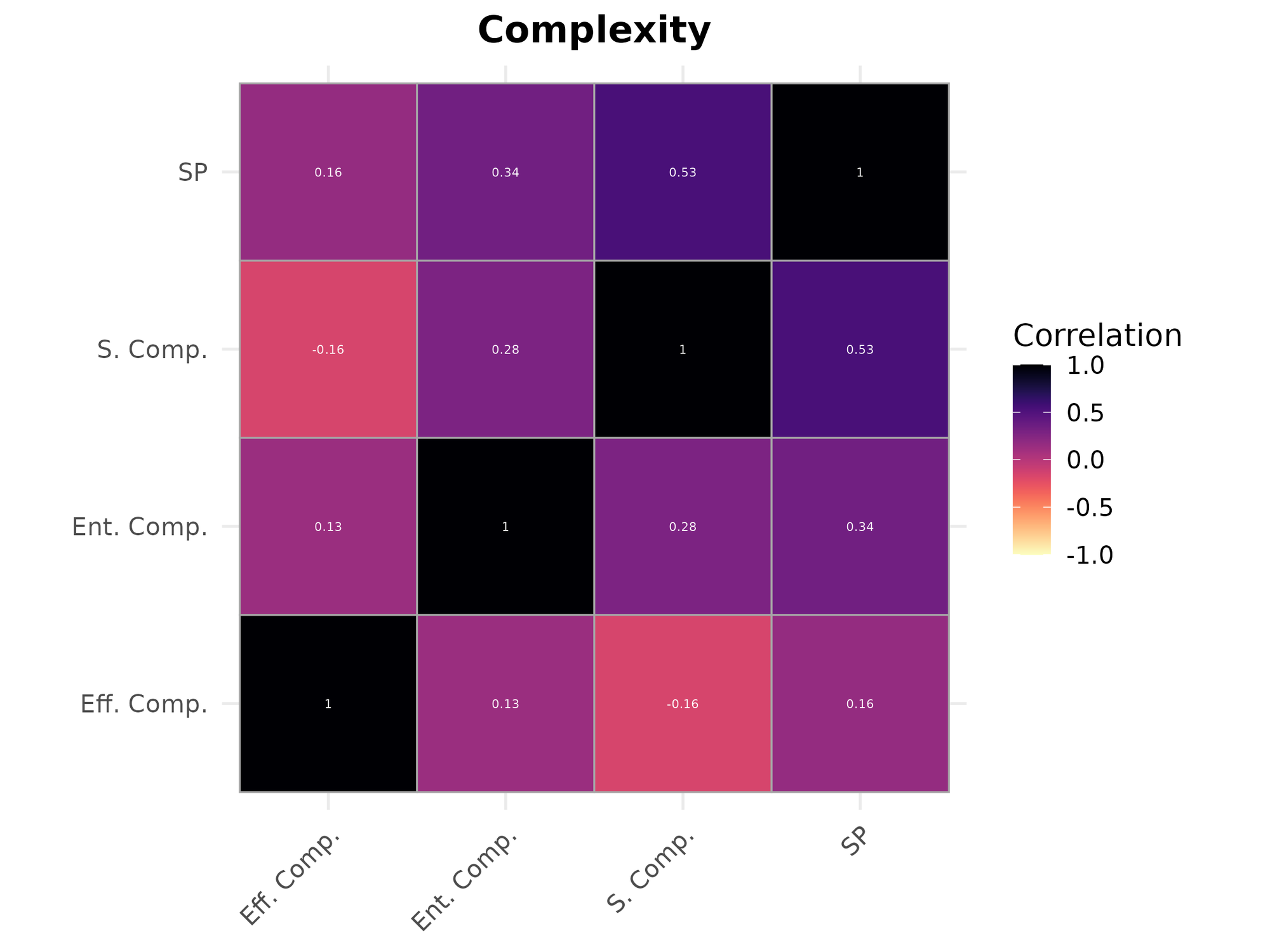}
        \caption{\cls{}}
        \label{fig:comp_corr_cls}
    \end{subfigure}
    \begin{subfigure}[t]{0.3\textwidth}
        \centering
        \includegraphics[width=\textwidth,draft=false]{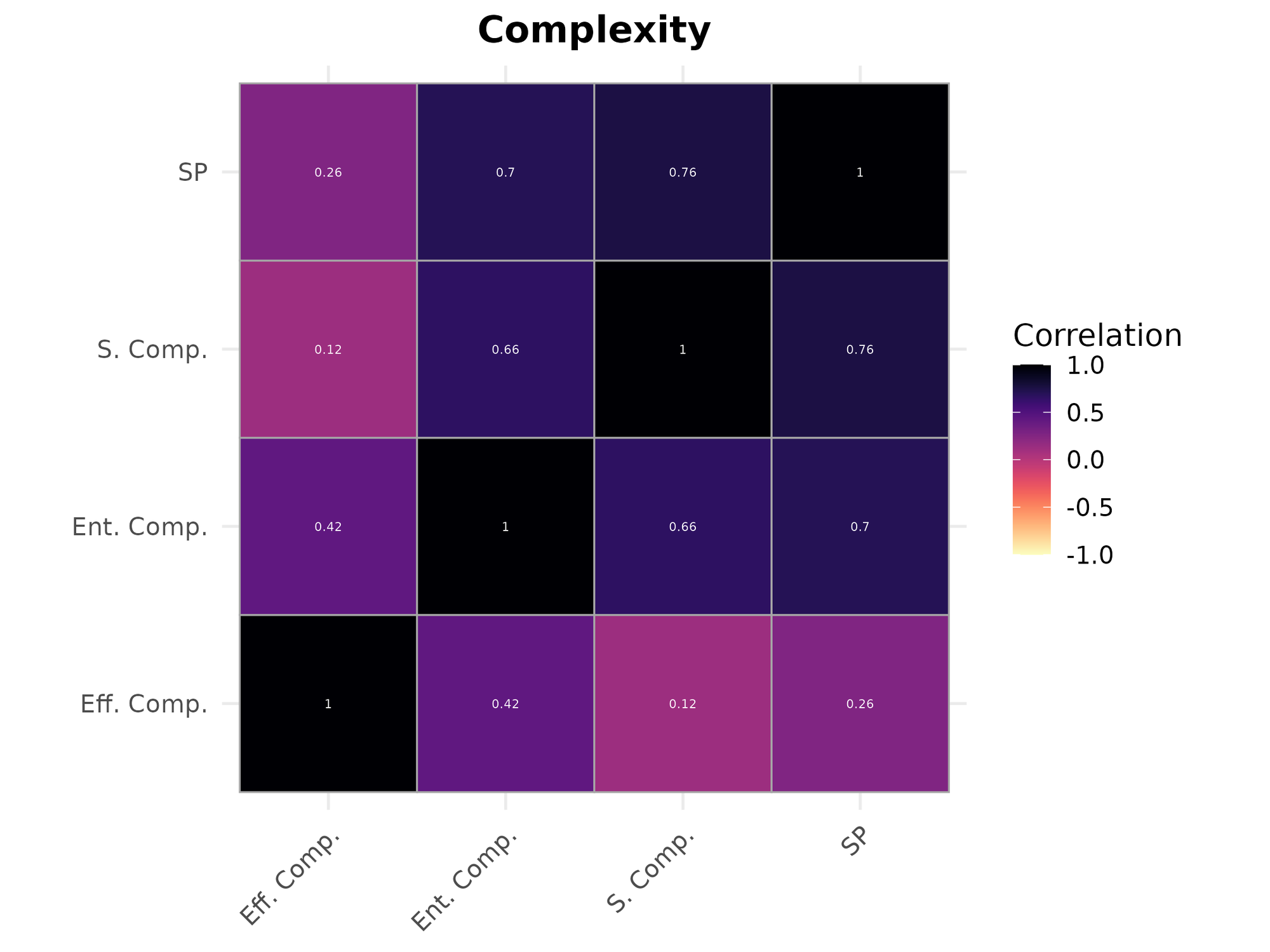}
        \caption{\ser{}}
        \label{fig:comp_corr_ser}
    \end{subfigure}
    \begin{subfigure}[t]{0.3\textwidth}
        \centering
        \includegraphics[width=\textwidth,draft=false]{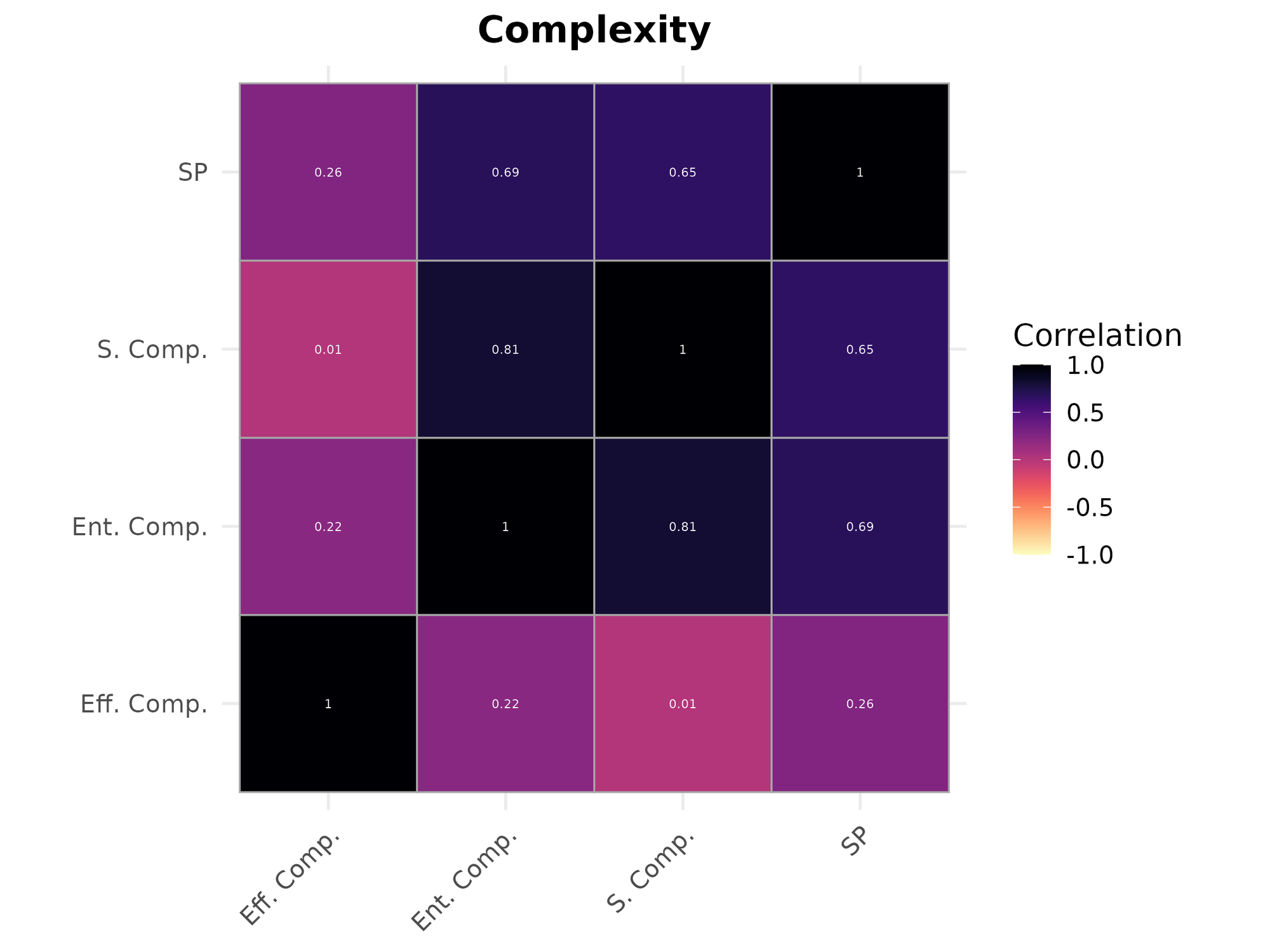}
        \caption{\qa{}}
        \label{fig:comp_corr_qa}
    \end{subfigure}
\caption{Per-task pairwise Spearman correlation between complexity metric rankings across respective datasets, models, and explainer settings.}
    \label{fig:corr_complexity_per_task}
\end{figure*}

\begin{figure*}[!t]
	\centering
	\includegraphics[width=0.7\textwidth]{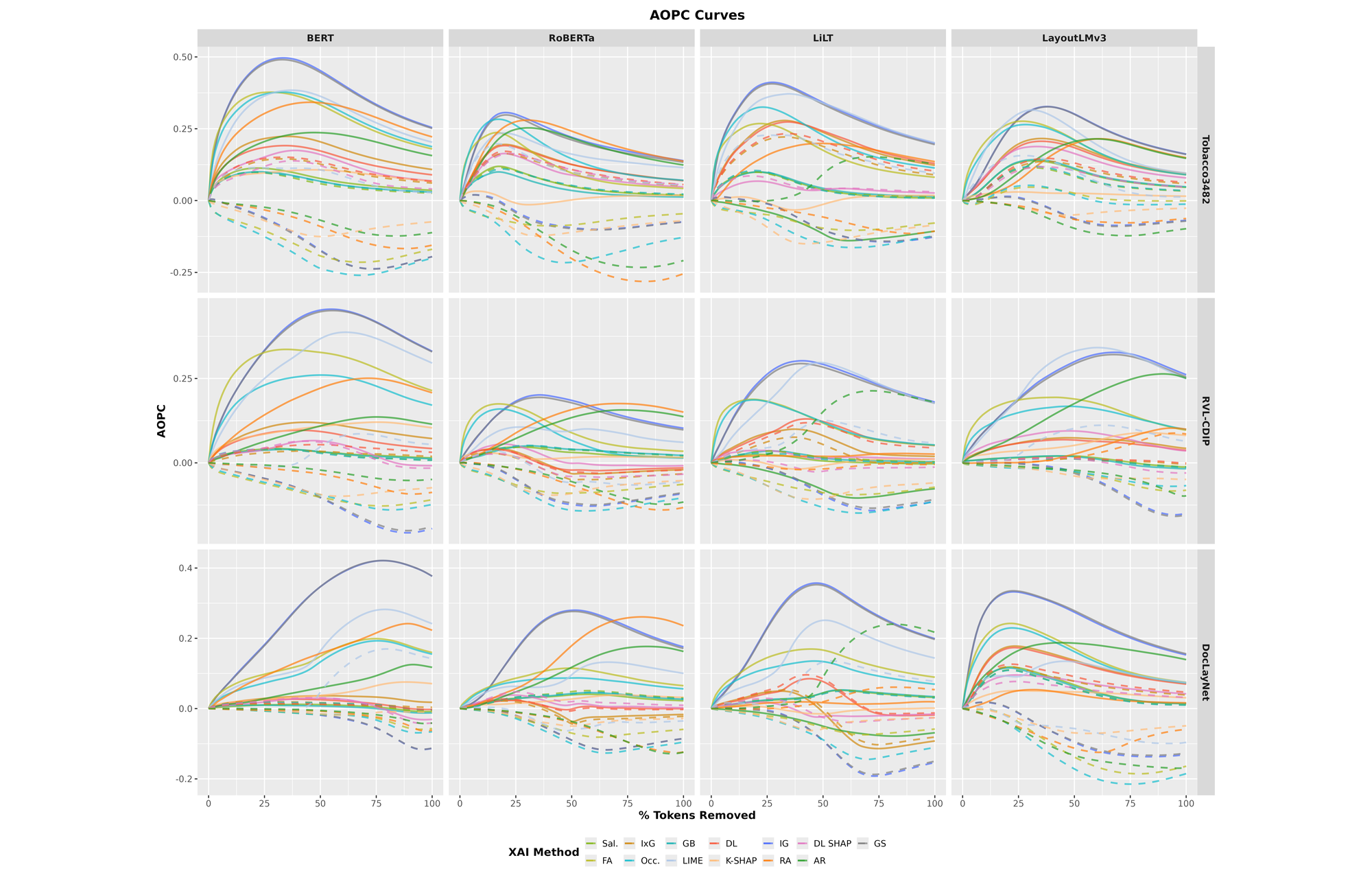}
	\caption{\aopc{} curves for all XAI methods across all dataset-model combinations on the \cls{} task. MoRF (solid) and LeRF (dotted) curves are shown relative to the random baseline.}
	\label{fig:aopc_cls}
\end{figure*}
\begin{figure*}[!t]
	\centering
	\includegraphics[width=0.7\textwidth]{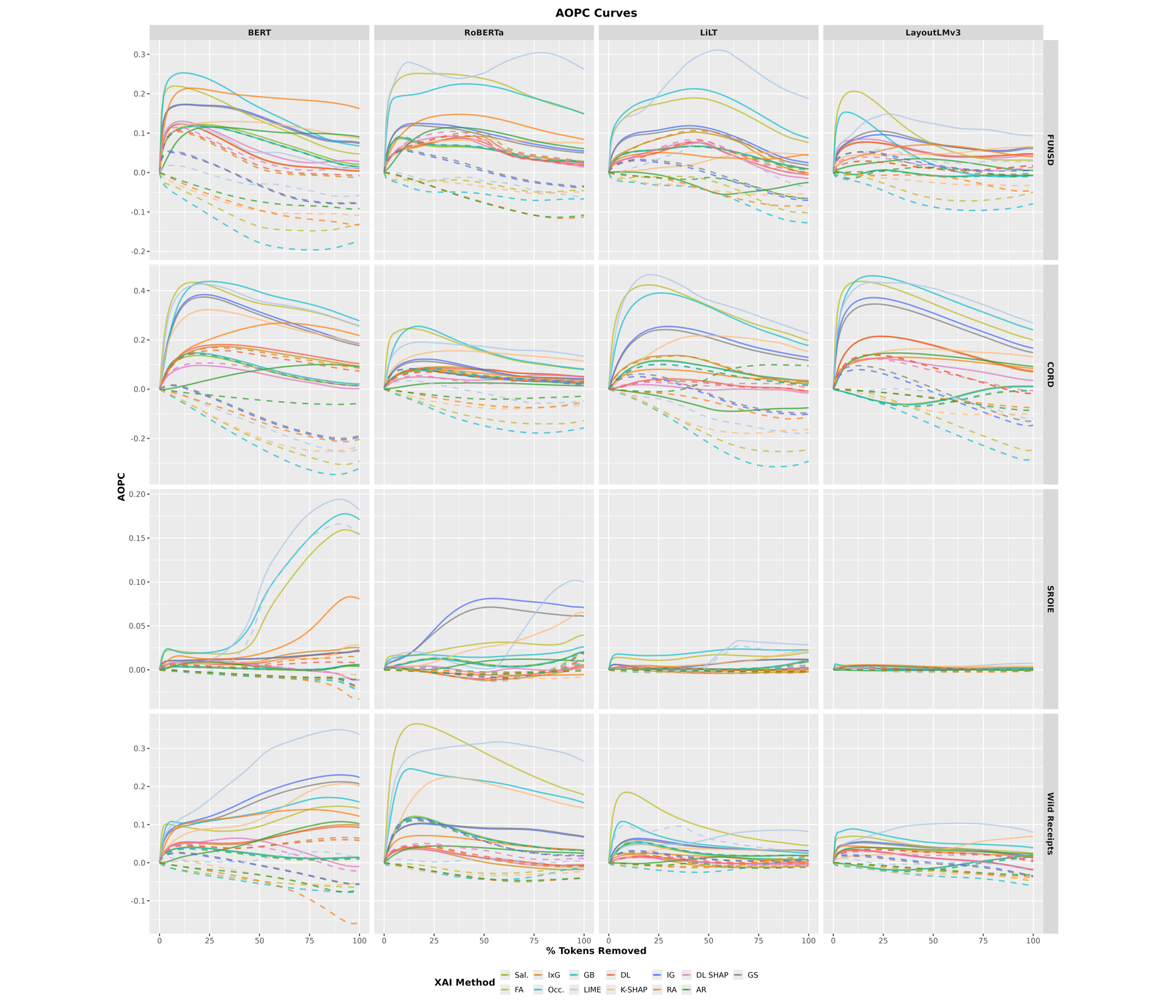}
	\caption{\aopc{} curves for all XAI methods across all dataset-model combinations on the \ser{} task. MoRF (solid) and LeRF (dotted) curves are shown relative to the random baseline.}
	\label{fig:aopc_ser}
\end{figure*}
\begin{figure*}[!t]
	\centering
	\includegraphics[width=0.7\textwidth]{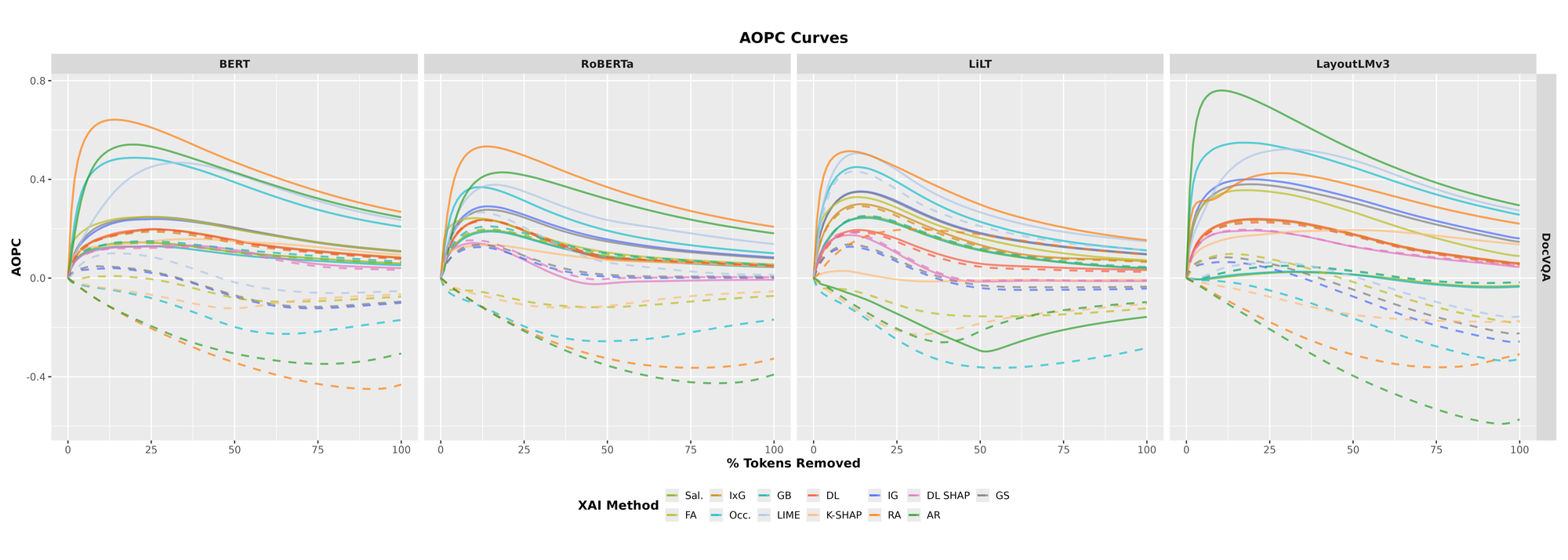}
	\caption{\aopc{} curves for all XAI methods across all dataset-model combinations on the \qa{} task. MoRF (solid) and LeRF (dotted) curves are shown relative to the random baseline.}
	\label{fig:aopc_vqa}
\end{figure*}
\begin{figure*}[htbp]
\centering
\noindent
\parbox[c]{1.5em}{\centering\rotatebox{90}{\small IG (RVL-CDIP)}}
\begin{subfigure}[c]{0.8\linewidth}
\includegraphics[width=\textwidth]{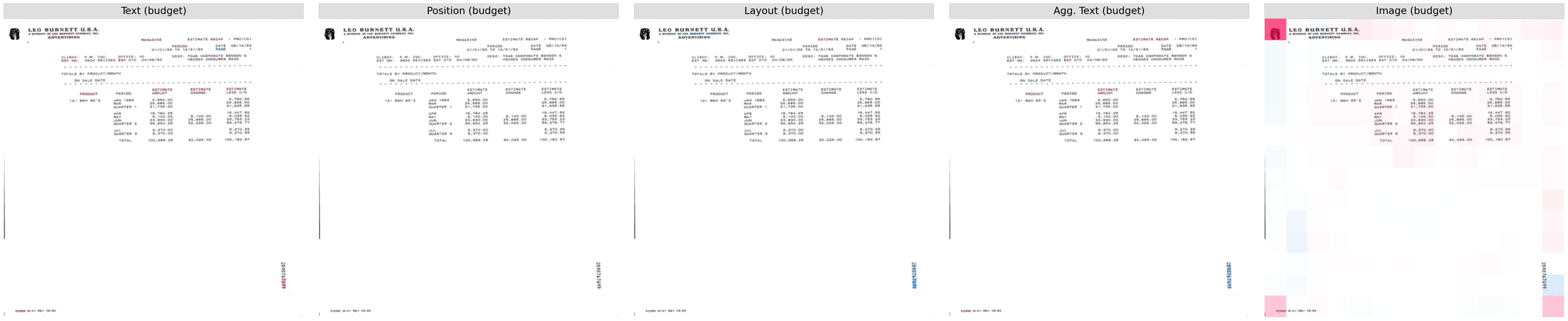}
\end{subfigure}
\par\smallskip
\noindent
\parbox[c]{1.5em}{\centering\rotatebox{90}{\small Occ. (RVL-CDIP)}}
\begin{subfigure}[c]{0.8\linewidth}
\includegraphics[width=\textwidth]{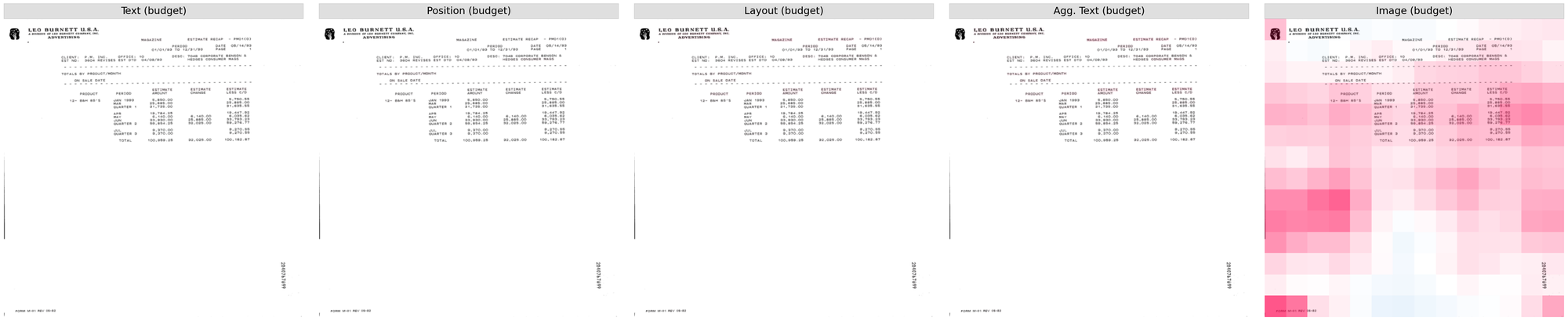}
\end{subfigure}
\par\smallskip
\noindent
\parbox[c]{1.5em}{\centering\rotatebox{90}{\small RA (RVL-CDIP)}}
\begin{subfigure}[c]{0.8\linewidth}
\includegraphics[width=0.5\textwidth]{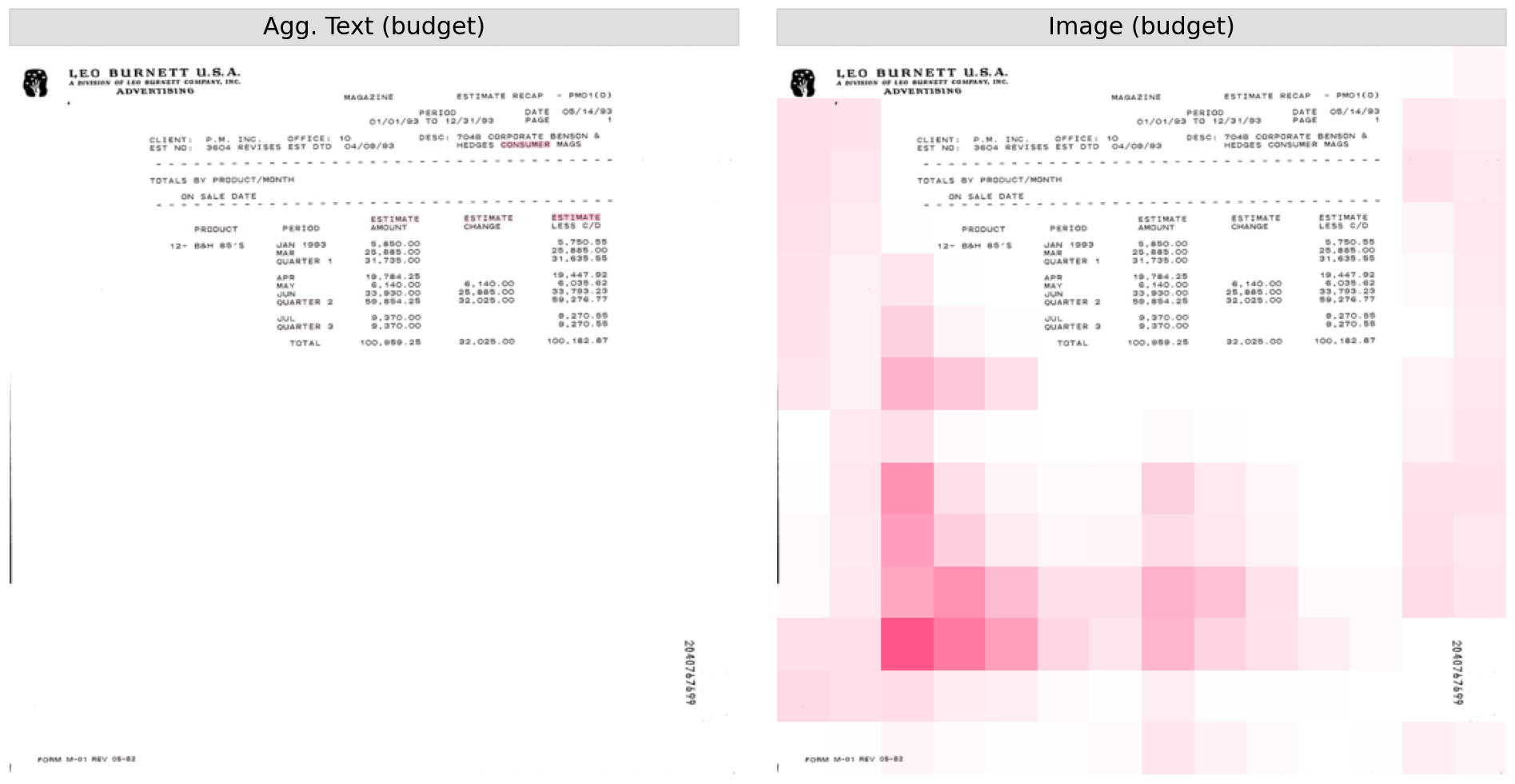}
\end{subfigure}
\par\smallskip
\noindent
\parbox[c]{1.5em}{\centering\rotatebox{90}{\small IG (DocLayNet)}}
\begin{subfigure}[c]{0.8\linewidth}
\includegraphics[width=\textwidth]{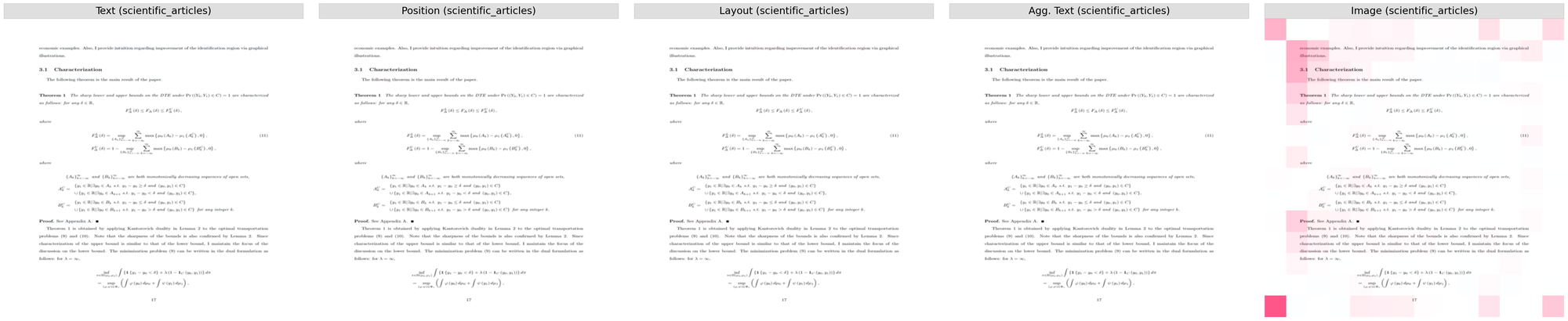}
\end{subfigure}
\par\smallskip
\noindent
\parbox[c]{1.5em}{\centering\rotatebox{90}{\small Occ. (DocLayNet)}}
\begin{subfigure}[c]{0.8\linewidth}
\includegraphics[width=\textwidth]{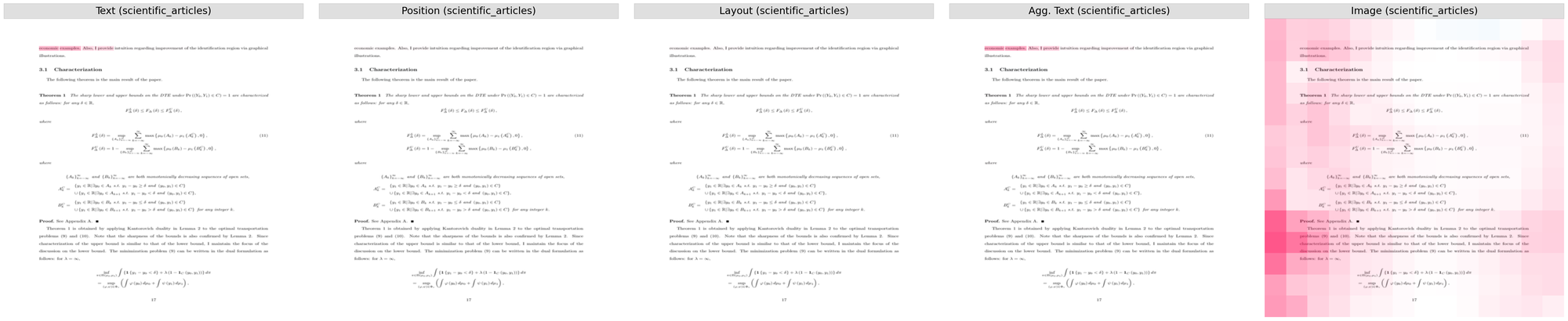}
\end{subfigure}
\par\smallskip
\noindent
\parbox[c]{1.5em}{\centering\rotatebox{90}{\small RA (DocLayNet)}}
\begin{subfigure}[c]{0.8\linewidth}
\includegraphics[width=0.5\textwidth]{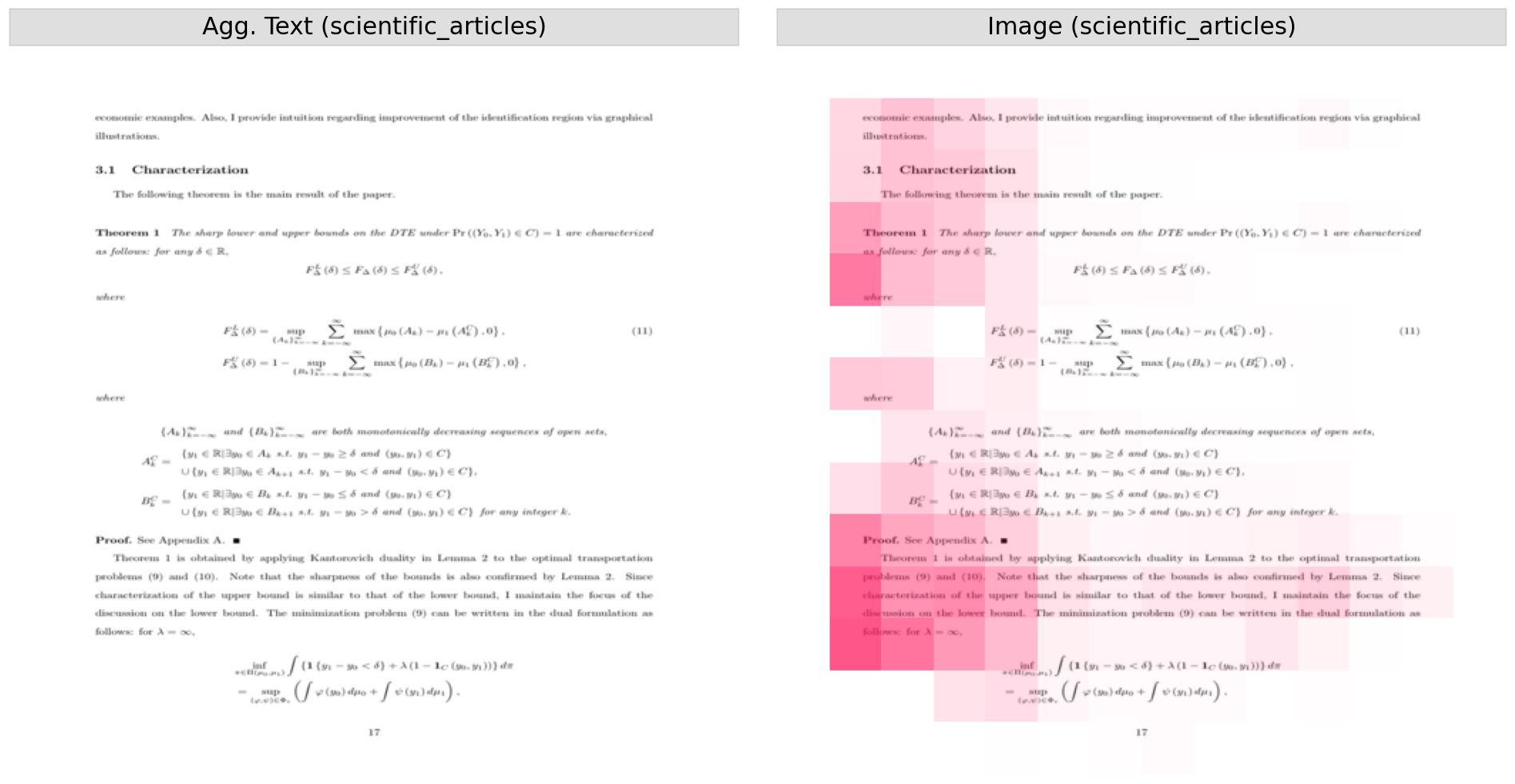}
\end{subfigure}
\caption{Modality-wise explanations of the \layoutlm{} model predictions for two randomly selected samples from \rvlcdip{} and \doclaynet{} (\cls{} task). Results are shown for three methods: \integratedgradientsf{}, \occlusionf{}, and \rawattnf{}.}
\label{fig:qual_cls_app_layoutlm}
\end{figure*}

\begin{figure*}[htbp]
\centering
\noindent
\parbox[c]{1.5em}{\centering\rotatebox{90}{\small IG (CORD)}}
\begin{subfigure}[c]{0.8\linewidth}
\includegraphics[width=\textwidth]{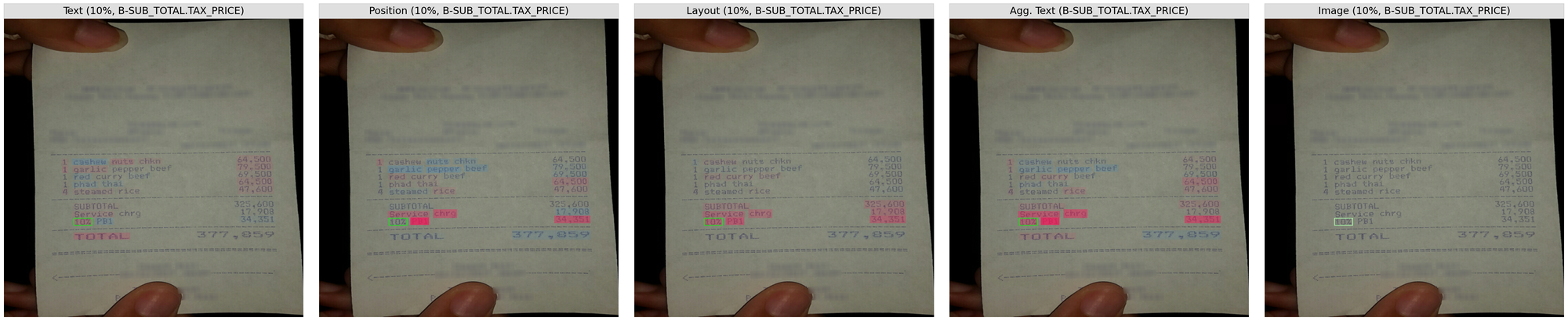}
\end{subfigure}
\par\smallskip
\noindent
\parbox[c]{1.5em}{\centering\rotatebox{90}{\small Occ. (CORD)}}
\begin{subfigure}[c]{0.8\linewidth}
\includegraphics[width=\textwidth]{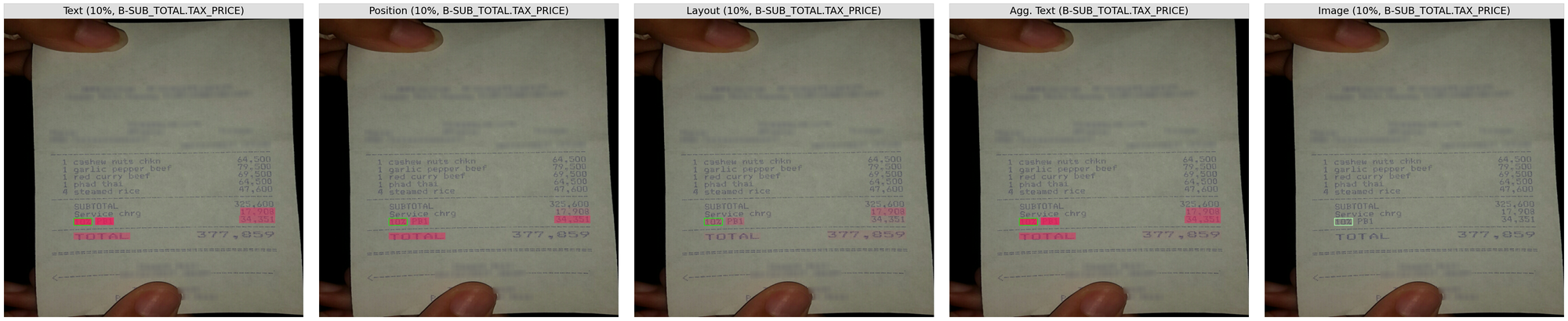}
\end{subfigure}
\par\smallskip
\noindent
\parbox[c]{1.5em}{\centering\rotatebox{90}{\small RA (CORD)}}
\begin{subfigure}[c]{0.8\linewidth}
\includegraphics[width=0.5\textwidth]{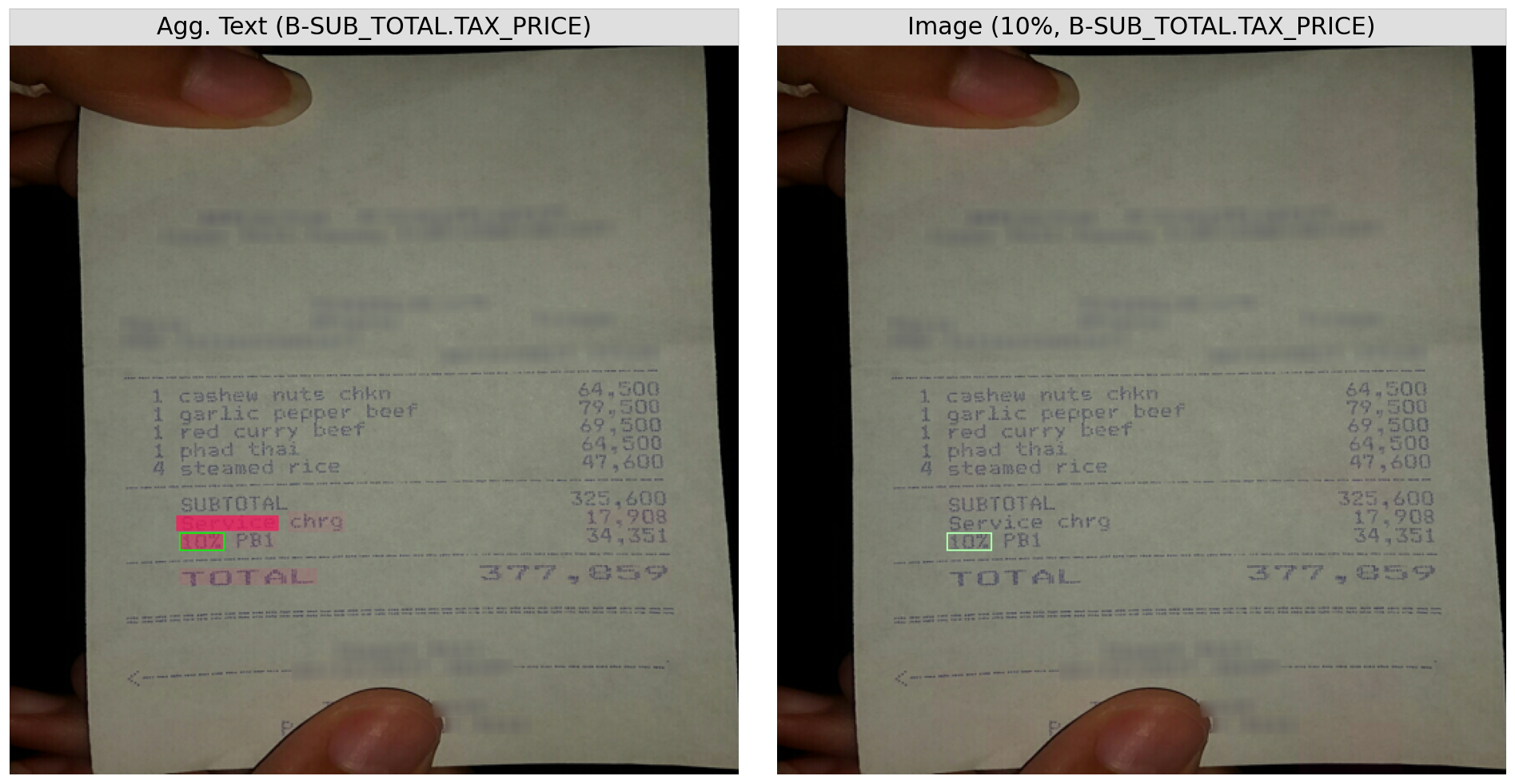}
\end{subfigure}
\par\smallskip
\noindent
\parbox[c]{1.5em}{\centering\rotatebox{90}{\small IG (FUNSD)}}
\begin{subfigure}[c]{0.8\linewidth}
\includegraphics[width=\textwidth]{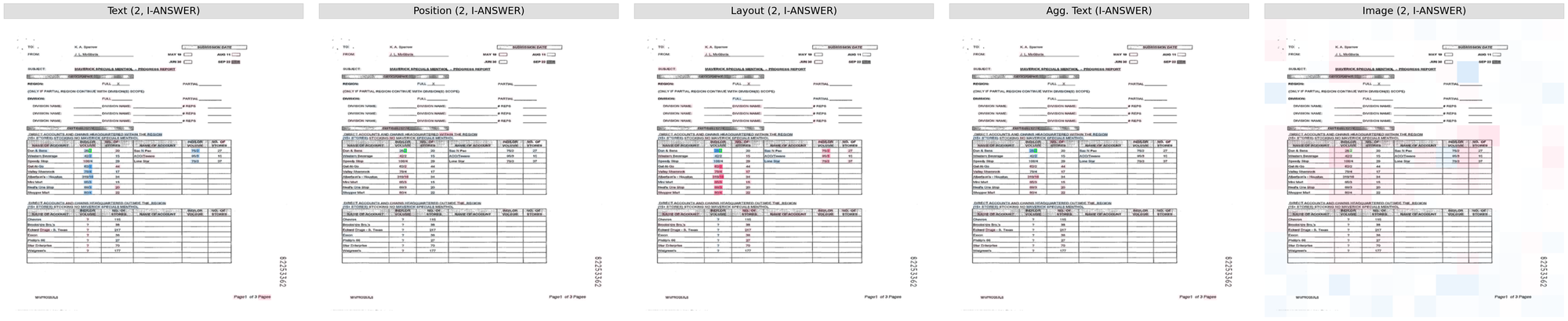}
\end{subfigure}
\par\smallskip
\noindent
\parbox[c]{1.5em}{\centering\rotatebox{90}{\small Occ. (FUNSD)}}
\begin{subfigure}[c]{0.8\linewidth}
\includegraphics[width=\textwidth]{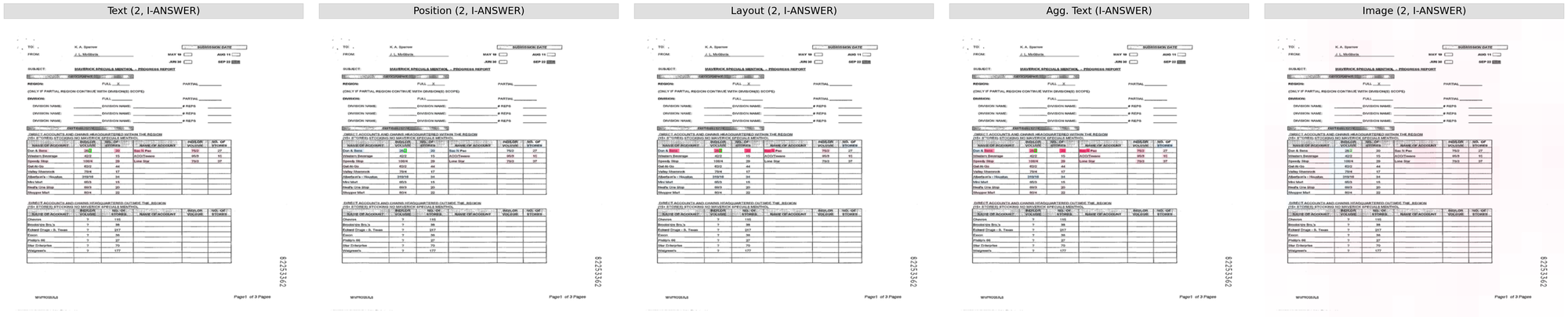}
\end{subfigure}
\par\smallskip
\noindent
\parbox[c]{1.5em}{\centering\rotatebox{90}{\small RA (FUNSD)}}
\begin{subfigure}[c]{0.8\linewidth}
\includegraphics[width=0.5\textwidth]{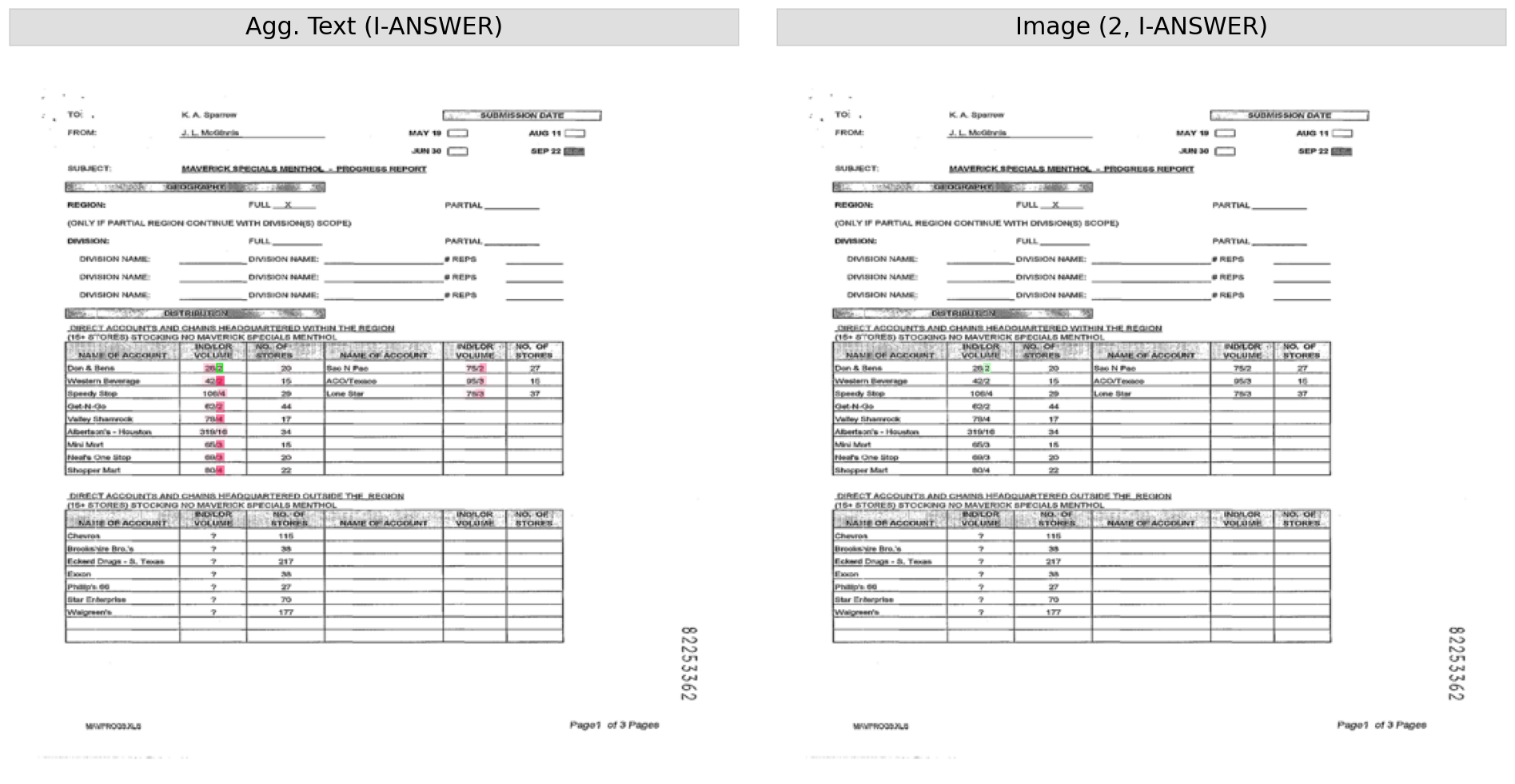}
\end{subfigure}
\caption{Modality-wise explanations of the \layoutlm{} model predictions for two randomly selected samples from \cord{} and \funsd{} (\ser{} task). Results are shown for three methods: \integratedgradientsf{}, \occlusionf{}, and \rawattnf{}.}
\label{fig:qual_ser_app_layoutlm}
\end{figure*}

\begin{figure*}[htbp]
\centering
\noindent
\parbox[c]{1.5em}{\centering\rotatebox{90}{\small IG (DocVQA)}}
\begin{subfigure}[c]{0.8\linewidth}
\includegraphics[width=\textwidth]{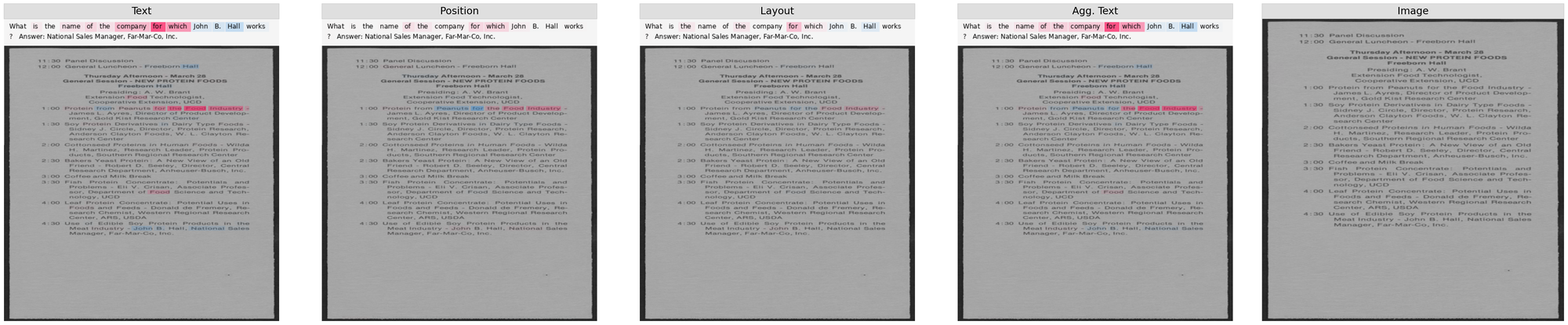}
\end{subfigure}
\par\smallskip
\noindent
\parbox[c]{1.5em}{\centering\rotatebox{90}{\small Occ. (DocVQA)}}
\begin{subfigure}[c]{0.8\linewidth}
\includegraphics[width=\textwidth]{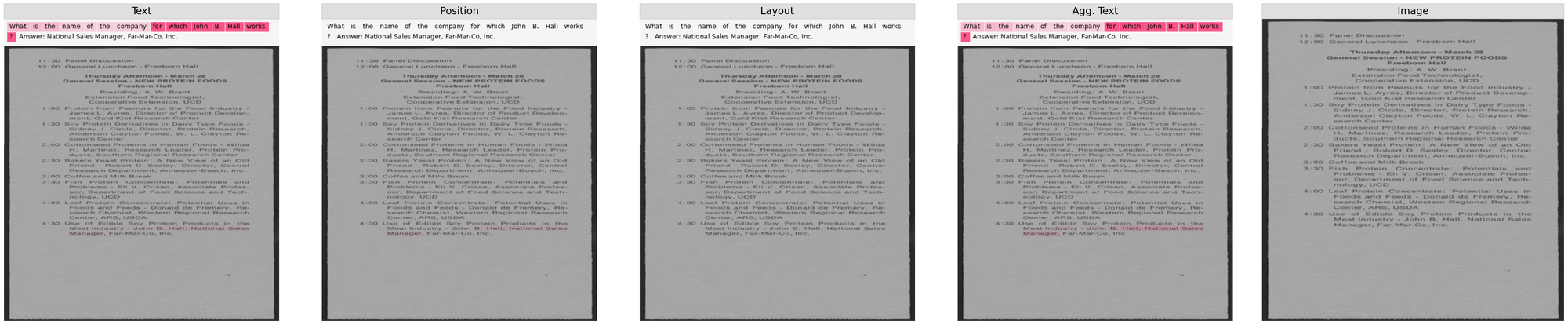}
\end{subfigure}
\par\smallskip
\noindent
\parbox[c]{1.5em}{\centering\rotatebox{90}{\small RA (DocVQA)}}
\begin{subfigure}[c]{0.8\linewidth}
\includegraphics[width=0.5\textwidth]{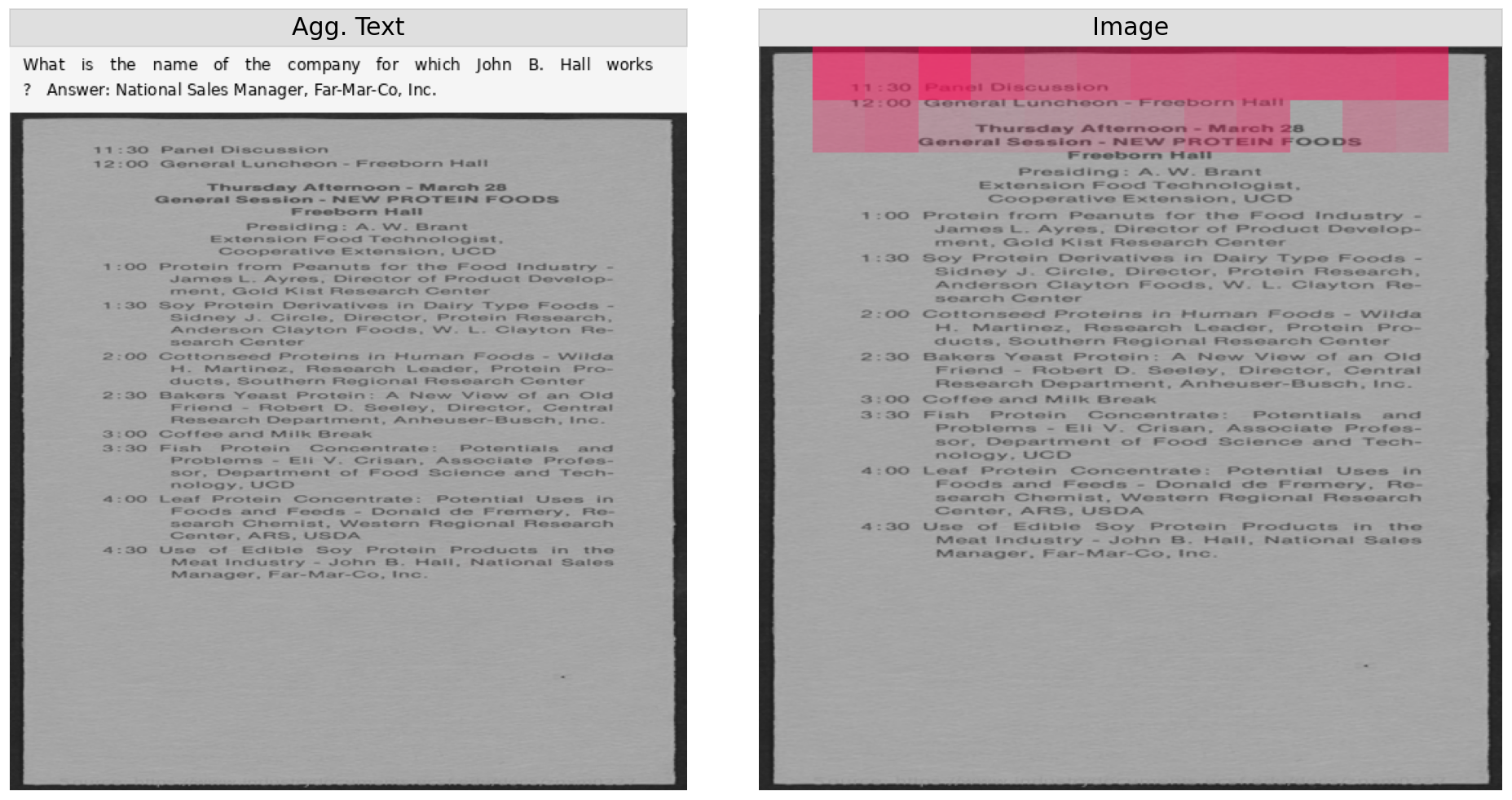}
\end{subfigure}
\par\smallskip
\noindent
\parbox[c]{1.5em}{\centering\rotatebox{90}{\small IG (DocVQA)}}
\begin{subfigure}[c]{0.8\linewidth}
\includegraphics[width=\textwidth]{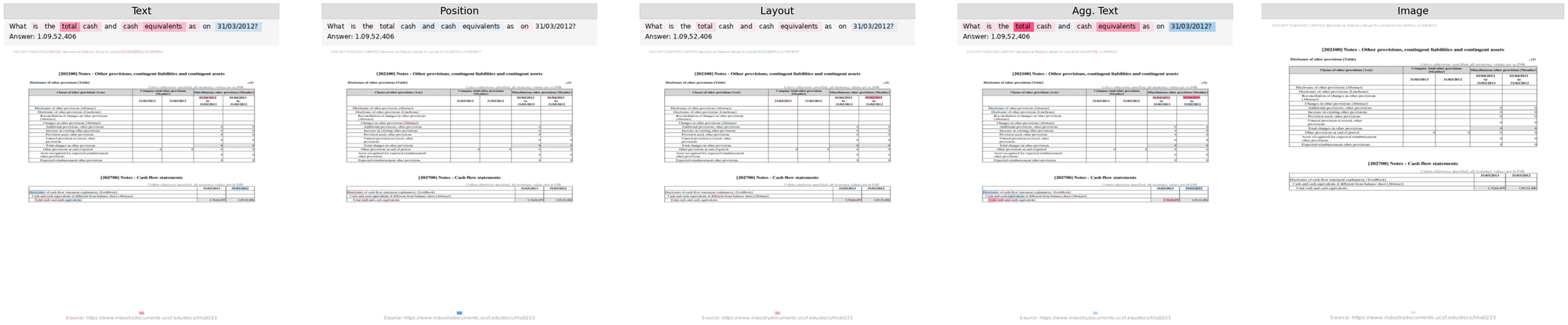}
\end{subfigure}
\par\smallskip
\noindent
\parbox[c]{1.5em}{\centering\rotatebox{90}{\small Occ. (DocVQA)}}
\begin{subfigure}[c]{0.8\linewidth}
\includegraphics[width=\textwidth]{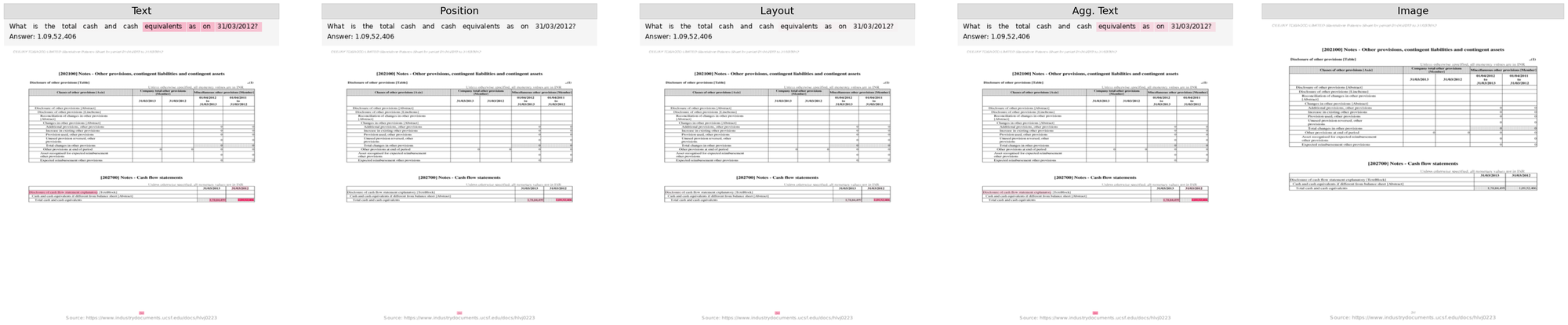}
\end{subfigure}
\par\smallskip
\noindent
\parbox[c]{1.5em}{\centering\rotatebox{90}{\small RA (DocVQA)}}
\begin{subfigure}[c]{0.8\linewidth}
\includegraphics[width=0.5\textwidth]{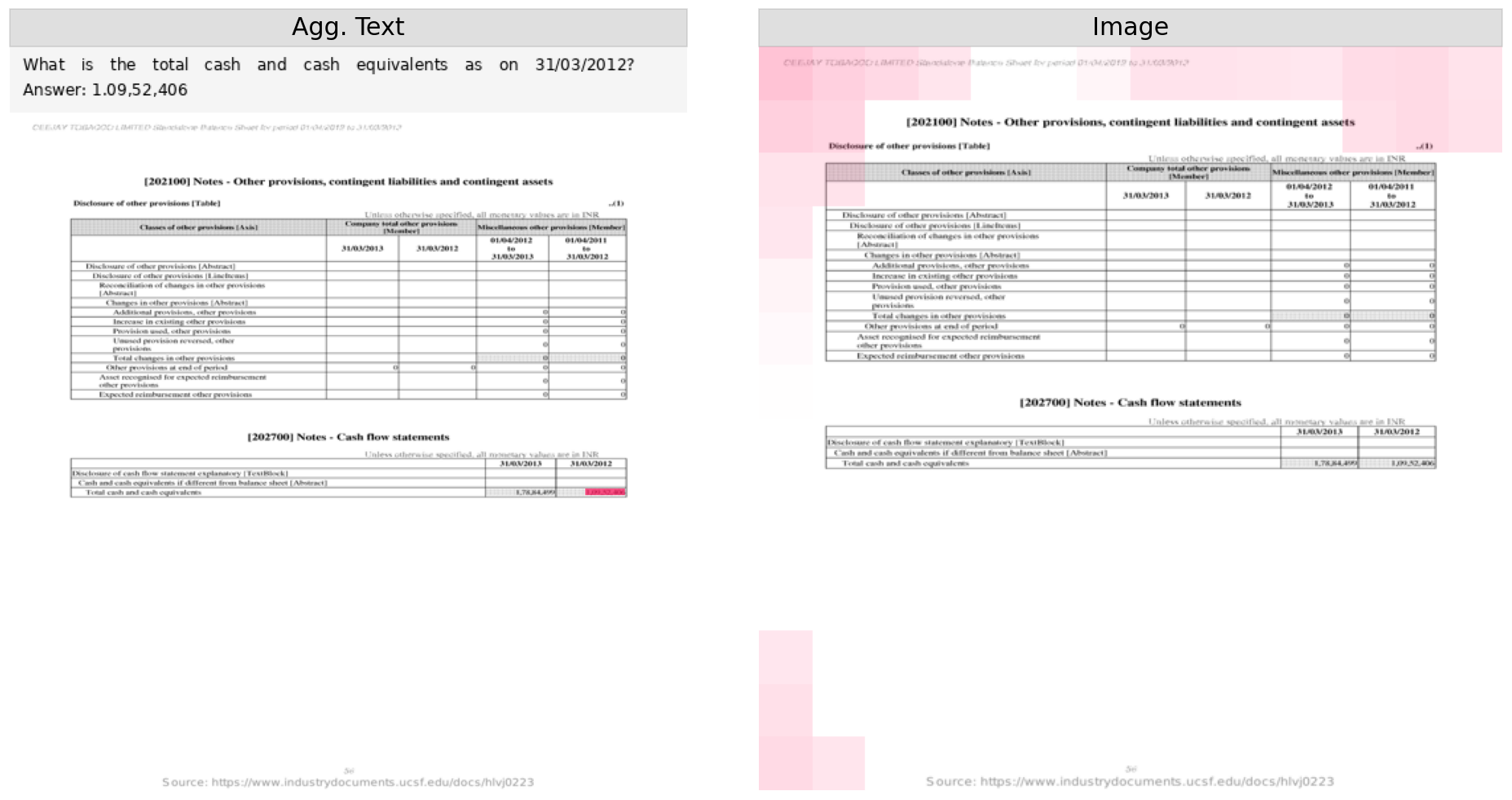}
\end{subfigure}
\caption{Modality-wise explanations of the \layoutlm{} model predictions for two randomly selected samples from \docvqa{} (\qa{} task). Results are shown for three methods: \integratedgradientsf{}, \occlusionf{}, and \rawattnf{}.}
\label{fig:qual_vqa_app_layoutlm}
\end{figure*}

\begin{figure*}[htbp]
\centering
\noindent
\parbox[c]{1.5em}{\centering\rotatebox{90}{\small IG (RVL-CDIP)}}
\begin{subfigure}[c]{0.8\linewidth}
\includegraphics[width=\textwidth]{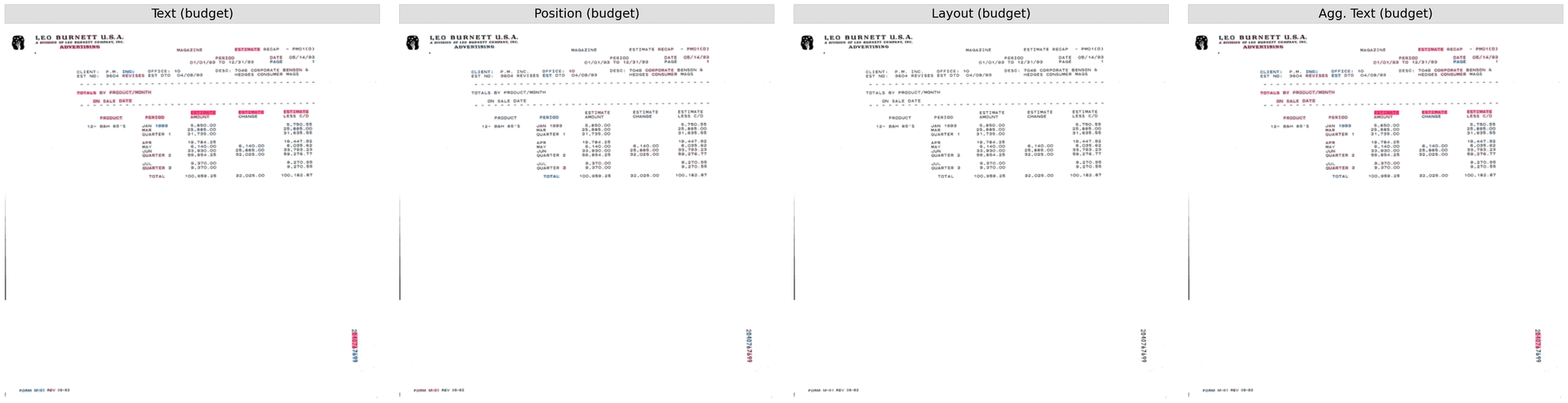}
\end{subfigure}
\par\smallskip
\noindent
\parbox[c]{1.5em}{\centering\rotatebox{90}{\small Occ. (RVL-CDIP)}}
\begin{subfigure}[c]{0.8\linewidth}
\includegraphics[width=\textwidth]{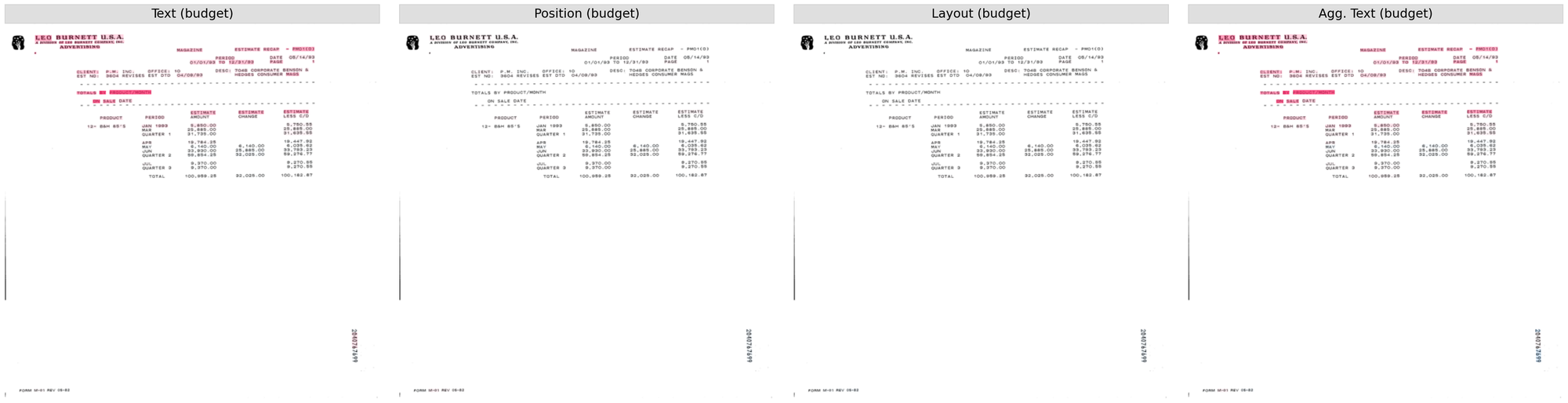}
\end{subfigure}
\par\smallskip
\noindent
\parbox[c]{1.5em}{\centering\rotatebox{90}{\small RA (RVL-CDIP)}}
\begin{subfigure}[c]{0.8\linewidth}
\includegraphics[width=0.25\textwidth]{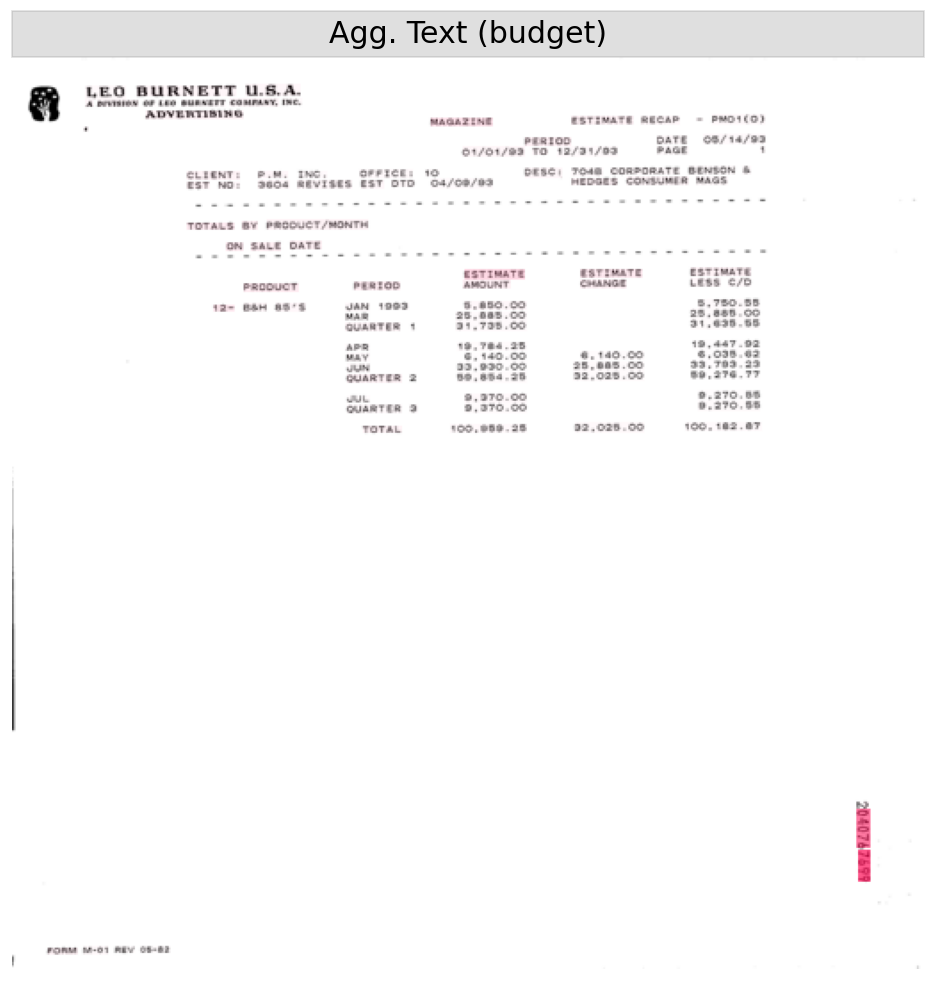}
\end{subfigure}
\par\smallskip
\noindent
\parbox[c]{1.5em}{\centering\rotatebox{90}{\small IG (DocLayNet)}}
\begin{subfigure}[c]{0.8\linewidth}
\includegraphics[width=\textwidth]{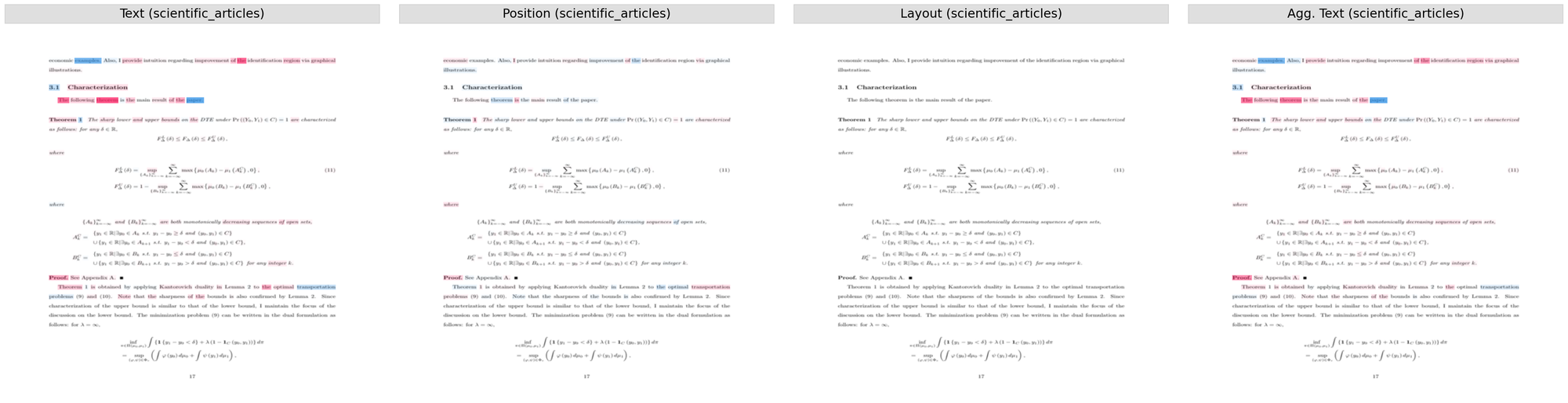}
\end{subfigure}
\par\smallskip
\noindent
\parbox[c]{1.5em}{\centering\rotatebox{90}{\small Occ. (DocLayNet)}}
\begin{subfigure}[c]{0.8\linewidth}
\includegraphics[width=\textwidth]{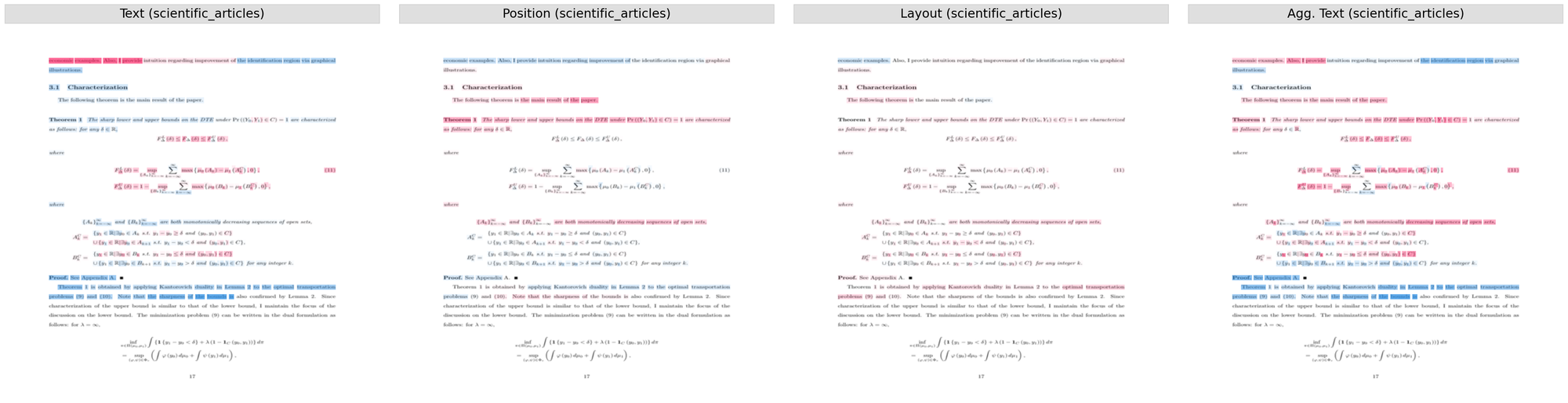}
\end{subfigure}
\par\smallskip
\noindent
\parbox[c]{1.5em}{\centering\rotatebox{90}{\small RA (DocLayNet)}}
\begin{subfigure}[c]{0.8\linewidth}
\includegraphics[width=0.25\textwidth]{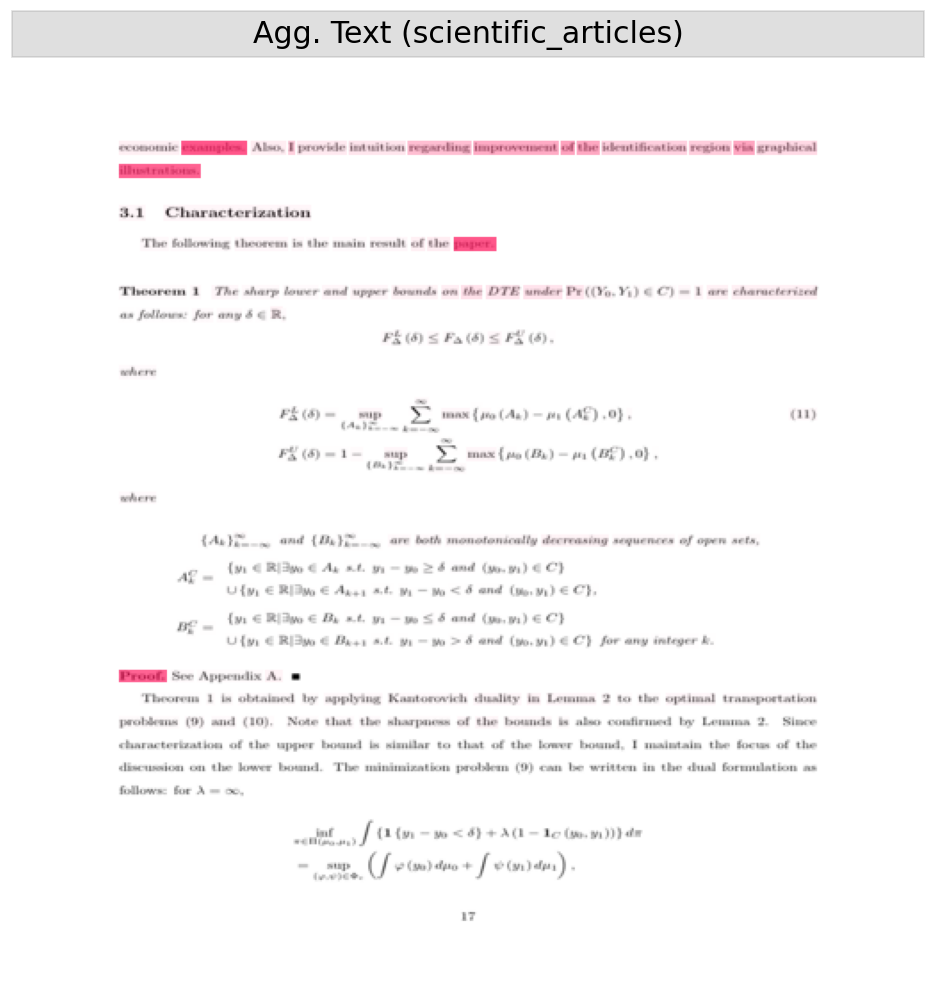}
\end{subfigure}
\caption{Modality-wise explanations of the \lilt{} model predictions for two randomly selected samples from \rvlcdip{} and \doclaynet{} (\cls{} task). Results are shown for three methods: \integratedgradientsf{}, \occlusionf{}, and \rawattnf{}.}
\label{fig:qual_cls_app_lilt}
\end{figure*}

\begin{figure*}[htbp]
\centering
\noindent
\parbox[c]{1.5em}{\centering\rotatebox{90}{\small IG (CORD)}}
\begin{subfigure}[c]{0.8\linewidth}
\includegraphics[width=\textwidth]{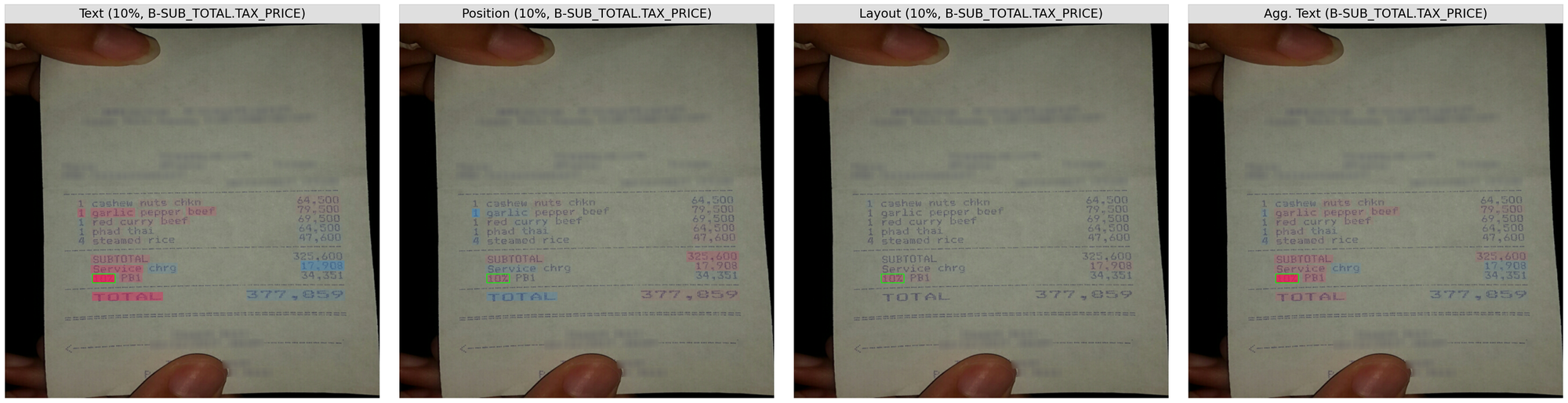}
\end{subfigure}
\par\smallskip
\noindent
\parbox[c]{1.5em}{\centering\rotatebox{90}{\small Occ. (CORD)}}
\begin{subfigure}[c]{0.8\linewidth}
\includegraphics[width=\textwidth]{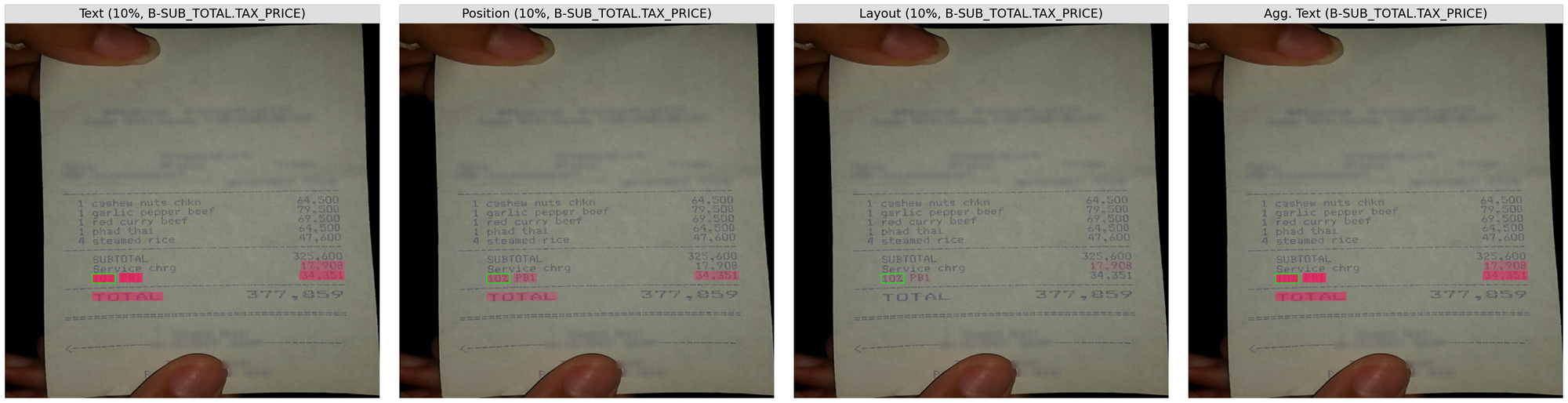}
\end{subfigure}
\par\smallskip
\noindent
\parbox[c]{1.5em}{\centering\rotatebox{90}{\small RA (CORD)}}
\begin{subfigure}[c]{0.8\linewidth}
\includegraphics[width=0.25\textwidth]{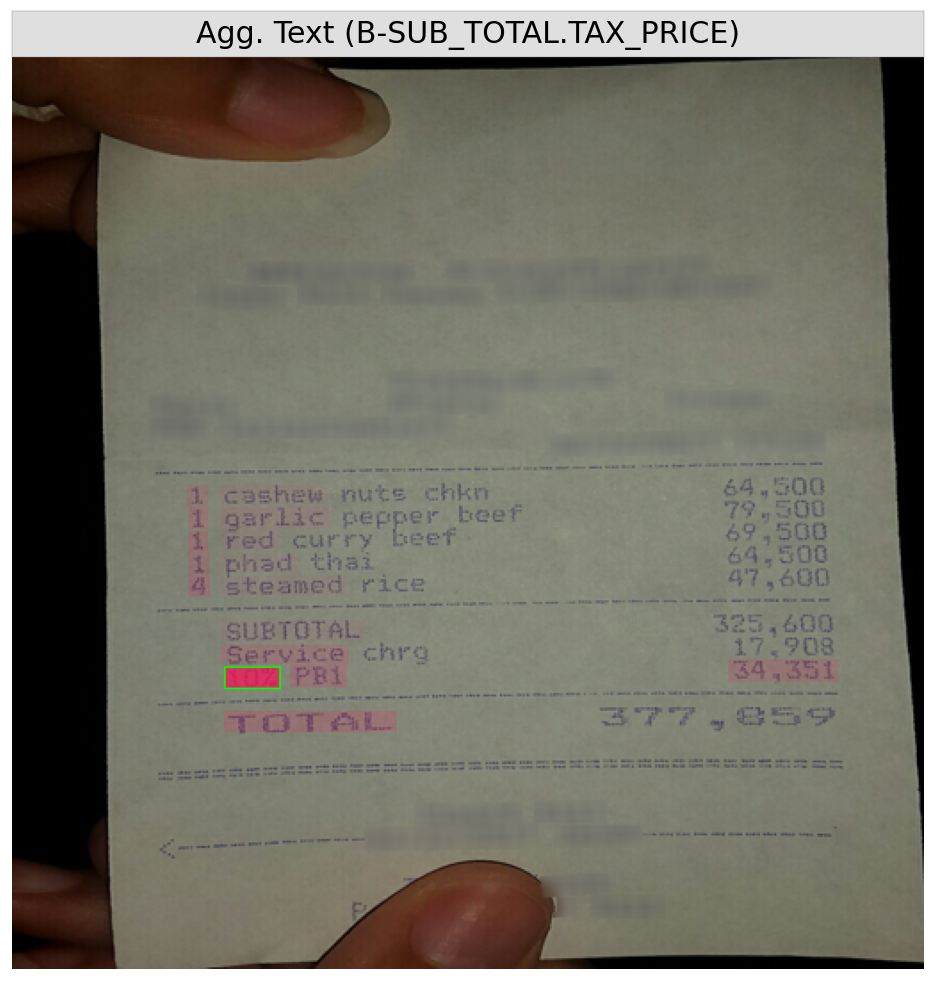}
\end{subfigure}
\par\smallskip
\noindent
\parbox[c]{1.5em}{\centering\rotatebox{90}{\small IG (FUNSD)}}
\begin{subfigure}[c]{0.8\linewidth}
\includegraphics[width=\textwidth]{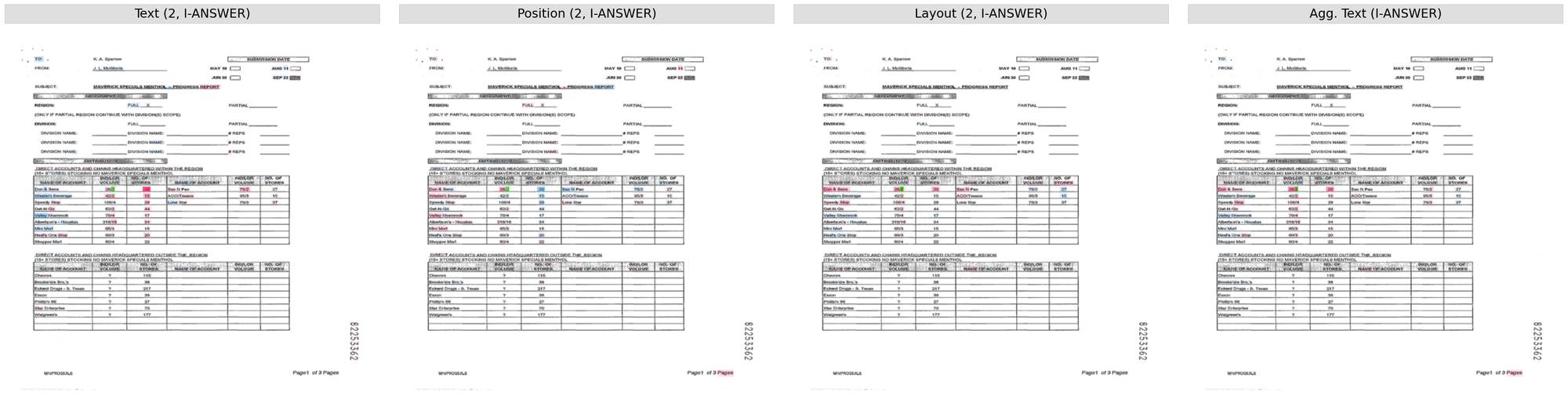}
\end{subfigure}
\par\smallskip
\noindent
\parbox[c]{1.5em}{\centering\rotatebox{90}{\small Occ. (FUNSD)}}
\begin{subfigure}[c]{0.8\linewidth}
\includegraphics[width=\textwidth]{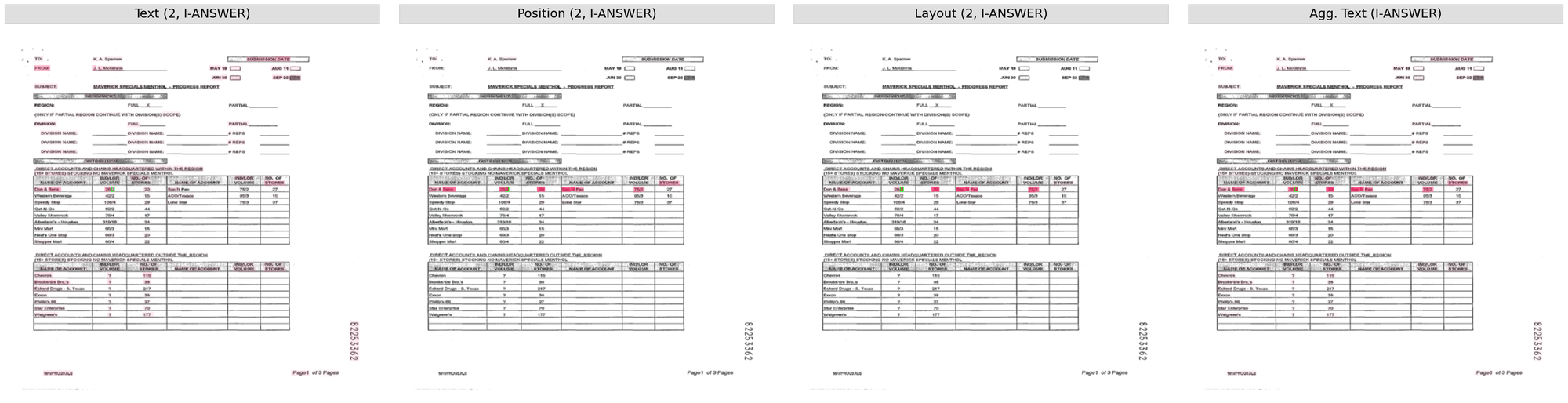}
\end{subfigure}
\par\smallskip
\noindent
\parbox[c]{1.5em}{\centering\rotatebox{90}{\small RA (FUNSD)}}
\begin{subfigure}[c]{0.8\linewidth}
\includegraphics[width=0.25\textwidth]{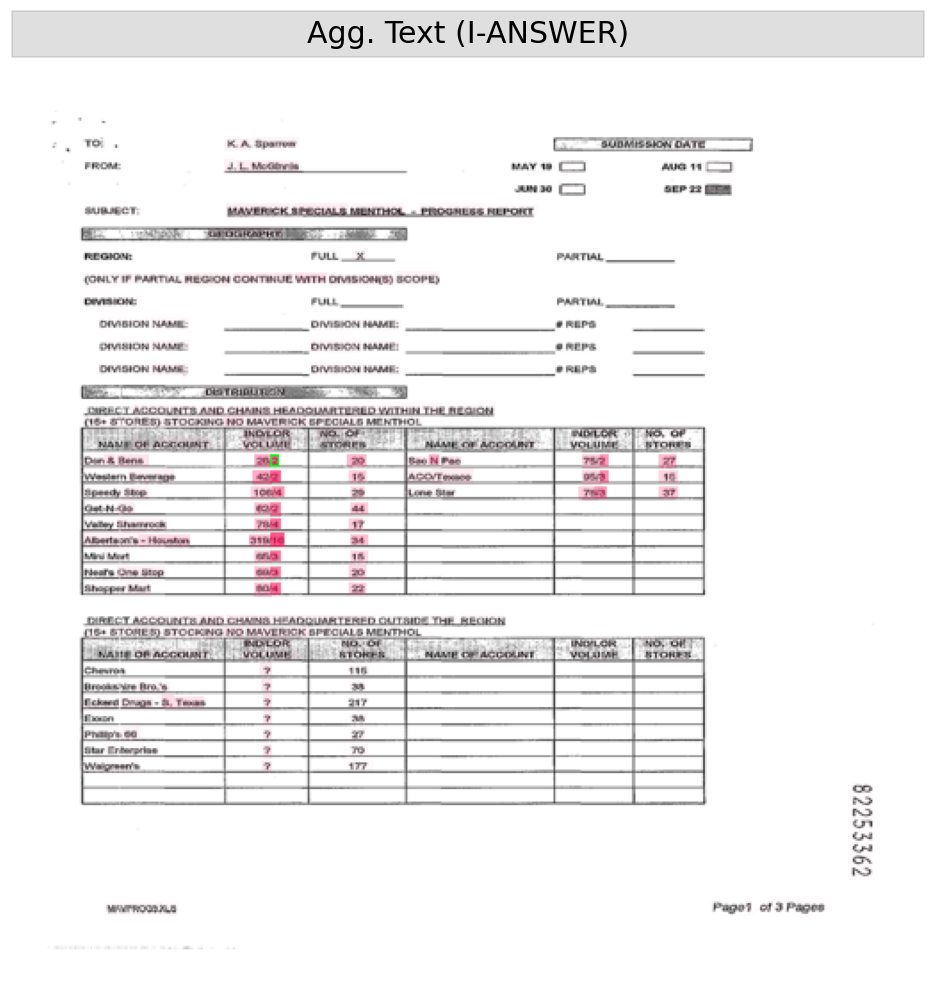}
\end{subfigure}
\caption{Modality-wise explanations of the \lilt{} model predictions for two randomly selected samples from \cord{} and \funsd{} (\ser{} task). Results are shown for three methods: \integratedgradientsf{}, \occlusionf{}, and \rawattnf{}.}
\label{fig:qual_ser_app_lilt}
\end{figure*}

\begin{figure*}[htbp]
\centering
\noindent
\parbox[c]{1.5em}{\centering\rotatebox{90}{\small IG (DocVQA)}}
\begin{subfigure}[c]{0.8\linewidth}
\includegraphics[width=\textwidth]{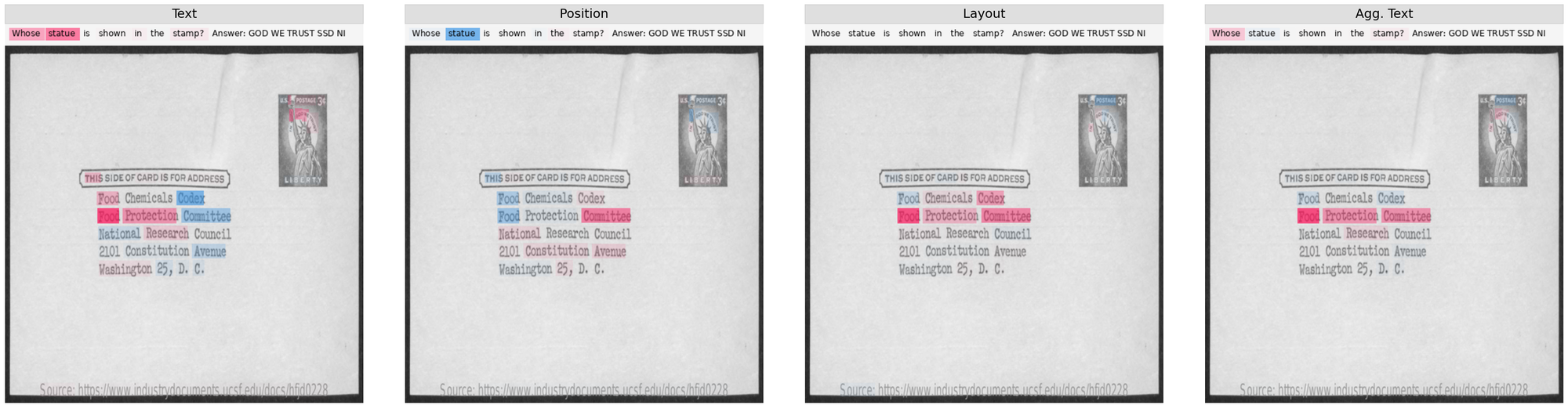}
\end{subfigure}
\par\smallskip
\noindent
\parbox[c]{1.5em}{\centering\rotatebox{90}{\small Occ. (DocVQA)}}
\begin{subfigure}[c]{0.8\linewidth}
\includegraphics[width=\textwidth]{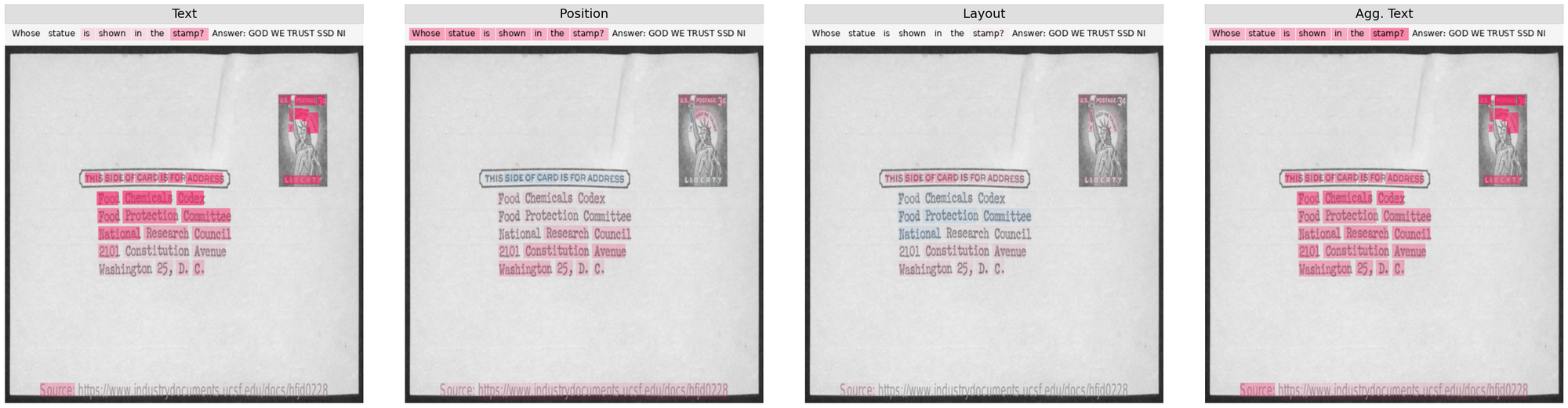}
\end{subfigure}
\par\smallskip
\noindent
\parbox[c]{1.5em}{\centering\rotatebox{90}{\small RA (DocVQA)}}
\begin{subfigure}[c]{0.8\linewidth}
\includegraphics[width=0.25\textwidth]{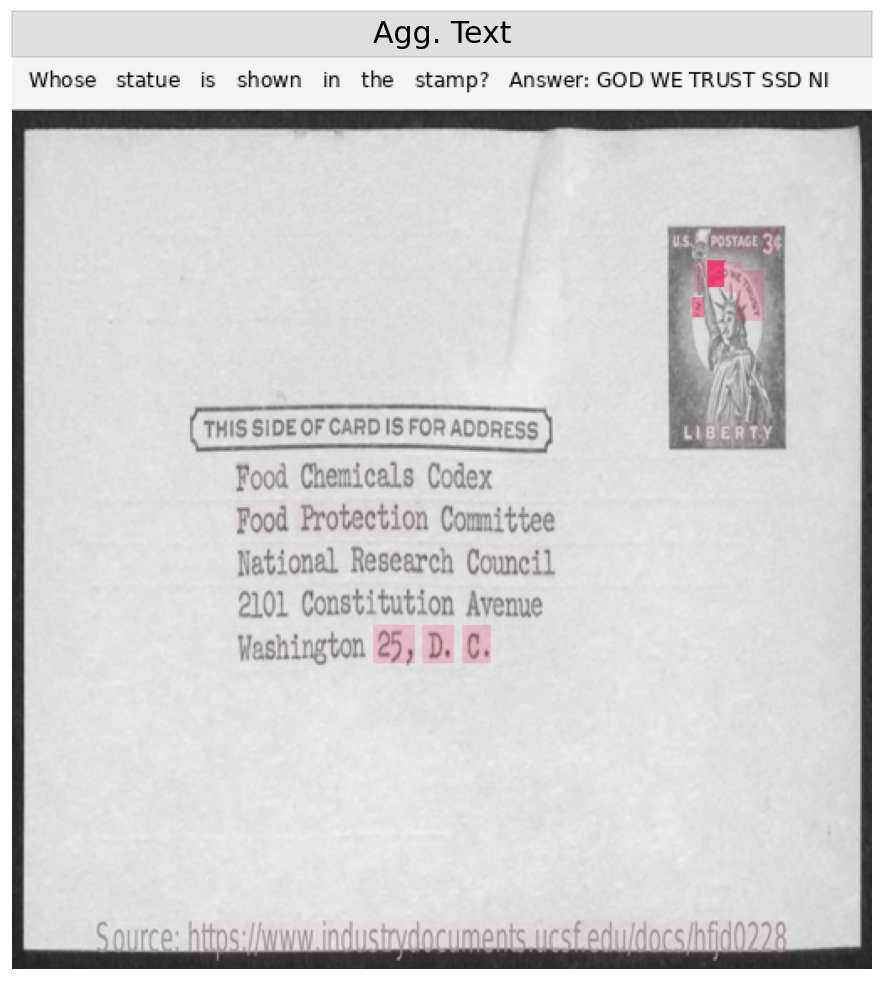}
\end{subfigure}
\par\smallskip
\noindent
\parbox[c]{1.5em}{\centering\rotatebox{90}{\small IG (DocVQA)}}
\begin{subfigure}[c]{0.8\linewidth}
\includegraphics[width=\textwidth]{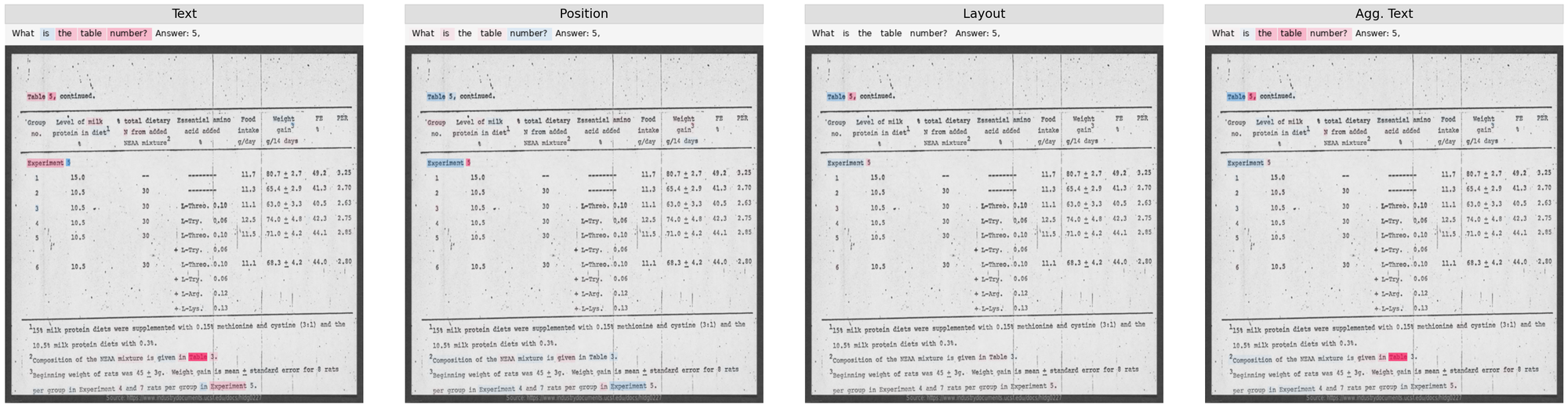}
\end{subfigure}
\par\smallskip
\noindent
\parbox[c]{1.5em}{\centering\rotatebox{90}{\small Occ. (DocVQA)}}
\begin{subfigure}[c]{0.8\linewidth}
\includegraphics[width=\textwidth]{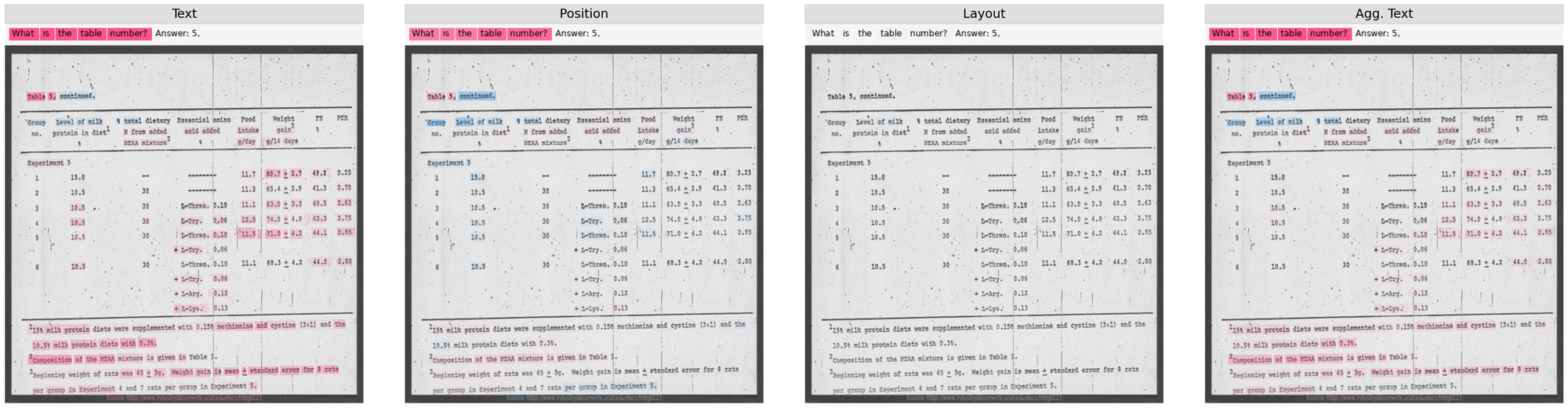}
\end{subfigure}
\par\smallskip
\noindent
\parbox[c]{1.5em}{\centering\rotatebox{90}{\small RA (DocVQA)}}
\begin{subfigure}[c]{0.8\linewidth}
\includegraphics[width=0.25\textwidth]{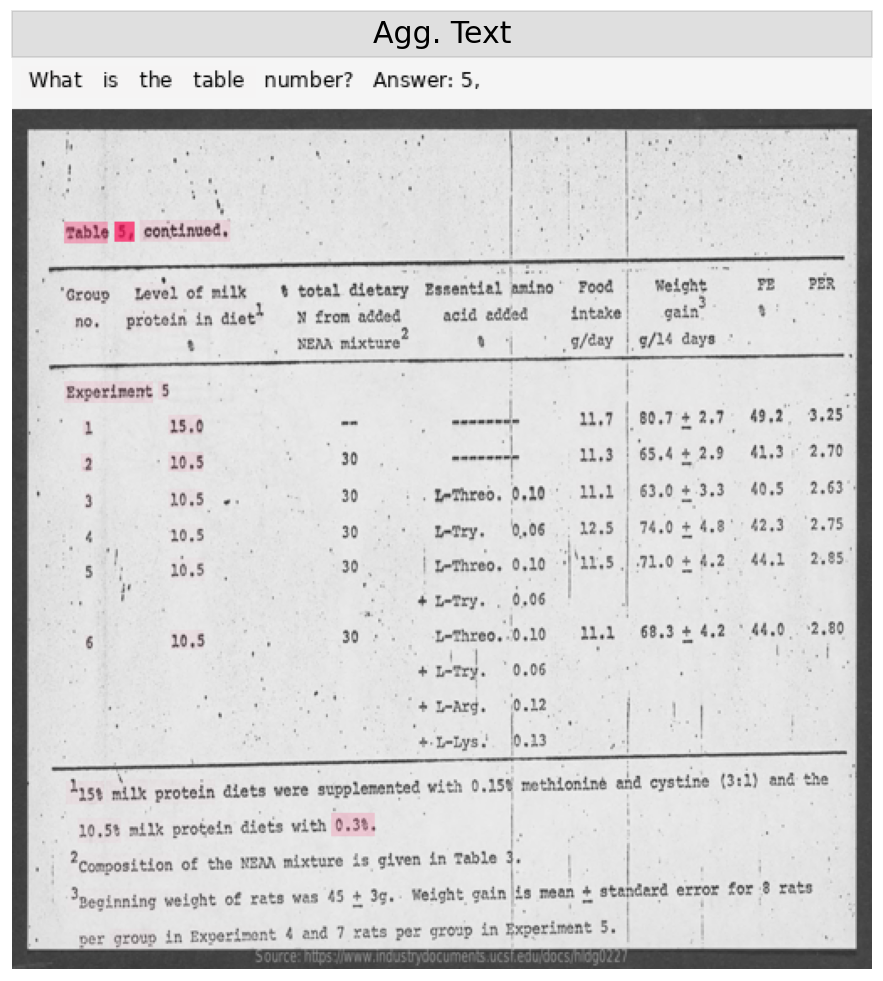}
\end{subfigure}
\caption{Modality-wise explanations of the \lilt{} model predictions for two randomly selected samples from \docvqa{} (\qa{} task). Results are shown for three methods: \integratedgradientsf{}, \occlusionf{}, and \rawattnf{}.}
\label{fig:qual_vqa_app_lilt}
\end{figure*}

\FloatBarrier
\begin{table*}[!t]
\caption{Modality contribution fractions computed for each model across all \cls{} datasets. \label{tab:modality_fractions_cls}}
\centering
\resizebox{0.5\textwidth}{!}{
\begin{tabular}{lll cccc}
Dataset & Model & Explainer & Token & Position & Layout & Image \\
\midrule
\multirow[t]{12}{*}{Tobacco3482} & \multirow[t]{3}{*}{BERT} & IG & 0.91 & 0.09 & N/A & N/A \\
 &  & Occ. & 0.88 & 0.12 & N/A & N/A \\
 &  & RA & 1.00 & N/A & N/A & N/A \\
\cmidrule{2-7}
 & \multirow[t]{3}{*}{RoBERTa} & IG & 0.80 & 0.20 & N/A & N/A \\
 &  & Occ. & 0.65 & 0.35 & N/A & N/A \\
 &  & RA & 1.00 & N/A & N/A & N/A \\
\cmidrule{2-7}
 & \multirow[t]{3}{*}{LiLT} & IG & 0.71 & 0.29 & 0.00 & N/A \\
 &  & Occ. & 0.64 & 0.26 & 0.10 & N/A \\
 &  & RA & 0.55 & N/A & 0.45 & N/A \\
\cmidrule{2-7}
 & \multirow[t]{3}{*}{LayoutLMv3} & IG & 0.12 & 0.02 & 0.13 & 0.74 \\
 &  & Occ. & 0.23 & 0.08 & 0.16 & 0.53 \\
 &  & RA & 0.00 & N/A & N/A & 1.00 \\
\cmidrule{1-7} \cmidrule{2-7}
\multirow[t]{12}{*}{RVL-CDIP} & \multirow[t]{3}{*}{BERT} & IG & 0.92 & 0.08 & N/A & N/A \\
 &  & Occ. & 0.67 & 0.33 & N/A & N/A \\
 &  & RA & 1.00 & N/A & N/A & N/A \\
\cmidrule{2-7}
 & \multirow[t]{3}{*}{RoBERTa} & IG & 0.76 & 0.24 & N/A & N/A \\
 &  & Occ. & 0.65 & 0.35 & N/A & N/A \\
 &  & RA & 1.00 & N/A & N/A & N/A \\
\cmidrule{2-7}
 & \multirow[t]{3}{*}{LiLT} & IG & 0.71 & 0.26 & 0.03 & N/A \\
 &  & Occ. & 0.57 & 0.27 & 0.17 & N/A \\
 &  & RA & 0.47 & N/A & 0.53 & N/A \\
\cmidrule{2-7}
 & \multirow[t]{3}{*}{LayoutLMv3} & IG & 0.55 & 0.07 & 0.21 & 0.18 \\
 &  & Occ. & 0.19 & 0.09 & 0.13 & 0.59 \\
 &  & RA & 0.17 & N/A & N/A & 0.83 \\
\cmidrule{1-7} \cmidrule{2-7}
\multirow[t]{12}{*}{DocLayNet} & \multirow[t]{3}{*}{BERT} & IG & 0.95 & 0.05 & N/A & N/A \\
 &  & Occ. & 0.90 & 0.10 & N/A & N/A \\
 &  & RA & 1.00 & N/A & N/A & N/A \\
\cmidrule{2-7}
 & \multirow[t]{3}{*}{RoBERTa} & IG & 0.74 & 0.26 & N/A & N/A \\
 &  & Occ. & 0.80 & 0.20 & N/A & N/A \\
 &  & RA & 1.00 & N/A & N/A & N/A \\
\cmidrule{2-7}
 & \multirow[t]{3}{*}{LiLT} & IG & 0.69 & 0.31 & 0.00 & N/A \\
 &  & Occ. & 0.70 & 0.22 & 0.09 & N/A \\
 &  & RA & 0.52 & N/A & 0.48 & N/A \\
\cmidrule{2-7}
 & \multirow[t]{3}{*}{LayoutLMv3} & IG & 0.35 & 0.07 & 0.27 & 0.31 \\
 &  & Occ. & 0.19 & 0.03 & 0.08 & 0.70 \\
 &  & RA & 0.02 & N/A & N/A & 0.98 \\
\bottomrule
\end{tabular}
}
\end{table*}

\begin{table*}[!t]

\caption{Modality contribution fractions computed for each model across all \ser{} datasets. \label{tab:modality_fractions_ser}}
\centering
\resizebox{0.5\textwidth}{!}{

\begin{tabular}{lll cccc}
Dataset & Model & Explainer & Token & Position & Layout & Image \\
\midrule
\multirow[t]{12}{*}{FUNSD} & \multirow[t]{3}{*}{BERT} & IG & 0.88 & 0.12 & N/A & N/A \\
 &  & Occ. & 0.76 & 0.24 & N/A & N/A \\
 &  & RA & 1.00 & N/A & N/A & N/A \\
\cmidrule{2-7}
 & \multirow[t]{3}{*}{RoBERTa} & IG & 0.46 & 0.54 & N/A & N/A \\
 &  & Occ. & 0.53 & 0.47 & N/A & N/A \\
 &  & RA & 1.00 & N/A & N/A & N/A \\
\cmidrule{2-7}
 & \multirow[t]{3}{*}{LiLT} & IG & 0.58 & 0.32 & 0.11 & N/A \\
 &  & Occ. & 0.53 & 0.34 & 0.13 & N/A \\
 &  & RA & 0.72 & N/A & 0.28 & N/A \\
\cmidrule{2-7}
 & \multirow[t]{3}{*}{LayoutLMv3} & IG & 0.25 & 0.29 & 0.42 & 0.04 \\
 &  & Occ. & 0.35 & 0.21 & 0.37 & 0.07 \\
 &  & RA & 0.98 & N/A & N/A & 0.02 \\
\cmidrule{1-7} \cmidrule{2-7}
\multirow[t]{12}{*}{CORD} & \multirow[t]{3}{*}{BERT} & IG & 0.94 & 0.06 & N/A & N/A \\
 &  & Occ. & 0.79 & 0.21 & N/A & N/A \\
 &  & RA & 1.00 & N/A & N/A & N/A \\
\cmidrule{2-7}
 & \multirow[t]{3}{*}{RoBERTa} & IG & 0.62 & 0.38 & N/A & N/A \\
 &  & Occ. & 0.45 & 0.55 & N/A & N/A \\
 &  & RA & 1.00 & N/A & N/A & N/A \\
\cmidrule{2-7}
 & \multirow[t]{3}{*}{LiLT} & IG & 0.64 & 0.32 & 0.04 & N/A \\
 &  & Occ. & 0.44 & 0.52 & 0.04 & N/A \\
 &  & RA & 0.76 & N/A & 0.24 & N/A \\
\cmidrule{2-7}
 & \multirow[t]{3}{*}{LayoutLMv3} & IG & 0.37 & 0.31 & 0.32 & 0.00 \\
 &  & Occ. & 0.31 & 0.38 & 0.31 & 0.00 \\
 &  & RA & 0.68 & N/A & N/A & 0.32 \\
\cmidrule{1-7} \cmidrule{2-7}
\multirow[t]{12}{*}{SROIE} & \multirow[t]{3}{*}{BERT} & IG & 0.85 & 0.15 & N/A & N/A \\
 &  & Occ. & 0.95 & 0.05 & N/A & N/A \\
 &  & RA & 1.00 & N/A & N/A & N/A \\
\cmidrule{2-7}
 & \multirow[t]{3}{*}{RoBERTa} & IG & 0.52 & 0.48 & N/A & N/A \\
 &  & Occ. & 0.63 & 0.37 & N/A & N/A \\
 &  & RA & 1.00 & N/A & N/A & N/A \\
\cmidrule{2-7}
 & \multirow[t]{3}{*}{LiLT} & IG & 0.50 & 0.47 & 0.04 & N/A \\
 &  & Occ. & 0.57 & 0.20 & 0.23 & N/A \\
 &  & RA & 0.75 & N/A & 0.25 & N/A \\
\cmidrule{2-7}
 & \multirow[t]{3}{*}{LayoutLMv3} & IG & 0.39 & 0.47 & 0.14 & 0.00 \\
 &  & Occ. & 0.60 & 0.17 & 0.18 & 0.06 \\
 &  & RA & 0.95 & N/A & N/A & 0.05 \\
\cmidrule{1-7} \cmidrule{2-7}
\multirow[t]{12}{*}{Wild Receipts} & \multirow[t]{3}{*}{BERT} & IG & 0.97 & 0.03 & N/A & N/A \\
 &  & Occ. & 0.81 & 0.19 & N/A & N/A \\
 &  & RA & 1.00 & N/A & N/A & N/A \\
\cmidrule{2-7}
 & \multirow[t]{3}{*}{RoBERTa} & IG & 0.69 & 0.31 & N/A & N/A \\
 &  & Occ. & 0.70 & 0.30 & N/A & N/A \\
 &  & RA & 1.00 & N/A & N/A & N/A \\
\cmidrule{2-7}
 & \multirow[t]{3}{*}{LiLT} & IG & 0.49 & 0.29 & 0.21 & N/A \\
 &  & Occ. & 0.50 & 0.34 & 0.16 & N/A \\
 &  & RA & 0.67 & N/A & 0.33 & N/A \\
\cmidrule{2-7}
 & \multirow[t]{3}{*}{LayoutLMv3} & IG & 0.67 & 0.23 & 0.10 & 0.00 \\
 &  & Occ. & 0.40 & 0.30 & 0.28 & 0.01 \\
 &  & RA & 0.96 & N/A & N/A & 0.04 \\
\end{tabular}
}
\end{table*}

\begin{table*}[!t]
\caption{Modality contribution fractions computed for each model on the \docvqa{} dataset. \label{tab:modality_fractions_qa}}

\centering

\resizebox{0.5\textwidth}{!}{
	
	\begin{tabular}{lll cccc}
		Dataset & Model & Explainer & Token & Position & Layout & Image \\
\midrule
\multirow[t]{12}{*}{DocVQA} & \multirow[t]{3}{*}{BERT} & IG & 0.83 & 0.17 & N/A & N/A \\
 &  & Occ. & 0.66 & 0.34 & N/A & N/A \\
 &  & RA & 1.00 & N/A & N/A & N/A \\
\cmidrule{2-7}
 & \multirow[t]{3}{*}{RoBERTa} & IG & 0.65 & 0.35 & N/A & N/A \\
 &  & Occ. & 0.59 & 0.41 & N/A & N/A \\
 &  & RA & 1.00 & N/A & N/A & N/A \\
\cmidrule{2-7}
 & \multirow[t]{3}{*}{LiLT} & IG & 0.55 & 0.29 & 0.17 & N/A \\
 &  & Occ. & 0.58 & 0.31 & 0.12 & N/A \\
 &  & RA & 0.61 & N/A & 0.39 & N/A \\
\cmidrule{2-7}
 & \multirow[t]{3}{*}{LayoutLMv3} & IG & 0.57 & 0.20 & 0.23 & 0.00 \\
 &  & Occ. & 0.61 & 0.20 & 0.16 & 0.03 \\
 &  & RA & 0.72 & N/A & N/A & 0.28 \\
\bottomrule
\end{tabular}
}
\end{table*}

\subsection{Per-Task Metric Agreement}
\label{sec:appendix_metric_agreement}
Fig.~\ref{fig:corr_faithfulness_per_task} and Fig.~\ref{fig:corr_complexity_per_task} present the task-wise pairwise Spearman correlations for the faithfulness and complexity metrics. Overall, the task-wise faithfulness and complexity correlation patterns were largely consistent with the aggregated results reported in Fig.~\ref{fig:faithfulness_corr_overall}.

\subsection{AOPC Curves}
\label{sec:appendix_aopc}
Figs.~\ref{fig:aopc_cls},~\ref{fig:aopc_ser}, and~\ref{fig:aopc_vqa} present the \aopc{} curves for all XAI methods across all dataset-model combinations for the \cls{}, \ser{}, and \qa{} tasks, respectively. We visualize both MoRF (solid) and LeRF (dotted) curves relative to the random baseline. Higher MoRF and lower LeRF curves indicate better faithfulness.

\subsection{Full Metric Results}
\label{sec:appendix_full_results}
The complete evaluation results across all 17 metrics, 13 XAI methods, 8 datasets, and 4 models are provided at \url{https://docs.google.com/spreadsheets/d/12oEqjm-a1fHae8TKfcVGBTjq0mFAMtpsEI-_CFQJeeU/edit?gid=0#gid=0}. These results include the per-dataset mean scores for each metric-method-model combination, which were used to compute the rankings and correlation analyses presented in Section~\ref{sec:results_ranking} and Section~\ref{sec:results_agreement}.

\subsection{Additional Qualitative Results}
\label{sec:appendix_qualitative}
Figs.~\ref{fig:qual_cls_app_layoutlm},~\ref{fig:qual_ser_app_layoutlm}, and~\ref{fig:qual_vqa_app_layoutlm} present additional qualitative results for the \layoutlm{} model, and Figs.~\ref{fig:qual_cls_app_lilt},~\ref{fig:qual_ser_app_lilt}, and~\ref{fig:qual_vqa_app_lilt} for the \lilt{} model. All explanations are generated using \occlusionf{}, \integratedgradientsf{}, and \rawattnf{}.

\subsection{Dataset-wise Modality Contributions}
\label{sec:appendix_modality}
Table~\ref{tab:modality_fractions_cls}, Table~\ref{tab:modality_fractions_ser}, and Table~\ref{tab:modality_fractions_qa} respectively present the overall modality contribution for \layoutlm{} and \lilt{} across the \cls{}, \ser{}, and \qa{} tasks, computed using \occlusion{}, \integratedgradients{}, and \rawattn{}.
\end{appendices}
\end{document}